\newif\ifpdf
\ifx\pdfoutput\undefined%
\pdffalse%
\else%
\pdfoutput=1%
\pdftrue%
\fi

\ifpdf
\documentclass[pdftex,a4paper]{illcdiss}
\else
\documentclass[a4paper,makeidx]{illcdiss}
\fi

\usepackage[utf8]{inputenc} 
\usepackage{url}

\ifpdf
\usepackage[pdftex, draft=false, colorlinks=true, urlcolor=black, linkcolor=black, citecolor=black, bookmarks=true, plainpages=false, hyperfootnotes=false, pdfauthor={Hans Mueller}, pdftitle={Multiple Classifier Systems Incorporating Uncertainty}, pdfsubject={Multiple Classifier Systems Incorporating Uncertainty, PhD Thesis, Ulm 2009},pdfkeywords={mcs, multiple classifiers systems, uncertainty, dempster-shafer, machine learning, classifier, certainty}]{hyperref} 
\fi

\usepackage{algorithm}
\usepackage{algorithmic}
\usepackage{mset} 
\usepackage{amsmath,amssymb,amsfonts} 
\usepackage{algorithmic}
\usepackage{amssymb} 
\usepackage{mathrsfs} 
\usepackage{amsthm} 
\usepackage{color} 
\usepackage{colortbl} 
\usepackage{graphicx}
\usepackage[english]{babel} 
\usepackage{setspace} 
\usepackage{tocloft} 
\usepackage{stdclsdv} 
\usepackage{float} 
\usepackage{fancyhdr}
\usepackage{tabularx}
\usepackage{lscape} 
\usepackage{multirow}
\usepackage{rotating} 
\usepackage{morefloats}
\usepackage[section, above, below]{placeins} 
\usepackage[bf]{caption}
\usepackage{booktabs}
\usepackage{subfig}
\usepackage[export]{adjustbox}

\newcommand{\ssymbol}[1]{^{\@fnsymbol{#1}}}

\definecolor{dark_grey}{rgb}{0.5,0.5,0.5}
\definecolor{light_grey}{rgb}{0.8,0.8,0.8}

\newcolumntype{C}{>{\centering\arraybackslash}X}
\newcolumntype{S}{>{\small\arraybackslash}X}
\newcolumntype{F}{>{\footnotesize\arraybackslash}X}
\newcolumntype{T}{>{\tiny\arraybackslash}X}
\newcolumntype{P}{>{\tiny\arraybackslash}p}
\newcolumntype{L}{>{\tiny\arraybackslash}l}

\usepackage{amsmath}
\DeclareMathOperator*{\argmax}{arg\,max}
\DeclareMathOperator*{\argmin}{arg\,min}
\theoremstyle{definition}

\usepackage{booktabs}
\usepackage[export]{adjustbox}
\usepackage{multirow}
\usepackage{url}
\usepackage{hyperref}  
\usepackage[utf8]{inputenc}
\usepackage[font=small,justification=justified]{caption}
\usepackage[hang,flushmargin]{footmisc}
\DeclareUnicodeCharacter{00A0}{~}
\DeclareUnicodeCharacter{05F3}{\'}

\begin{document}
\pagestyle{plain}
\pagenumbering{alph}


{\pagestyle{empty}
\newcommand{\printtitle}{%
{\Large\bf
Deep Learning for Robust and Explainable \\
Models in Computer Vision
\\}}    



%
%
\begin{titlepage}
\clearpage

\includegraphics[height=1.8cm]{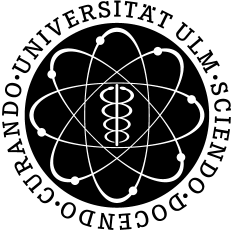}
\hfill
\includegraphics[height=1.8cm]{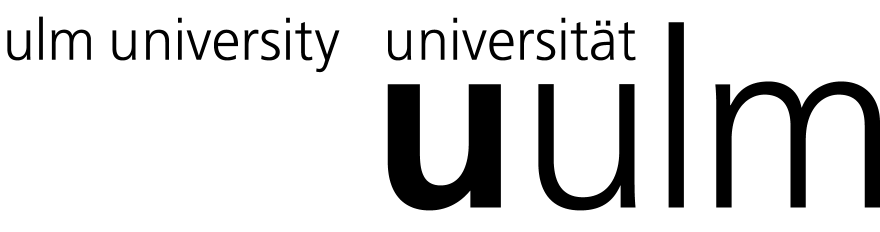}\\[1em]

{\footnotesize
{\bfseries Universit\"at Ulm} \textbar ~89069 Ulm \textbar ~Germany
\hspace*{0.35cm}{\bfseries Fakult\"at f\"ur Ingenieurwissenschaften,}\\
\hspace*{7.1cm}{\bfseries Informatik und Psychologie}\\
\hspace*{7.1cm}Institut f\"ur Neuroinformatik\\
\hspace*{7.1cm}Direktor: Prof. Dr. Dr. Daniel Alexander Braun \\
~\\~\\
}



\par\vskip 2cm
\noindent \rule{0.9\textwidth}{2pt}
\par\vspace{0.3cm}
\noindent \printtitle
\par\vspace {3cm}
\noindent
Dissertation zur Erlangung des Doktorgrades\\
Doktor der Naturwissenschaften (Dr. rer. nat.)\\
der Fakult\"at f\"ur Ingenierwissenschaft, Informatik und Psychologie \\
der Universit\"at Ulm\\
\par\vspace {1.5cm}
\noindent vorgelegt von\\
Mohammadreza Amirian\\
aus Shiraz
\par\vfill
\noindent Ulm 2023
\end{titlepage}

%
\mbox{}\newpage

\clearpage \mbox{}
 \vfill
\noindent%
Amtierende Dekanin: Prof. Dr. Anke Huckauf \\
~\\
\par
\noindent
Gutachter: Prof. Dr. Friedhelm Schwenker\\            
Gutachter: Prof. Dr. Thilo Stadelmann   \\            
Gutachter: Prof. Dr. Martin Jaggi       \\            
\par
\noindent
Tag der Promotion: 20\textsuperscript{th} October 2023\\



\clearpage

%
\mbox{}\newpage \setcounter{page}{1}
} 


\cleardoublepage
\pagenumbering{roman}
\section*{Acknowledgements}
I want to extend my heartfelt gratitude to the individuals who have played essential roles in my academic journey while completing my Ph.D. dissertation. Their support, mentorship, and presence have been invaluable, and I am deeply appreciative. I must begin by acknowledging the support of my family: Edalat, Parivash, and Milad. Their belief in my aspirations and constant encouragement has strengthened me throughout this demanding journey.

I am profoundly grateful to my distinguished Ph.D. committee, particularly Professor Friedhelm Schwenker, to whom I owe special thanks for his mentorship. The wisdom and expertise of Professors Thilo Stadelmann, Martin Jaggi, Hans Kestler, Daniel Alexander Braun, Günther Palm, and Matthias Tichy have been instrumental in shaping my research and academic growth. My gratitude extends to project managers Stefan Scheib and Frank-Peter Schilling, whose guidance and collaboration significantly enriched my research endeavors. 

The diverse and talented group of coauthors, who have made invaluable contributions to my track record, have enriched the academic discourse and expanded my horizons within the realm of research. Their collaborative spirit, combined with our shared dedication to the subject matter, has played a pivotal role in advancing my research. I want to extend my gratitude to each coauthor for their instrumental contributions to my academic journey, including Lukas Tuggener, Patrick Thiam, Viktor Kessler, Markus Kächele, Stefan Scheib, Javier Montoya, Alexander Züst, Ivo Herzig, Peter Eggenberger Hotz, Rudolf Marcel Füchslin, Pascal Paysan, Igor Peterlik, Samuel Wehrli, Corinna Hertweck, Stefan Glüge, Lukas Lichtensteiger, Marco Morf, Fernando Benites, Pius von Däniken, Peter Bellmann, Georg Layher, Yan Zhang, Maria Velana, Sascha Gruss, Steffen Walter, Harald Christhelm Traue, Daniel Schork, Jonghwa Kim, Elisabeth André, Heiko Neumann, Taye Girma Debelee, Abraham Gebreselasie, Dereje Yohannes, Kamran Kazemi, Mohammad Javad Dehghani, Jonathan Gruss, Yves D. Stebler, Ahmet Selman Bozkir, Marco Calandri, Ricardo Chavarriaga, Yvan Putra Satyawan, and Dandolo Flumini. Each has left an indelible mark on my academic journey, and I am grateful for their collaborative efforts.

I also want to acknowledge the colleagues I've worked with at Ulm University, Zurich University of Applied Sciences (ZHAW), and the Swiss Federal Institute of Technology in Lausanne (EPFL). Special thanks to Patrick, Viktor, and Heinke from Ulm. Your inspiration and shared commitment to academic excellence have impacted my Ph.D. journey. Special thanks to my colleagues and collaborators at ZHAW: Jonas, Katrin, Yasmin, Sean, Adhiraj, Yvan, Sebastiano, Susanne, Norman, Raphael, Pascal, Peng, Katsiaryna, Daniel, Philipp, Jonathan, Catherine (for her invaluable proofreading), Claude, Gabriel, and Ahmad. And to my colleagues at EPFL: Mary-Anne, Anastasia, Lie, Matteo, Amirkeivan, El Mahdi, Atli, Thijs, Jean-Baptiste, Tao, Prakhar, Praneeth, and Seyed-Mohsen. Your shared experiences, diverse backgrounds, and academic commitment have enriched my academic journey.

I expand my appreciation to the administrative staff at Ulm University, ZHAW, and EPFL, including Traude, Birgit, Annette, Cornelia, Regula, Pamela, and Jennifer, whose efficient and supportive work has facilitated my academic journey.

Special mention to my flatmates, Patrick, Draženka, Patrick, Valentina, and Sandro, who have contributed to my social life and created a supportive and nurturing environment. Close friends and community in Ulm, Winterthur, and Zurich area, including Abbas, Ramanjeet, Alper, Meissam, Gholamhossein, Mahsa, Maryam, Fatemeh, Sajad, Mahdi, Elham, Timo, Ronak, Reza, Romina, Farhad, Najmeh, Pooya, Helmut, Javier, Giani, and Roshan, have been a source of joy and balance, significantly enhancing my social life. I would also like to express my gratitude to those who have enriched my life in Switzerland beyond academia, including the FC Kreuzlingen football club, the Keller family, and the Bailamos Salsa Club.

The support, mentorship, and friendship of countless individuals have profoundly shaped my Ph.D. dissertation journey. I am deeply thankful for the lessons, experiences, and relationships that contributed to my academic and personal growth. Your contributions have been invaluable in making this academic journey possible, and I am sincerely grateful to each of you.
\abstract
Recent breakthroughs in machine and deep learning (ML and DL) research have provided excellent tools for leveraging enormous amounts of data and optimizing huge models with millions of parameters to obtain accurate networks for image processing. These developments open up tremendous opportunities for using artificial intelligence (AI) in the automation and human assisted AI industry. However, as more and more models are deployed and used in practice, many challenges have emerged. This thesis presents various approaches that address robustness and explainability challenges for using ML and DL in practice.

Robustness and reliability are the critical components of any model before certification and deployment in practice. Deep convolutional neural networks (CNNs) exhibit vulnerability to transformations of their inputs, such as rotation and scaling, or intentional manipulations as described in the adversarial attack literature. In addition, building trust in AI-based models requires a better understanding of current models and developing methods that are more explainable and interpretable a priori. 

This thesis presents developments in computer vision models' robustness and explainability. Furthermore, this thesis offers an example of using vision models' feature response visualization (models' interpretations) to improve robustness despite interpretability and robustness being seemingly unrelated in the related research. Besides methodological developments for robust and explainable vision models, a key message of this thesis is introducing model interpretation techniques as a tool for understanding vision models and improving their design and robustness. In addition to the theoretical developments, this thesis demonstrates several applications of ML and DL in different contexts, such as medical imaging and affective computing.

\cleardoublepage
\tableofcontents
\clearpage
\listoffigures
\clearpage
\listoftables



\cleardoublepage
\pagestyle{headings}
\pagenumbering{arabic}

\chapter{Introduction}
\label{chap:introduction}
\vspace{-0.5cm}
As a result of the widespread interest in applying artificial intelligence (AI) in practice, several intriguing challenges and research topics have recently emerged due to the trustworthiness of models in various circumstances being brought into question~\cite{huang2020survey}. Researchers have cited explainability, robustness, and fairness among other hindrances in developing trustworthy AI~\cite{kaur2022trustworthy}. 
Therefore, understanding the reasons for failure and creating ability to explain the inner workings of neural networks has attracted researchers' attention~\cite{dovsilovic2018explainable, tjoa2020survey}. Furthermore, developing interpretable and explainable models has become a research focus in its own right~\cite{gilpin2018explaining}.

The three terms \textit{robustness}, \textit{explanability}, and \textit{interpretabilty} are the fundamental concepts behind this thesis. 
The term \textit{robustness} refers to a clearer concept compared with \textit{explanability} and \textit{interpretabilty}. Model \textit{robustness} is proportional to the consistency of the model's performance against naturally-induced or manually-computed corruption and alterations affecting the data to deviate from the training distribution~\cite{drenkow2021robustness, madry2017towards}.
However, the other two terms, \textit{explainability} and \textit{interpretability}, and their boundaries and overlaps are still subjects of research at the taxonomy level~\cite{graziani2022global}.
In this thesis, the terms \textit{explanability} and \textit{interpretabilty} are used analogously to their usage in~\cite{rudin2019stop}. 
Accordingly, \textit{explanability} refers to the explanation of models' decisions (even though these models can be intrinsically black boxes), 
         and \textit{interpretability} refers to the design patterns that are inherently interpretable and understandable by humans.

The scope of this work is narrowed down from AI in general to focus on computer vision models, including convolutional neural networks (CNNs) and vision transformers (ViTs). 
This thesis is motivated by practical applications and presents relevant research concerning neural networks' robustness, fairness, interpretability, and explainability. 
Moreover, the thesis provides not only theoretical and fundamental advances but also offers several applications in which computer vision models have been successfully used.
The remainder of this chapter explains the motivations for this research and describes the scientific problem we address. 
Finally, this chapter provides a list of papers and publications related to this research, followed by the thesis organization.

\section{Motivation}

This thesis aims at using ML and DL in practical applications, and presents several success stories in Chapter~\ref{chap:motion} and Chapter~\ref{chap:applications}. Despite the numerous successful applications of DL, there are deterrents for putting it into practice in sensitive applications where humans are involved. This thesis presents relevant research tackling such challenges as follows: 1) adversarial robustness in Chapter~\ref{chap:trace}, 2) explainability via interpretable classifiers in Chapter~\ref{chap:rbfs}. The remainder of this section describes the motivation of the researchers, elaborates on related efforts, and describes the niches to which this thesis contributes.  

Researchers attempted to expose the complications of using DL in practice by studying robustness~\cite{bai2021transformers}, fairness~\cite{mehrabi2021survey}, explainability~\cite{dovsilovic2018explainable}, interpretability~\cite{chakraborty2017interpretability}, accountability~\cite{kim2020transparency}, reliability~\cite{dos2018analyzing}, safetly~\cite{ahmed2021intelligent}, and privacy~\cite{huang2019lightweight} etc. Although these themes were independently the subject of research and scientific concern, they have only recently been grouped under the overarching topic referred to as \textit{trustworthy AI} in the literature~\cite{kaur2022trustworthy}. Trustworthy AI literature summarizes the research effort as developing models which are effective in practice and aligned with positive societal effects. The following paragraphs explain the key components of trustworthy AI which are considered in this thesis,\textit{explanability}, \textit{interpretability}, \textit{robustness} and \textit{fairness}, and describe the goals of the related research.

After researchers found flaws in the preciseness and biasedness in vision models~\cite{szegedy2013intriguing, crawford2016artificial} as well as natural language processing methods~\cite{bolukbasi2016man, wolf2017we}, the European Union introduced the ``right to explain'' in the general data protection regulation (GDPR) as an attempt to protect human rights when decisions are automated~\cite{dovsilovic2018explainable}. This human right relates to the human's ability to understand the AI-based agent logic in human-machine interaction~\cite{rosenfeld2020and}. Some of the research work related to \textit{explainability} targets explaining the models' decisions even if researchers treat the models as black boxes~\cite{crabbe2022label}. The concept of \textit{interpretability} includes research attempting to open the black box of neural networks with revealing patterns about the inner mechanism of models~\cite{zablocki2022explainability}. Saliency map visualizations~\cite{bojarski2018visualbackprop} and feature response visualization methods~\cite{springenberg2014striving} are examples of researchers' endeavors in targeting interpretability.

The topic of \textit{robustness} is directly related to the performance of vision models. Vision models have shown a drop in performance as the consequence of changes in data distribution based on naturally-induced variations~\cite{drenkow2021robustness} or manually-computed perturbations~\cite{madry2017towards}. Robustness is a significant hurdle to overcome in life-long deployments of vision models, which has inspired many recent research works. An example of the output of such research is equivariant CNNs, which aim to improve robustness against rotation and translation of input images~\cite{cohen2016group, romero2020group, hutchinson2021lietransformer}. Moreover, adversarial training research attempts to neutralize the targeted attack's effect in fooling vision models~\cite{tramer2019adversarial}. Still, there are several gaps in dataset collection and robust model development for naturally-induced variations and different illumination conditions~\cite{li2019illumination}.

Social activists and computer vision researchers recently raised concerns regarding \textit{fairness} in automated decisions~\cite{hill2020wrongfully, bellamy2019ai}. Automated decisions are considered unfair if they rely on sensitive variables such as gender, ethnicity, sexual orientation, or disability~\cite{verma2018fairness}. The researchers identified sources of bias leading to unfair decisions, which can be divided into two general categories of algorithmic biases, and biases in the training datasets~\cite{mehrabi2021survey}. This topic gained much attention and media coverage after the deployment of face recognition (FR) systems in public surveillance~\cite{hill2020wrongfully, mac2021facebook, lohr2018facial, crawford2016artificial}. Since then, many researchers have attempted to tackle the problem of biasedness at the algorithmic level~\cite{robinson2020face, serna2022sensitive} and collect diverse datasets to achieve fair modeling for all genders and races~\cite{wang2019racial, smith2019ibm}.

This thesis explores the obstacles to using AI, particularly machine learning (ML) and deep learning (DL), in practical applications of computer vision in which robustness and explainability are of high importance. The ultimate goal of this work is the successful application of ML and DL algorithms, although this is not a trivial task and raises many additional questions that require further research. A short phrase that summarizes the long-term goal alongside the focus of this work is the development of \textit{trustworthy AI}. Although the term \textit{trustworthy AI} has only recently come into vogue, this thesis addresses its various components, including fairness, robustness, interpretability, and explainability. Trustworthiness includes other elements outside the scope of this research, such as security, privacy, and accountability. This thesis presents several applications besides fundamental research that support robust and explainable AI (XAI). 

Understanding the behavior of computer vision models (explainability) has always been a subject of curiosity for scientific endeavors. The first group of researchers who analyzed current models considered them to be black boxes and predicted their performance by changing the input and observing the behavior of the models' output. The second group of researchers proposed intrinsically more interpretable and explainable models. This thesis presents a chapter on using radial basis function networks (RBFs) as classifiers on top of CNNs to improve the interpretability of decision-making in computer vision models which contributes to XAI research.

Early in the development of CNNs, researchers found that computer vision models were only robust in a limited range of rotation and scaling in the input images\footnote{\url{http://yann.lecun.com/exdb/lenet}}. In addition, lighting conditions and other environmental disturbances caused errors in recognizing image patterns. Last but not least, the researchers found that optimizing images made it possible to fool the computer vision models into interpreting two images, which humans perceive to be identical, differently, leading to the computation of so-called adversarial perturbations. Since then, improving the robustness of computer vision models has become a popular research topic. Researchers have made enormous efforts to understand the models, identify the reasons behind failures, and improve the computer vision models. Interpreting the behavior of computer vision models can serve as a tool to monitor the reasons for failures. Therefore, interpretability and robustness are closely related in the literature. For example, researchers have found that computer vision models can focus on the wrong features or background information when classifying an object. This thesis includes a chapter on using feature response maps--where a computer vision focuses its attention in response to the visual input--for identifying adversarial examples.

In addition to the above theoretical developments, this thesis presents many applications inspired by its original goal. It targets the vulnerabilities (motion artifacts) found in classical computed tomography (CT) reconstruction methods as the main practical contribution. Moreover, it presents many other side contributions to affective computing, pain estimation, AutoML, AutoDL, medical data homogenization, and fairness in face recognition systems.

\section{Problem Statement}
This thesis is motivated by solving real-world problems using computer vision methodology. The applications presented in Chapter~\ref{chap:motion} and Chapter~\ref{chap:applications} are derived from several real-world problems where ML and DL are useful. However, there are still gaps in the current methodologies that must be addressed in order to achieve trustworthy models for general applications. Chapter~\ref{chap:rbfs} and Chapter~\ref{chap:trace} present the fundamental research targeting the explainability and robustness gaps required to apply computer vision models in practice. The remainder of this section details the scientific problems addressed in each chapter individually.

Current architectures for computer vision models based on CNNs and ViTs use a stack of convolutional or self-attention layers to develop a representation of the inputs. Despite the different architectures in the image encoder of most computer vision models, all of these models use a stack of multiple fully connected (FC) layers or multilayer perceptrons (MLP), on top of learned latent representations~\cite{bianco2018benchmark}. Researchers commonly used FC layers as the optimal classifiers for deep models because of their efficiency in gradient backpropagation~\cite{lecun1998gradient}. MLPs divide the embedding space in their last layer into multiple classes using hyperplanes. The distance of the input image representations from the decision boundary drawn by the hyperplanes in the last layer of such classifiers determines the decision confidence of the models. Researchers found that such classification is not optimal for outliers, since models developed using MLP classifiers demonstrate low reliability for random (garbage) classes. This is due to the outliers being far from the classifier's decision boundary, which contributes to the models' flawed high confidence in these samples. In addition, computer vision models trained using linear classifiers are vulnerable to optimized perturbations (adversarial attacks). Ian Goodfellow has attempted to thoroughly analyze various classifiers to evaluate their robustness to adversarial attackers and garbage classes~\cite{goodfellow2014explaining}. His preliminary speculations hint that RBFs might be more robust than MLPs. However, the study was inconclusive due to the difficulties in optimizing RBF networks, even for simple tasks such as classifying handwritten digits. This thesis proposes modifications to RBF networks that improve the optimization of RBFs and shows how RBF classifiers are beneficial for interpreting the decision-making of computer vision models. 

After the advent of CNNs, researchers were very skeptical and curious about their functionality. They work as highly accurate models but seem to appear as black boxes. The result of this curiosity and scientific venture is a vast amount of literature analyzing the behavior of CNNs by computing the models' feature response maps through the inversion of the forward path. As extensive as the techniques for visualizing models are, their applications are rare. This thesis presents an example of detecting adversarial attacks using feature response maps. The intention is that this idea inspires researchers to use their knowledge of interpreting computer vision models for architecture development and debugging. 

Neural networks have outperformed their competitors in approximating arbitrary functions and learning patterns from enormous amounts of data. However, AI projects still face high risks due to not achieving the intended goal, unforeseen delays, extensions, and application failures. This thesis presents several successful applications that allow the reader to understand where using ML and DL-based methods are beneficial. For example, we address motion artifacts in cone beam computed tomography (CBCT) scans. Volumetric (3D) data from CBCT scans are reconstructed from hundreds of 2D X-ray images from different angles. The analytical reconstruction algorithms are robust when the target volumes are constant and free of motion. However, this assumption does not hold due to respiratory or cardiac motions present in the human body. In this work, we demonstrate how CNNs can be used to compensate for motion artifacts in CBCT scans. Along with this application, this thesis offers several other applications to show some possible and successful venues in which ML- and DL-based models are superior to classical computer vision methods in practice.

\section{Contributions}

The main contributions of this thesis to ML and DL research are as follows:
\begin{itemize}
    \item Chapter~\ref{chap:rbfs}: Modern vision architectures use multilayer perceptrons (MLPs) in the form of fully connected layers as classifiers, as researchers have largely abandoned radial basis function networks (RBFs) due to optimization problems. This thesis provides the following developments in training RBFs as classifiers for convolutional neural networks (CNN) backbones: 1) Presentation of the first successful attempt to use RBFs as the classifier of modern computer vision models for object classification. 2) Introduction of a novel quadratic activation function to build a linear computational graph with RBFs. 3) Simultaneous optimization of supervised loss for classification and unsupervised loss for clustering~\cite{amirian2020radial}.
    \item Chapter~\ref{chap:rbfs}: Solving the technical problems of optimizing RBFs as classifiers for computer vision models opens several possibilities for training computer vision models: 1) Combining supervised and unsupervised learning by simultaneously optimizing two target losses. 2) Learning a similarity distance metric to find similar images by optimizing the covariance matrix in the embedding space. 3) Improving the interpretability of the computer vision models by visualizing the data using prototypes and learning more about the models' decision-making~\cite{amirian2020radial}. 
    \item Chapter~\ref{chap:trace}: This thesis presents findings on a well-known vulnerability in the robustness of computer vision models referred to as adversarial attacks in related literature. First, the research presented in Chapter~\ref{chap:trace} shows how adversarial perturbations leave a detectable trace on the feature response map of CNNs, even though the input image remains identical. Then, feature response maps of CNNs are used with a simple and effective algorithm to detect adversarial attacks with a very competitive accuracy compared to state-of-the-art methods~\cite{amirian2018trace}.  
    \item Chapter~\ref{chap:motion}: Motion artifacts in medical images are a common problem, especially for lengthy acquisition times. This work provides a data-driven solution based on supervised learning to reduce motion artifacts where no analytical solution exists. The proposed solution addresses motion reduction in two reconstruction methods (analytical and iterative) and reduces artifacts in raw data (acquired projections) and reconstructed scans (volume domain). The target domain of this method is cone-beam computed tomography (CBCT) scans, which are used for automatic segmentation and dose calculation in cancer therapy. In this thesis, we present techniques for training models on simulated data that achieve an improvement of over $6~dB$ in terms of signal-to-noise ratio (PSNR). Moreover, the proposed models generalize to real-world data, and clinical experts have verified their performance in the first attempt at motion compensation for CBCT scans. 
    \item Chapter~\ref{chap:applications}: Optimizing ML and DL models and finding the best models and architectures for small datasets is an intriguing area of research. This thesis presents the most relevant findings from research in automated machine and deep learning (AutoML and AutoDL). The experiments in the context of automated machine learning show that optimizing the parameters of Gaussian processes as surrogate models for hyperparameter spaces (HPs) is the most successful method for HP tuning and meta-learning in AutoML. Moreover, the experimental results in the context of automated deep learning show that regularization and augmentation are the keys for fitting computer vision models to small datasets, that pre-trained models consistently outperform randomly initialized ones, and that large classifiers train faster than smaller ones~\cite{tuggener2019automated, tuggener2020design}.
    \item Chapter~\ref{chap:applications}: Domain adaptation and merging datasets from multiple data sources in medical imaging is a current research challenge. This thesis proposes an autoencoder-based architecture trained using an adversarial loss to preprocess 2D computed tomography scans for merging multiple datasets with minimal changes in the original scans. The proposed method extends classical training, validation, and testing performance to evaluate cross-dataset generalization and improves the cross-dataset performance for COVID detection from lung CT scans by over $10\%$~\cite{amirian2021prepnet}.
    \item Chapter~\ref{chap:applications}: This thesis presents relevant findings on the measurement of different sorts of biases in face recognition (FR) systems and the relationship between algorithmic bias and awareness. First, after analyzing the results of different models and network embeddings, this work concludes that awareness is not a good proxy for measuring racial bias in FR systems. Second, this thesis presents evidence that models which are designed to be unaware of race are not necessarily unbiased and suggest that further measures are critical for achieving fairness in FR systems~\cite{gluge2020not, wehrli2021bias}.
\end{itemize}
\newpage
\section{Publications}
This section presents the list of peer-reviewed and published research papers connected to this thesis, divided based on the publication venue into two groups of journal and conference contributions.
\subsection{Journal Papers}
The following is a list of peer-reviewed and published research papers in scientific journals contributing to this thesis:
\begin{itemize}
    \item \textbf{Mohammadreza Amirian}, Javier Montoya, Thilo Stadelmann, Frank-Peter Schilling, Rudolf Marcel F\"uchslin, Ivo Herzig, Peter Eggenberger Hotz, Lukas Lichtensteiger, Marco Morf, Alexander Z\"ust, Pascal Paysan, Igor Peterlik, and Stefan Scheib. ``Mitigation of motion-induced artifacts in Cone Beam Computed Tomography using Deep Convolutional Neural Networks." Journal of Medical Physics, pp. 6228-6242 (2023)~\cite{amirian2023mitigation}.
    \item Ivo Herzig, Pascal Paysan, Stefan Scheib, Alexander Züst, Frank-Peter Schilling, Javier Montoya, \textbf{Mohammadreza Amirian}, Thilo Stadelmann, Peter Eggenberger Hotz, Rudolf Marcel Füchslin, Lukas Lichtensteiger. ``Deep learning-based simultaneous multi-phase deformable image registration of sparse 4D-CBCT". Medical Physics, pp. e325-e326 (2022)~\cite{herzig2022deep}.
    \item Samuel Wehrli, Corinna Hertweck, \textbf{Mohammadreza Amirian}, Stefan Gl\"uge, and Thilo Stadelmann. ``Bias, awareness, and ignorance in deep-learning-based face recognition." AI and Ethics, pp. 1-14 (2022)~\cite{wehrli2021bias}.
    \item Lukas Tuggener, \textbf{Mohammadreza Amirian}, Fernando Benites, Pius von D\"aniken, Prakhar Gupta, Frank-Peter Schilling, and Thilo Stadelmann. ``Design patterns for resource-constrained automated deep-learning methods." AI, pp. 510-538 (2020)~\cite{tuggener2020design}.
    \item \textbf{Mohammadreza Amirian}, and Friedhelm Schwenker. ``Radial basis function networks for convolutional neural networks to learn similarity distance metric and improve interpretability." IEEE Access, pp. 123087-123097 (2020)~\cite{amirian2020radial}.
    \item Patrick Thiam, Viktor Kessler, \textbf{Mohammadreza Amirian}, Peter Bellmann, Georg Layher, Yan Zhang, Maria Velana, Sascha Gruss, Steffen Walter, Harald Christhelm Traue, Daniel Schork, Jonghwa Kim , Elisabeth Andr\'e, Heiko Neumann, and Friedhelm Schwenker. ``Multi-modal pain intensity recognition based on the senseemotion database." IEEE Transactions on Affective Computing, pp. 743-760 (2021)~\cite{thiam2021multi}.
    \item Taye Girma Debelee, Abrham Gebreselasie, Friedhelm Schwenker, \textbf{Mohammadreza Amirian}, Dereje Yohannes. ``Classification of mammograms using texture and cnn based extracted features." Journal of Biomimetics, Biomaterials and Biomedical Engineering, pp. 79-97 (2019)~\cite{debelee2019}.
    \item Markus K\"achele, \textbf{Mohammadreza Amirian}, Patrick Thiam, Philipp Werner, Steffen Walter, G\"unther Palm, and Friedhelm Schwenker. ``Adaptive confidence learning for the personalization of pain intensity estimation systems." Evolving Systems, pp. 71-83 (2017)~\cite{kachele2017adaptive}.
    \item Markus K\"achele, Patrick Thiam, \textbf{Mohammadreza Amirian}, Friedhelm Schwenker, Günther Palm. ``Methods for person-centered continuous pain intensity assessment from bio-physiological channels." IEEE Journal of Selected Topics in Signal Processing, pp. 854-864 (2016)~\cite{kachele2016methods}.
    \item Kamran Kazemi, \textbf{Mohammadreza Amirian}, Mohammad Javad Dehghani. ``The S-transform using a new window to improve frequency and time resolutions." Signal, Image and Video Processing, pp. 533-541 (2014)~\cite{kazemi2014s}.
\end{itemize}

\subsection{Conference Papers}
Here is the list of the peer-reviewed and presented research papers in scientific conferences contributing to this thesis:
\begin{itemize}
    \item \textbf{Mohammadreza Amirian}, Javier A. Montoya-Zegarra, Jonathan Gruss, Yves D. Stebler, Ahmet Selman Bozkir, Marco Calandri, Friedhelm Schwenker, and Thilo Stadelmann. ``PrepNet: A Convolutional Auto-Encoder to Homogenize CT Scans for Cross-Dataset Medical Image Analysis." In 2021 14th International Congress on Image and Signal Processing, BioMedical Engineering and Informatics (CISP-BMEI), pp. 1-7 (2021)~\cite{amirian2021prepnet}.
    \item \textbf{Mohammadreza Amirian}, Lukas Tuggener, Ricardo Chavarriaga, Yvan Putra Satyawan, Frank-Peter Schilling, Friedhelm Schwenker, and Thilo Stadelmann. ``Two to trust: Automl for safe modelling and interpretable deep learning for robustness." In International Workshop on the Foundations of Trustworthy AI Integrating Learning, Optimization and Reasoning, pp. 268-275 (2021)~\cite{amirian2021two}.
    \item Stefan Gl\"uge, \textbf{Mohammadreza Amirian}, Dandolo Flumini, and Thilo Stadelmann. ``How (not) to measure bias in face recognition networks." In Proceedings of the IAPR Workshop on Artificial Neural Networks in Pattern Recognition, pp. 125-137 (2020)~\cite{gluge2020not}.
    \item \textbf{Mohammadreza Amirian}, Katharina Rombach, Lukas Tuggener, Frank-Peter Schilling, Thilo Stadelmann. ``Efficient deep CNNs for cross-modal automated computer vision under time and space constraints." In Proceedings of the ECML-PKDD 2019, pp. 16-19 (2019)~\cite{amirian2019efficient}.
    \item Lukas Tuggener, \textbf{Mohammadreza Amirian}, Katharina Rombach, Stefan Lör- wald, Anastasia Varlet, Christian Westermann, and Thilo Stadelmann. ``Automated machine learning in practice: state of the art and recent results." In Proceedings of the 6th Swiss Conference on Data Science (SDS), pp. 31-36 (2019)~\cite{tuggener2019automated}.
    \item Thilo Stadelmann, \textbf{Mohammadreza Amirian}, Ismail Arabaci, Marek Arnold, Gilbert François Duivesteijn, Ismail Elezi, Melanie Geiger, Stefan Lörwald, Benjamin Bruno Meier, Katharina Rombach, and Lukas Tuggener. ``Deep learning in the wild." In IAPR Workshop on Artificial Neural Networks in Pattern Recognition, pp. 17-38 (2018)~\cite{stadelmann2018deep}.
    \item Benjamin Bruno Meier, Ismail Elezi, \textbf{Mohammadreza Amirian}, Oliver D\"urr, and Thilo Stadelmann. ``Learning neural models for end-to-end clustering." In Proceedings of the IAPR Workshop on Artificial Neural Networks in Pattern Recognition, pp. 126-138 (2018)~\cite{meier2018learning}.
    \item \textbf{Mohammadreza Amirian}, Friedhelm Schwenker, and Thilo Stadelmann. ``Trace and detect adversarial attacks on CNNs using feature response maps." In Proceedings of the IAPR Workshop on Artificial Neural Networks in Pattern Recognition, pp. 346-358 (2018)~\cite{amirian2018trace}.
    \item Viktor Kessler, Patrick Thiam, \textbf{Mohammadreza Amirian}, Friedhelm Schwenker. ``Pain recognition with camera photoplethysmography." In Proceedings of the  Seventh International Conference on Image Processing Theory, Tools and Applications (IPTA), pp. 1-5 (2017)~\cite{Kessler2017pain}.
    \item Viktor Kessler, Patrick Thiam, \textbf{Mohammadreza Amirian}, Friedhelm Schwenker. ``Multimodal fusion including camera photoplethysmography for pain recognition." In Proceedings of the International Conference on Companion Technology (ICCT), pp. 1-4. (2017)~\cite{Kessler2017Multimodal}.
    \item Taye Girma Debelee, \textbf{Mohammadreza Amirian}, Achim Ibenthal, Günther Palm, Friedhelm Schwenker. ``Classification of mammograms using convolutional neural network based feature extraction." International Conference on Information and Communication Technology for Development for Africa, pp. 89-98 (2017)~\cite{debelee2017classification}.
    \item \textbf{Mohammadreza Amirian}, Markus K\"achele, G\"unther Palm, and Friedhelm Schwenker. ``Support vector regression of sparse dictionary-based features for view-independent action unit intensity estimation." In Proceedings of the 12th IEEE International Conference on Automatic Face \& Gesture Recognition (FG 2017), pp. 854-859 (2017)~\cite{amirian2017support}.
    \item \textbf{Mohammadreza Amirian}, Markus Kächele, Patrick Thiam, Viktor Kessler, Friedhelm Schwenker. ``Continuous multimodal human affect estimation using echo state networks." In Proceedings of the 6th International Workshop on Audio/Visual Emotion Challenge. pp. 67–74 (2016)~\cite{amirian2016continuous}. 
    \item \textbf{Mohammadreza Amiria}, Markus Kächele, Friedhelm Schwenker. ``Using radial basis function neural networks for continuous and discrete pain estimation from bio-physiological signals." In Proceedings of the IAPR Workshop on Artificial Neural Networks in Pattern Recognition. pp. 269-284 (2016)~\cite{amirian2016using}.
    \item Markus K\"achele, Patrick Thiam, \textbf{Mohammadreza Amirian}, Philipp Werner, Steffen Walter, Friedhelm Schwenker, Günther Palm. ``Multimodal data fusion for person-independent, continuous estimation of pain intensity." In Proceedings of the International Conference on Engineering Applications of Neural Networks, pp. 275-285 (2015)~\cite{kachele2015multimodal}.
\end{itemize}

\subsection{Book Chapter}
Here is the contribution published as a book chapter in conjunction with this thesis:
\begin{itemize}
    \item Lukas Hollenstein, Lukas Lichtensteiger, Thilo Stadelmann, \textbf{Mohammadreza Amirian}, Lukas Budde, Jürg Meierhofer, Rudolf M Füchslin, Thomas Friedli. ``Unsupervised learning and simulation for complexity management in business operations." Applied Data Science. pp. 313-331 (2019)~\cite{hollenstein2019unsupervised}.
\end{itemize}
\newpage
\section {Organization of Thesis}
The remainder of this thesis is organized as follows:
\begin{itemize}
    \item Chapter~\ref{chap:theory} provides an overview of the necessary prerequisites and theoretical background for understanding this thesis, summarizes the related work, and relates the following chapters to the current literature. This chapter begins with preliminary content, such as the fundamentals of convolution operation and self-attention. The chapter continues with best practices in architecture search and hyperparameter tuning methods. 
    \item Chapter~\ref{chap:rbfs} introduces the main theoretical contribution of this thesis, namely the use of RBF networks as classifiers of CNNs for interpretable decisions. This chapter also proposes changing the training process and introduces a novel quadratic activation function to adapt RBFs for optimization with conventional CNNs.
    \item Chapter~\ref{chap:trace} demonstrates how understanding neural networks using feature response map visualizations can improve their robustness by detecting adversarial attacks. In addition, this chapter explains guided backpropagation, a well-known technique for inverting CNN architectures and visualizing the regions of input images which are relevant to the model's classification, and shows the application of feature responses in detecting adversarial attacks.  
    \item Chapter~\ref{chap:motion} presents the main practical contribution of this thesis, in which neural networks reduce motion artifacts from CBCT scans for various reconstruction techniques. This chapter describes the first attempt to reduce motion artifacts in CBCT scans. It explains the architecture and underlying idea of how supervised learning with simulated data can address a solution for a real-world problem in which there are no ground-truth labels. This chapter includes a clinical evaluation of this method using real-world data and shows how improvement in numerical measures translates to the preferences of clinical experts. 
    \item Chapter~\ref{chap:applications} provides an overview of several applications in conjunction with this thesis. This chapter aims to draw the readers' attention to several practical problems with ongoing research, present related solutions, and suggest promising areas for future research in these applications.
    \item Chapter~\ref{chap:conclusion} concludes the thesis and discusses a roadmap for future research opportunities in the niches to which this research work contributes.
\end{itemize}

\chapter{Theoretical Foundations}
\label{chap:theory}

This chapter summarizes the general prerequisites and theoretical background necessary to understand the remainder of the thesis. Specific niche techniques used in each part of the scientific and applied contributions are explained in each chapter individually. Therefore, reading this chapter is recommended only for those interested in refreshing their fundamental knowledge of computer vision techniques. Furthermore, the following chapters contain a more detailed theoretical overview of the related concepts, a knowledgeable reader can read them independently of this chapter.

This thesis's main fundamental and methodological contributions are related to computer vision techniques for object recognition. First, this chapter explains the basics of convolutional neural networks (CNNs) in Section~\ref{chap:theory_sec:cnns}, which have revolutionized computer vision research by outperforming the classical methods. Second, this chapter briefly reviews the history of CNNs, including their major architectures and exciting recent developments. CNN-based models are the most recurring theoretical theme in this thesis, and understanding these models is important for being able to follow Chapter~\ref{chap:rbfs}, Chapter~\ref{chap:trace}, and parts of Chapter~\ref{chap:applications}. The brief introduction to 3D-CNNs at the end of Section~\ref{chap:theory_sec:cnns} is also necessary for understanding Chapter~\ref{chap:motion}.

Moreover, this chapter briefly summarizes computer vision techniques using vision transformers (ViTs) in Section~\ref{chap:theory_sec:vits}. Researchers investigating machine and deep learning (ML and DL) methods have long sought efficient methods for modeling attention-inspired mechanisms to focus on the most relevant information in time series, images, and to fuse information from several data modalities. Transformers and self-attention provide an excellent solution for attention in natural language processing (NLP). Transformers have recently been applied to computer vision problems by adapting self-attention for object recognition and segmentation. ViTs belong to more recent research compared to CNNs. Although their underlying theory only supports Section~\ref{chap:applications_sec:vits} in this thesis. ViTs are more likely to gain more attention in future computer vision research because of their ability to train on very large datasets compared to CNNs. 

Deep learning has opened a great opportunity to distill information from massive datasets and optimize millions of parameters. However, these methods depend on optimization techniques that converge rapidly to an optimum which generalizes well. Therefore, understanding optimization techniques is necessary to bring the computer vision models to their optimal performance. The computer vision models presented in this thesis can overcome challenging problems that occur when using enormous datasets. These models are prone to overfitting, but they can be optimized to make correct predictions for the training data. However, the models cannot generalize to the unseen data, as is expected and observed in human vision. The last two sections of this chapter discuss the optimization methods, how to avoid overfitting, and how to improve the performance of computer vision architectures in generalization tasks with unseen data. The optimization and generalization of vision models are not the direct subjects of any chapter in this thesis; they are running themes throughout all chapters, especially in Chapter~\ref{chap:rbfs}, Chapter~\ref{chap:motion} and parts of Chapter~\ref{chap:applications}.

\section{Convolutional Neural Networks}
\label{chap:theory_sec:cnns}
This section reviews the basics as well as recent advances in developing CNNs for computer vision. It begins with explanations of the building blocks used in CNNs. The computer vision community initially focused on manually improving these models' internal building blocks and introduced novel and suitable building blocks. The focus changed to automated model developments and architecture search when compute resources became widely available. After explaining the basics, this section presents some of the architectural breakthroughs that have improved the accuracy of computer vision models.

Automated neural architecture search has replaced manual architecture development attempts in the next generation of image processing models. Therefore, this section also describes some of the efforts in the automated search for optimal computer vision neural architectures. Finally, this section concludes with an explanation of the basics of 3D-CNNs and UNet architectures, which are necessary for understanding the content in the final chapters of this thesis.

CNN backbones have replaced hand-crafted feature extraction techniques such as scale-invariant features (SIFT)~\cite{lowe1999object} because they can automatically learn representations of images during the optimization process. Based on this analogy, a model can be divided into two parts: 1) an encoder that converts the visual information (e.g., images) into a set of latent (intermediate) representations (model embeddings), and 2) a classifier that identifies the existing objects or segment patterns in the images. The main advantage of CNNs over manual feature extraction is the ability to optimize and fine-tune millions of parameters for encoding visual information into discriminative representations using large datasets. Much computer vision research focuses on optimizing models' architecture to compute more generic representations (embeddings) of images. The ultimate goal of this area of research is not only to develop models that can learn image representations but also to train models that generalize well to unseen images from the same data distribution as the training data. A computer vision model's ultimate goal is learning representations that generalize to new categories of images outside the data sets used for optimization. Although the encoder part of neural networks has been the subject of much recent research, feed-forward neural networks have often been chosen for classifiers in the literature because of their efficiency in optimization.

\subsection{Convolution Operator}
The convolution operator of two functions shows how two input functions change their shape when shifted against each other for all possible shift values. For two one-dimensional real-valued time series ($x$ and $w$), their convolution ($s$) can be defined as follows:

\begin{flalign}\label{chap:theory_eq:conv_1d}
s(t) = \int x(a)w(t-a)da = (x*w)(t)
\end{flalign}
where $t$ is the time, and $*$ denotes the convolution operator. For a given time shift ($t$), the convolution of two time series is equal to the dot product of one multiplied by the mirrored and shifted version of the other. The convolution operator is commutative due to the time inversion in the definition of the function ($f*g=g*f$). Similarly, the convolution operator can be defined for two 1D discrete functions ($X$ and $W$) with time stamps $i$ and $j$ within the validity range determined by $m$ as follows: 
\begin{flalign}\label{chap:theory_eq:conv_1dd}
S(i) = \sum_{j=1}^m X(j)W(i-j) = (X*W)(i)
\end{flalign}
Based on this interpretation of the 1D convolution operator, we can define the 2D convolution ($S$) for the two-dimensional image as follows:
\begin{flalign}\label{chap:theory_eq:conv_2d}
S(i, j) = (I*W)(i, j) = \sum_m \sum_n I(m, n)W(i-m, j-n)
\end{flalign}
where $I$ and $W$ represent two images, $i$ and $j$ define the spatial coordinates of these images. The valid range for images is indicated by $m$ and $n$, respectively. The convolution function shows how a given kernel ($W$) changes an input image ($I$) after the kernel is applied. The commutativity properties also apply to two-dimensional convolutions because of the mirroring the images. Since commutativity is not an essential property of neural networks, most libraries use \textit{cross-correlation} instead of convolutions for implementation:
\begin{flalign}\label{chap:theory_eq:cross_cor}
\hat{S}(i, j) = (I*W)(i, j) = \sum_m \sum_n I(m, n)W(i+m, j+n)
\end{flalign}
The \textit{cross-correlation} function computes the dot product of an image patch and the kernel ($W$) by shifting the kernel vertically and horizontally over the input image in the range of the images' definition. The step size at which the kernel shifts after each convolutional step is called stride and is a parameter of a convolutional layer in neural networks.  

\begin{figure}[htb!]
    \centering
    \includegraphics[width=0.75\textwidth]{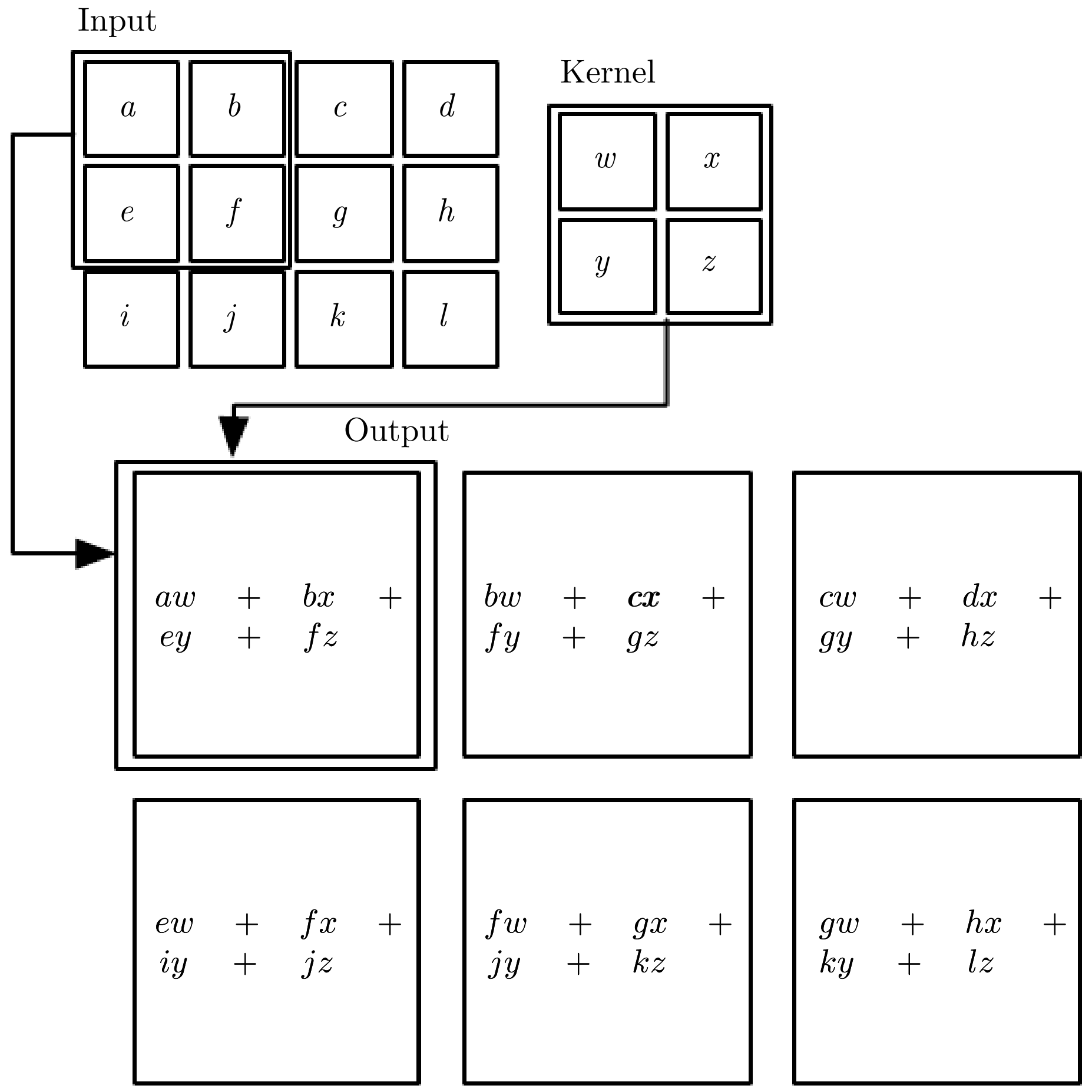}
    \hspace{-0.35cm}
    \caption{The cross-correlation function is often implemented in deep learning libraries for convolutional neural networks. For an input image, the output (kernel response) is the dot product of the vectorized kernel with a field which is the same size as the input image. The kernel slides over the entire image area with a given step size (figure adopted from~\cite{lecun2015deep}).}
 	\label{chap:theory_fig:cross_corelation}
\end{figure}

\subsection{Feature Maps}
Applying a kernel with the cross-correlation function in equation~\ref{chap:theory_eq:cross_cor} to an image leads to computing a so-called feature map. Depending on the type of kernels, the feature maps contain different information (see figure~\ref{chap:theory_fig:cross_corelation}). Feature maps are the first representations of the images computed in the CNNs, and visually inspecting them, along with the first layer inputs, is crucial for understanding the behavior of the entire network. The filters have the same depth as the input images (three for RGB and one for grayscale), and it is possible to visualize them along with the feature maps without complications.

\subsection{Pooling Layers}
\begin{figure}[htb!]
    \centering
    \includegraphics[width=0.3\textwidth]{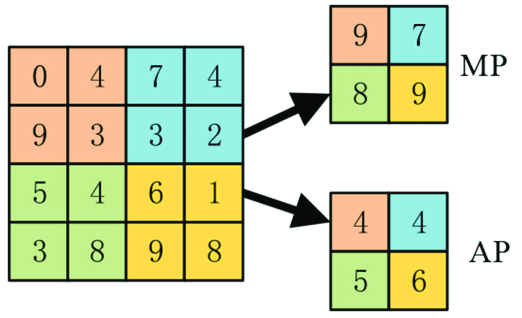}
    \hspace{-0.35cm}
    \caption{Max-pooling (MP) and average-pooling (AP) layers with kernel size and stride of $2$ for CNNs (figure adopted from~\cite{wang2018multiple}).}
 	\label{chap:theory_fig:pooling_layer}
\end{figure}

The pooling layers aim to summarize the previous layer's output by merging the information in a given neighborhood. Two standard techniques for pooling local information are using the maximum or average value around the center of the kernel. The output of a pooling operator, as shown in Figure \ref{chap:theory_fig:pooling_layer}, is independent of the order of values in that specific region. The pooling layer and the cross-correlation function are the key components of the CNN for translation invariance as \textit{inductive bias}\footnote{Inductive biases are a set of assumptions encoded in a learning algorithm to counter hypothetical input data and cases.} in CNNs.

\subsection{FeedForward Neural Networks}
Deep feedforward neural networks, also called multilayer perceptrons (MLPs), are often used in deep architectures to approximate functions. The goal of these networks is to approximate a function ($f$) that maps a set of input features ($\boldsymbol{x}$) to ground truth labels ($y\approx f(\boldsymbol{x})$). Feedforward networks, as shown on the right side of the figure \ref{chap:theory_fig:convnet}, consist of an input, an output, and several hidden layers. Each hidden layer of the feedforward network is a \textit{fully connected} layer that contains an intermediate representation of features by computing the weighted combination of all features in the previous layer. Each fully connected layer in the architecture of an MLP can have a trainable bias term, denoted by $x_0$ in Figure \ref{chap:theory_fig:convnet} and trainable weights. Feedforward networks are optimized using \textit{backpropagation} as described in the next sections.   

\subsection{Convolutional Neural Networks}
\begin{figure}[htb!]
    \centering
    \resizebox{\textwidth}{!}{
    \includegraphics[width=0.7\textwidth, valign=t]{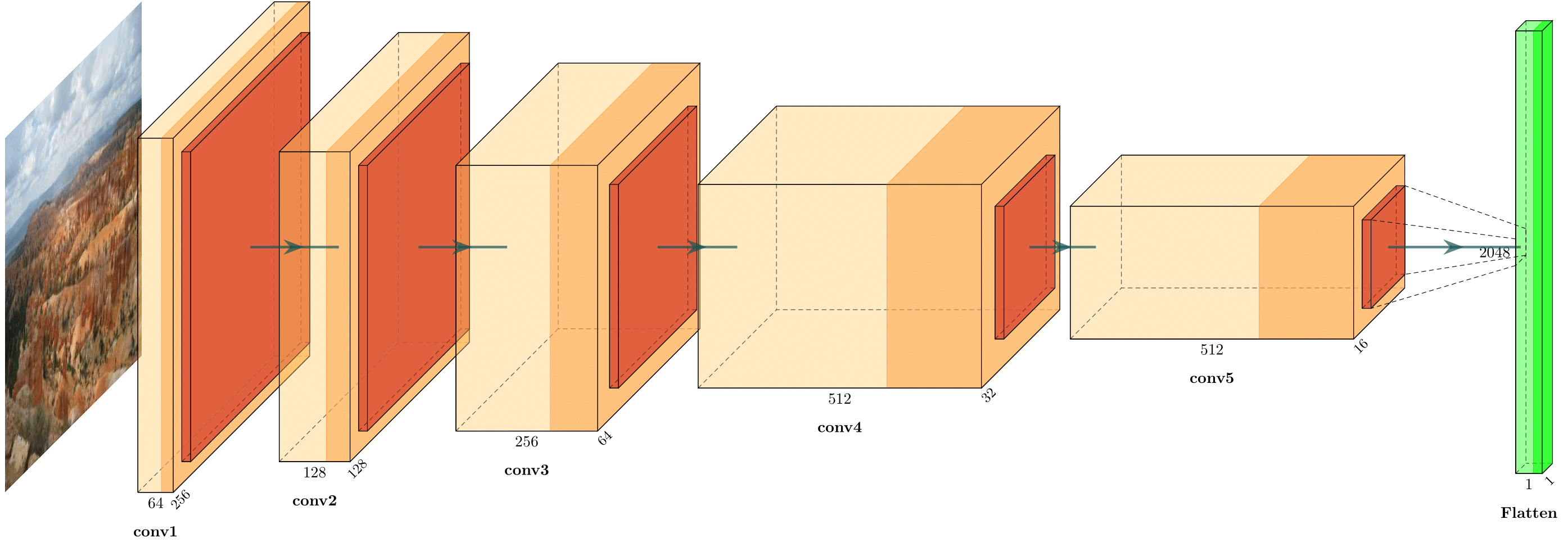}
    \raisebox{-0.05cm}{\includegraphics[width=0.4\textwidth, valign=t]{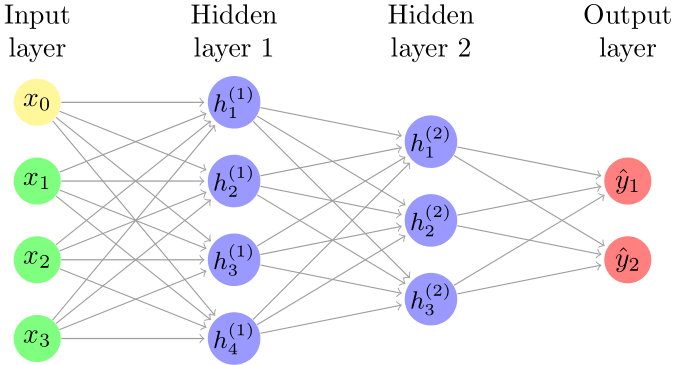}}}
    \caption{A convolutional neural network for representation learning from an input image, followed by a feedforward network for object classification.}
 	\label{chap:theory_fig:convnet}
\end{figure}

The simplest form of convolutional nets consists of the two basic layers (convolution and pooling) explained in the previous sections. Basic networks can be constructed using successive convolutional layers to compute input representations and a pooling layer to summarize the information. However, modern convolutional architectures for vision use much more than these two layers. Figure \ref{chap:theory_fig:convnet} illustrates a simple convolutional network. The original image's pixel values reveal only the mapped object's information for the given pixel size. \textit{Feature maps}, which show the representations of the first convolutional layer, combine the local information for a given filter size (usually $3\times3$). The pooling layer combines more local information from multiple filter activations (typically $2\times2$) and increases the size of the input region that contributes to a single activation value. The region's size in the original input image contributing to a single value at each network's layer determines the so-called \textit{receptive field} at a given layer. 

Finally, multiple convolutional and pooling layers are connected to a feedforward network for classification. A convolutional net, also called convolutional \textit{backbone}, aims to compute discriminative (latent) representations of the images for each image class. These representations are finally transformed into the form of a feature vector in the last layer by flattening or global pooling. Flattening rearranges all the activations of a convolutional layer into a single vector, whereas global pooling applies the maximum and average functions to the spatial dimensions of the representations. The one-dimensional vector computed for each image is often referred to in the literature as \textit{embeddings}. The \textit{embeddings} of a convolutional neural network are passed to a feedforward network for object classification (Figure \ref{chap:theory_fig:convnet}). 

\subsection{Advanced Blocks}
The recent history of convolutional neural networks has had many exciting breakthroughs. However, one of the first modern convolutional neural network prototypes (LeNet-5) was only able to classify handwritten digits~\cite{lecun1998gradient}. Training AlexNet~\cite{krizhevsky2012imagenet}, a relatively small model compared to currently available networks, was only made possible by splitting the model between two graphics processing units (GPUs). Even two years after introducing AlexNet, it was impossible to train very deep VGG models end-to-end without pretraining the model layer by layer~\cite{simonyan2014very}. Given the limitations of resources and algorithms prior to the feasibility of automatic neural architecture search, computer vision researchers mainly tried to incorporate \textit{inductive biases} to develop better convolutional architectures. These techniques are inspired by image processing tasks and failure cases in the classification task or improvement of the optimization process and gradient flow. The following sections review three interesting architectural advances in computer vision.

\subsubsection{Residual Connections}
\begin{figure}[htb!]
    \centering
    \includegraphics[width=0.3\textwidth]{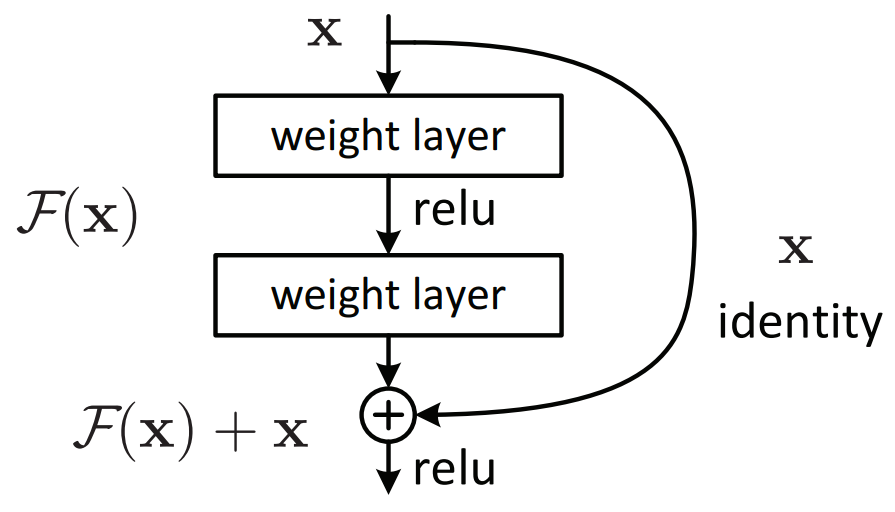}
    \hspace{-0.35cm}
    \caption{The residual connection between a layer's input and output improves the gradient flow (figure is adopted from~\cite{he2016deep}).}
 	\label{chap:theory_fig:resblock}
\end{figure}

Residual connections in CNNs establish a bridge between the input and output of a layer~\cite{he2016deep}. Although using a stack of multiple convolutional layers and forming deep models showed better generalization properties than shallow networks, the researchers have traditionally proposed residual connections to improve gradient flow on the backward path. Researchers introduced the idea of using residual connections at the same time that gradient vanishing was the focus of research in computer science for long short-term memory (LSTM) models ~\cite{hochreiter1997long}. The improvement of the gradient flow also led to the breakthrough of highway networks at the same time~\cite{srivastava2015highway}. However, in the following years, residual networks were more commonly used in the research community. Figure~\ref{chap:theory_fig:resblock} from the original paper shows one of the most straightforward and practical ideas in the history of deep learning. 

\subsubsection{Inception blocks}
\begin{figure}[htb!]
    \centering
     \subfloat[Naiive inception block]{\includegraphics[width=0.45\textwidth, keepaspectratio]{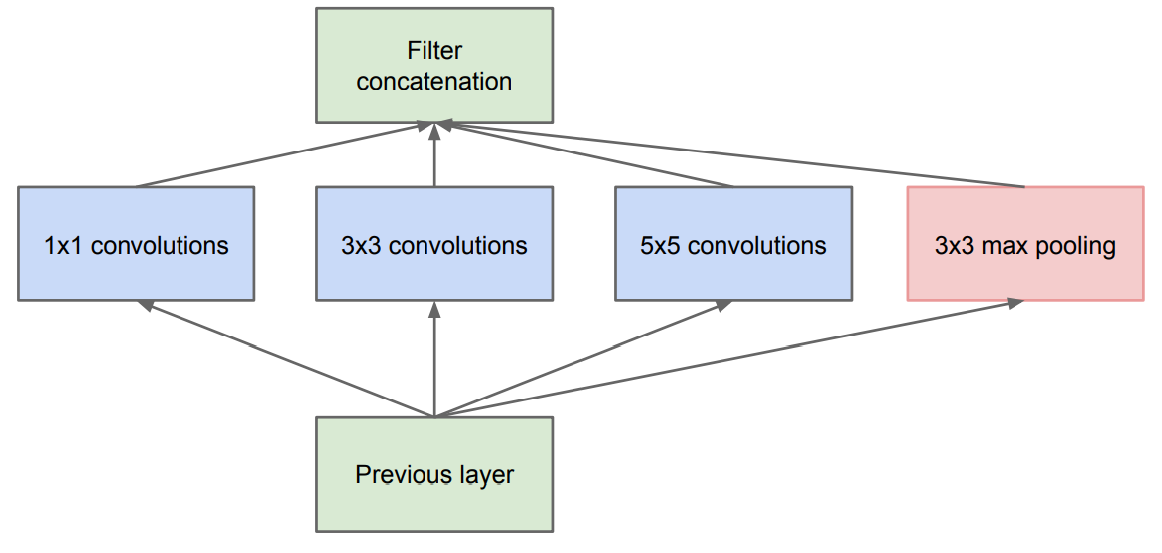}\label{chap:theory_fig:inception_naiive}}
     \subfloat[Practical inception block]{\includegraphics[width=0.45\textwidth, keepaspectratio]{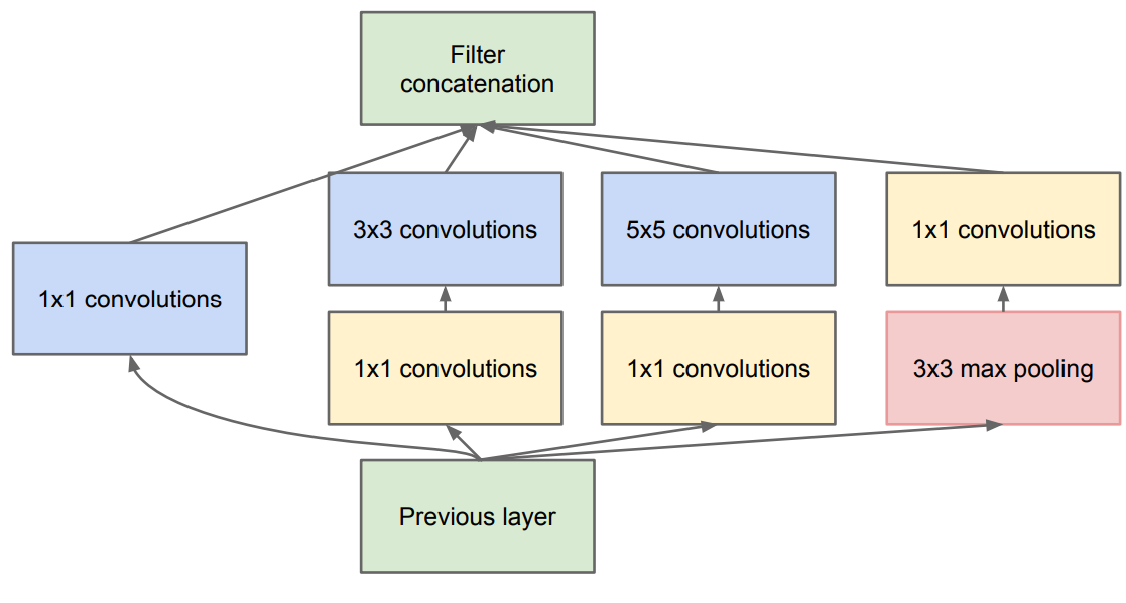}\label{chap:theory_fig:inception_practical}} \\
    \caption{This figure, adopted from~\cite{szegedy2015going}, depicts the idea of the inception blocks and their practical implementation.}
    \label{chap:theory_fig:inception}
\end{figure}

The original idea of inception blocks is to summarize the sparse latent representations of image patches into a dense form and cluster the relevant samples using convolutional filters with different patch sizes. Inspired by Arora et al.~\cite{arora2014provable}, the naive inception block finds the correlations between image patches or representations and clusters them into groups and units of highly correlated samples. Szegedy et al.~\cite{szegedy2015going} suggested using a layer of $1\times1$ convolutions to cover a small region with many clusters, which is practical for regions where clusters are densely distributed. Furthermore, larger convolutions of size $3\times3$ and $5\times5$ are used for the more spatially spread clusters. Inception blocks also include a pooling operator to maintain the translation invariance property (see Figure~\ref{chap:theory_fig:inception_naiive}). 

The concept of a naive inception block is immensely appealing; however, it suffers from practical feasibility since the computational cost of such blocks blows up within the first few layers. Thus, to reduce the computational complexity of naive inception blocks, they are implemented with $1\times1$ filters to downsample the input representations in practice, while the outputs of the layers are computed by concatenating the representations of the input computed using all four sets of filterbanks depicted in Figure~\ref{chap:theory_fig:inception_practical}. As a result, the inception models improved state-of-the-art performance in image recognition tasks after their advent.

\subsubsection{Convolutional Block Attention Module}
\begin{figure}[htb!]
     \centering
     \subfloat[Convolutional block attention module]{\includegraphics[width=0.85\textwidth, keepaspectratio]{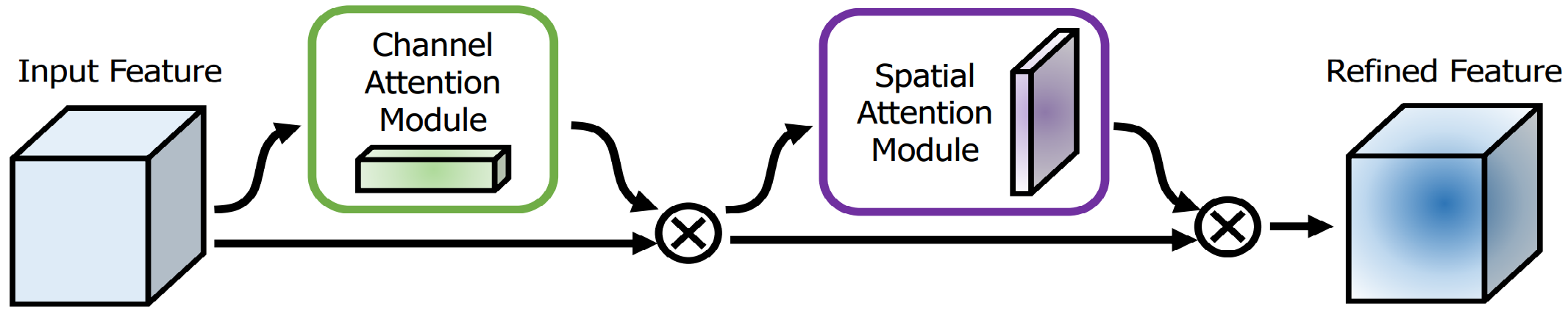}\label{chap:theory_fig:cbam_all}} \\
     \subfloat[Channel attention module]{\includegraphics[width=0.85\textwidth, keepaspectratio]{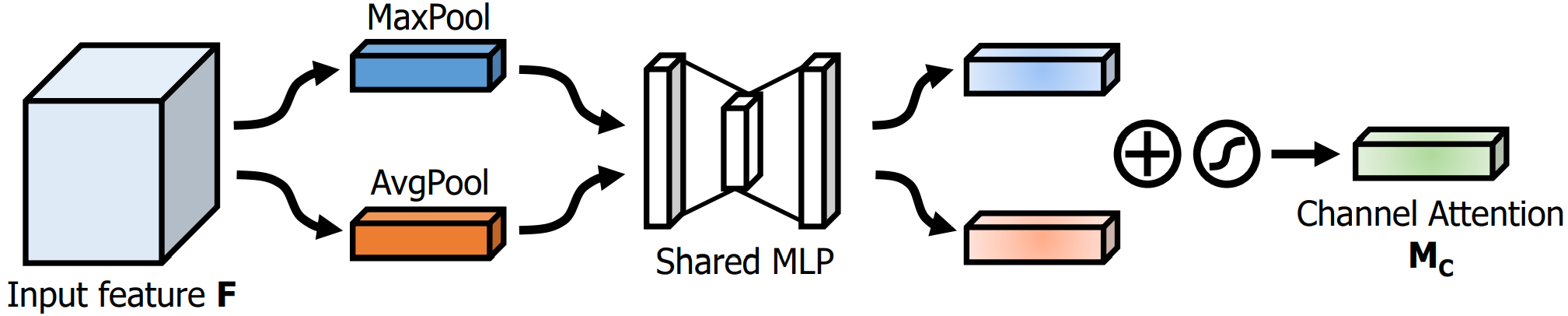}\label{chap:theory_fig:cbam_c}} \\
     \subfloat[Spatial attention module]{\includegraphics[width=0.6\textwidth, keepaspectratio]{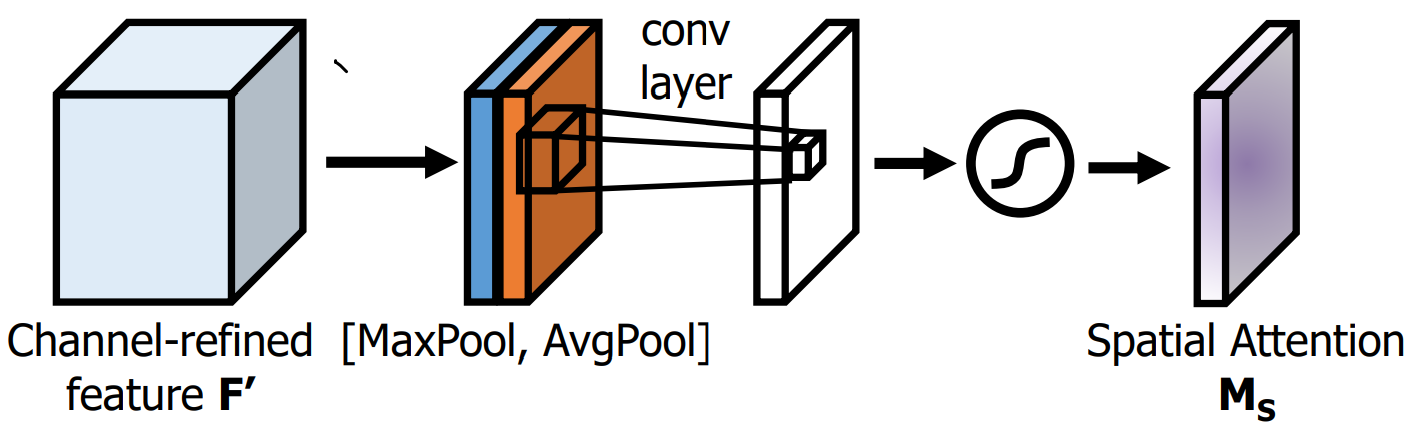}\label{chap:theory_fig:cbam_s}}
    \caption{Convolutional block attention module (CBAM) with its two main components for refining channel and spatial features (figures are adopted from~\cite{szegedy2015going}).}
    \label{chap:theory_fig:cbam}
\end{figure}

The main goal of the attention module in convolutional layers is to provide the ability to focus on a specific channel as well as spatial information~\cite{woo2018cbam}. Therefore, this module uses a channel attention module similar to squeeze and excitation techniques~\cite{hu2018squeeze} in addition to a very similar spatial attention module. The high-level idea, shown in Figure~\ref{chap:theory_fig:cbam_all}, is to compute a channel and a spatial attention map for the input of a given layer and multiply the activation values by these maps to focus on specific channel-spatial information. The whole convolutional block attention module (CBAM) can profit from residual connections to improve gradient flow and allow the model to skip the attention modules. 

The computation of the attention maps is relatively straightforward, as shown in Figure~\ref{chap:theory_fig:cbam}. To compute the channel attention maps (see Figure~\ref{chap:theory_fig:cbam_c}), we first employ a mean and a max pooling over the input feature maps to obtain a vector of the average and maximum of the activation values for each channel. Then, these two vectors are passed through a trainable MLP with shared weights for both pooling outputs. Finally, the activations of the MLPs are averaged and passed through a sigmoid activation to form the final channel attention maps. A similar system is used for spatial attention mechanisms by computing the average and max-pooling over the spatial information instead of the channels and by replacing the MLP with a convolutional layer (see Figure~\ref{chap:theory_fig:cbam_s}).

\newpage
\subsection{Architecture Search}
\label{chap:theory_sec:ArchtictureSearch}
With the rapid increase in computer resources, computer vision researchers began to find ways to expand their search space, from optimizing hyperparameters to exploring new model architectures. Scientists who intuitively invented novel blocks for imaging models began to use their intuition to find the best search space for computer vision architectures and to optimize search techniques. The remainder of this section presents two breakthroughs in neural architecture search. 

\subsubsection{NASNets}
Zoph et al.~\cite{zoph2018learning} introduced NASNets in the first famous attempt to search for the optimal architecture for image recognition. They performed the architecture search on a dataset with images of size $32\times32$ pixels from $10$ object classes (CIFAR10~\cite{krizhevsky2009learning}). However, the heuristics and inductive biases allowed successful scaling of the sought after architectures to a large dataset with $299\times299$ images of $1000$ classes (ImageNet~\cite{deng2009imagenet}). Zoph et al. designed a controller using recurrent neural networks (RNN) to find the optimal architecture of two motifs named \textit{normal cell} and \textit{reduction cell}. The convolutional architecture is constructed using a stack of such searched architectures. They considered the number of initial convolutions and motif repetitions as free parameters to solve the problem of scaling from a small dataset (CIFAR10) to a larger dataset (ImageNet). Although this research improved state-of-the-art image recognition performance by $1.2\%$ and reduced the number of best model parameters by $28\%$, it was only the beginning of more exciting research in this area. 

\subsubsection{EfficientNets}
Tan and Le made the next breakthrough in the search for a neural architecture with a model that achieved the same performance as state of the art in image recognition, importantly it was $8.4\times$ smaller and $6.1\times$ faster~\cite{tan2019efficientnet} than competitor models. Their research focused on two directions: 1) improving architectural search and 2) introducing a compound scaling technique. As in their previous research on developing mobile neural architectures for searched networks (MnasNet~\cite{tan2019mnasnet}), Tan and Le used a reinforcement learning (RL) based method to optimize their objective function. Their objective function is to find a Pareto-optimal mobile network called EfficientNet. It includes two components: 1) maximizing the accuracy of the network, similar to MnasNets (mobile NasNets), and 2) minimizing the number of FLOPs (required floating point computations) instead of latency, which is considered in MnasNets. Inspired by the architecture of mobile networks (MobileNet and MobileNetV2~\cite{andrew2017efficient, sandler2018mobilenetv2}), the authors used the mobile inverted bottleneck (MBCon~\cite{sandler2018mobilenetv2}) as the building block of EfficinetNet. This work's second breakthrough was introducing a compound approach to scaling neural architectures, while maintaining a balance between their height, width, and depth. Their scaling method demonstrates improvements in scaling EfficientNets, MobileNets, and ResNets.
\subsection{3D Convolutional Neural Networks}
\begin{figure}[htb!]
     \centering
     \subfloat[2D-convolution]{\includegraphics[width=0.45\textwidth, keepaspectratio]{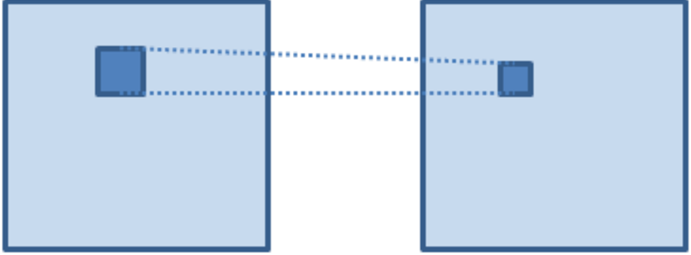}\label{chap:theory_fig:conv3d_2d0}}
     ~~~~~~
     \subfloat[3D-convolution]{\includegraphics[width=0.45\textwidth, keepaspectratio]{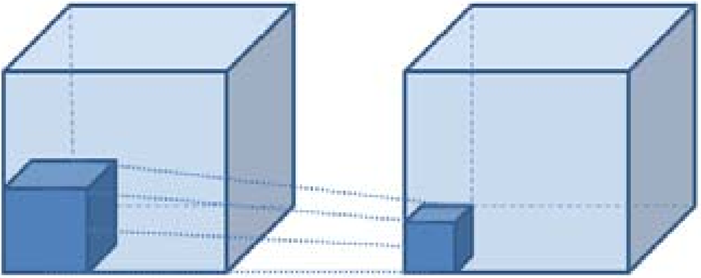}\label{chap:theory_fig:conv3d_2d1}}
     \caption{Two- and three-dimensional convolutions. 2D convolutions target images, while 3D convolutions are suitable for volumetric data (figures are adopted from~\cite{7950542}).}
    \label{chap:theory_fig:conv3d}
\end{figure}

Researchers have extended the idea of two-dimensional convolutions to three-dimensional spaces where data samples span multiple images (slices) per input instance, such as in videos and volumetric medical images~\cite{milletari2016v}. Although the goal of video processing and 3D medical image processing is different, both can utilize 3D convolutional neural networks with the same architectures as in Figure~\ref{chap:theory_fig:conv3d}. 3D convolutional neural networks (3D-CNNs) aim to find temporal dependencies in video processing~\cite{huang2020efficient} and 3D spatial dependencies in medical imaging and point clouds~\cite{zhang2018efficient}. As an extension of 2D filters, 3D filters have one dimension higher - a size of $3\times3\times3$ voxels\footnote{Voxel is a single value in a data volume analogous to pixels in images.} is a common choice - to find spatial or temporal information in (3D) volumetric data. The feature responses of 3D filters are computed similarly by finding the correlation between the filter and a particular spatial position of the data volume. Applying a single 3D filter to a data volume results in a 3D feature map calculation. Stacking the feature maps of multiple 3D filters results in a four-dimensional feature map at each layer of a 3D model. The pooling operation is extended to find a volume's average or maximum value with the typical size of $3\times3\times3$. Similar to techniques used with 2D-CNNs, such as flattening and global spatial pooling, the feature maps of the last layer can be converted into embeddings for classification. 
The high dimensionality of the feature maps of 3D-CNNs makes their implementation very memory intensive, and processing the 3D inputs increases the computational complexity of the 3D-CNNs. However, researchers have recently explored 3D-CNNs for medical applications as the memory limits of modern GPUs have increased significantly. It seems that 3D CNNs will receive more attention in the future as computational resources continue to develop.
\vspace{-0.5cm}

\section{Vision Transformers}
Vision transformers (ViTs) for computer vision have emerged through the adaption of the self-attention mechanism developed in the field of natural language processing (NLP)~\cite{vaswani2017attention}. Researchers have searched for efficient attention mechanisms to optimally focus on the most relevant information to recognize patterns from different information sources. However, researchers in the field of NLP have only recently discovered a practical and efficient implementation of attention. The use of attention in NLP was so successful that the models developed in several NLP applications quickly outperformed the state-of-the-art~\cite{devlin-etal-2019-bert, liu2019roberta}. The core of these recent breakthroughs in NLP promptly found its way to image processing applications. The remainder of this section aims to review ViTs and explain the main components of these models used in computer vision. This section lays the theoretical foundation for the ViTs used in the following chapters of this thesis.

\subsection{Preliminaries}
The theoretical background of ViTs and attention mechanisms is grounded in machine translation. A brief explanation of the basic concepts is necessary to understand the rest of this section. Let us use a simple example from daily life to explain these concepts. Imagine that we make a text \textit{query} to find a relevant research paper in a search engine. The search engine evaluates the query based on several \textit{keys} that summarize the titles of the available papers and return the most relevant papers (\textit{values})\footnote{The terms \textit{key}, \textit{query}, and \textit{values} are used frequently in this section with very similar meanings to those used in the information retrieval literature.}. Upon receiving a query, the search engine may use a \textit{tokenizer} to segment the query sentence or break it into multiple \textit{tokens} (words or punctuations). The model maps the tokens to their token IDs based on a particular tokenizer and pads it with zeros up to a certain length to form the \textit{embedding} vectors of queries, keys, and values. 

Dosovitskiy et al. introduced information conversion into tokens in image processing for the first time~\cite{dosovitskiy2020image}. Based on this definition, commonly used in recent studies, they divided the input images into smaller patches of size $16\times16$ with three color channels. A random projection of the vectorized shape of these image patches is computed, and then the tokens that form the input of the vision transformer are generated. After tokenizing the information from any source, including images or text, the model focuses on the most relevant information in a query and relates them with a key to find the most appropriate values.

\subsection{Attention}
The concept of attention, as first described in~\cite{bahdanau2014neural}, is nothing more complicated than a weighted average of values ($h$) defined as follows:

\begin{flalign}\label{chap:theory_eq:attn}
\boldsymbol{c} = \sum_j \boldsymbol{\alpha}_j \boldsymbol{h}_j
\end{flalign}
where $\sum_j \boldsymbol{\alpha}_j=1$. $\boldsymbol{\alpha}_j=1$ corresponds to the importance of each element in the vector $\boldsymbol{h}$. 

Attention has been the key component in training outstanding models in NLP, such as BERT and RoBERTa~\cite{devlin-etal-2019-bert, liu2019roberta}, through the use of keys, queries, and values from different sources in a supervised scenario. In addition, the attention mechanism is also used in self-supervised training of language models to predict missing information in training models such as GPT~\cite{brown2020language}. Attention can estimate dependencies between two sequences and can be extended to self-attention (SA) for modeling dependencies within a text sequence. Self-attention techniques are commonly used in image processing to find local correlations between tokens computed from the same image.

\subsection{Self-Attention}
The following steps describe how to compute the self-attention (SA) layer's output for an image ($\boldsymbol(X) \in \rm I\!R ^{N\times T}$) converted into $N$ projected patches (tokens): 1) Calculate the projections of all tokens based on three different matrices to compute the keys, queries, and values based on all tokenized image patches. 2) Compute the attention matrix by multiplying keys and queries and normalizing the results using the softmax operation, which is defined for a vector $x$ as follows: $softmax(\boldsymbol{x}) = \frac{exp(\boldsymbol{x}_i)}{\sum_{j}^{ }exp(\boldsymbol{x}_j)}$. 3) Multiply the attention matrix by the values to calculate the SA matrix. In practice, ViTs consist of a stack of several such SA layers, which provide the opportunity to compute the dependencies of each pixel to every other one and combine the correlations based on their importance using the attention matrix.

\subsection{Postional Encoding}
SA layers, as described, are an effective tool for finding correlations between individual pixels. However, after converting an image to patches and computing the tokens, the SA layer is invariant to the order of the input tokens. In other words, the SA layer is independent of the order of the patches in the input images and ignores the order of the tokens in the input data. To address this deficiency of the transformers, researchers added an (absolute) positional encoding that considers the order of the input tokens in NLP models and the image patches in the ViTs. The vector of positional encoding contains additive information proportional to the absolute position of words in a phrase or patches in an image.    

\subsection{Relative Postional Encoding}
Absolute positional encoding in transformers retains the spatial information of a single patch, but fails to account for the relative distances between various patches. Shaw et al. used relative positional coding to address this shortcoming of self-attention in ViTs~\cite{shaw2018self}. First, relative positional coding computes a distance function between image patches. Then, it applies a function based on these distances to the attention matrix, instead of absolute positional encoding, which adds the positional encoding to the input tokens. 

\subsection{Vision Transformers for Classification}
\label{chap:theory_sec:vits}
\begin{figure}[htb!]
    \centering
    \includegraphics[width=\textwidth]{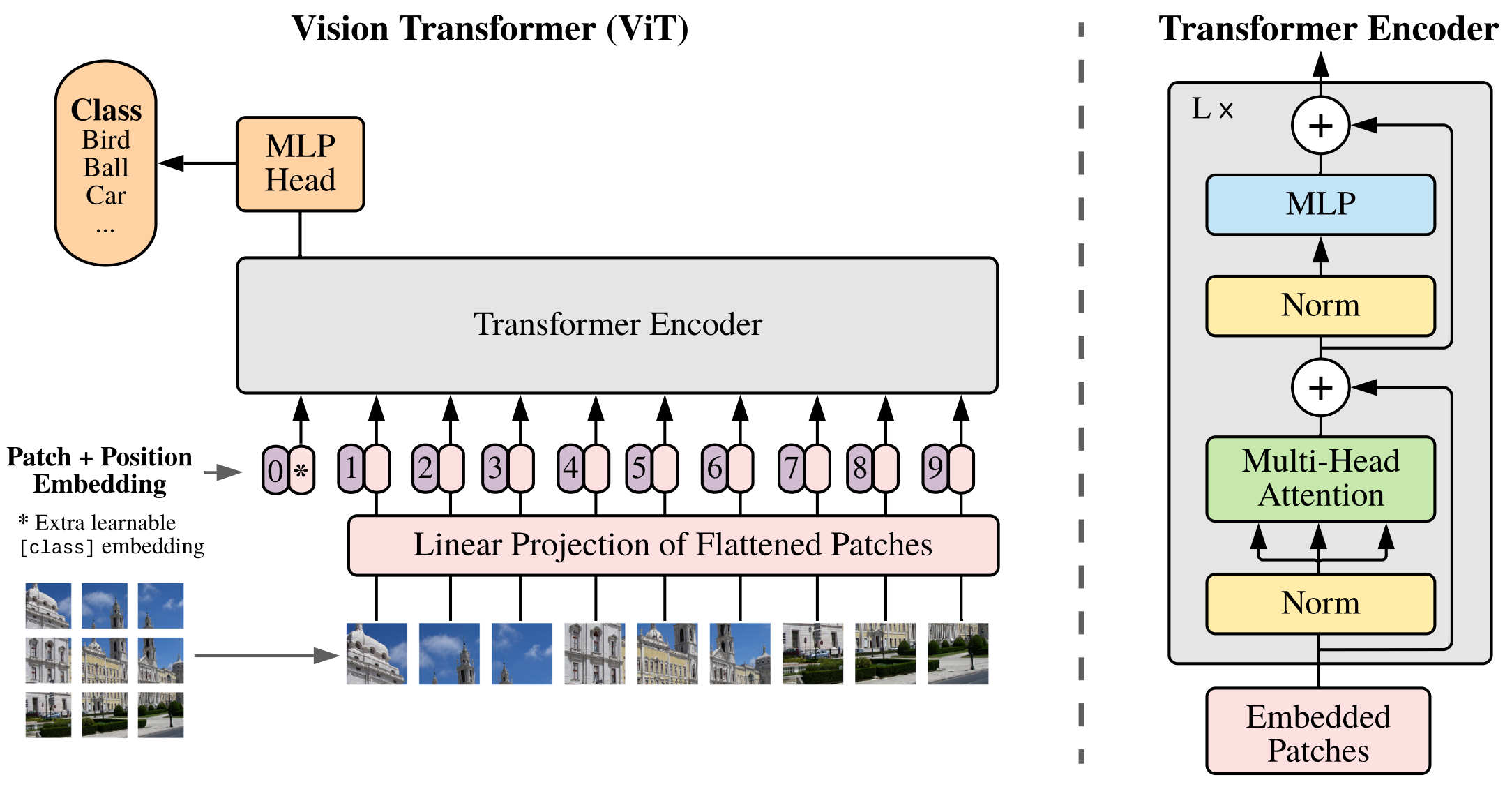}
    \caption{Vision transformers (ViTs) for image classification (figure is adopted from~\cite{dosovitskiy2020image}).}
    \label{chap:theory_fig:vit}
\end{figure}

Dosovitskiy et al. trained the first vision of ViTs on ImageNet~\cite{dosovitskiy2020image}, three years after Vaswani et al. introduced the attention mechanism ~\cite{vaswani2017attention}. 
The architecture of their vision transformer, as shown in Figure \ref{chap:theory_fig:vit}, consists of a transformer encoder (backbone) with a multilayer perceptron (MLP) head for classification. 
The input images used for image recognition are divided into patches of size $16\times16$.
Then, each patch is converted (flattened) into a vector, and a linear projection is applied to compute the input tokens for the transformer architecture.
The backbone of the transformer contains multiple layers of multi-head attention\footnote{
    Multi-head attention is an extension of the attention mechanism that computes multiple attention matrices with different weights from keys and queries and combines the results of these many self-attention layers~\cite{vaswani2017attention}.}.
Positional embeddings corresponding to the position of the patches in the original image are added to the computed tokens. A stack of multiple layers of multi-head attention computes a deep representation of the input images.
These latent representations of the input images are passed to the classifier to classify them into distinct categories, such as dog, cat, and car. The possibility of computing the correlation between each pixel in the input images via an attention matrix makes ViTs more powerful than CNNs for a given dataset; nevertheless, ViTs are prone to overfitting.
However, the higher learning capacity of ViTs provides the opportunity to use more data for training.
The larger version of ImageNet with more than $21,000$ classes (ImageNet-21k) is useful for pre-training ViTs (usually, optimal performance of pre-trained CNN models is achieved with ImageNet-1k. Additional data was not as helpful as in the case of ViTs).

\subsection{Vision Transformers for Segmentation}
\begin{figure}[htb!]
     \centering
     \includegraphics[width=\textwidth]{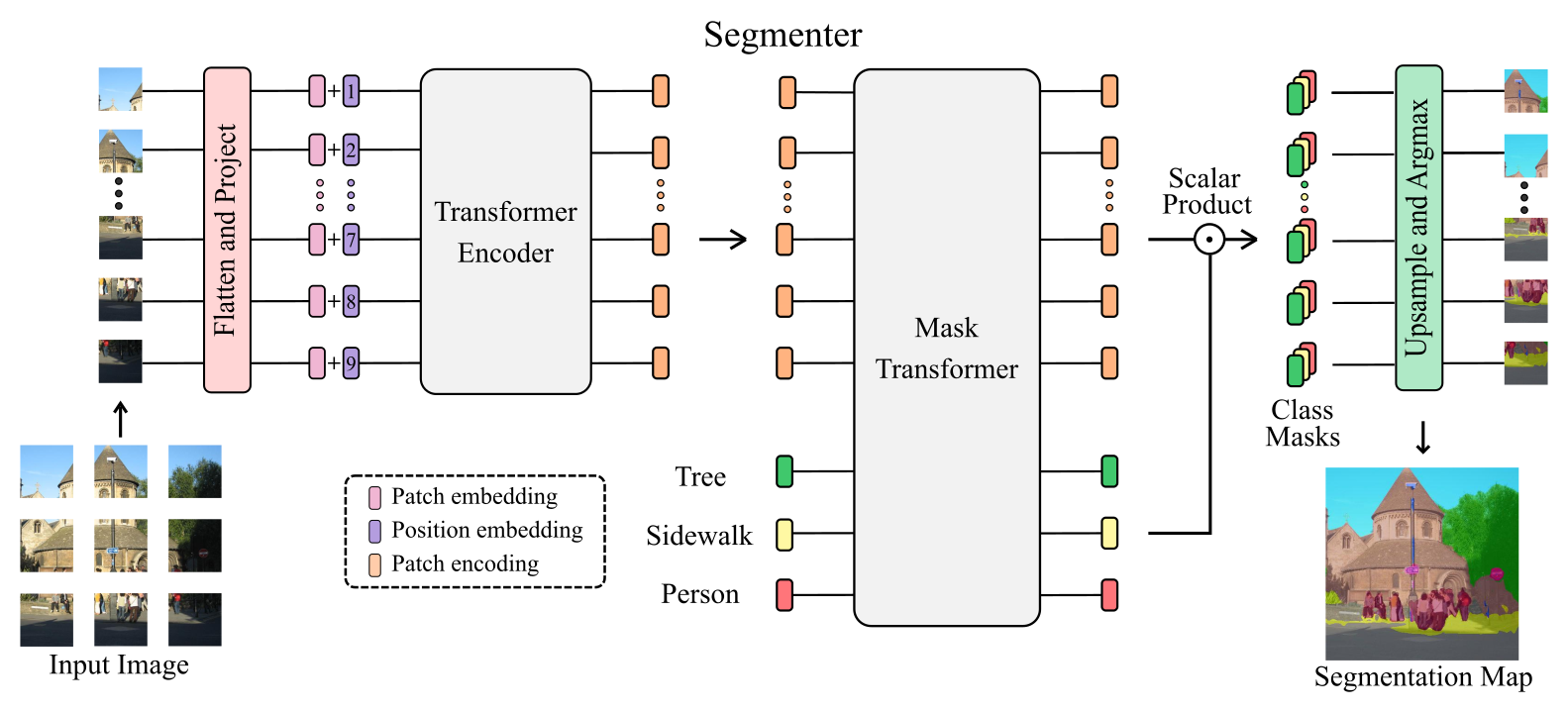}
    \caption{Vision transformers (ViTs) for image semantic segmentation (figure is adopted from~\cite{strudel2021segmenter}).}
    \label{chap:theory_fig:vitseg}
\end{figure}

ViTs can be extended from classification to semantic segmentation using similar encoding backbones.
Figure~\ref{chap:theory_fig:vitseg} shows how pre-trained ViTs for classification can serve as the first building block of ViTs for semantic segmentation. The main difference between classification and segmentation ViTs starts after encoding the images into patch embeddings ($\boldsymbol{z_L}\in\mathbb{R}^{N\times D}$ with $N$ patches and tokens of $D$ dimensions).
Classifiers then predict a vector with elements that sum to one, with the values being proportional to the probability of the predicted class.
The segmenter ViTs, on the other hand, approximate a segmentation map $s\in\mathbb{R}^{H\times W\times K}$ that represents the segmentation predictions for each pixel of $K$ classes in an image of a given height ($H$) and width ($W$).
Two additional components of randomly initialized class embeddings and a mask transformer support the adaptation of ViT architecture to compute the segmentation map. After calculating the patch embeddings of the input images using pre-trained classification ViTs, the patch embeddings are concatenated with class embeddings (($[cls_1,...,cls_k]\in\mathbb{R}^{K\times D}$)). The mask transformer includes several multi-headed self-attention layers where each class embedding attends each patch's pixel. At the end of the mask transformer, the normalized patch embeddings $\boldsymbol{z'_L}\in\mathbb{R}^{N\times D}$ and the class embeddings are separated, normalized based on their $\ell_2$, and a scalar dot product of each class embedding and patch embedding is computed to create a mask for each class. To predict the final image masks, the segmentation model includes an $argmax$ function to find the most likely class per pixel and reduce the predictions to the same size as the input image. 
   
\section{Optimizing Neural Networks}
So far, this chapter has explained several aspects of different models for image processing. This section describes how these models are optimized to fit a given dataset and find patterns in the images within a dataset. The goal of the optimization process is tunning the trainable parameters of the models ($\boldsymbol{\theta}$) for an objective function ($\boldsymbol{L}$) such that the model can generalize to unseen data. 
\subsection{Optimizing Trainable Parameters}
\label{chap:theory_sec:opt}
The optimization's objective function, the so-called loss function, reflects the dataset's target task. For example, a classifier has a loss function that provides the highest probability for the presence of the correct class in an input image. Likewise, the segmenter computes the highest probability for the object surrounding a single pixel. The goal of the optimization algorithms is to minimize the expected value ($\mathbb{E}$) of a loss function over the entire training dataset ($\hat{p}_{data}$) as follows~\cite{Goodfellow-et-al-2016}:

\begin{flalign}\label{chap:theory_eq:optimization}
J(\boldsymbol{\theta}) = \mathbb{E}_{\boldsymbol{x},y\sim\hat{p}_{data}}~ L(f(\boldsymbol{x};\boldsymbol{\theta}), y)
\end{flalign}

where $\boldsymbol{x}$ and $y$ denote the pair of data samples and ground truth labels. Although the training process of ML and DL models uses the training data ($\hat{p}_{data}$) for optimization, the main goal is to find a model that fits the data distribution ($p_{data}$): 

\begin{flalign}\label{chap:theory_eq:generalization}
J_{\boldsymbol{\theta}}^* = \mathbb{E}_{\boldsymbol{x},y\sim{p}_{data}}~ L(f(\boldsymbol{x};\boldsymbol{\theta}), y)
\end{flalign}

where $J(\boldsymbol{\theta})^*$ is the expected value of the error over the data distribution (not just the training set). The main difference between ML and DL optimization and classical problems is that the loss function depends on the training data. The search for optimal parameters for an ML ($\boldsymbol{\theta}_{ML}$) for a maximum likelihood problem can be described as follows:    

\begin{flalign}\label{chap:theory_eq:thetha_ml}
\boldsymbol{\theta}_{ML} = \argmax_{\boldsymbol{\theta}} \sum_{i=1}^N p_{model}(\boldsymbol{x}^{(i)}, y^{(i)}; \boldsymbol{\theta})
\end{flalign}

Maximizing the likelihood of predictions with ground truth labels is equivalent to minimizing the prediction error. For discrete pairs of data samples and labels ($\boldsymbol{x}$ and $y$), generalization is expressed as follows:

\begin{flalign}\label{chap:theory_eq:generalization_descrite}
J_{\boldsymbol{\theta}}^* = \sum_{\boldsymbol{x}}\sum_{y}p_{data}~ L(f(\boldsymbol{x};\boldsymbol{\theta}), y)
\end{flalign}

The gradient of the loss function ($\boldsymbol{g}$) is calculated for all parameters using training data in practice to find the optimal model's parameters:

\begin{flalign}\label{chap:theory_eq:gradient}
\boldsymbol{g} = \nabla _{\boldsymbol{\theta}} J_{\boldsymbol{\theta}} = \mathbb{E}_{\boldsymbol{x},y\sim\hat{p}_{data}}(\boldsymbol{x}, y)~ \nabla_{\boldsymbol{\theta}} L(f(\boldsymbol{x};\boldsymbol{\theta}), y)
\end{flalign}

Theoretically, to use a gradient descent algorithm to optimize the neural networks based on the gradients of the parameters with respect to the loss function, we need to compute the average of gradients over the entire training dataset before updating the parameters. However, this method is computationally very expensive, and the optimization algorithm converges faster when multiple updates are made from subsets of the training dataset (\textit{mini-batches}). Therefore, the gradient of the parameters with respect to the loss function is calculated for a mini-batch with a size of $m$ samples:

\begin{flalign}\label{chap:theory_eq:gradient2}
\hat{\boldsymbol{g}} = \frac{1}{m}\nabla_{\boldsymbol{\theta}} \sum_i L(f(\boldsymbol{x}^{(i)};\boldsymbol{\theta}), y^{(i)})
\end{flalign}

Larger mini-batch sizes result in cleaner gradients toward the minima of the objective function, and gradients computed with smaller batch sizes are noisier. However, this gradient noise can regularize the training process and improve the models' generalization. The use of mini-batches became a practical optimization approach due to their advantages in generalization and convergence speed. Neural networks and deep vision models consist of many layers with a depth of more than a hundred. The algorithm for computing the gradient of parameters for all layers is based on the chain rule, which is called \textit{backpropagation}. After calculating the loss function at the end of the models' computational graph, its partial derivatives are calculated with respect to the trainable variables of the models' last layer. These gradients are backpropagated toward input images to compute the gradients of all parameters minimizing the loss function. Then, the gradients are multiplied by a learning rate, and the parameters are updated based on these multiplications. The iterative training process continues until a stopping criterion is met.  

\subsection{Optimization to Generalization}
Training neural networks and computer vision models, as described in Section \ref{chap:theory_sec:opt}, focuses on minimizing the prediction error on the training set. However, the main goal of training is to optimize models that generalize well to the entire data distribution outside the training set. The art of bringing computer vision models to optimal performance involves many techniques and a lot of empirical trial and error. The main goal of these techniques is to limit the networks' capacity or artificially increase the amount of data by presenting different variations of the original dataset, which forces the model to learn more generic patterns instead of memorizing individual images from the training dataset. The remainder of this section describes some well-known methods for improving generalization after optimization.

\subsubsection*{Dropout}
There are several reasons for poor generalization (overfitting) in deep neural networks described in the literature. These reasons include neurons' coadaptation to a particular image with poor generalization and learning dense representations of the input images. Srivastava et al.~\cite{srivastava2014dropout} proposed dropout as an effective technique to improve the neural network's generalization to address the above reasons for overfitting. Dropout is equivalent to randomly setting the activations of a layer for a given input image to zero with a certain probability during training. The random suppression of activations via dropout prevents the model from coadapting neurons. Furthermore, dropout leads to learning sparse representations of input images and consequently improving the generalization. Although the usage of dropout in the classifiers (fully connected layers or MLPs) is more common, it is possible to use dropout in convolutional layers as well. Srivastava et al.~\cite{srivastava2014dropout} originally introduced dropout to reduce overfitting during training. Nonetheless, Gal and Ghahramani proposed a framework for estimating neural network uncertainty using dropout in test time as an additional application of dropout~\cite{gal2016dropout}. In their theoretical framework, dropout in neural networks has been successfully used for Bayesian inference in Gaussian processes to estimate uncertainty.

\subsubsection*{Regularization}
The other technique to limit the neural networks' capacity is adding a penalty to the neural network loss function that increases with the absolute value of the trainable weights. Researchers used different functions of the trainable weights as additive penalties to the original (classification or segmentation) loss function. Tibshirani and Zheng used the $\ell_1-norm$ of the weights to calculate such a penalty~\cite{tibshirani1996regression,NIPS2003_0a65e195}. The $\ell_1-norm$ regularization, also called Lasso regression regularization, keeps the sum of the absolute values of the trainable parameters small. Lasso regularization leads to sparse weight vectors by setting some weights to zero. An alternative to Lasso regularization is to compute the penalty term using the square of the weights ($\ell_2-norm$), which leads to Ridge regression regularization~\cite{nigam1999using}. The computed regularization penalty (loss) is multiplied by a regularization factor and then added to the loss value of the training. Loshchilov and Hutter have shown that decoupling the regularization penalty from the classification loss by defining an independent weight decay from the learning rate for adaptive gradient algorithms improves the generalization performance~\cite{loshchilov2018decoupled}.

\subsubsection*{Augmentation}
Besides dropping neurons from the neural network architecture and regularizing the weights, neural network generalization improves by presenting the models with different variations of the input data. Since the early days of deep learning research, studies have shown that CNNs have limited robustness to rotation and scaling~\cite{lecun1998gradient}\footnote{\url{http://yann.lecun.com/exdb/lenet/}}. However, researchers quickly found that computer vision models can recover such weaknesses by presenting variations of input images to the network during training time. Thus, computer vision models trained with rotated versions of the input images are robust against rotations. \textit{Augmentation} is the term proposed in the literature for training computer vision models with transformed images. Since then, researchers have used various strategies to augment their input images depending on the application. Techniques for augmentation such as shearing, translation, rotation, rectification, and changing contrast, color, brightness, and sharpness are broadly used in the computer vision research community~\cite{ciregan2012multi, sato2015apac, simard2003best, wan2013regularization, krizhevsky2012imagenet}. 

Similar to the architectural design literature explained in section \ref{chap:theory_sec:ArchtictureSearch}, augmentation methods have also evolved towards automation. An important work of research on this subject resulted in \textit{AutoAugment}~\cite{cubuk2019autoaugment}, which presents optimal strategies that are automatically checked against one of the largest computer vision datasets (ImageNet~\cite{deng2009imagenet}). Further improvement in the speed of such a resource-exhaustive search led to the development of \textit{Fast AutoAugment}, an algorithm that is lighter in terms of computational complexity and more suitable for exploring optimal augmentation strategies on private datasets~\cite{lim2019fast}.

\section{Related Work}
Two critical components of the rapid increase in computational resources and datasets' size have revived machine and deep learning (ML and DL) techniques in practical applications for image pattern classification~\cite{lecun1989backpropagation}. Initially, raw pixel values for simple tasks such as classifying handwritten digits were sufficient to train neural networks and support vector machines for pattern classification~\cite{cortes1995support}. However, the urge to extract robust features for more complex computer vision problems led researchers to develop advanced methods for representation learning~\cite{lowe2004distinctive}. 

ML pipelines became increasingly complicated in the first decade of the twentieth century as problem-oriented feature extraction techniques grew rapidly~\cite{kumar2014detailed}. Classical computer vision researchers, who moved away from Fourier transforms and brought image-based prior knowledge to multiscale wavelet transform, began training sparse dictionaries from data~\cite{rubinstein2010dictionaries}, and robust hand-crafted features~\cite{lowe1999object} for pattern recognition received enormous attention. Moreover, the growth of datasets quickly saturated the performance of classical ML models, and task diversity made the search space for the best priors exhaustive. DL appeared as the next breakthrough to increase the capacity of models for massive datasets and automate feature extraction and representation learning in a wide range of different tasks~\cite{lecun2015deep}.

The first DL milestone in computer vision was reviving the convolutional neural networks (CNNs) for pattern recognition~\cite{lecun1989backpropagation}. CNNs introduced in the late 1980s finally found their way into practice by overcoming their high computational complexity. LeNet5~\cite{lecun1998gradient} is one of the first CNN models applied to handwritten digit classification algorithms, and AlexNet~\cite{gabor1946theory} is among the first successes of deep CNNs in image classification on large datasets. The second half of the $2010$s is the most prosperous time in the history of DL and CNNs in computer vision with a lot of exciting research and developments regarding various architectures~\cite{simonyan2014very, szegedy2016rethinking, sandler2018mobilenetv2, tan2019mnasnet} and optimization techniques~\cite{robbins1951stochastic, kiefer1952stochastic, kingma2014adam, loshchilov2018decoupled}. Researchers trained CNNs using neural architectures, search became fashionable, optimization methods evolved based on large datasets such as ImageNet~\cite{ILSVRC15}, and this distilled knowledge was successfully applied to smaller datasets using transfer learning~\cite{pan2009survey}.

After CNNs matured in image classification~\cite{bianco2018benchmark} and segmentation~\cite{lateef2019survey}, the next generation of research focused on automatic neural architecture search~\cite{elsken2019neural} and finding optimal search spaces for efficient networks with minimal delay~\cite{tan2019mnasnet} and computational power consumption for mobile applications~\cite{tan2019efficientnet}. However, the parallel increase in computational resources and the presence of massive datasets, motivated by data-driven artificial intelligence (AI) research, created the opportunity for the next breakthrough in computer vision. The next breakthrough occurred in the early 2020s by introducing vision transformers (ViTs) ~\cite{dosovitskiy2020image}, and adapting self-attention, originally discovered in natural language processing (NLP) literature~\cite{vaswani2017attention}, for computer vision tasks. ViTs are widely used for image classification and segmentation, and these models have improved their performance with a larger version of the ImageNet dataset called ImageNet-21k with over $21,000$ classes~\cite{ridnik2021imagenet}. 

Researchers have expressed doubts and concerns about the robustness of computer vision models using CNNs~\cite{goodfellow2014explaining} and their explainability~\cite{DARPA2016explainable} from their inception. CNNs lack some basic properties of classical methods, such as rotation equivariance\footnote{A rotationally equivariant representation of an image rotates with the same angle as its input rotates. Edge detection filters, for example, are rotationally equivariant}, despite their capability to learn translational equivariant features~\cite{lecun1998gradient}. Due to the lack of rotation equivariance, the performance of CNNs decreases when the input images rotate~\cite{kanbak2018geometric}. Similar instabilities and inaccuracies have been reported due to changes in lighting conditions, contrast, image acquisition techniques, and overall data distribution drift~\cite{disabato2019learning}. Studies even show that CNNs focus more on the texture than the shapes when classifying objects~\cite{geirhos2018imagenet}. Researchers also discovered that they could compute minimal perturbations, called adversarial attacks, for an input image to fool CNN models with images that are indistinguishable from one another to the human eye~\cite{goodfellow2014explaining}. They even optimized a so-called universal adversarial attack that generalizes to many images~\cite{moosavi2017universal}. Among all of the challenges mentioned in computer vision research related to robustness, this work presents a solution for rotational invariance in ViTs and adversarial attack detection in CNNs~\cite{amirian2018trace}.

CNNs were known as powerful black-box models following a similar trend to many other ML and DL-based techniques in information processing~\cite{buhrmester2021analysis}. These models can be used in many applications without additional reasoning; however, understanding these high-precision decisions is critical for applications that affect human safety and health, such as autonomous driving systems~\cite{atakishiyev2021explainable} or healthcare~\cite{amann2020explainability}. The literature that has developed around explainable AI (XAI)~\cite{tjoa2020survey} and interpretation of computer vision models~\cite{zhang2018visual} is the result of researchers' concerns about the usage of black-box models in critical applications. Researchers have two main approaches to the interpretability and explainability of neural networks. The first group analyzes the trained models to understand the predictions using post-processing and post-hoc techniques~\cite{crabbe2022label}. Another group disagrees with the idea of interpretability solely as an add-on for neural networks. Instead, these researchers point to changing the design of the neural architectures so that the decisions are transparent and explainable~\cite{marcinkevivcs2021interpretable}. Concerning the issue of explainability, this thesis proposes using radial basis neural networks as classifiers on top of the CNNs to provide more understandable information to humans about the decision-making of the models~\cite{amirian2020radial}.

Despite all the challenges mentioned above, computer vision breakthroughs have found their way into numerous applications~\cite{alzubaidi2021review}. CNNs have outperformed all other methods in the majority of applications, such as object detection, recognition, and segmentation~\cite{bianco2018benchmark, lateef2019survey}. Furthermore, CNNs perform well in other classical image processing tasks such as image denoising~\cite{dong2018denoising}, super-resolution~\cite{yamanaka2017fast}, and motion deblurring~\cite{sun2015learning}. Moreover, the applications of CNNs extend not only to medical imaging for diagnosis~\cite{sarvamangala2021convolutional} and automatic segmentation~\cite{lei2020medical}, but also to image quality enhancement and motion artifact reduction~\cite{lin2019dudonet}. In addition to all the above theoretical contributions, this paper presents practical applications of ML and DL, especially in medical imaging, and demonstrates a variety of empirical findings~\cite{stadelmann2018beyondimagenet}.


\chapter{RBF Classifiers for Explainable Computer Vision Using CNNs}
\label{chap:rbfs}
\chaptermark{RBF Classifiers for Explainable CNNs}
Radial basis function neural networks (RBFs) are prime candidates for pattern classification and regression and have been used extensively in classical machine learning applications. However, RBFs have not been integrated into contemporary deep learning research, and computer vision has continued using conventional convolutional neural networks (CNNs) because of technical difficulties. This chapter presents the techniques to adapt RBF networks as a classifier on top of CNNs by modifying the training process and introducing a new activation function to train modern vision architectures end-to-end for image classification. The specific architecture of RBFs enables them to learn a similarity distance metric to compare and categorize similar and dissimilar images. Furthermore, this chapter demonstrates that using an RBF classifier on top of any CNN architecture provides new human-interpretable insights about the decision-making process of the vision models. Finally, RBFs are successfully applied to a range of CNN architectures, and their performance on benchmark computer vision datasets is presented in this chapter. This chapter is adopted from the research published in~\cite{amirian2020radial}, licensed under CC BY 4.0~\footnote{\url{https://creativecommons.org/licenses/by/4.0}}.
\newpage
\section{Introduction}

\begin{figure*}[bth!]
    \centering
    \begin{minipage}{\textwidth}
    \includegraphics[width=\linewidth]{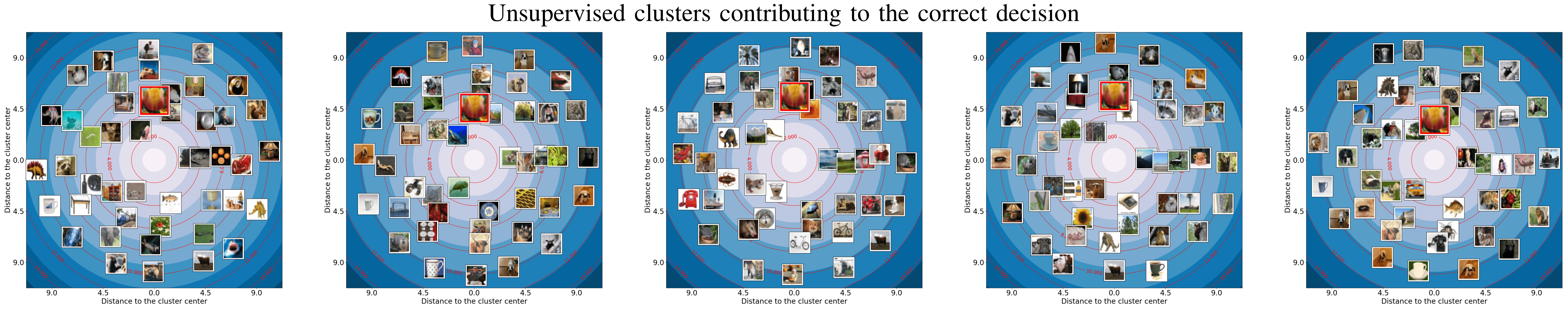} \\
    \includegraphics[width=0.5\linewidth]{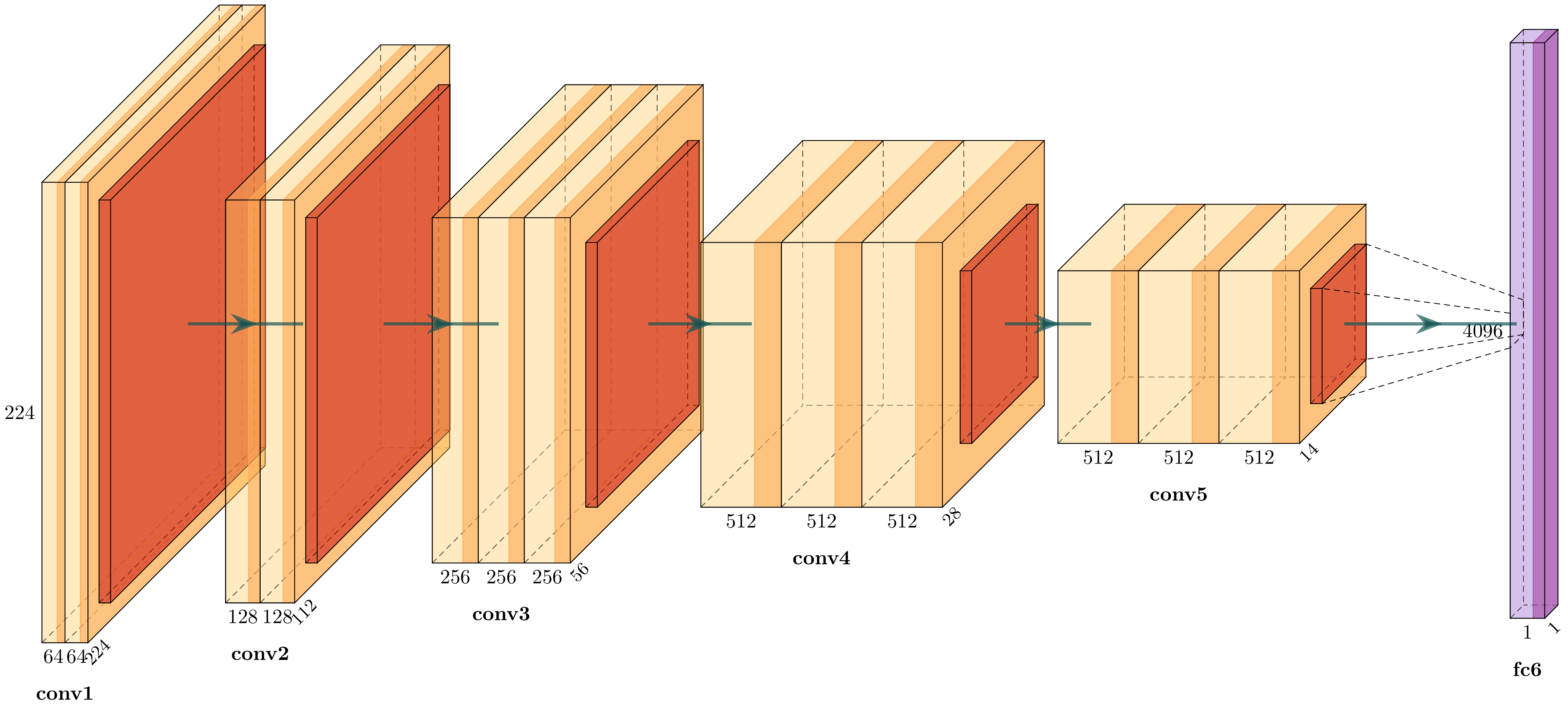}
    \includegraphics[width=0.10\linewidth]{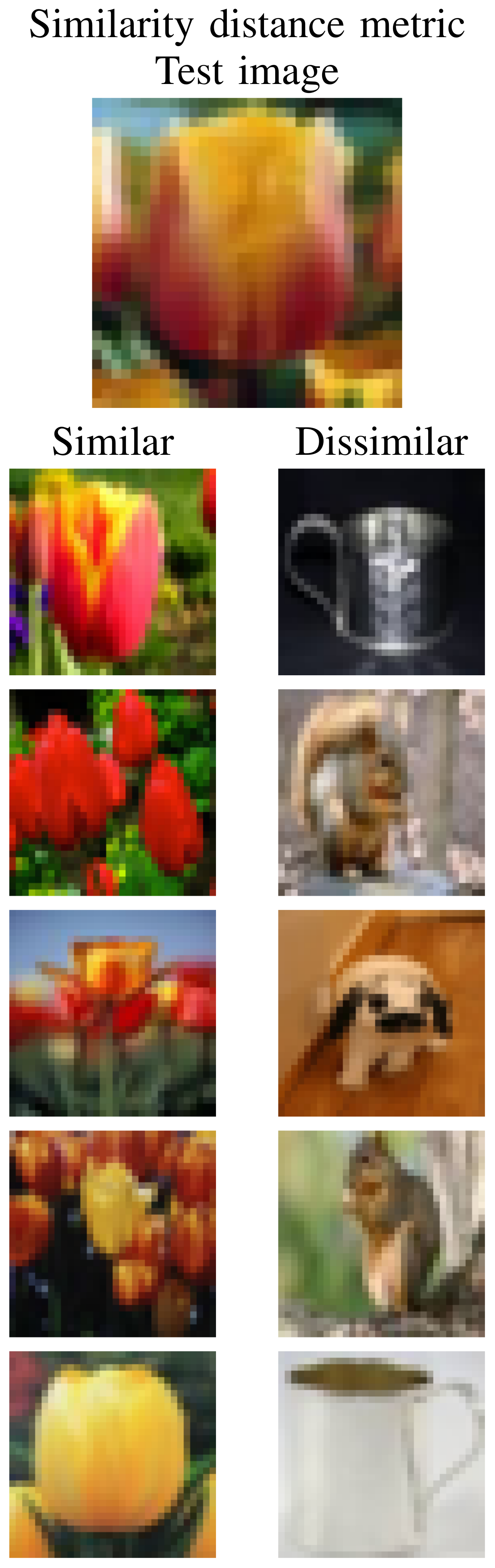}
    \includegraphics[width=0.3\linewidth]{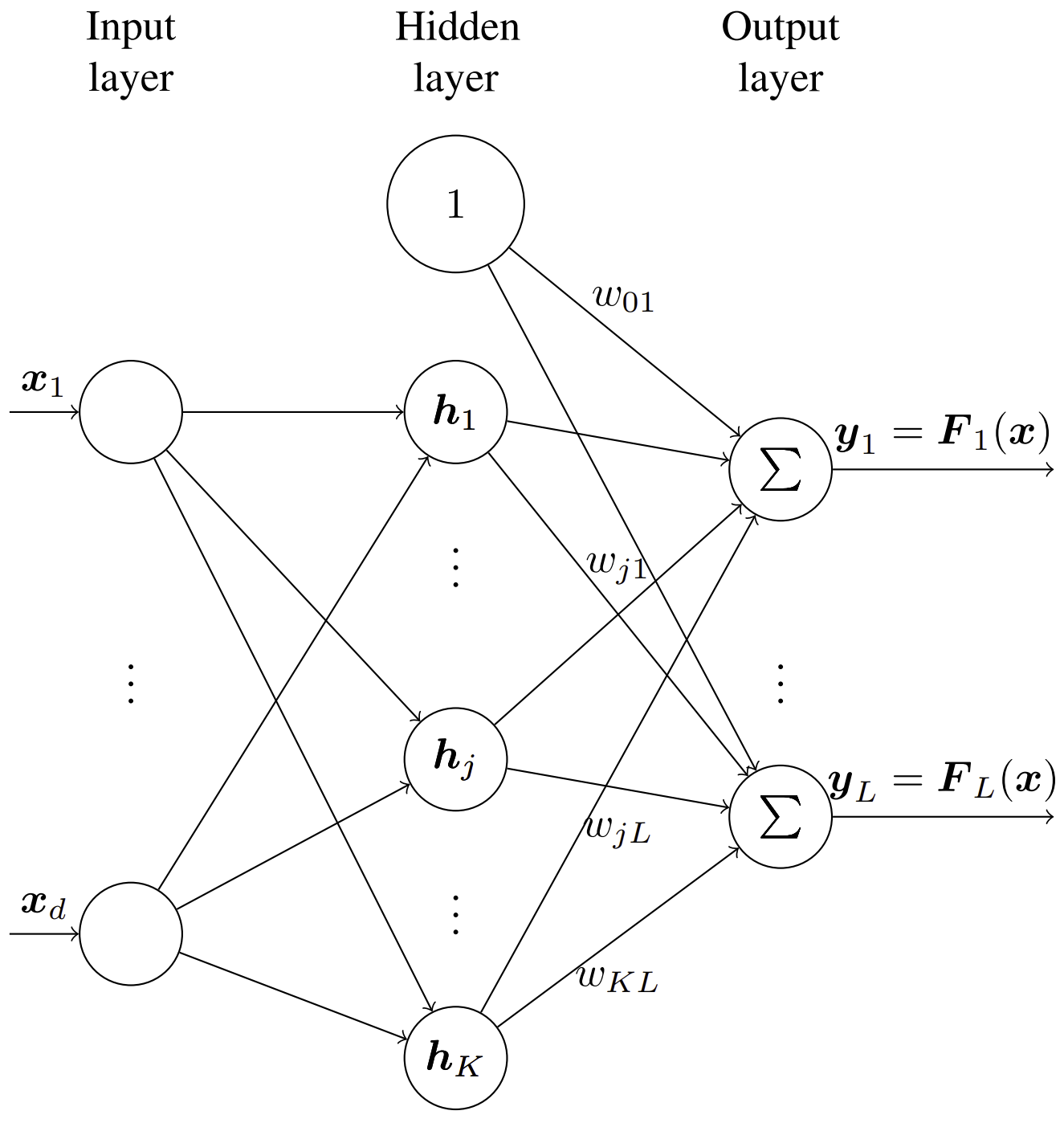} \\
    \includegraphics[width=\linewidth]{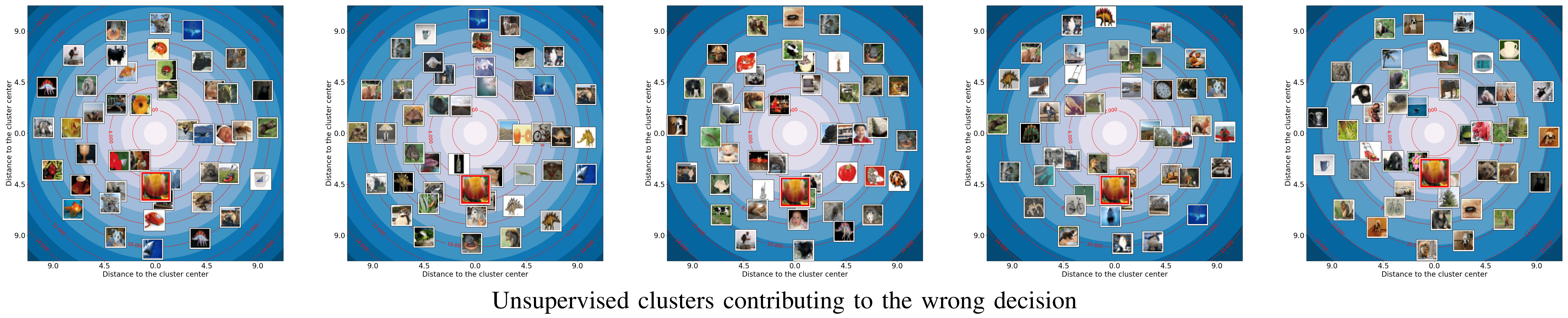} \\
    \end{minipage}
    \caption{Figures on the top and bottom rows visualize the position of a test image in the clusters optimized using the unsupervised loss function. The output of CNN backbones is connected to RBFs' input through a fully connected layer, and the input features of the RBFs are referred to as embeddings in this chapter. The model compares the embeddings of each image with cluster centers using a trainable similarity distance metric. The same distance metric can be used to find similar and dissimilar images to a test sample amongst training images (visualized in the table in the middle row). The RBFs apply an activation function to the distance of the training images from the cluster centers to compute activation values. The output layer of the RBF is optimized for classification based on these activation values. The entire CNN-RBF architecture is optimized end-to-end with a specific initialization (figure adopted from~\cite{amirian2020radial}).}
    \label{fig:CNN-RBF}
\end{figure*}

Inspired by the locally tuned response of biological neurons, Broomhead and Lowe introduced radial basis function neural networks (RBFs) in 1988~\cite{broomhead1988multivariable}. The modeling concept behind RBFs is a combination of unsupervised and supervised learning for pattern classification and regression. However, due to structural deficiencies, RBFs have not been integrated into contemporary research in computer vision using Convolutional Neural Networks (CNNs). This chapter presents developments in a new area of research and lays the foundation for using RBFs in deep learning and computer vision by modifying their architecture and learning process. The results demonstrate that integrating RBFs into CNN models for computer vision provides a similarity distance metric and an interpretable decision-making process.

This chapter is motivated by RBF architectures' unique opportunities when used with CNN models because of their explainability and robustness compared to linear classifiers. The new training process introduced for RBFs in this chapter provides the opportunity to use labeled and unlabeled data by optimizing two loss functions combining supervised and unsupervised learning. Moreover, the training process of RBF architectures includes optimizing a distance metric that serves as a similarity distance metric to find similar and dissimilar images. Additionally, this chapter proposes visualization techniques to illustrate the clusters and activations with training and test images to gain more insights about the reason behind the decisions made by the networks, thus improving interpretability. The contributions of this chapter to computer vision literature can be summarized as follows:       
\begin{itemize}
    \item Combining supervised and unsupervised learning.
    \item Learning a similarity distance metric to find similar images.
    \item Improving the interpretability of decision-making.
\end{itemize}

Despite the advantages of combining RBFs with modern CNN architectures, two factors in the architecture and training process of RBFs hinder their integration into CNNs. First, the nonlinear activations and computational graphs of RBFs used in the literature prevent efficient gradient flow. Secondly, RBFs assume that the training features are fixed, so the cluster centers are initialized accordingly. Nonetheless, CNN architecture dynamically learns the embeddings used as input features of RBFs. This chapter tackles the limitations of the original RBFs and presents the following contributions to RBF literature:

\begin{itemize}
    \item Introducing a quadratic activation function and a linear computational graph for end-to-end learning.
    \item Adding an unsupervised loss term to update the cluster centers in the training process with the learned embeddings.
    \item Applying the RBFs to computer vision in a first attempt at using deep CNN architectures. 
\end{itemize}


The remainder of the chapter covers the related work in Section~\ref{chap:rbf_sec:related} followed by the theoretical background of RBFs in Section~\ref{chap:rbf_sec:RBF}.
Then, Section~\ref{chap:rbf_sec:CNN-RBF} presents the original research and contributions with the proposed modifications to RBFs, followed by a visual explanation of the new proposed training and decision-making process in Section~\ref{chap:rbf_sec:training}.
The experimental results of applying the proposed RBF-CNN architectures using a range of CNN backbones on benchmark datasets are presented in Section~\ref{chap:rbf_sec:experimental}.
The potential contributions of the proposed similarity distance metric on computer vision to enhance the transparency of the decision-making process is demonstrated in Section~\ref{chap:rbf_sec:interpret}.
This chapter concludes with discussions and conclusions in Section~\ref{chap:rbf_sec:conclusion}.

\section{Related Work}
\label{chap:rbf_sec:related}
The research followed two approaches to optimize RBF architectures. The first approach concentrates on the training process and initialization of the networks, while the second aims to find superior activation functions. This chapter presents improvements in both research directions to integrate the RBFs into contemporary computer vision models using CNNs.

RBFs were originally introduced as supervised models for classification and regression tasks. Broomhead and Lowe initially proposed drawing the cluster centers either from a uniform distribution or randomly from the training samples and then optimizing the output weights using a pseudo-inverse analytic solution~\cite{broomhead1988multivariable}. Initializing the cluster centers randomly and only training the output weights is called a one-phase training process for RBFs. Two-phase training for RBFs uses various methods to initialize the cluster centers before optimizing the output weights. Research since 1988 has used supervised and unsupervised methods to initialize the cluster centers. Moody and Darken proposed an unsupervised algorithm to initialize these cluster centers~\cite{moody1989fast}, while Schwenker et al. proposed supervised vector quantization~\cite{schwenker1994similarities}. Decision trees were used to find centers independently by~\cite{kubat1998decision} and~\cite{schwenker2000initialisation} before training the output weights. Finally, Schwenker et al. proposed the third phase to optimize the entire RBF network end-to-end, including output weights, the cluster center, and trainable parameters of activation functions using gradient descent~\cite{schwenker2001three}.   

These methods for cluster center initialization assume a fixed feature space for the input layer. However, CNNs learn the embeddings automatically and develop the feature space of the images during the training process. Therefore, this research suggests optimizing an unsupervised learning loss during the training to cope with this change in the feature space. This work differs from previous research as it combines supervised and unsupervised learning by optimizing two separate losses simultaneously using gradient descent.

The technical requirements of new applications and implementations have motivated the use of several activation functions presented in the literature of RBFs~\cite{du2002fast}. The Gaussian function is the kernel developed by modeling the data through a multivariate Gaussian distribution~\cite{broomhead1988multivariable}. Other functions adapted in the RBF architecture include linear kernels, thin-plate splines, logistic functions, and multiquadratic functions~\cite{franke1979critical, poggio1990networks,liao2003relaxed,chen1990practical}.
Hardy's multiquadratic functions motivated an activation function for RBFs used by Karimi et al., and Zhao et al.~\cite{karimi2020generalized, zhao2019prediction}. Du et al. proposed a kernel for digital signal processing (DSP) units~\ref{eq:kernels7}. This chapter presents a novel quadratic kernel to build a linear computational graph for efficient gradient flow and RBF integration for end-to-end training with CNN architectures.

Besides the mature fundamental research, RBFs have been applied to many applications for pattern classification and regression in recent years. For example, Nicodemou et al. used RBF networks for 3D hand pose estimation~\cite{nicodemou2020single}, Dehghan and Mohammadi estimated a numerical solution for Fokker-Planck differential equations with RBFs~\cite{dehghan2014numerical}, Li et al. used sparse multiscale RBFs for seizure detection in EEG signals~\cite{li2019epileptic}. Furthermore, Zhao et al. predicted interfacial interactions by training RBFs~\cite{zhao2019prediction}, and Geng et al. introduced deep RBF networks and applied the method to food safety inspection data. Finally, RBFs are used to train models for classification and regression in discrete and continuous pain quantification~\cite{amirian2016using}.

RBFs can be applied to computer vision tasks and image classification as well. Schwenker et al. used raw images as feature vectors to classify hand-written digits~\cite{schwenker2001three}. Er et al. extracted the features from facial images using principal component analysis (PCA) and processed these features using Fisher's linear discriminant (FLD) technique before classifying the faces using RBFs~\cite{er2002face}. However, the successful rise of modern CNNs, such as LeNet-5~\cite{lecun1998gradient} and AlexNet~\cite{russakovsky2015imagenet}, led to a paradigm shift from using hand-crafted features to automated deep CNN-based feature and representation learning. In recent years, most computer vision tasks, like facial recognition~\cite{masi8614364survey}, are dominated by modern CNN architectures as they present superior performance compared to classical methods for image processing. To the best of our knowledge, this chapter presents the first attempt to integrate RBFs into modern CNN architectures for computer vision.

This chapter relates to literature focusing on deep metric learning since RBFs automatically optimize a similarity distance metric during training based on their architecture. Euclidean distance, Mahalanobis distance, and cosine similarity have been used to evaluate the similarity between the embeddings (the features extracted from CNNs) of two images in the literature~\cite{hoffer2015deep, xing2003distance, wojke2018deep}. Researchers have applied different strategies and loss functions to optimize these similarity metrics for same-class images while also maximizing the distance of different-class images. The research in this area concentrates on the training process and the design of a loss function which brings similar images closer in the embedding space based on a similarity measure. Hu et al. proposed minimizing the inter-class scores and maximizing the intra-class scores based on Euclidian distances~\cite{hu2015deep}. Hoffer and Ailon suggested optimizing a similarity-based loss function defined for selected triplets of images~\cite{hoffer2015deep}. Song et al. used the pairwise distances between images of an entire batch and proposed a structured loss function for metric learning~\cite{oh2016deep}. Similar research work has aimed at optimizing angular distance, cosine distance, and large-margin Euclidean distance of similar and dissimilar images~\cite{wang2017deep, wojke2018deep, cheng2018deep}.

This chapter presents a method to retrieve a ranked list of similar and dissimilar images, leading to visually appealing similarity metric learning results. However, the proposed similarity metric learned by the RBFs does not require any complicated triplet sample section or loss design. Instead, these results have been obtained using a typical supervised loss function for classification (softmax cross-entropy). Furthermore, RBFs can not only optimize for Euclidean and Mahalanobis distances but also for the entire covariance matrix.

\section{Radial Basis Function Networks}

\label{chap:rbf_sec:RBF}
This section briefly reviews and explains the theoretical foundation of radial basis function networks. RBFs are presented in the literature as a global approximation method for learning a mapping $\boldsymbol{F}$ from a given feature space with the dimensionality of $d$ to a label space with $K$ dimensions ($\boldsymbol{F}: {\rm I\!R}^d \rightarrow {\rm I\!R}^K$)~\cite{broomhead1988multivariable}. In this chapter, the function $\boldsymbol{F}$ of features $\boldsymbol{x}$ approximates the one-hot encoded labels $\boldsymbol{y}$. The features used to train the RBFs in this chapter are the embeddings of deep CNNs, which are used to predict the class labels using end-to-end optimization. A fully connected layer connects the CNN architectures and RBFs to provide compatibility between the two architectures. The architecture of the RBF consists of input layers, a single trainable hidden layer with $C$ cluster centers ($\boldsymbol{c}_j$)~\ref{fig:CNN-RBF}, and an output layer.

\begin{figure*}[htb!]
    \centering
    \includegraphics[width=0.4\textwidth, valign=t]{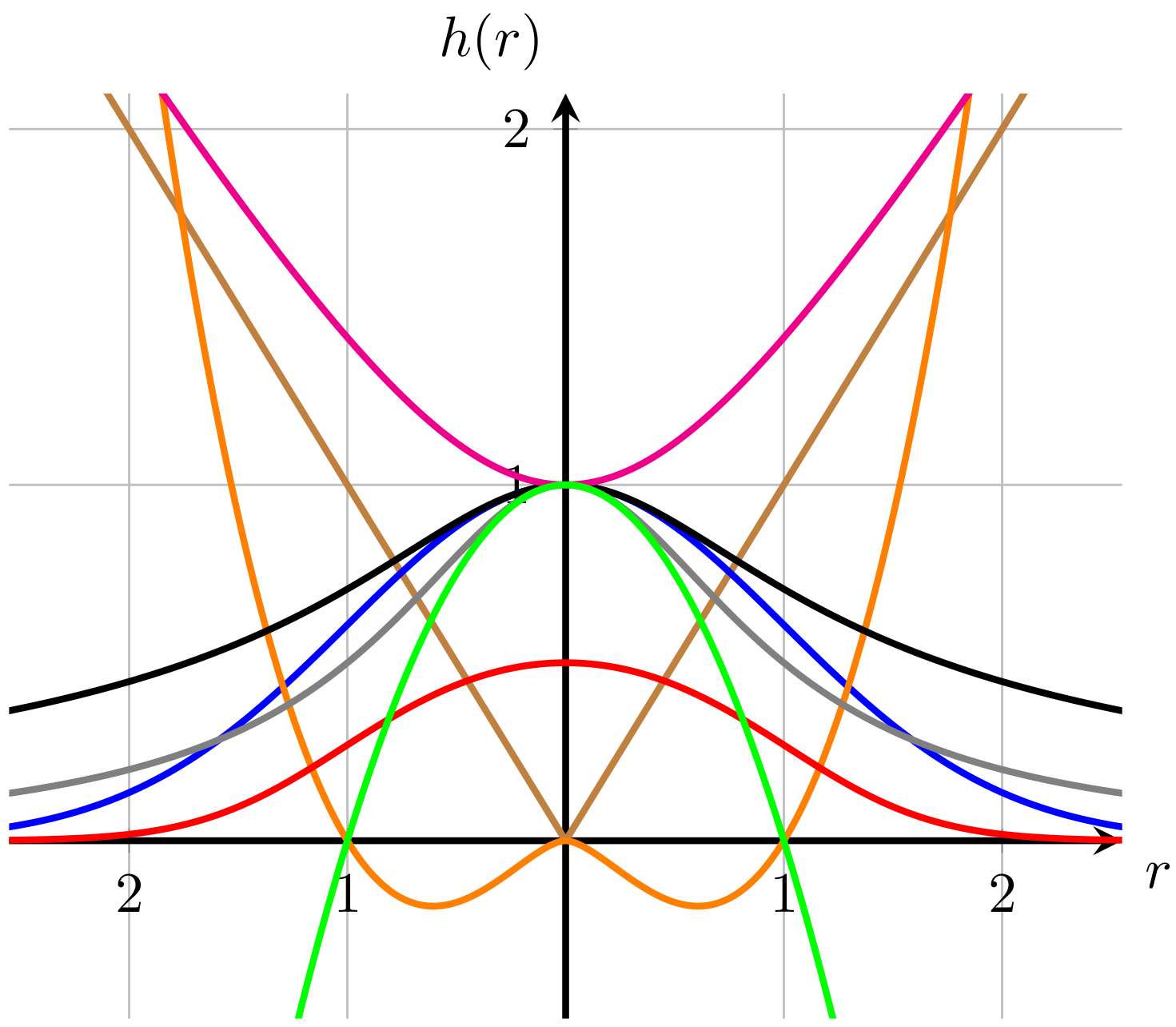}
    \hspace{-0.35cm}
    \raisebox{-0.75cm}{\includegraphics[width=0.17\textwidth, valign=t]{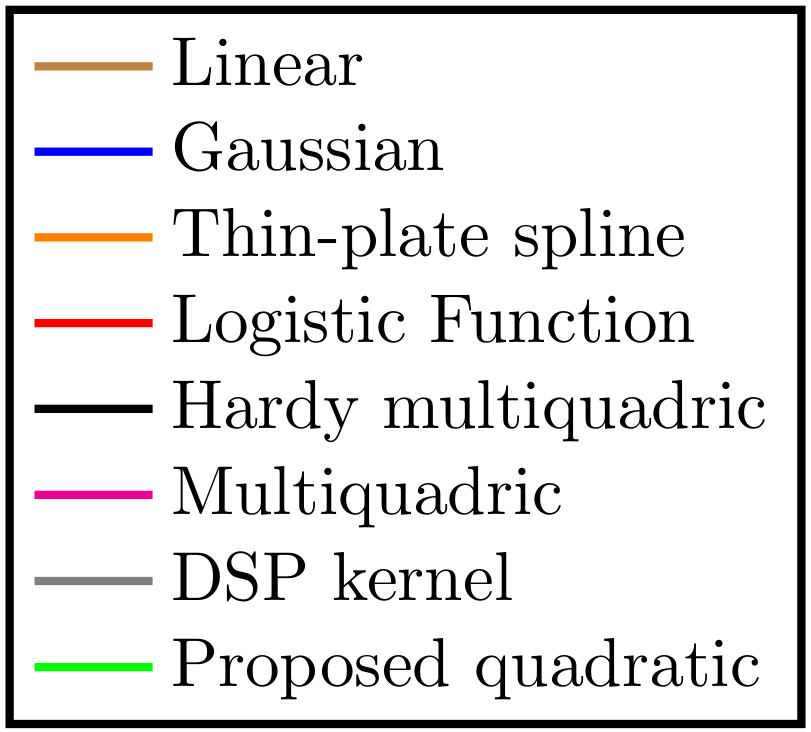}}
    \includegraphics[width=0.4\textwidth, valign=t]{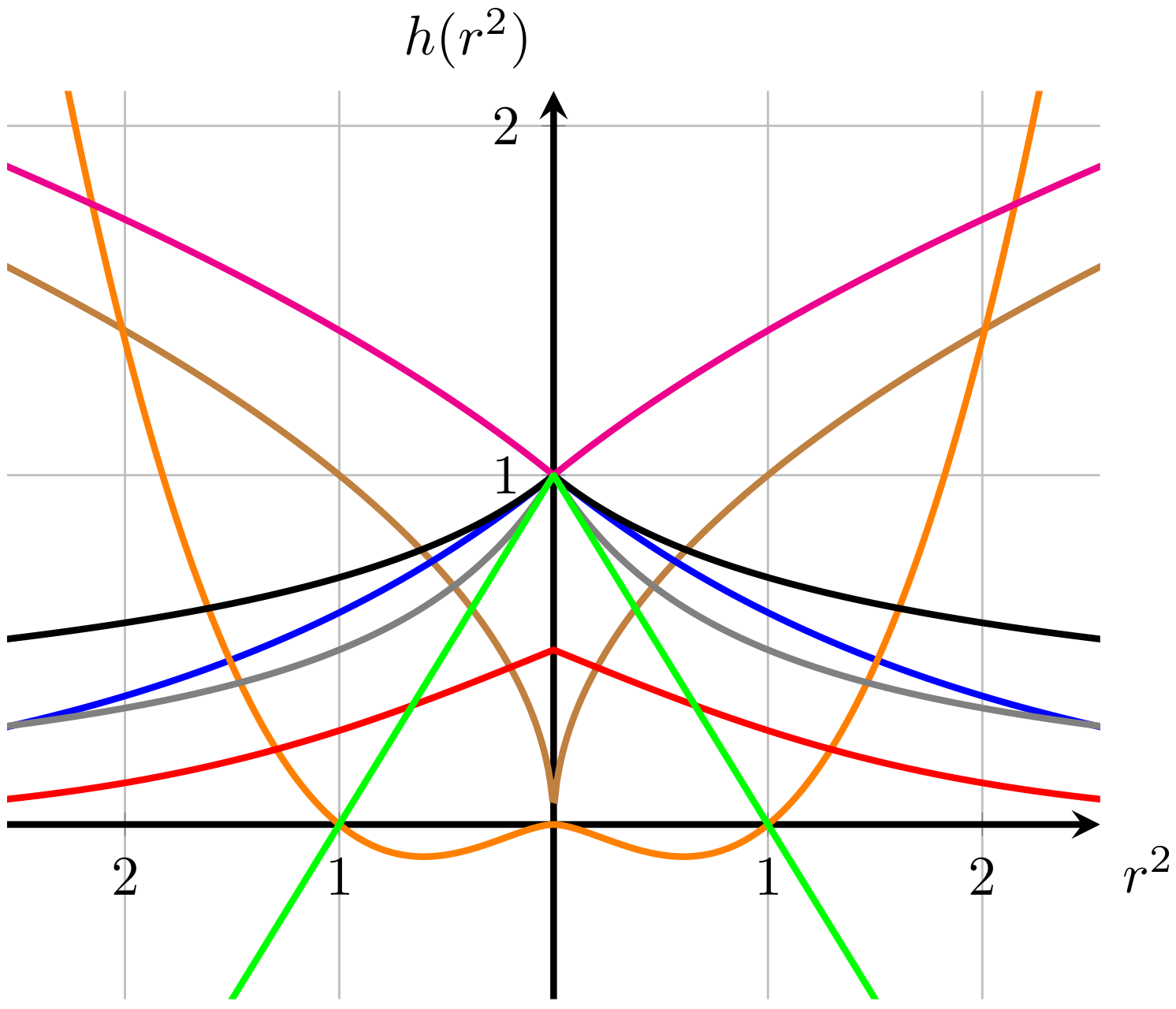}
    \caption{Activation functions for RBF networks. Here is the list of the parameters for depicting the kernels: $\sigma=1$, $\alpha=1/2$, and $\beta=1/2$. The proposed quadratic activation kernel is linear based on the $r^2$. Consequently, the CNN goes through a completely linear forward path, and thus, gradients are computed and backpropagated efficiently (figure adopted from~\cite{amirian2020radial}).}
 	\label{fig:activation}
  \vspace{-0.2cm}
\end{figure*}

During the evaluation, also known as inference in deep learning, the RBF computes a distance between embeddings of CNNs and the cluster centers and applies an activation function to this distance. The network outputs are then computed by multiplying the output layer weights with the activation values. This forward path of RBFs is formally defined as:
\begin{flalign}\label{eq:distance}
r^2 = 
(\boldsymbol{x}-\boldsymbol{c}_j)^T\boldsymbol{R}_j(\boldsymbol{x}-\boldsymbol{c}_j) &&
\end{flalign}
\begin{flalign}\label{eq:output}
\boldsymbol{y}_k = \boldsymbol{F}_k(\boldsymbol{x}) = \sum_{j=1}^{C}w_{jk} h(\parallel \boldsymbol{x}^\mu-\boldsymbol{c}_j \parallel^2_{\boldsymbol{R}_j}) + w_{0k} &&
\end{flalign} 
where $r$ represents the distance, $R_j$ is the positive definite covariance matrix (trainable distance metric), $T$ denotes the matrix transposition, $ w_{jk} $ shows the weights of the output layers, $h$ is the activation function, and $w_{0k}$ are the biases. In these Equations, $\mu$, $j$, and $k$ enumerate the number of samples, cluster centers, and classes. Trainable parameters in Equation~\ref{eq:distance} and~\ref{eq:output} are the output weights, cluster centers, and covariance matrix. 

Optimizing the RBF networks with an identity covariance matrix is equivalent to training in Euclidean space. It is possible to optimize a Mahalanobis distance~\cite{de2000mahalanobis} by training the main diagonal on the covariance matrix. Any arbitrary distance metric can be trained by optimizing the entire covariance matrix, while projecting the matrix to the space of positive definite matrices. The distance, $r$, computed in Equation~\ref{eq:distance} is not only a measure of the proximity of an image to a cluster center; it can also be used to compare images and find similar and dissimilar images in the embedding space.

The linear and nonlinear activation functions used in RBFs are as follows~\cite{poggio1990networks,liao2003relaxed,chen1990practical}: 
\begin{flalign}
	&\text{Linear}:            & h(r) &= r                                           && \label{eq:kernels1}\\
	&\text{Gaussian}:          & h(r) &= e^{-r^2/2\sigma^2}                          && \label{eq:kernels2}\\
	&\text{Thin-plate spline}: & h(r) &= r^2 \ln{r}                                  && \label{eq:kernels3}\\
	&\text{Logistic function}: & h(r) &= \frac{1}{1+e^{(r^2-r^2_0)/\sigma^2}}        && \label{eq:kernels4}\\
    &                          & h(r) &= \frac{1}{(r^2+\sigma^2)^\alpha},~\alpha > 0 && \label{eq:kernels5}\\
    &                          & h(r) &= (r^2+\sigma^2)^\beta,~0 < \beta < 1         && \label{eq:kernels6}\\
	&                          & h(r) &= \frac{1}{1+r^2/\sigma^2}                    && \label{eq:kernels7}
\end{flalign}

In addition to the standard machine learning activation kernels in Equations~\ref{eq:kernels1} to~\ref{eq:kernels4}, the kernel presented in Equation~\ref{eq:kernels5} is derived from the generalized Hardy's multiquadratic function~\cite{franke1979critical}. Du et al.~\cite{du2002fast} proposed the kernel in Equation~\ref{eq:kernels7} because of its convenience for implementation on DSP units. Various activation functions for RBFs are depicted in Figure~\ref{fig:activation}.

The complete process of training RBFs was introduced by Schwenker et al.~\cite{schwenker2001three} as a three-phase process:

\textbf{Unsupervised learning}: This step aims to find cluster centers that are representative of the given dataset. The k-means~\cite{anderberg1973cluster} clustering algorithm is widely used for this purpose. K-means iteratively finds a set of cluster centers and minimizes the overall distance between cluster centers and members over the entire dataset. The target of the k-means algorithm can be written in the following form:
\begin{flalign}\label{eq:clus}
    \text{Loss}_{unsupervised} = \sum_{j=1}^{K}\sum_{\boldsymbol{x}^\mu\in\vartheta_j}^{} \parallel \boldsymbol{x}^\mu-\boldsymbol{c}_j \parallel^2 &&
\end{flalign} 
where $\boldsymbol{x}^\mu\in\vartheta_j$ denotes the members of the $j^{th}$ cluster shown by $\vartheta_j$. 

\textbf{Computing weights}: The output weights of an RBF network can be computed using a closed-form solution. The matrix of activation of the samples is defined from the training set ($H$) as follows:
\begin{flalign}\label{eq:activationMatrix}
    \boldsymbol{H} = h(\parallel \boldsymbol{x}^\mu-\boldsymbol{c}_j \parallel^2_{\boldsymbol{R}_j})_{\mu=1,~...~,M, j=1,~...~,C} &&
\end{flalign}

Based on Equation \ref{eq:output}, the matrix of output weights ($W$), which estimates the matrix of labels ($Y$), is computed using the following equation:
\begin{flalign}\label{eq:computeWeight}
    \boldsymbol{Y} \approx \boldsymbol{H}\boldsymbol{W} \Rightarrow \boldsymbol{W} \approx \boldsymbol{H}^\dagger\boldsymbol{Y} &&
\end{flalign}
where $\boldsymbol{H}^\dagger$ is the Moore–Penrose pseudo-inverse matrix~\cite{penrose1955generalized} of $\boldsymbol{H}$ and is computed as:
\begin{flalign}\label{eq:inverse}
    \boldsymbol{H}^\dagger = \lim\limits_{\alpha \to 0^+}(\boldsymbol{H}^T\boldsymbol{H} + \alpha \boldsymbol{I})^{-1} \boldsymbol{H}^T &&
\end{flalign}

\textbf{End-to-end optimization}: After initializing the RBF weights and cluster centers with clustering algorithms such as k-means, it is possible to optimize the network end-to-end via backpropagation and gradient descent. Schwenker et al. computed the gradients of the loss function for a Gaussian activation function in~\cite{schwenker2001three}.
\newpage
\section{Adapting RBFs for CNNs}
\label{chap:rbf_sec:CNN-RBF}
\vspace{-0.2cm}
This section presents the adaptation steps for using RBF classifiers for CNNs as depicted in Figure~\ref{fig:CNN-RBF}. The deep embeddings of the CNNs, computed using standard convolutional layers and inception blocks, are flattened and fed to RBFs after a fully connected layer in the architecture. The network ends in an output layer with softmax activation and is optimized end-to-end. Integrating the RBFs into deep structures and using them in conjunction with CNNs presents three challenges:

\textbf{Initialization}: Training the RBFs from scratch with randomly initialized weights using gradient descent is quite inefficient due to inconvenient initial cluster centers. The large initial distances in high dimensional spaces lead to small activation values, and the gradients attenuate considerably after the RBF hidden layer during backpropagation. Therefore, the k-means algorithm initializes the cluster centers before starting the training. Furthermore, computing the weights from Equation~\ref{eq:computeWeight} is not feasible due to the scale of computer vision datasets such as ImageNet~\cite{deng2009imagenet}, which has over $14$ million images and $1000$ classes. Hence, using gradient descent and optimizing randomly initialized output layer weight is the optimal way to proceed.

\textbf{Dynamic input features}: The input features of classical RBFs are fixed, but this assumption is not valid with respect to CNNs. As the embeddings of CNNs develop during the training process, the cluster centers initialized by the k-means algorithm are no longer optimal after a few epochs of training. This research work proposes the optimization of the k-means algorithm's target with unsupervised loss during the training process as defined in Equation~\ref{eq:clus}.

\textbf{Activation}: The nonlinear computational graph drawn by computing the distance in Equation~\ref{eq:distance} and applying the activations in equations~\ref{eq:kernels1}-\ref{eq:kernels7} leads to inefficient gradient flow. Therefore, this research attempts to build a linear computational graph in RBFs through the introduction of a new activation function.

This section presents two modifications to classical RBFs to make them suitable for deep CNNs. First, it introduces an additional loss term to the RBFs' hidden layer. This term is based on the target function of the k-means algorithm defined in Equation~\ref{eq:clus} and continues in the unsupervised learning process during the development of the embeddings. The second contribution of this section is the introduction of a new quadratic kernel to build a linear computational graph for efficient optimization using backpropagation.

\subsection{Introducing Unsupervised Learning Loss}
The embeddings of CNNs change during the training process, which necessitates updating the cluster centers with an unsupervised loss. Therefore, introducing an additional term to the RBFs' supervised loss function to optimize the cluster centers during training using the k-means unsupervised loss in Equation~\ref{eq:clus} is crucial for optimizing CNNs with an RBFs classifier (CNN-RBFs) end-to-end. The final loss of a CNN-RBF network is computed as follows:

\begin{flalign} \label{eq:LossRBF}
    \text{Loss}_\text{rbf} = \text{Loss}_\text{supervised}+\lambda\text{Loss}_\text{unsupervised}  &&
\end{flalign}
where the classification loss $\text{Loss}_\text{classification}$ is any arbitrary loss function, for instance categorical cross-entropy.

It is conventional to use clustering algorithms such as the k-means or expectation-maximization (EM) algorithms to initialize the cluster centers. The loss function in Equation~\ref{eq:LossRBF} is optimized using gradient descent by minimizing the distance of the embeddings for each sample from its nearest cluster center regardless of the class labels. The distance from the nearest cluster center is computed using the distance metric $\boldsymbol{R_j}$ defined in Equation \ref{eq:distance}.

\subsection{Quadratic Kernel}
The kernels used for classical RBFs are nonlinear and increase the model's complexity. The architectures proposed in Figure~\ref{fig:CNN-RBF} profit from using the state-of-the-art models for representation learning, i.e., CNNs, as a backbone. Therefore, CNN-RBF architectures can be trained with simpler linear models to improve the gradient flow during backpropagation. The proposed quadratic activation function is linear in the space of $r^2$ and is defined as follows:
\begin{flalign}\label{eq:kernel_propose}
	h(r) = 1-r^2/\sigma^2  &&
\end{flalign}
where $\sigma$ is the trainable parameter that determines the width of the kernel. The proposed kernel is depicted in Figure~\ref{fig:activation} alongside the conventional activation functions. The proposed quadratic kernel reduces the nonlinearity of the CNN-RBF computational graph for backpropagation. The squares of the distances between cluster centers and samples are computed by linear matrix multiplication in Equation~\ref{eq:distance} and the application of the proposed linear activation for $r^2$. Thus, the gradients of deep embeddings propagate backward through a distance computation with matrix multiplication and linear activations.
\section{Experimental Results}
\label{chap:rbf_sec:experimental}
This section presents the experimental results that reinforce the applicability of RBFs to CNNs on several standard computer vision benchmark datasets and investigates the effect of tweaking various hyperparameters of the CNN-RBF architectures in the training phase and generalization to test data. Three convolutional backbones EfficientNet-B0~\cite{tan2019efficientnet}, InceptionV2~\cite{szegedy2016rethinking}, and ResNet50~\cite{he2016deep} compute the embedding of CNN-RBFs in this section. A list of the benchmark computer vision datasets used in this section is presented in Table \ref{table:datasets}.

\begin{table}[!htb]
    \centering
    \resizebox{0.55\textwidth}{!}{
    \begin{tabular}{@{}lrrr@{}}
        \toprule
        Dataset                                     & Train Size & Test Size & \# Classes \\ \midrule
        CIFAR-10~\cite{krizhevsky2009learning}      & 50 000     & 10 000    & 10         \\
        CIFAR-100~\cite{krizhevsky2009learning}     & 50 000     & 10 000    & 100        \\
        Oxford-IIIT Pets~\cite{parkhi2012cats}      & 3 680      & 3 369     & 37         \\
        Oxford Flowers~\cite{nilsback2008automated} & 1 020      & 6 140     & 102        \\
        FGVC Aircraft~\cite{maji2013fine}           & 6 667      & 3 333     & 100        \\
        Caltech Birds~\cite{WelinderEtal2010}       & 5 996      & 5 794     & 200        \\ \bottomrule
    \end{tabular}}
    \caption{An overview of computer vision benchmark datasets used to evaluate the performance of CNN-RBFs (table adopted from~\cite{amirian2020radial}).}
    \label{table:datasets}
\end{table}

\begin{figure}[htb!]
    \begin{center}
    \begin{tabular}{c }
    \toprule
    \includegraphics[width=0.7\textwidth]{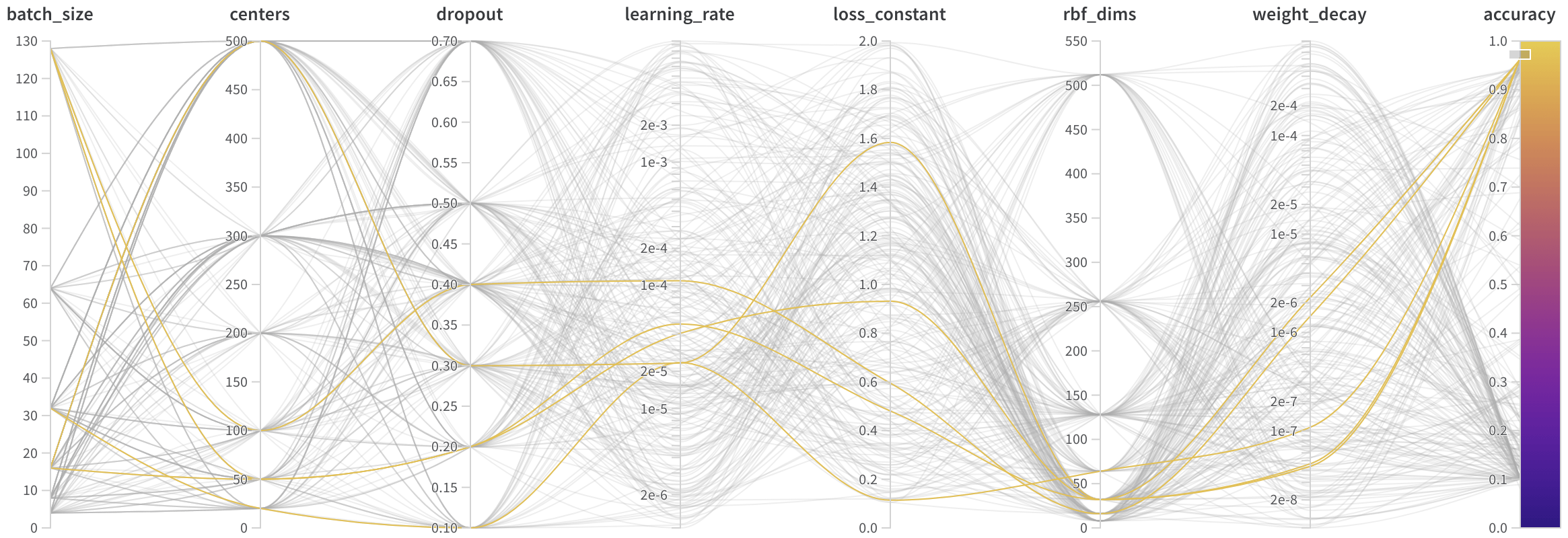}   \\ \midrule
    \includegraphics[width=0.7\textwidth]{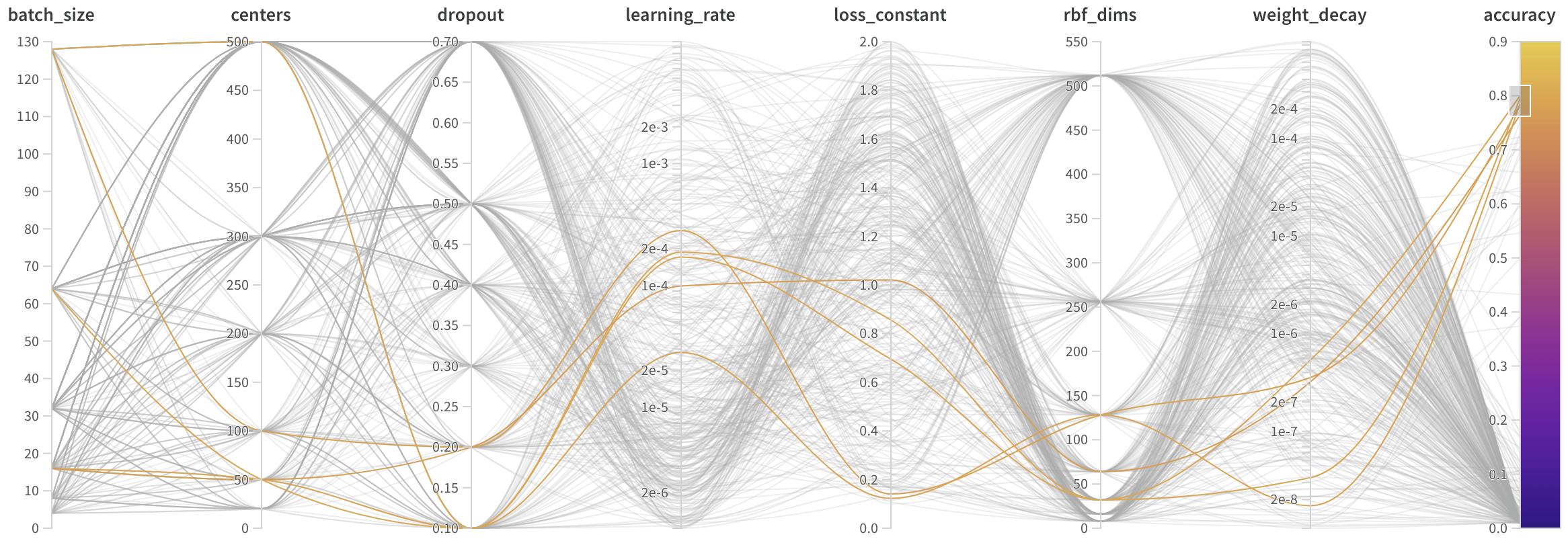} \\ 
    \toprule
    \end{tabular}
    \end{center}
    \caption{Hyperparameter search results from CIFAR-10 (top) and CIFAR-100 (bottom). The top five performing sets of hyperparameters for each dataset are highlighted in yellow (figures adapted from~\cite{amirian2020radial}).}
    \label{fig:hyperparam}
\end{figure}

Figure~\ref{fig:hyperparam} shows the hyperparameter search results for object classification on two benchmark computer vision datasets: CIFAR-10 and CIFAR-100. The backbone CNN model in these experiments is EfficientNet-B0, with a layer of RBFs for classification. The image preprocessing pipeline, called AutoAugment~\cite{cubuk2019autoaugment}, consists of a set of optimal and automatically discovered augmentation policies for the ImageNet~\cite{deng2009imagenet} dataset. The CNN-RBF architecture demonstrated in Figure~\ref{fig:CNN-RBF} has two further hyperparameters: the number of cluster centers and the input dimensions of the RBF network. The models are optimized using an AdamW~\cite{loshchilov2018decoupled} optimizer with different learning rates and weight decay serving as tunable hyperparameters. The other hyperparameters are the loss constant ($\lambda$) from Equation~\ref{eq:LossRBF}, dropout rate, and batch size.

The hyperparameter searches in Figure~\ref{fig:hyperparam} are conducted using the hyperband~\cite{li2017hyperband} algorithm with $4$ agents running in parallel on two \textit{Tesla V100} graphic processing units (GPUs) for approximately $10$ days. It should be noted that dropout is only applied after the CNN backbone and before the fully connected layer in Figure~\ref{fig:CNN-RBF}. The output of the fully connected layer, without any activation function, is used as input features of the RBFs. The results in Figure~\ref{fig:hyperparam} show that training CNN-RBF architectures leads to high performances, even with a wide range of hyperparameters. However, achieving good test performance with a high dropout rate and a large input dimension is challenging. CNN-RBF architectures show a better performance without dropout and rectified linear unit (ReLU) activations in the input layer of RBFs. Thus, dropout is neglected for further hyperparameter searches conducted on the other datasets in Table~\ref{table:datasets}. The list of optimal hyperparameters for all datasets is presented in Table~\ref{table:list_of_hyperparameters}. 

\begin{table}[htb!]
    \begin{center}
    \resizebox{0.8\textwidth}{!}{
    \begin{tabular}{l | c c c c c c}
     \toprule
      & Loss & Learning & Embeddings & Batch & Number of & Weight \\
     Dataset & constant & rate & dimensions & size & centers & decay \\
     \midrule
     CIFAR-10           & 0.1141 &  2.355e-5  &  64  &  32  &  20  &  1.090e-7 \\
     CIFAR-100          & 0.8557 &  1.873e-4  &  32  &  64  &  50  &  5.369e-7 \\
     Oxford-IIIT Pets   & 1.067  &  7.487e-5  &  64  &  16  &  50  &  1.150e-7 \\
     Oxford Flowers     & 1.562  &  1.076e-4  &  16  &  64  & 100  &  3.843e-6  \\     
     FGVC Aircraft      & 0.5471 &  1.103e-4  &  8   &   8  &  50  &  1.222e-6  \\
     Caltech Birds      & 0.5156 &  2.603e-4  &  32  &  32  &  50  &  1.416e-8  \\
    \toprule
    \end{tabular}}
    \end{center}
    \caption{List of the final hyperparameters used for each computer vision benchmark dataset to achieve the performance of CNN-RBF architectures (table adapted from~\cite{amirian2020radial}).}
    \label{table:list_of_hyperparameters}
\end{table}

Several CNN-based backbone architectures are used within the CNN-RBFs to train models for all computer vision datasets. The experimental results of applying the CNN-RBFs to the computer vision benchmark datasets with the standard train and test splits are presented in Table~\ref{table:efficientnet_dataset}. CNN-RBFs show the capacity to learn the entire training dataset in all of the cases. There is, however, a small gap between the best-reported performances in computer vision literature and CNN-RBF architectures. Using dropout with CNN-RBFs for regularization does not lead to desirable results. Reducing the number of parameters of the RBFs while limiting their input size is the best empirically proven regularization strategy for RBFs besides data augmentation. Developing regularization methods for RBFs to improve generalization is an open research topic for reducing the gap between current performances and state-of-the-art computer vision models.

\begin{table}[htb!]
\begin{center}
\resizebox{0.8\textwidth}{!}{  
\begin{tabular}{ l | c | c c c | c }
\toprule
\multirow{2}{*}{Dataset} & & \multicolumn{3}{c|}{CNN-RBFs} & Best \\ 
& Backbone & EfficientNet-B0 & InceptionV2 & ResNet50 & result \\ 
\midrule
\multirow{2}{*}{CIFAR-10} & No-Augment & 0.966 & 0.963 & 0.969 & 0.993   \\
& Auto-Augment  & 0.975 & \textbf{0.977} & 0.942    \\ \midrule

\multirow{2}{*}{CIFAR-100} & No-Augment & 0.797 & 0.752 & 0.693 & 0.936   \\
& Auto-Augment & \textbf{0.822} & 0.805 & 0.778   \\ \midrule

\multirow{2}{*}{Oxford-IIIT Pets} & No-Augment  & 0.840 & 0.804 & 0.622 & 0.967   \\
& Auto-Augment  & \textbf{0.887} & 0.820 & 0.829  \\ \midrule

\multirow{2}{*}{Oxford Flowers} & No-Augment & 0.609 & 0.659 & 0.595 & 0.997   \\
& Auto-Augment  & \textbf{0.828} & 0.757 & 0.667  \\ \midrule

\multirow{2}{*}{FGVC Aircraft} & No-Augment & 0.723 & 0.717 & 0.665 & 0.945  \\
& Auto-Augment & 0.842 & \textbf{0.843} & 0.828 \\ \midrule

\multirow{2}{*}{Caltech Birds} & No-Augment  & 0.613 & 0.428 & 0.281 & 0.904  \\
& Auto-Augment  & \textbf{0.618} & 0.587 & 0.503  	 \\ \midrule

\end{tabular}}
\end{center}
\caption{Comparing the performance of various CNN-RBF architectures with pretraining and augmentation on benchmark computer vision datasets. The best results column is the top performance of the current state-of-the-art architecture on the benchmark dataset (table adapted from~\cite{amirian2020radial}).}
\label{table:efficientnet_dataset}
\end{table}

\section{Visualization of the RBF Classifiers}
This section attempts to visually explain the training and test processes of vision models using RBF classifiers. First, it demonstrates the training process for the simple task of classifying hand-written digits. Then, the distribution and training samples of active clusters for a test sample are depicted for more complicated object detection tasks described in the previous section.

\subsection{Visualization of the Training Process}
\label{chap:rbf_sec:training}
This section visualizes the performance of the RBFs classifiers for CNNs on a simple dataset. The experiments are conducted on the modified national institute of standards and technology (MNIST) dataset~\cite{lecun1998gradient}, a dataset of hand-written digits including ten classes. Learning the dataset is considered a simple task in computer vision. The simplicity of the dataset and learning task allows for the visualization of the training process at a fine level of detail. CNN-RBF architecture has the same number of cluster centers as classes (ten) in the dataset to depict the training process in this experiment. The network architecture in this section consists of a four-layer CNN, and the output of these layers is connected to the RBF after a global average pooling layer and a fully connected layer.

\begin{figure}[htb!]
     \begin{center}
     \resizebox{\textwidth}{!}{
     \begin{tabular}{c c c c c}
     \toprule
     
     \includegraphics[width=0.175\linewidth]{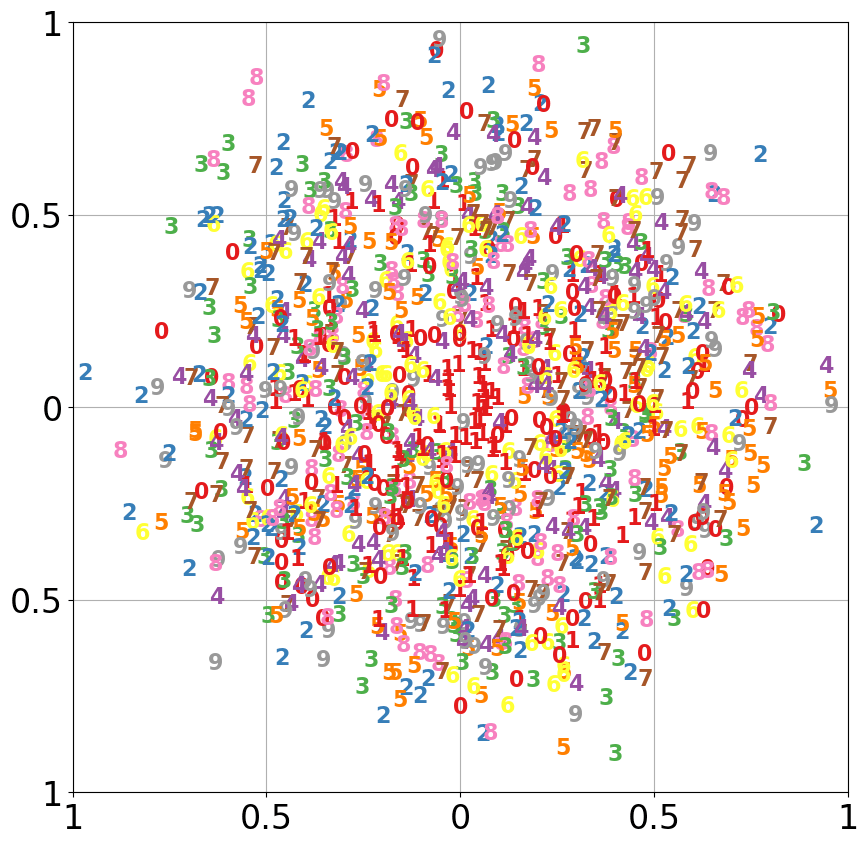} & \includegraphics[width=0.175\linewidth]{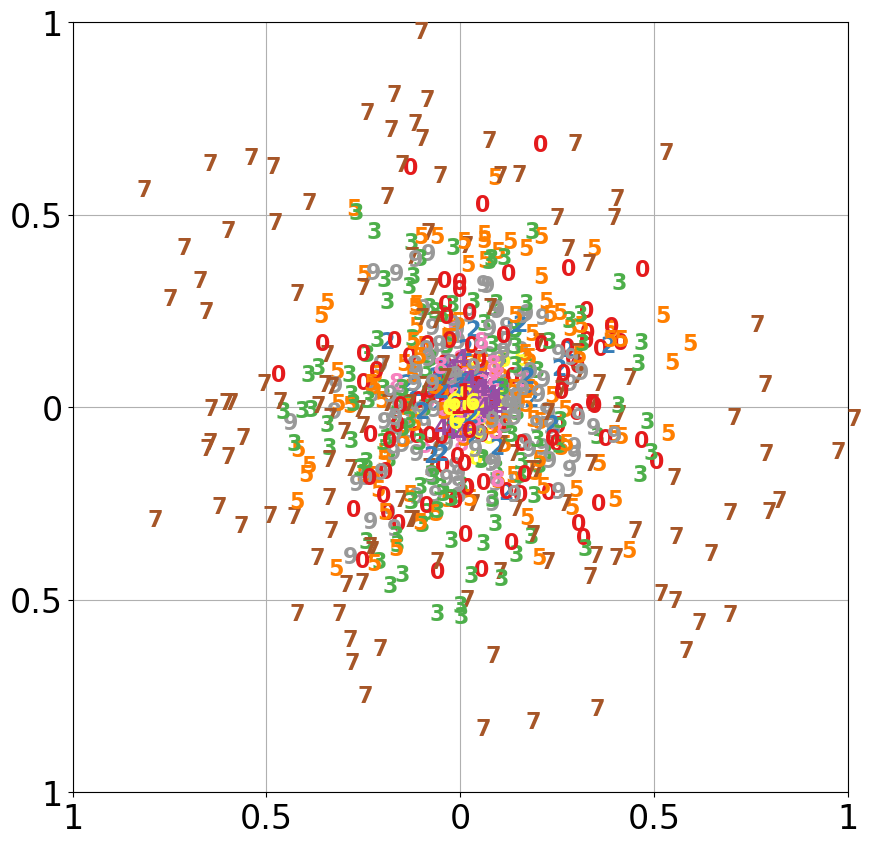} & \includegraphics[width=0.175\linewidth]{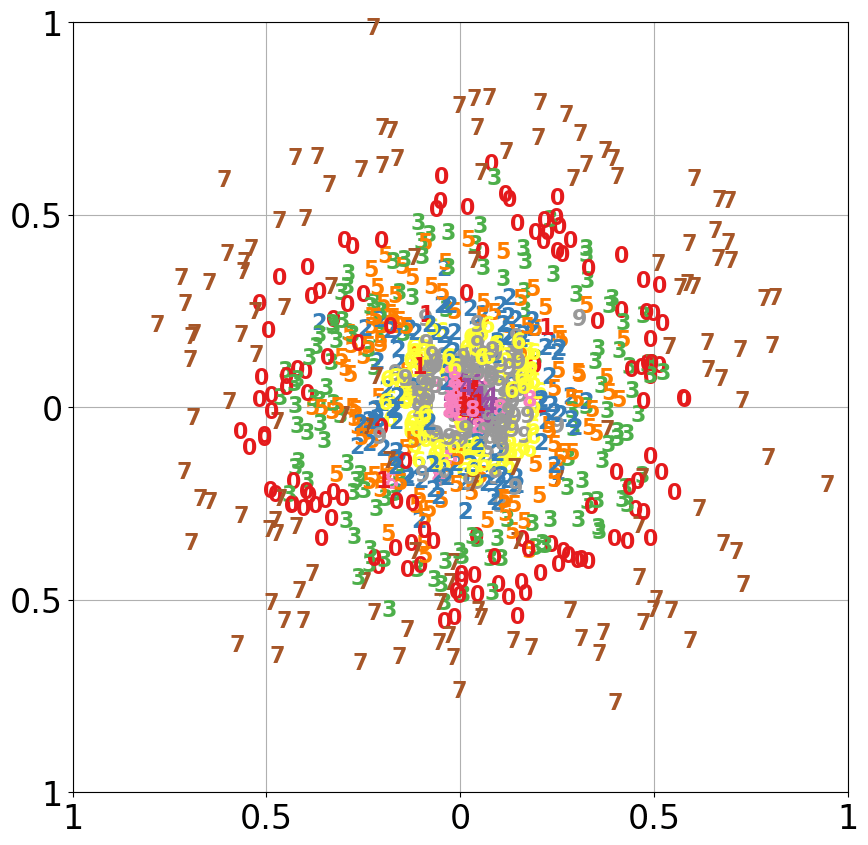} & \includegraphics[width=0.175\linewidth]{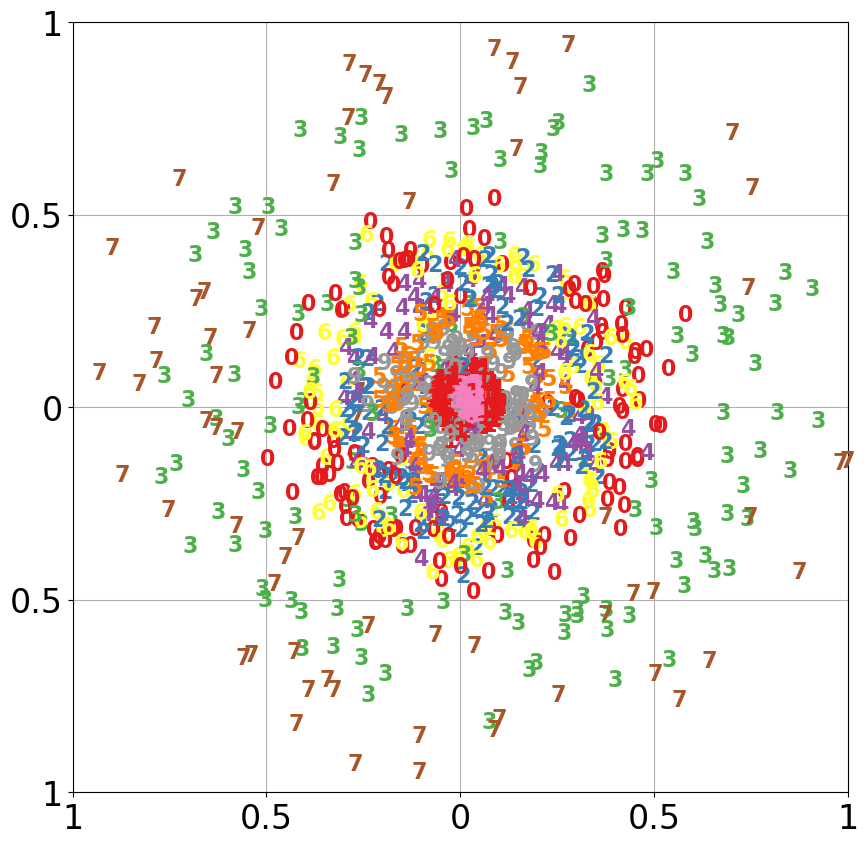} & \includegraphics[width=0.175\linewidth]{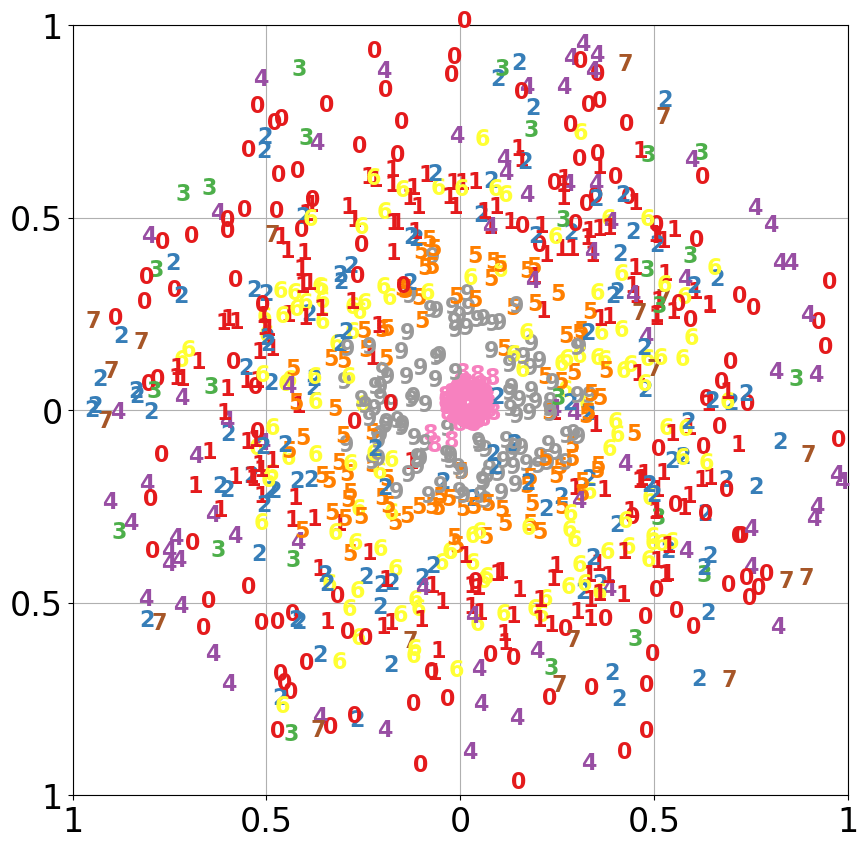} \\ \midrule
      
      \includegraphics[width=0.175\linewidth]{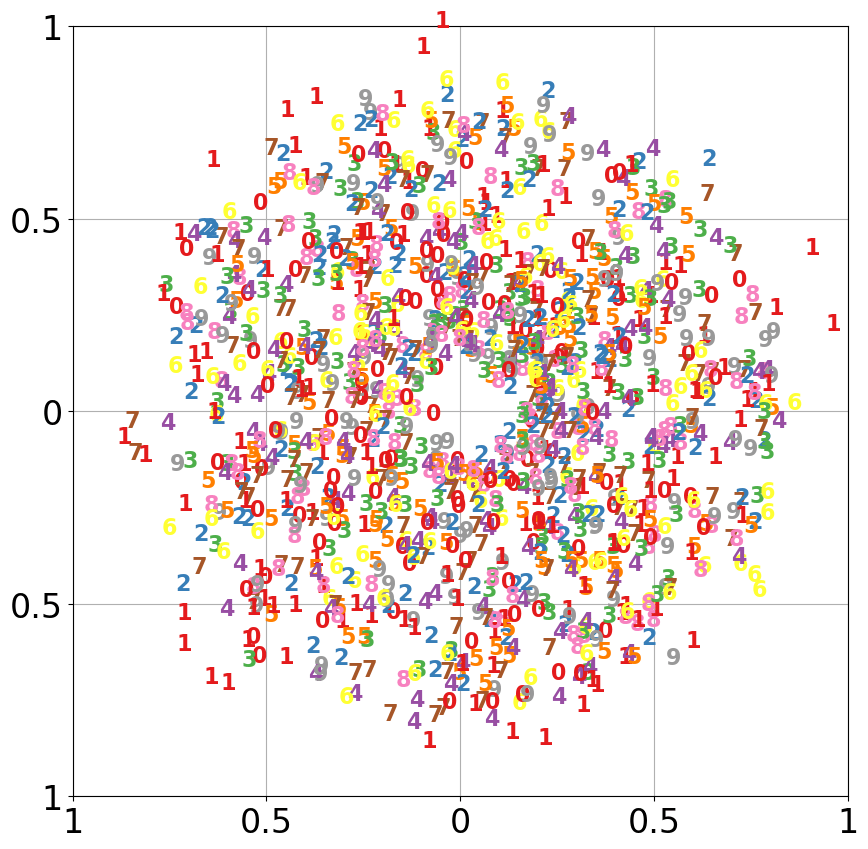} & \includegraphics[width=0.175\linewidth]{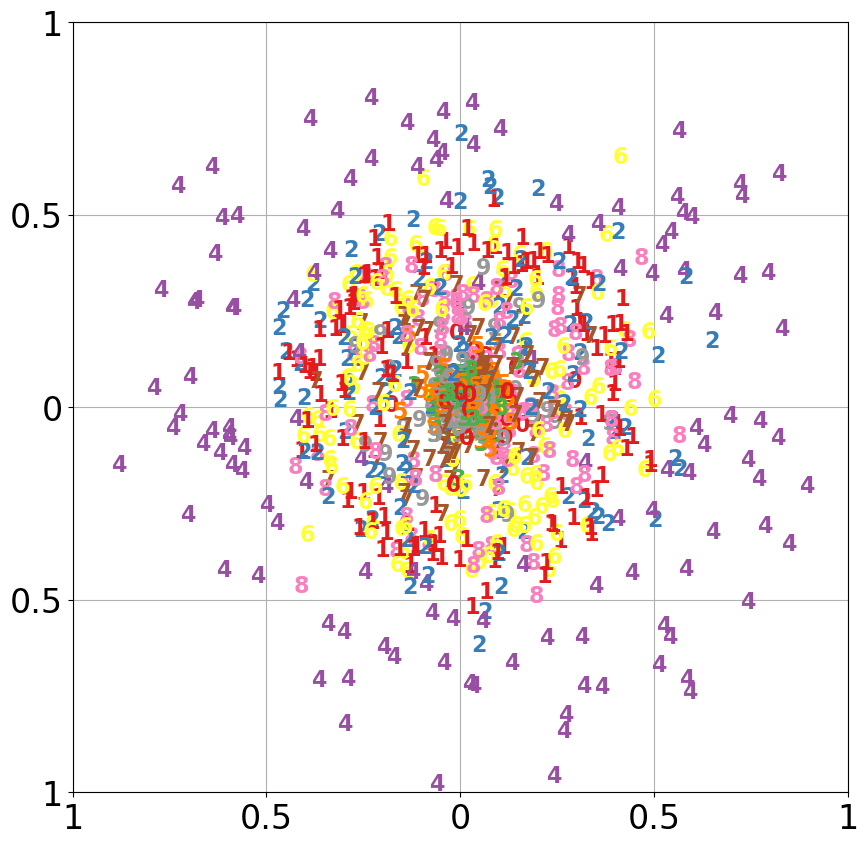} & \includegraphics[width=0.175\linewidth]{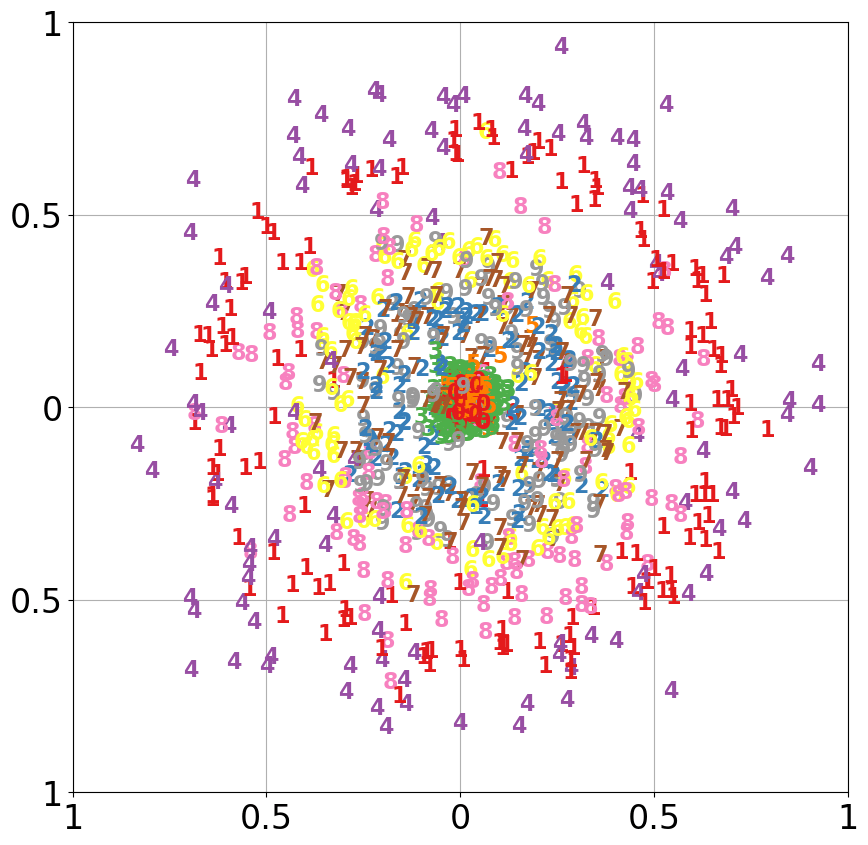} & \includegraphics[width=0.175\linewidth]{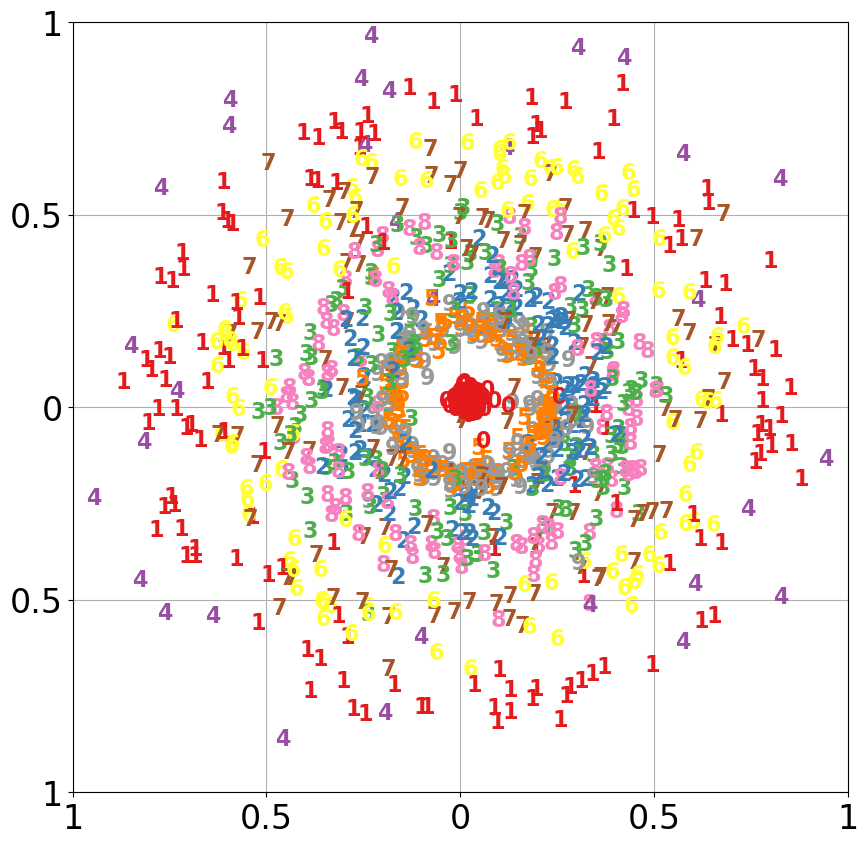} & \includegraphics[width=0.175\linewidth]{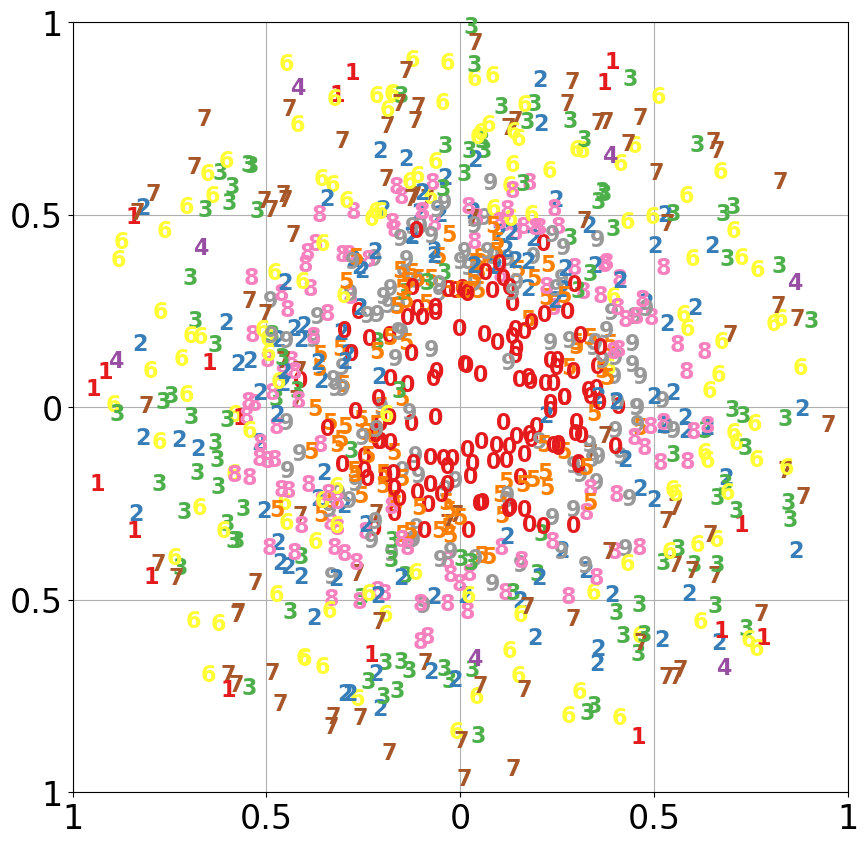} \\ \midrule 
      
      Before training & $50$\% of first epoch & First epoch & Third epoch & Fifth epoch  \\
      Test accuracy: $8.9$\% & Test accuracy: $20.3$\% & Test accuracy: $77.1$\% & Test accuracy: $97.5$\% & Test accuracy: $98.8$\%  \\
        
     \toprule
     \end{tabular}}
     \end{center}
     \caption{This figure presents the location of data samples compared to the cluster centers during the training process. The centers of the clusters are in the middle of the figures. The training samples are located at a random angle based on their distance from the center of the clusters. The vertical and horizontal axes show the normalized distances (figure adopted from~\cite{amirian2020radial}).}
     \label{fig:single_prototype_in_training}
      
\end{figure}

\begin{figure}[htb!]
     \begin{center}
     \resizebox{\textwidth}{!}{
     \begin{tabular}{c c c c c}
     \toprule
     \includegraphics[width=0.175\linewidth]{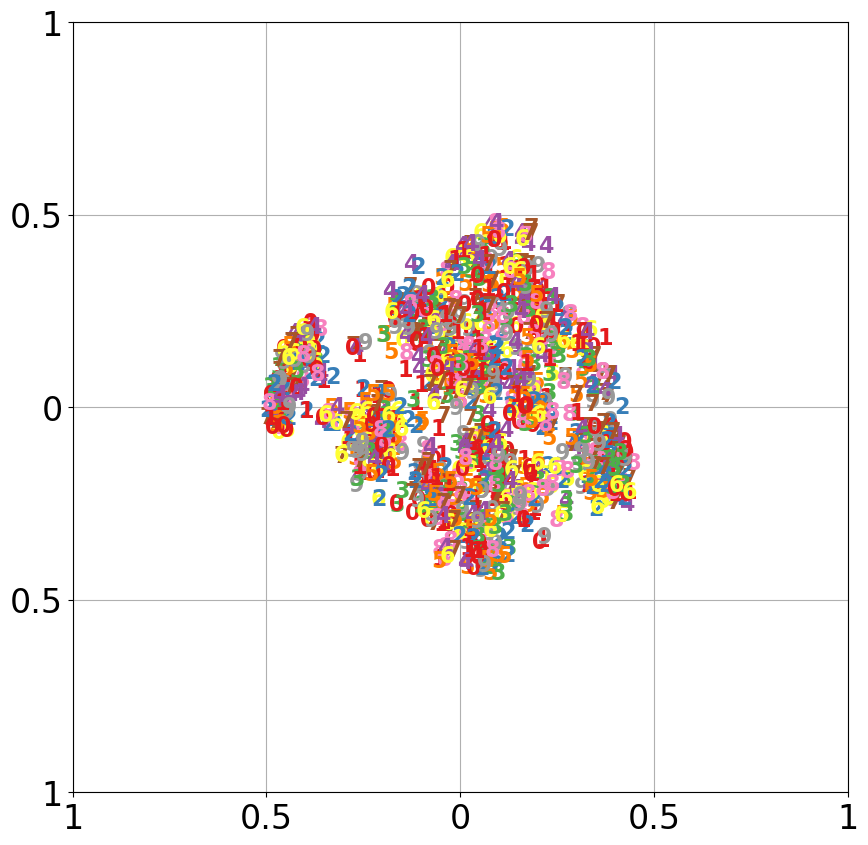} & \includegraphics[width=0.175\linewidth]{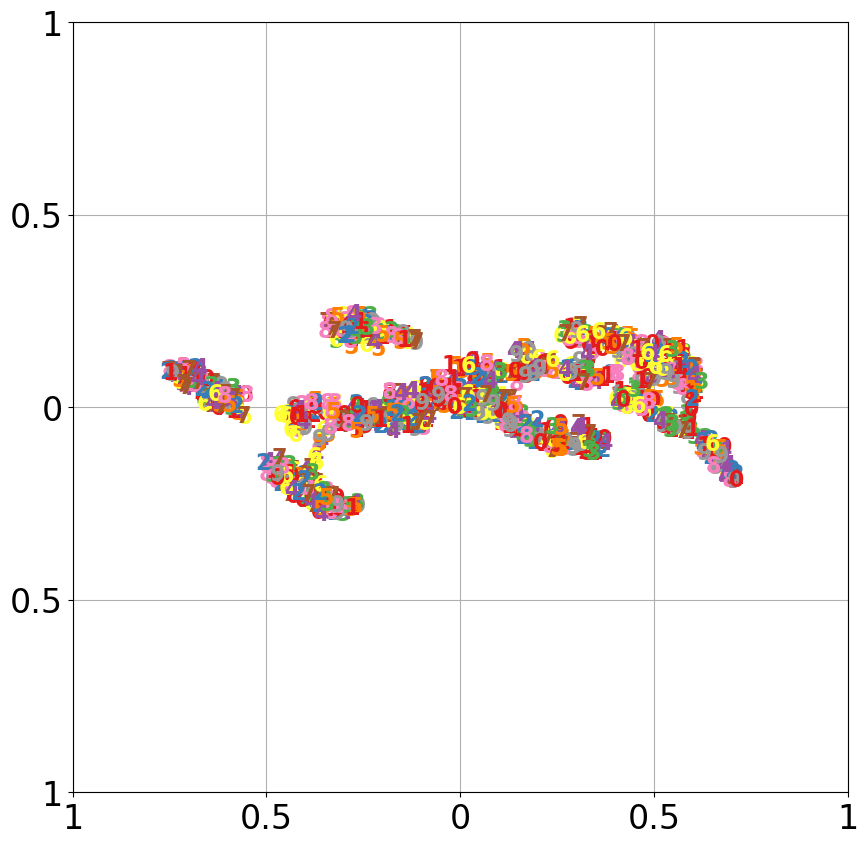} & \includegraphics[width=0.175\linewidth]{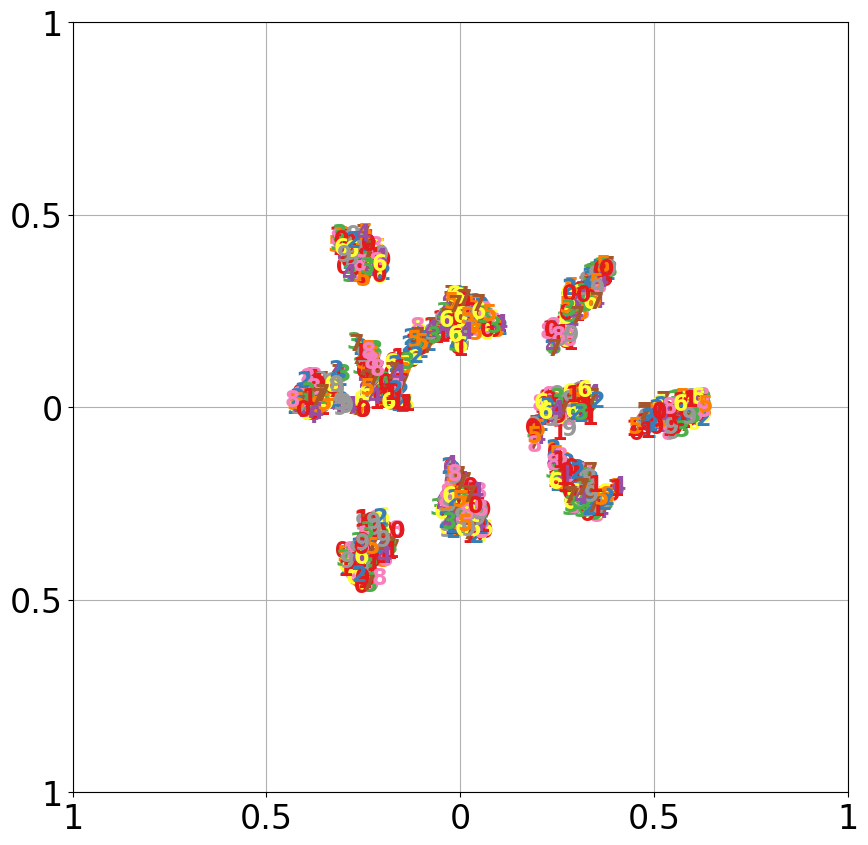} & \includegraphics[width=0.175\linewidth]{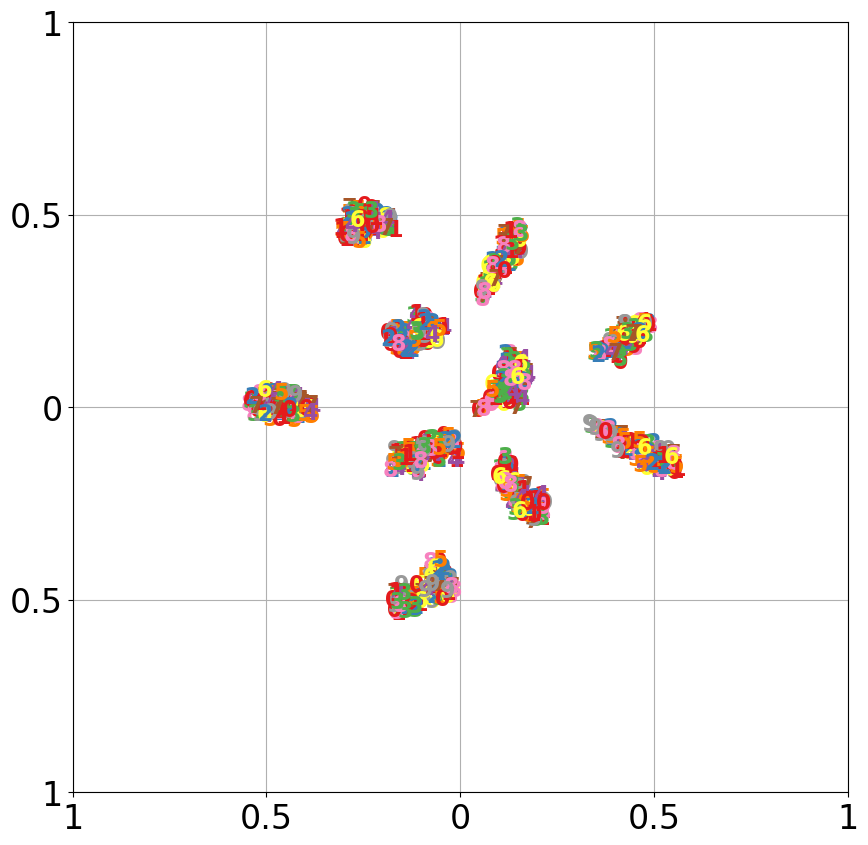} & \includegraphics[width=0.175\linewidth]{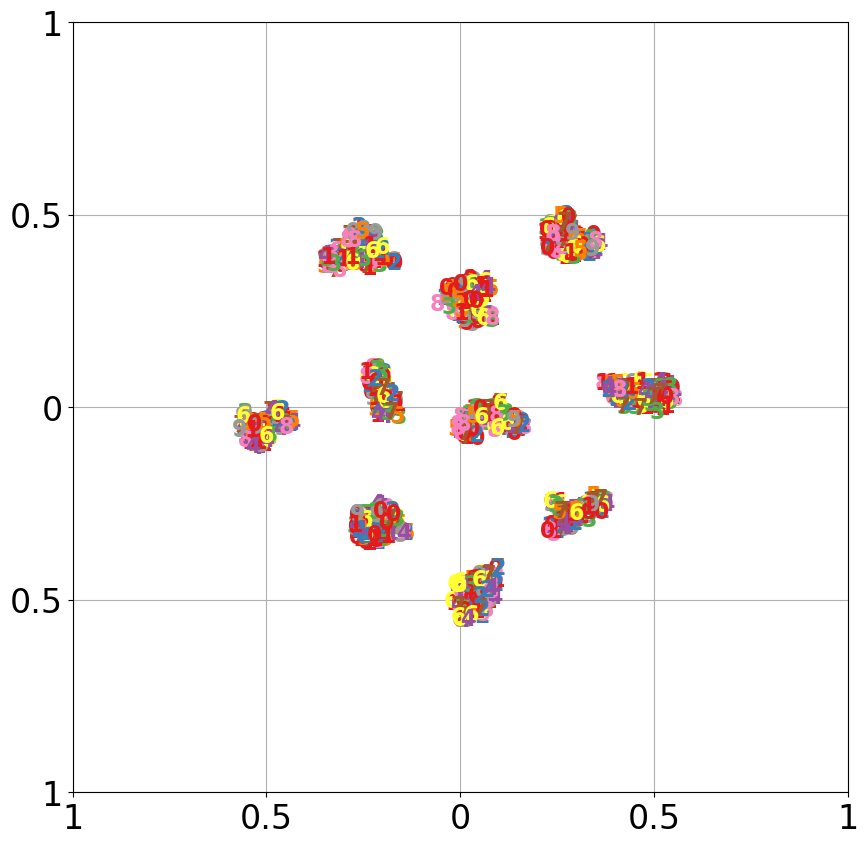} \\ \midrule
     \includegraphics[width=0.175\linewidth]{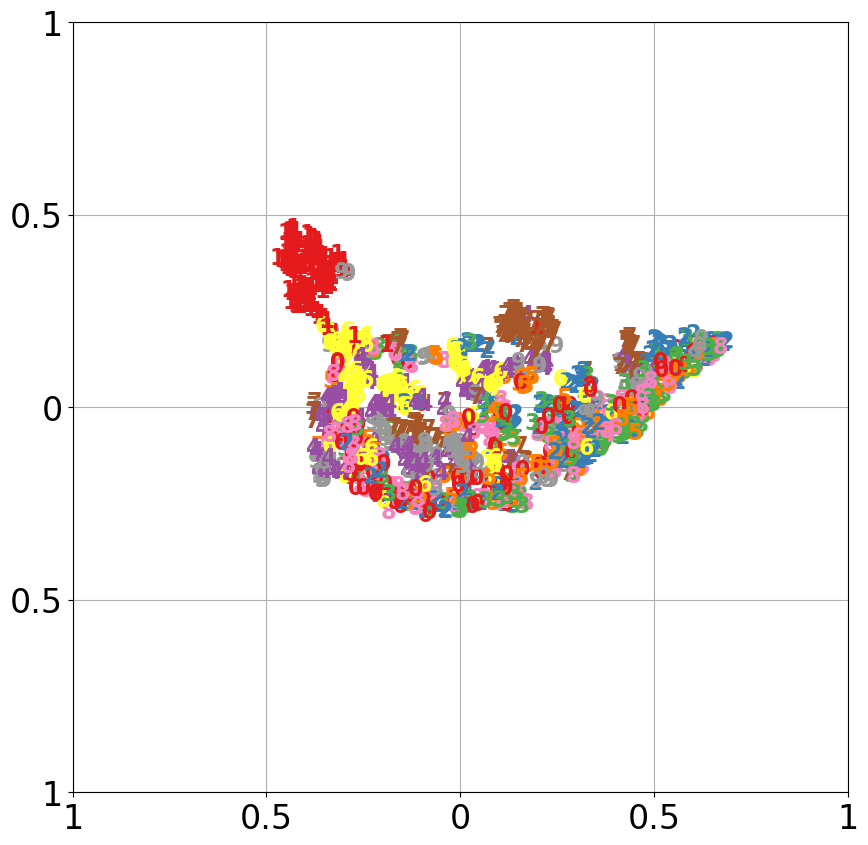} & \includegraphics[width=0.175\linewidth]{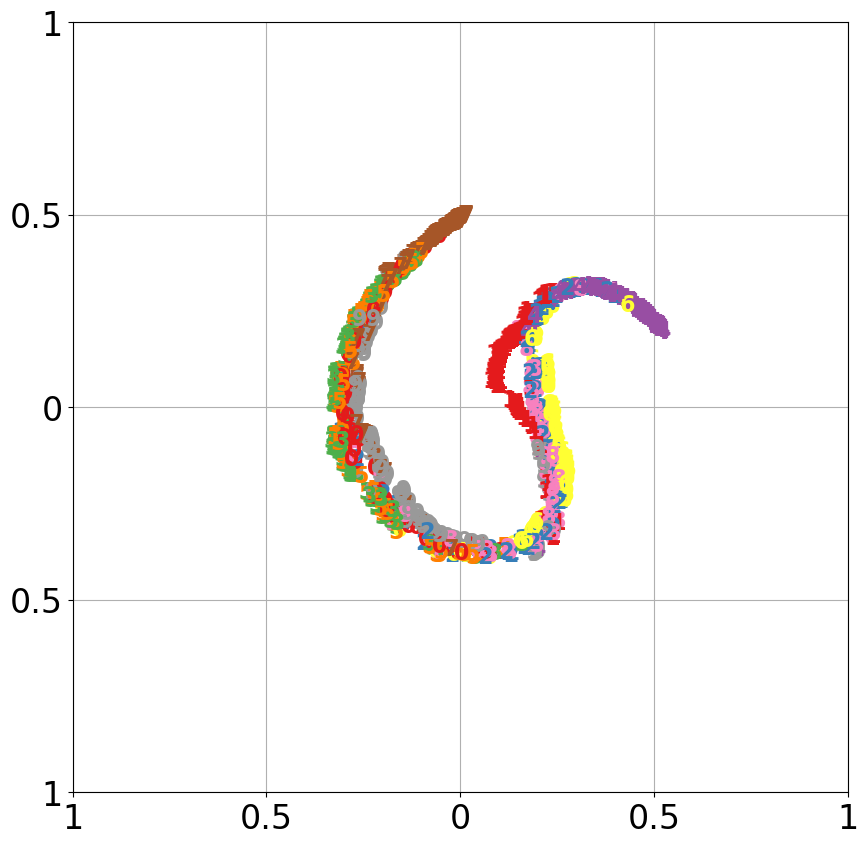} & \includegraphics[width=0.175\linewidth]{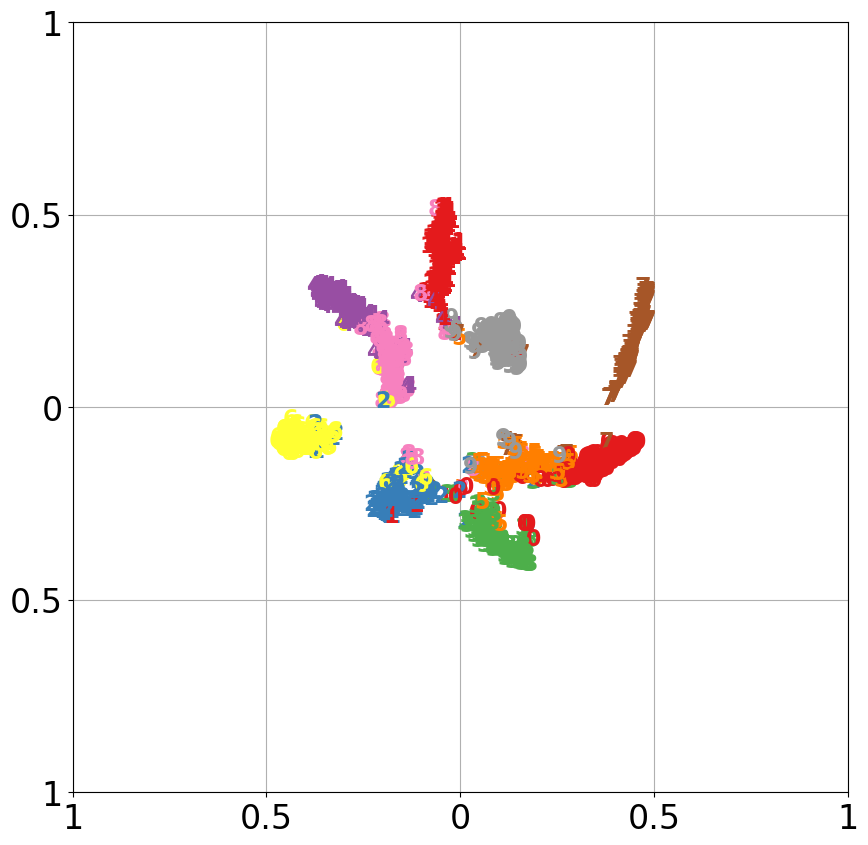} & \includegraphics[width=0.175\linewidth]{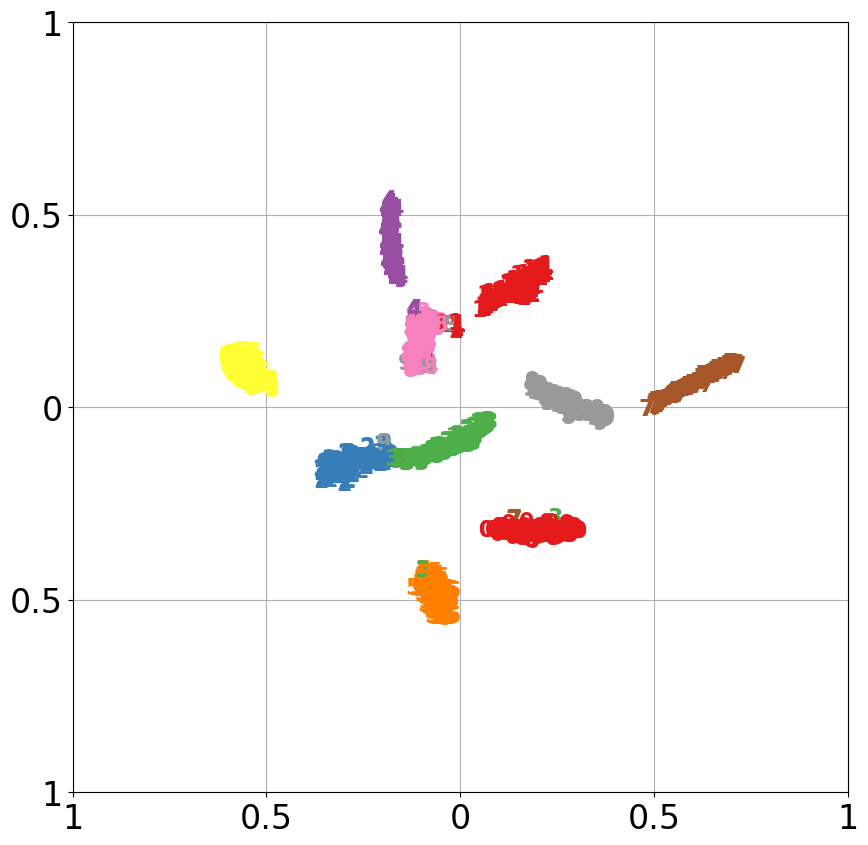} & \includegraphics[width=0.175\linewidth]{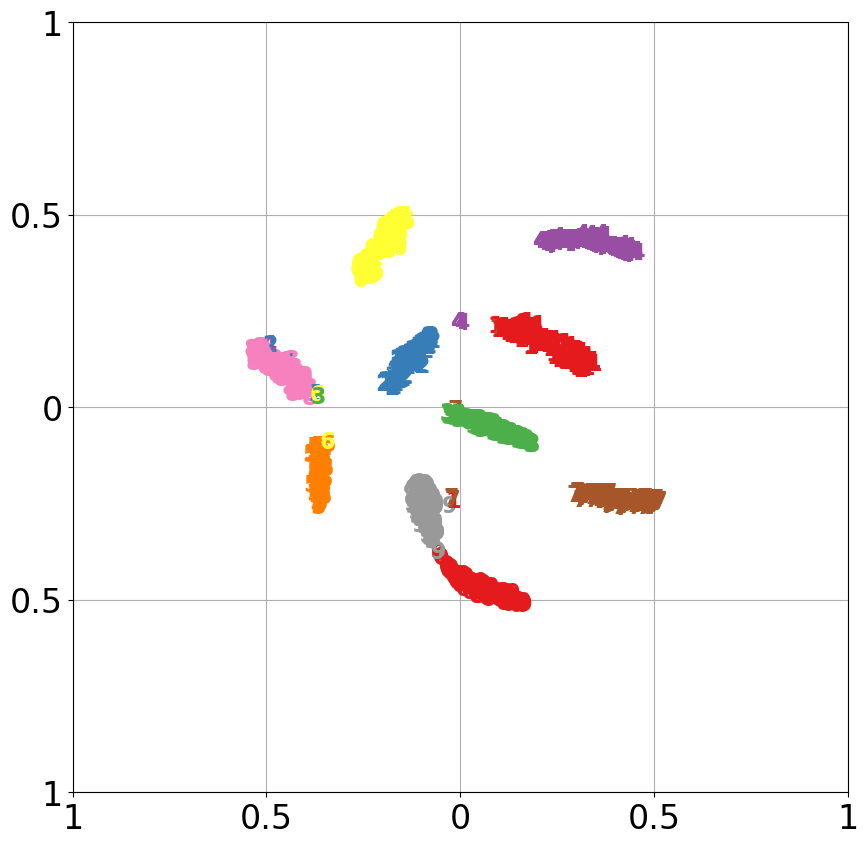} \\ \midrule      
     Before training & 50\% of first epoch & First epoch & Third epoch & Fifth epoch \\
     \toprule
     \end{tabular}}
     \end{center}
     \caption{Two-dimensional representation of the training process: the figure presents the embeddings of the convolutional backbone (top row), and the activations of the RBFs (bottom row) mapped to a two-dimensional space using t-SNE~\cite{van2012visualizing}. The vertical and horizontal axes depict the normalized values; however, all sub-figures use the same normalization factors (figure adopted from~\cite{amirian2020radial}).}
     \label{fig:tsne}
\end{figure}

Figure~\ref{fig:single_prototype_in_training} demonstrates the evolution of the representations around the cluster center during the training process. The data samples in Figure~\ref{fig:single_prototype_in_training} are placed according to their distance from the center and at a random angle. The samples are shown with a number corresponding to their class, and the color is similar for samples of the same class in Figure~\ref{fig:single_prototype_in_training}. To reduce the overlap between close samples, a random uniform noise of amplitude $0.1$ is added to the distance of the samples from the cluster centers.

Minimizing the unsupervised loss in Equation~\ref{eq:clus} reduces the distance of the data samples from the cluster centers. Furthermore, the supervised loss enforces the samples of the same class to maintain the same distance from cluster centers, as the activations are the only information for the network's final decision. The circles with samples of the same class around the cluster centers demonstrate the effect of supervised loss in training. Notably, the clusters presented in Figure~\ref{fig:single_prototype_in_training} are selected to illustrate the concepts underlying training CNN-RBFs optimally.

Figure~\ref{fig:tsne} illustrates the two-dimensional mapping of the CNN embeddings (top row) and RBF activations (bottom row) using t-SNE~\cite{van2012visualizing}. The effect of both supervised and unsupervised loss from Equation~\ref{eq:LossRBF} is evident in this figure. The data samples split into clusters regardless of their class labels in the embedding space of CNN due to the unsupervised loss (top row in Figure \ref{fig:tsne}). The activation values divide into clusters corresponding to the class labels, a process encouraged by the supervised loss.

\subsection{Similarity Metric Learning and Interpretability}
\label{chap:rbf_sec:interpret}

Using a different approximation strategy compared with fully connected layers provides CNN-RBFs with the chance to probe the decision-making process based on the following visual clues:
\begin{itemize}
    \item Similar images as measured by the similarity distance metric of RBFs trained on CNN embeddings
    \item Clusters visualization with a higher contribution to the network's decision and distance of the samples from the centers of these clusters
\end{itemize}

\begin{figure}[H]
    \centering
     \resizebox{0.87\textwidth}{!}{ \begin{tabular}{c | c c c c c c c}
 \toprule
 Test image & \multicolumn{7}{c}{Similar and dissimilar images} \\
 \midrule
 \includegraphics[width=2cm]{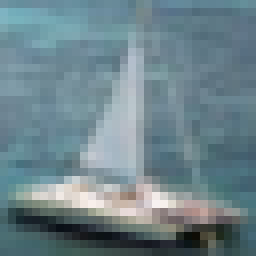} &    
 \includegraphics[width=2cm]{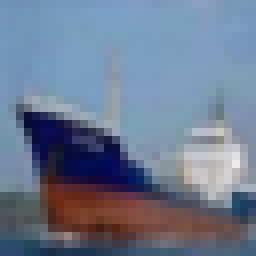} &
 \includegraphics[width=2cm]{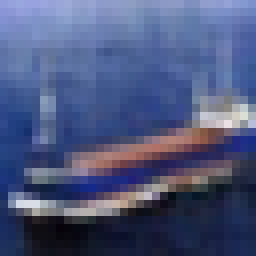} &
 \includegraphics[width=2cm]{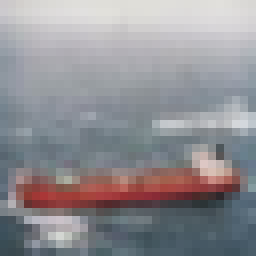} &
 \includegraphics[width=2cm]{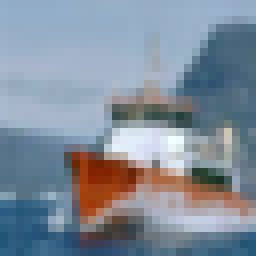} &
 \includegraphics[width=2cm]{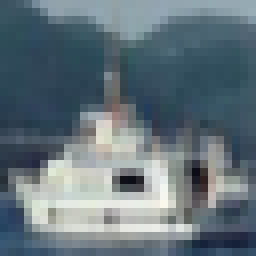} &
 \includegraphics[width=2cm]{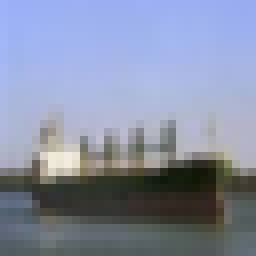} &
 \includegraphics[width=2cm]{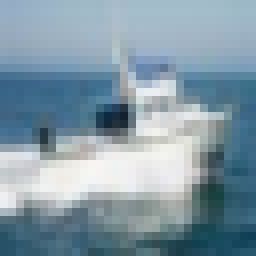} \\
 \midrule
 \includegraphics[width=2cm]{figures/rbf_for_cnn/interpret_metric/metric/test_sample_1.png} &    
 \includegraphics[width=2cm]{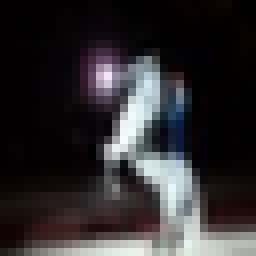} &
 \includegraphics[width=2cm]{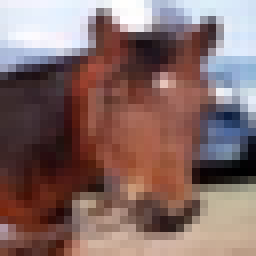} &
 \includegraphics[width=2cm]{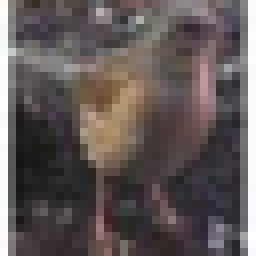} &
 \includegraphics[width=2cm]{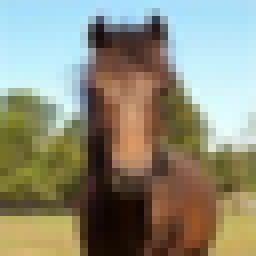} &
 \includegraphics[width=2cm]{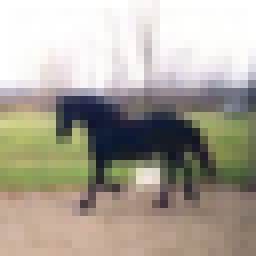} &
 \includegraphics[width=2cm]{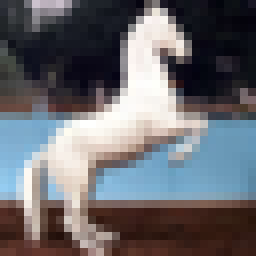} &
 \includegraphics[width=2cm]{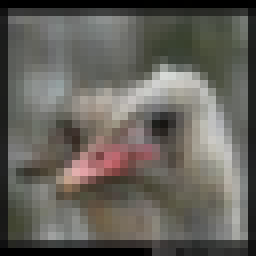} \\
 
 \toprule
 \includegraphics[width=2cm]{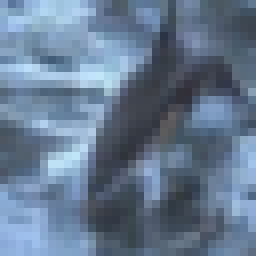} &    
 \includegraphics[width=2cm]{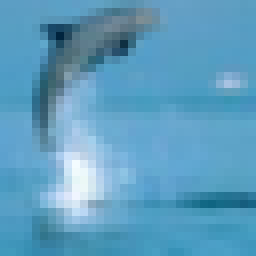} &
 \includegraphics[width=2cm]{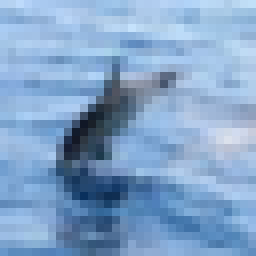} &
 \includegraphics[width=2cm]{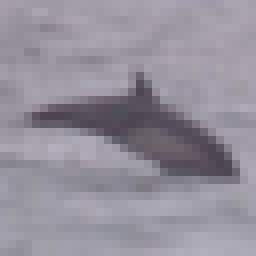} &
 \includegraphics[width=2cm]{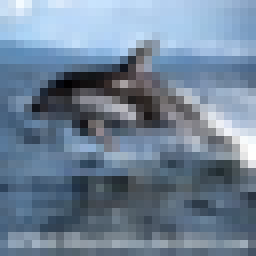} &
 \includegraphics[width=2cm]{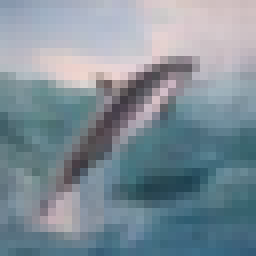} &
 \includegraphics[width=2cm]{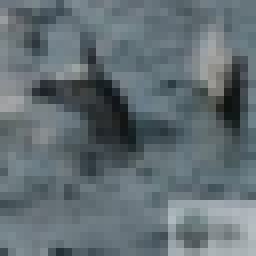} &
 \includegraphics[width=2cm]{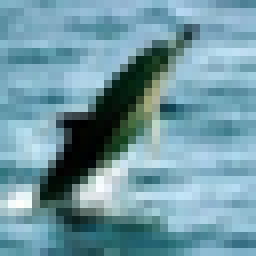} \\
 \midrule
\includegraphics[width=2cm]{figures/rbf_for_cnn/interpret_metric/metric/test_sample_2.png} &    
 \includegraphics[width=2cm]{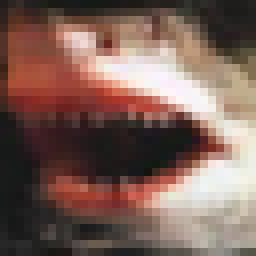} &
 \includegraphics[width=2cm]{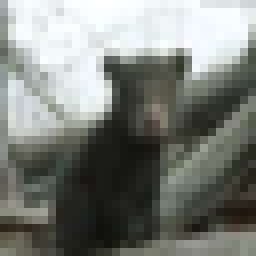} &
 \includegraphics[width=2cm]{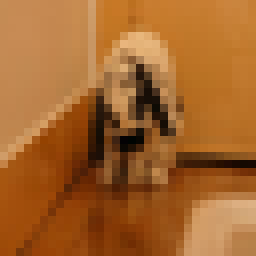} &
 \includegraphics[width=2cm]{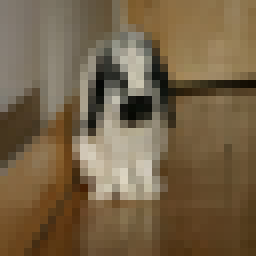} &
 \includegraphics[width=2cm]{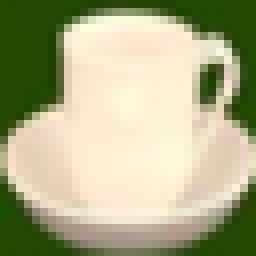} &
 \includegraphics[width=2cm]{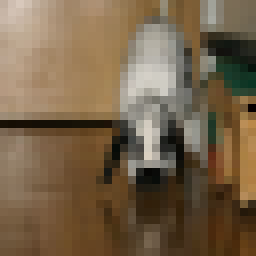} &
 \includegraphics[width=2cm]{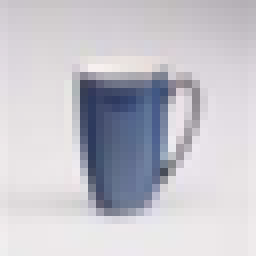} \\
 
 \toprule
 \includegraphics[width=2cm]{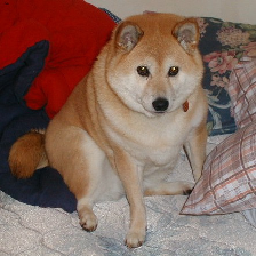} &    
 \includegraphics[width=2cm]{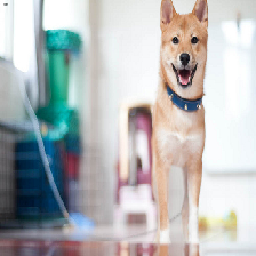} &
 \includegraphics[width=2cm]{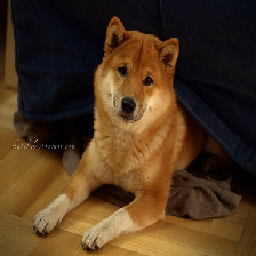} &
 \includegraphics[width=2cm]{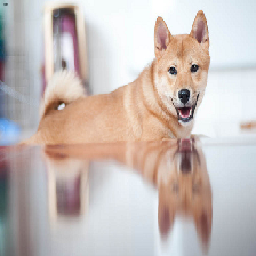} &
 \includegraphics[width=2cm]{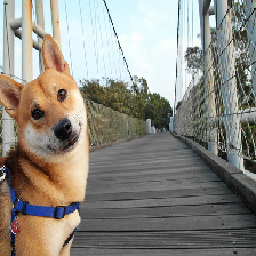} &
 \includegraphics[width=2cm]{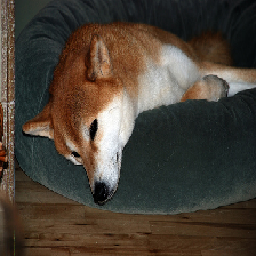} &
 \includegraphics[width=2cm]{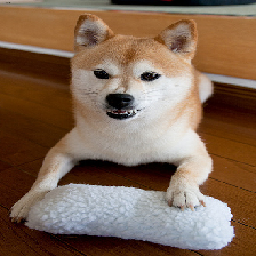} &
 \includegraphics[width=2cm]{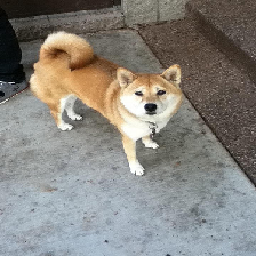} \\
 \midrule
\includegraphics[width=2cm]{figures/rbf_for_cnn/interpret_metric/metric/test_sample_3.png} &    
 \includegraphics[width=2cm]{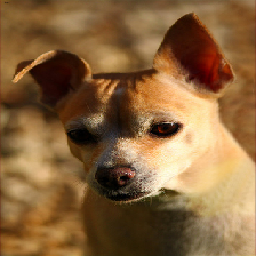} &
 \includegraphics[width=2cm]{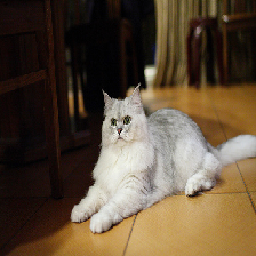} &
 \includegraphics[width=2cm]{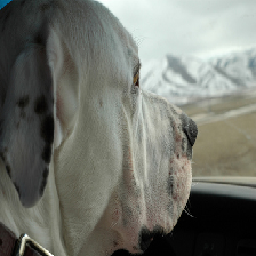} &
 \includegraphics[width=2cm]{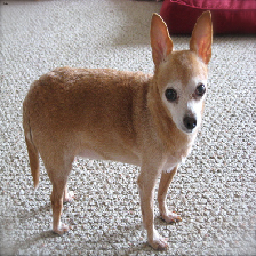} &
 \includegraphics[width=2cm]{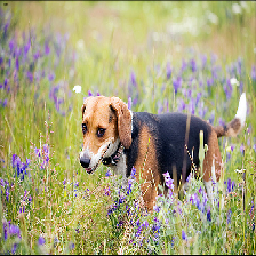} &
 \includegraphics[width=2cm]{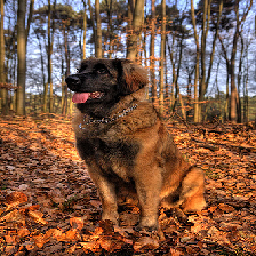} &
 \includegraphics[width=2cm]{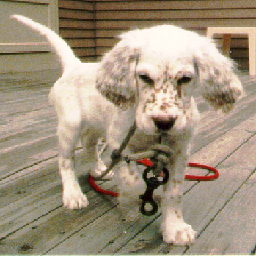} \\
 
 \toprule
 \includegraphics[width=2cm]{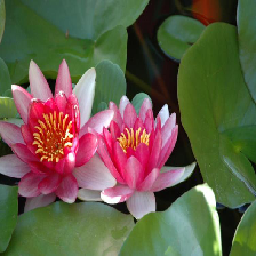} &    
 \includegraphics[width=2cm]{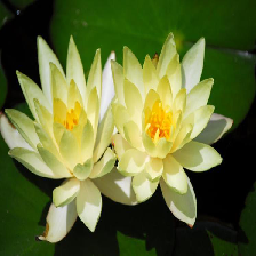} &
 \includegraphics[width=2cm]{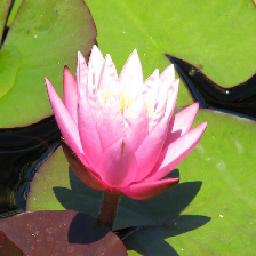} &
 \includegraphics[width=2cm]{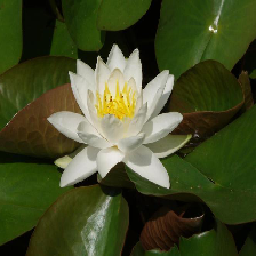} &
 \includegraphics[width=2cm]{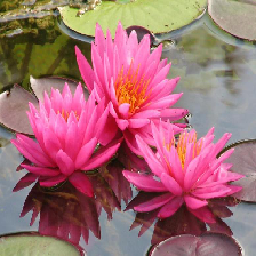} &
 \includegraphics[width=2cm]{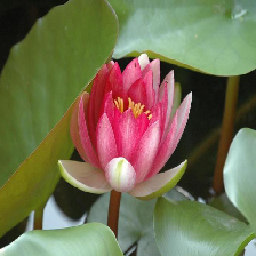} &
 \includegraphics[width=2cm]{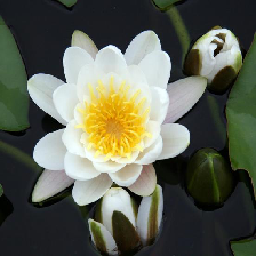} &
 \includegraphics[width=2cm]{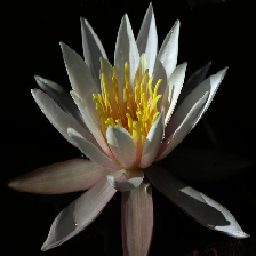} \\
 \midrule
\includegraphics[width=2cm]{figures/rbf_for_cnn/interpret_metric/metric/test_sample_4.png} &    
 \includegraphics[width=2cm]{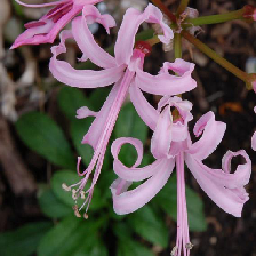} &
 \includegraphics[width=2cm]{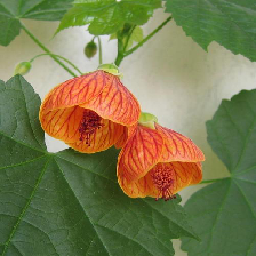} &
 \includegraphics[width=2cm]{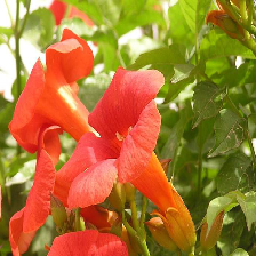} &
 \includegraphics[width=2cm]{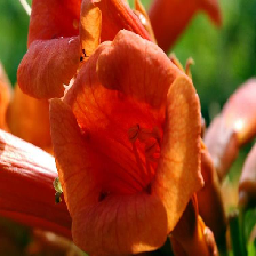} &
 \includegraphics[width=2cm]{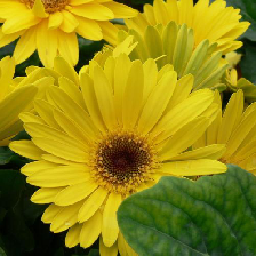} &
 \includegraphics[width=2cm]{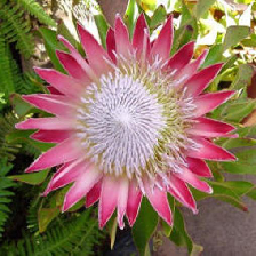} &
 \includegraphics[width=2cm]{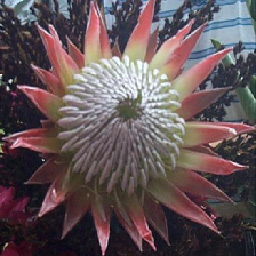} \\
 
 \toprule
 \includegraphics[width=2cm]{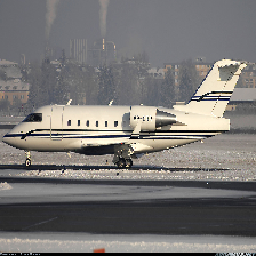} &    
 \includegraphics[width=2cm]{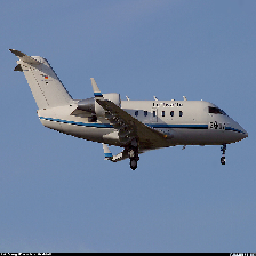} &
 \includegraphics[width=2cm]{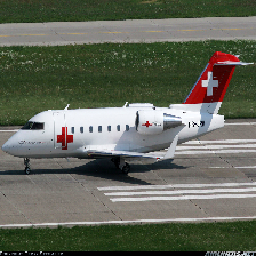} &
 \includegraphics[width=2cm]{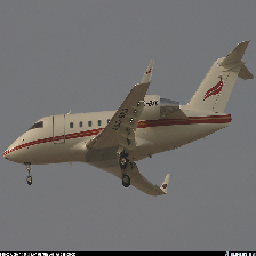} &
 \includegraphics[width=2cm]{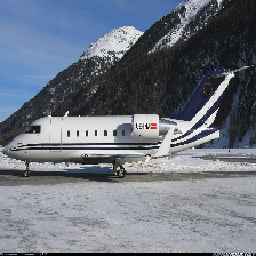} &
 \includegraphics[width=2cm]{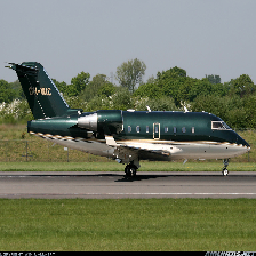} &
 \includegraphics[width=2cm]{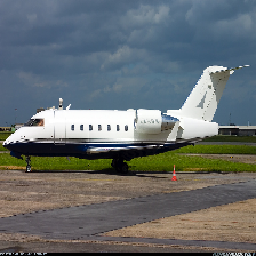} &
 \includegraphics[width=2cm]{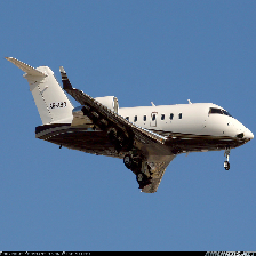} \\
 \midrule
\includegraphics[width=2cm]{figures/rbf_for_cnn/interpret_metric/metric/test_sample_5.png} &    
 \includegraphics[width=2cm]{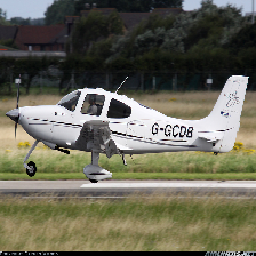} &
 \includegraphics[width=2cm]{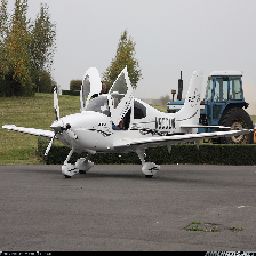} &
 \includegraphics[width=2cm]{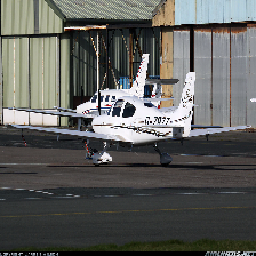} &
 \includegraphics[width=2cm]{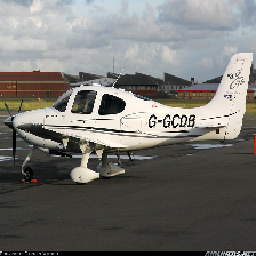} &
 \includegraphics[width=2cm]{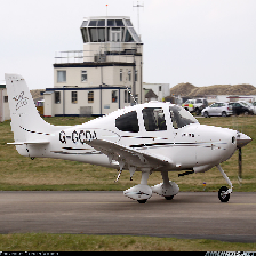} &
 \includegraphics[width=2cm]{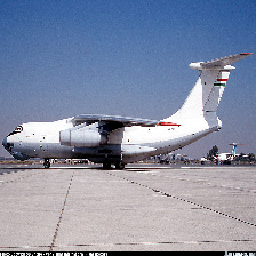} &
 \includegraphics[width=2cm]{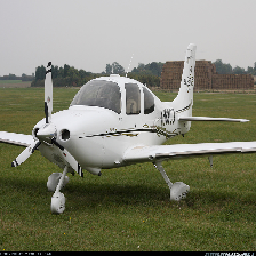} \\
 
 \toprule
 \includegraphics[width=2cm]{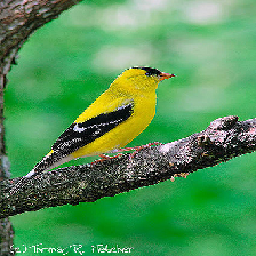} &    
 \includegraphics[width=2cm]{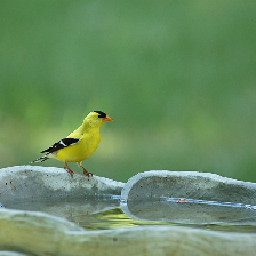} &
 \includegraphics[width=2cm]{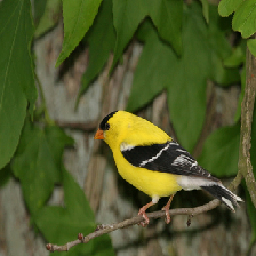} &
 \includegraphics[width=2cm]{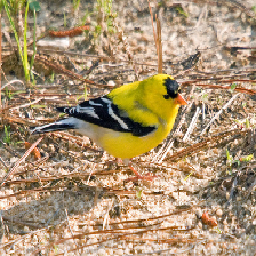} &
 \includegraphics[width=2cm]{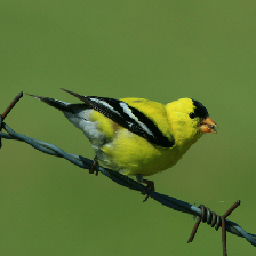} &
 \includegraphics[width=2cm]{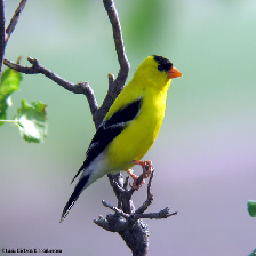} &
 \includegraphics[width=2cm]{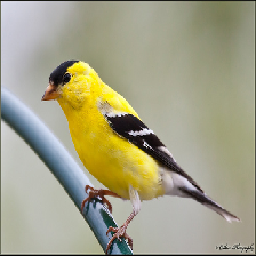} &
 \includegraphics[width=2cm]{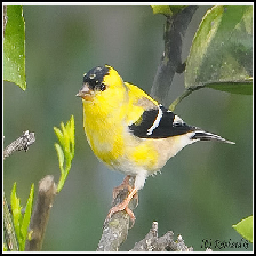} \\
 \midrule
\includegraphics[width=2cm]{figures/rbf_for_cnn/interpret_metric/metric/test_sample_6.png} &    
 \includegraphics[width=2cm]{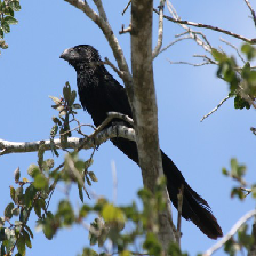} &
 \includegraphics[width=2cm]{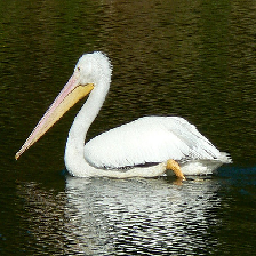} &
 \includegraphics[width=2cm]{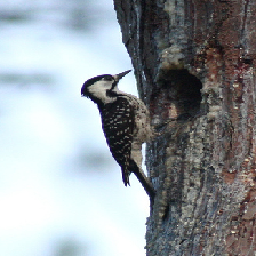} &
 \includegraphics[width=2cm]{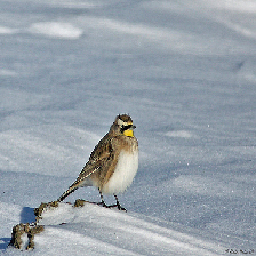} &
 \includegraphics[width=2cm]{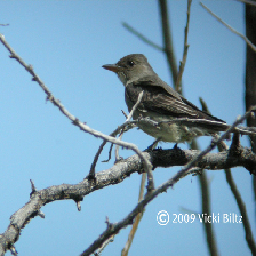} &
 \includegraphics[width=2cm]{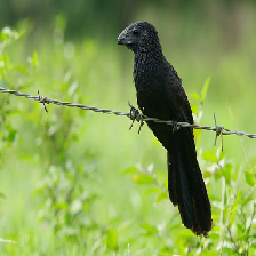} &
 \includegraphics[width=2cm]{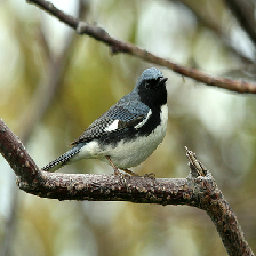} \\
 
 \toprule     
 \end{tabular}

}
     \caption{This figure depicts similar and dissimilar training images for given test images based on the similarity metric computed in Equation~\ref{eq:distance}. The figure depicts the top $7$ most similar and dissimilar training images for a given test image in every two rows. The images shown in every two consecutive rows belong to one of the datasets in Table~\ref{table:datasets} in the same order (figure adopted from~\cite{amirian2020radial}).}
     \label{fig:similar_dissimilar}
\end{figure}

The embeddings of the CNN are compared with the position of the cluster centers using the learned distance metric from the RBFs. The learned distance metric of the RBFs can measure the similarity between a test image and similar images from the training data. Figure~\ref {fig:similar_dissimilar} shows the similar images found in the training dataset for a given test sample as determined by the similarity distance metric in Equation~\ref{eq:distance}. The most similar and dissimilar images are computed using the following criteria:


\vspace{-0.5cm}
\begin{flalign}
&\boldsymbol{x}_{similar} = \argmin_{\boldsymbol{x}_{train}} \parallel \boldsymbol{x}^\mu_{train}-\boldsymbol{x}_{test} \parallel^2_{\boldsymbol{R}}  && \label{eq:similar} \\
&\boldsymbol{x}_{dissimilar} = \argmax_{\boldsymbol{x}_{train}} \parallel \boldsymbol{x}^\mu_{train}-\boldsymbol{x}_{test} \parallel^2_{\boldsymbol{R}} && \label{eq:dissimliar}
\end{flalign} 

where $x_{test}$ presents the input of RBFs for a given test image, $x^\mu_{train}$ shows the input vector for training samples, and $\mu$ enumerates the training samples from $1$ to $N$. $\boldsymbol{x}_{simliar}$ and $\boldsymbol{x}_{dissimliar}$ represent the most similar and dissimilar images to the given test image ($x_{test}$) respectively, and $\boldsymbol{R}$ denotes the positive definite covariance matrix similar to Equation~\ref{eq:distance}. The same similarity metrics in Equation~\ref{eq:similar} and Equation~\ref{eq:dissimliar} allow computing a ranked list of similar and dissimilar images for a given test sample.

\begin{figure}[H]
    \centering
     \resizebox{\textwidth}{!}{     \begin{tabular}{c | c c c c c c c}
     \toprule
     Test image & \multicolumn{7}{c}{Similar images from training set in the embedding space} \\
     \midrule
     \includegraphics[width=2cm]{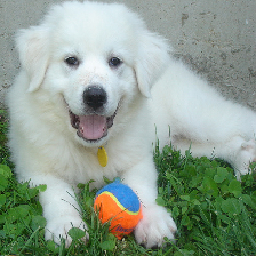} &    
     \includegraphics[width=2cm]{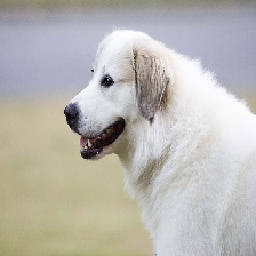} &
     \includegraphics[width=2cm]{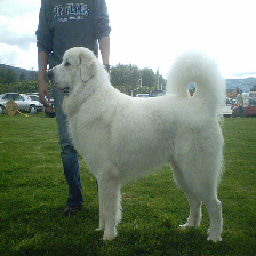} &
     \includegraphics[width=2cm]{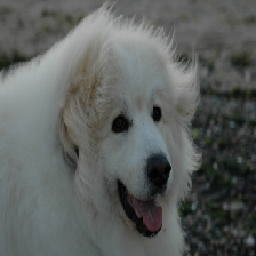} &
     \includegraphics[width=2cm]{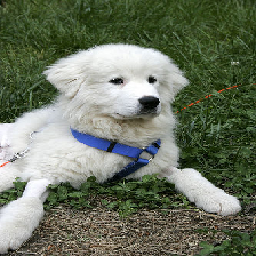} &
     \includegraphics[width=2cm]{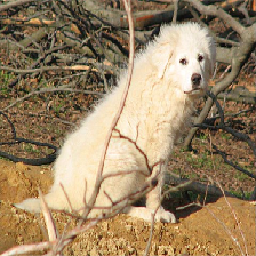} &
     \includegraphics[width=2cm]{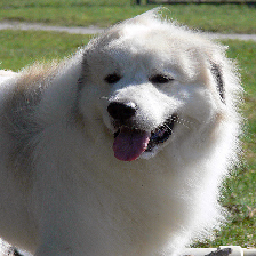} &
     \includegraphics[width=2cm]{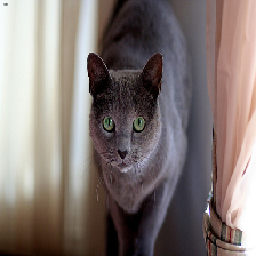} \\
     Learned metric &    
     \includegraphics[width=2cm]{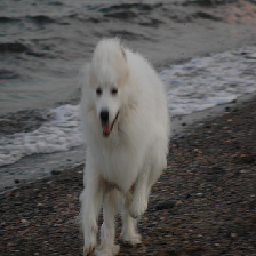} &
     \includegraphics[width=2cm]{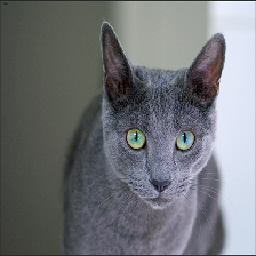} &
     \includegraphics[width=2cm]{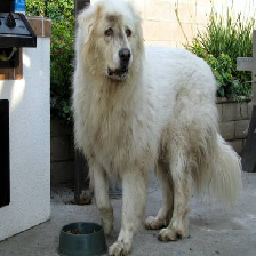} &
     \includegraphics[width=2cm]{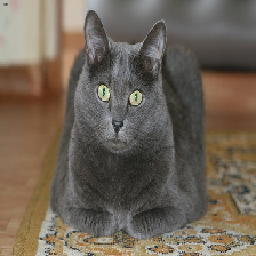} &
     \includegraphics[width=2cm]{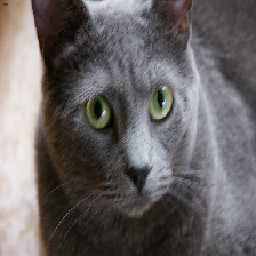} &
     \includegraphics[width=2cm]{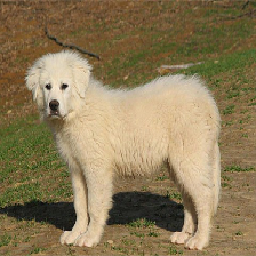} &
     \includegraphics[width=2cm]{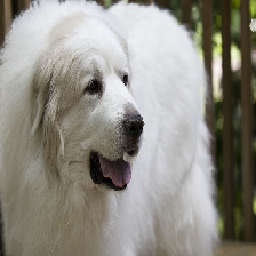} \\
     
     \toprule
     \includegraphics[width=2cm]{figures/rbf_for_cnn/distances/test_image.png} &    
     \includegraphics[width=2cm]{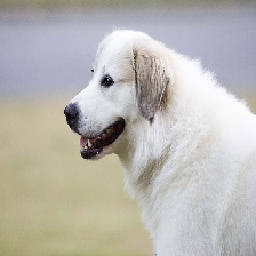} &
     \includegraphics[width=2cm]{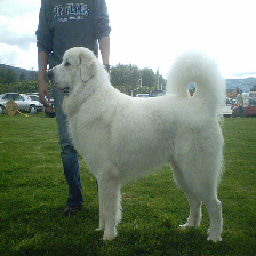} &
     \includegraphics[width=2cm]{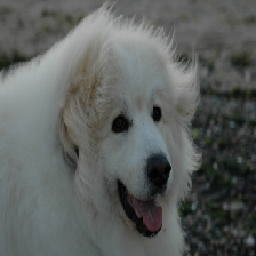} &
     \includegraphics[width=2cm]{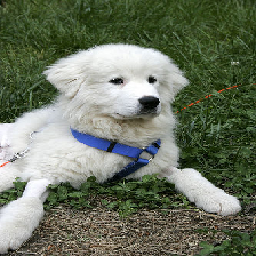} &
     \includegraphics[width=2cm]{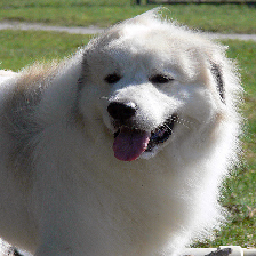} &
     \includegraphics[width=2cm]{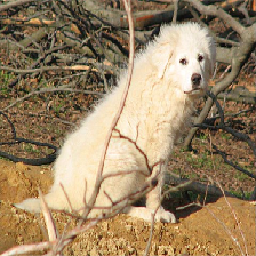} &
     \includegraphics[width=2cm]{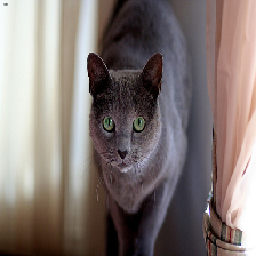} \\
     Euclidean distance &    
     \includegraphics[width=2cm]{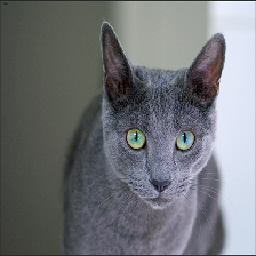} &
     \includegraphics[width=2cm]{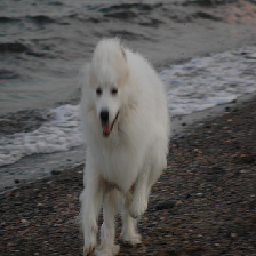} &
     \includegraphics[width=2cm]{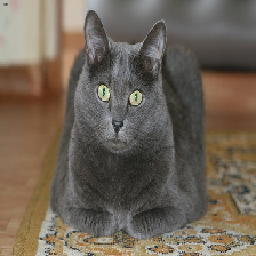} &
     \includegraphics[width=2cm]{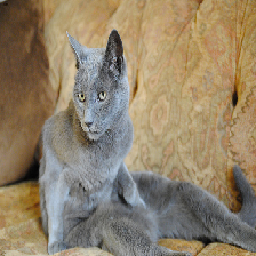} &
     \includegraphics[width=2cm]{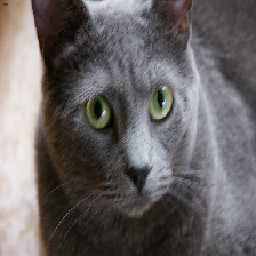} &
     \includegraphics[width=2cm]{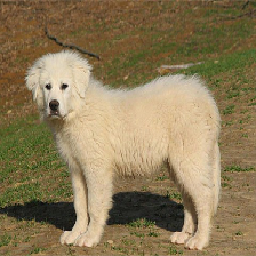} &
     \includegraphics[width=2cm]{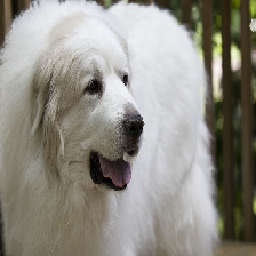} \\
     
     \toprule
     \includegraphics[width=2cm]{figures/rbf_for_cnn/distances/test_image.png} &    
     \includegraphics[width=2cm]{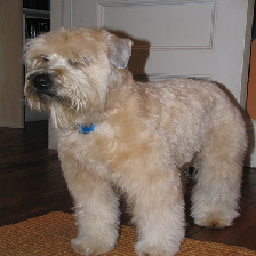} &
     \includegraphics[width=2cm]{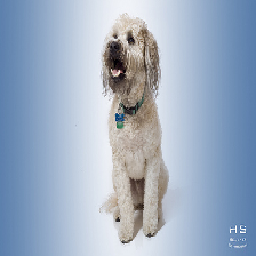} &
     \includegraphics[width=2cm]{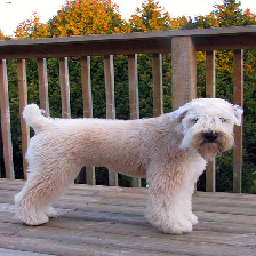} &
     \includegraphics[width=2cm]{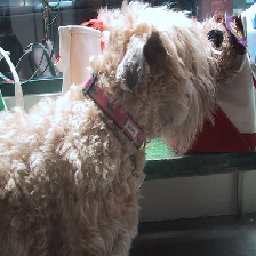} &
     \includegraphics[width=2cm]{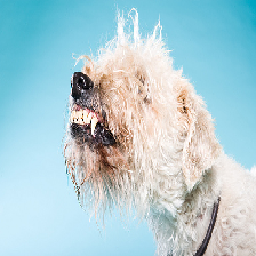} &
     \includegraphics[width=2cm]{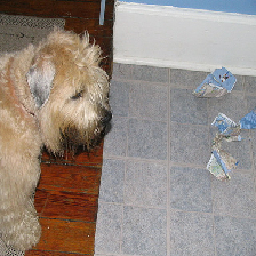} &
     \includegraphics[width=2cm]{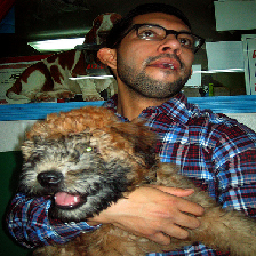} \\
     Cosine distance &    
     \includegraphics[width=2cm]{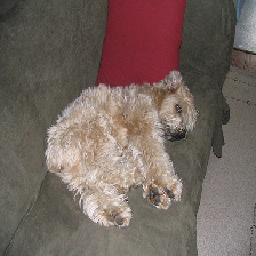} &
     \includegraphics[width=2cm]{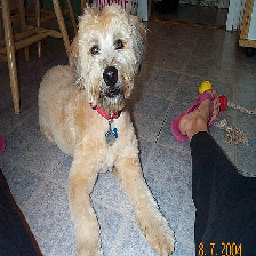} &
     \includegraphics[width=2cm]{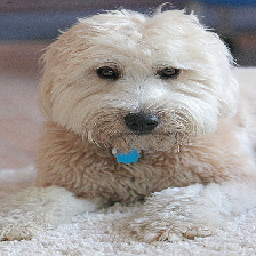} &
     \includegraphics[width=2cm]{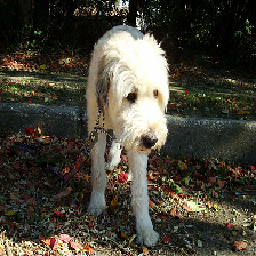} &
     \includegraphics[width=2cm]{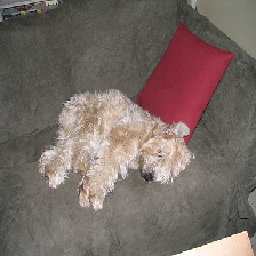} &
     \includegraphics[width=2cm]{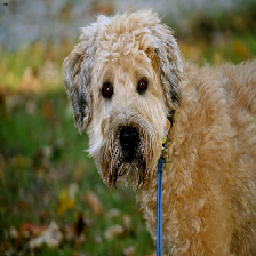} &
     \includegraphics[width=2cm]{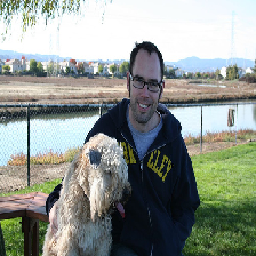} \\
     
     \toprule     
     \end{tabular}
}
     \caption{The presented figure visualizes the top $14$ images selected using different distance metrics in the embedding space for a given test image (figure adopted from~\cite{amirian2020radial}).}
     \label{fig:similar_distances}
\end{figure}

Figure~\ref{fig:similar_distances} compares the performance of the similar sample selection for given test images. The figure suggests that the learned metric and Euclidean distances outperform the cosine distance for similar sample selection. Furthermore, the learned metric slightly outperforms the Euclidean distance in these specific cases.

\begin{figure*}[htb!]
     \begin{center}
     \begin{tabular}{c c c c c}
\toprule
\includegraphics[width=0.175\linewidth]{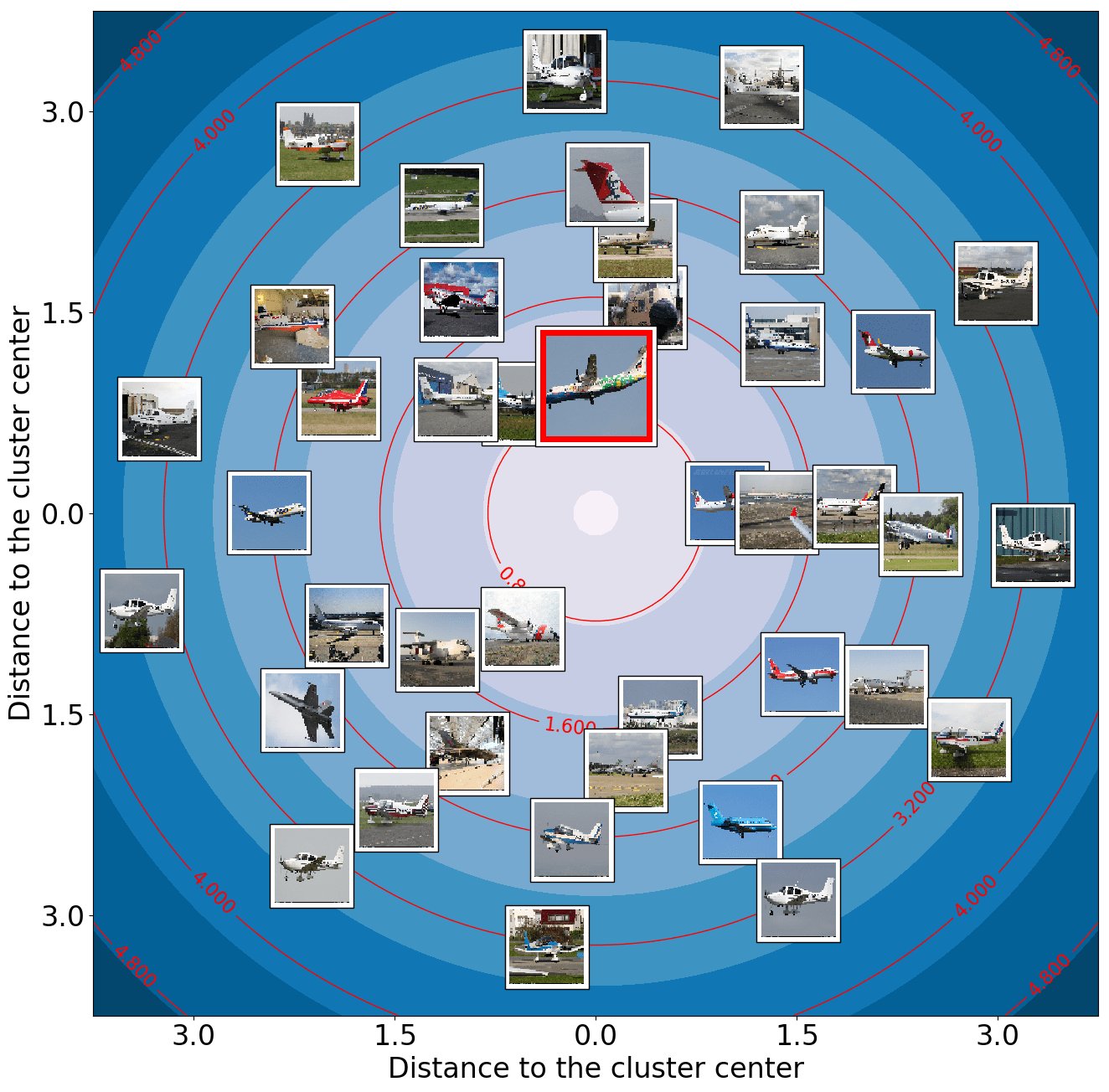} & \includegraphics[width=0.175\linewidth]{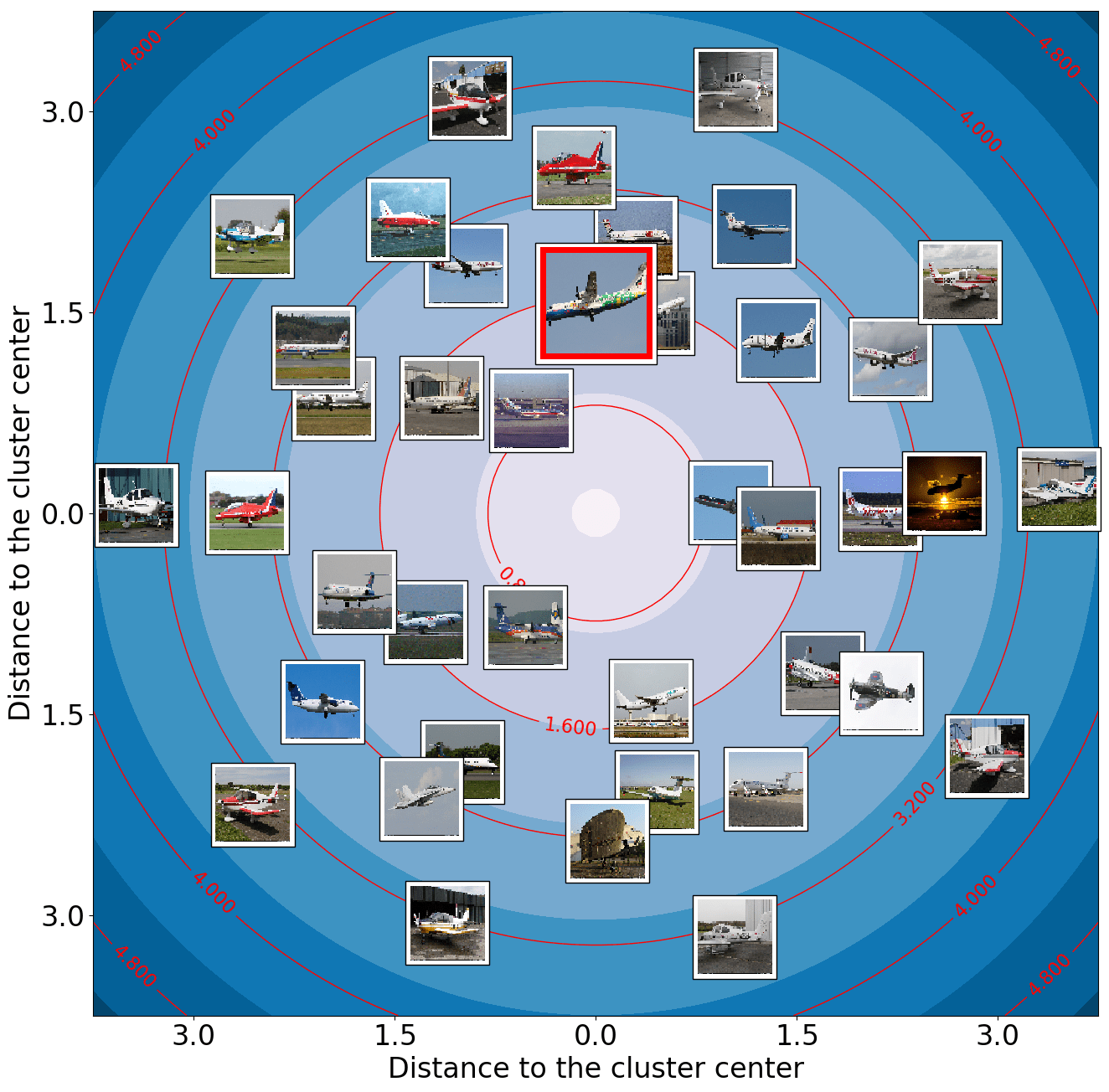} & \includegraphics[width=0.175\linewidth]{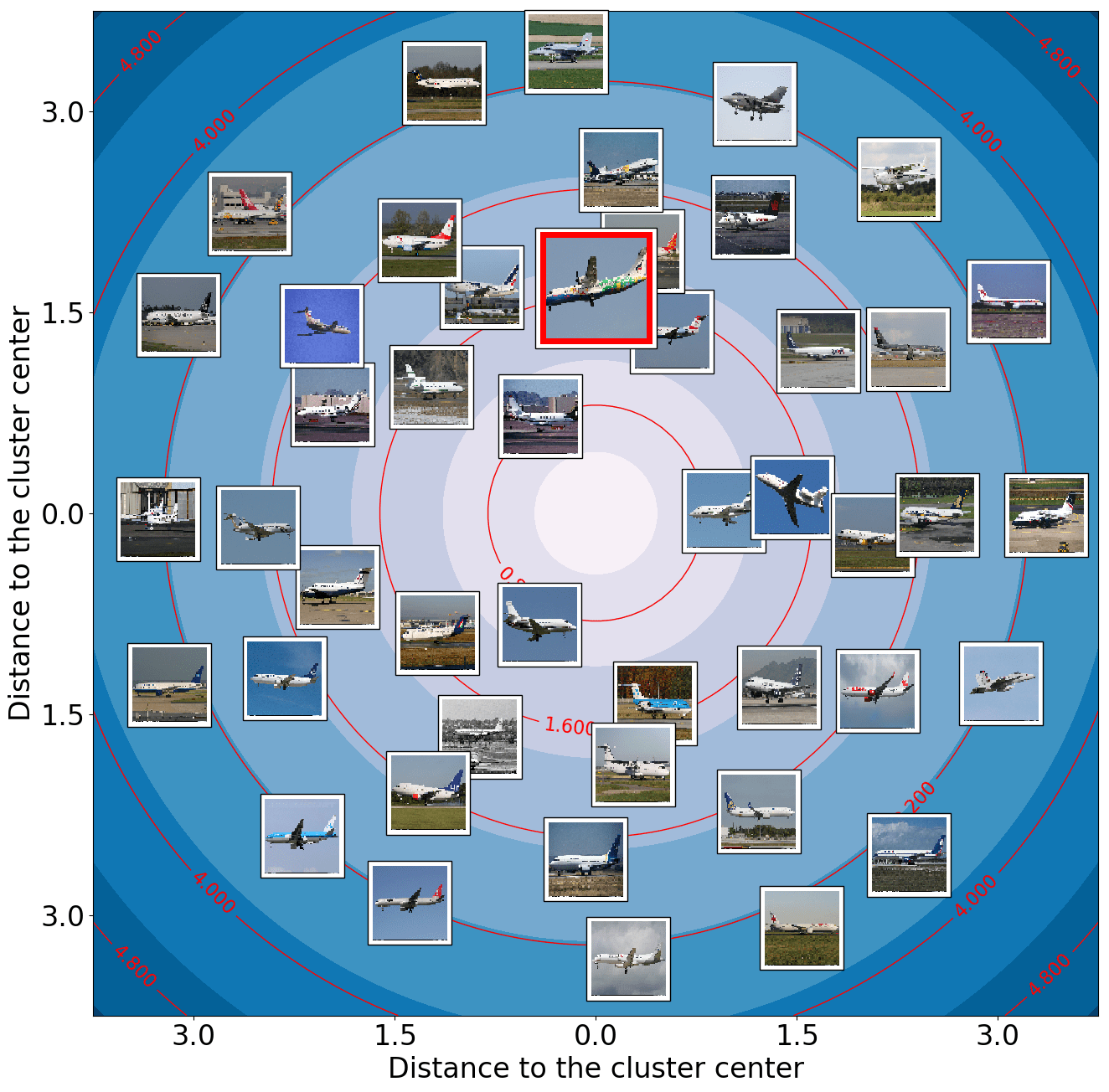} & \includegraphics[width=0.175\linewidth]{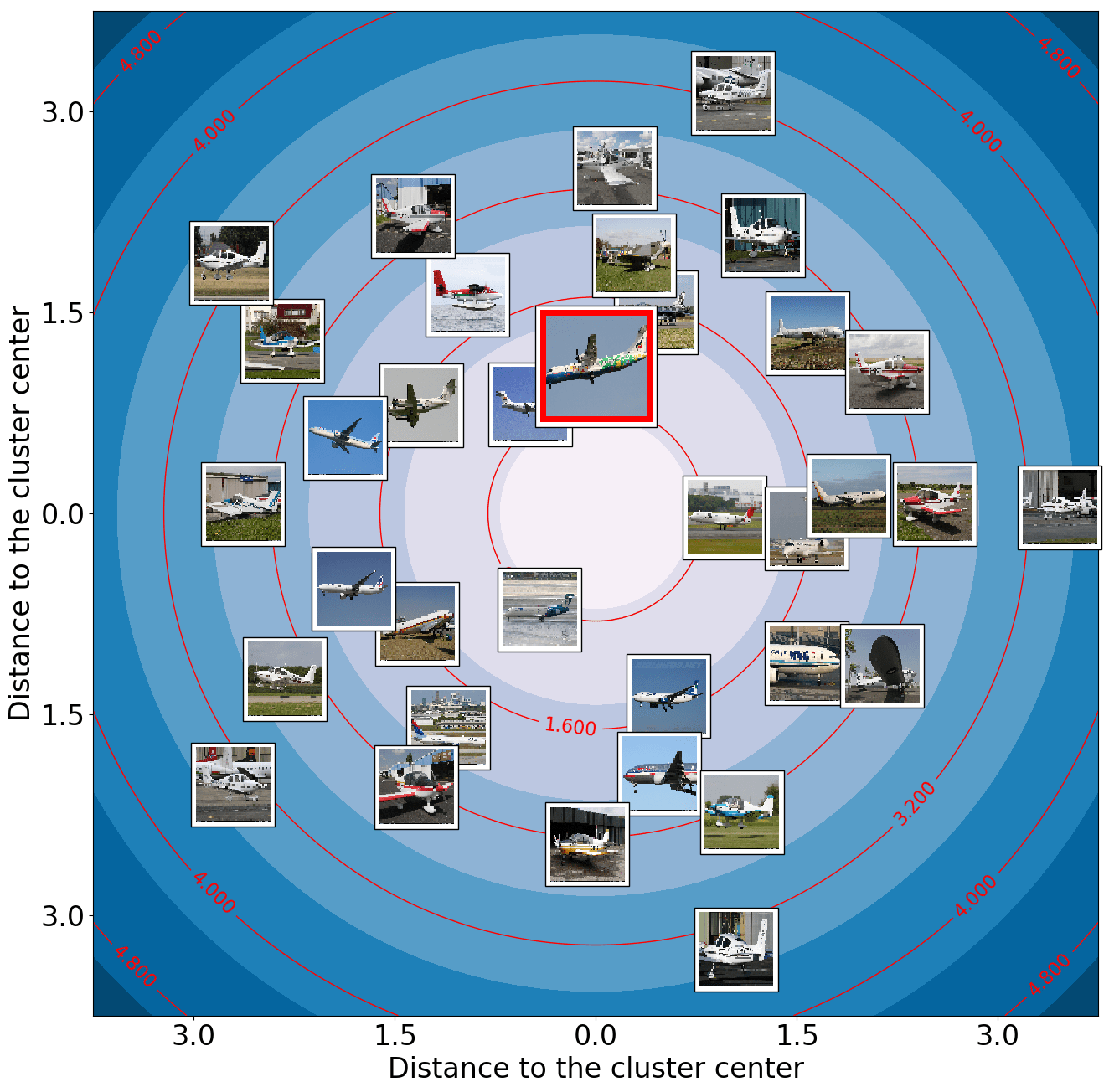} & \includegraphics[width=0.175\linewidth]{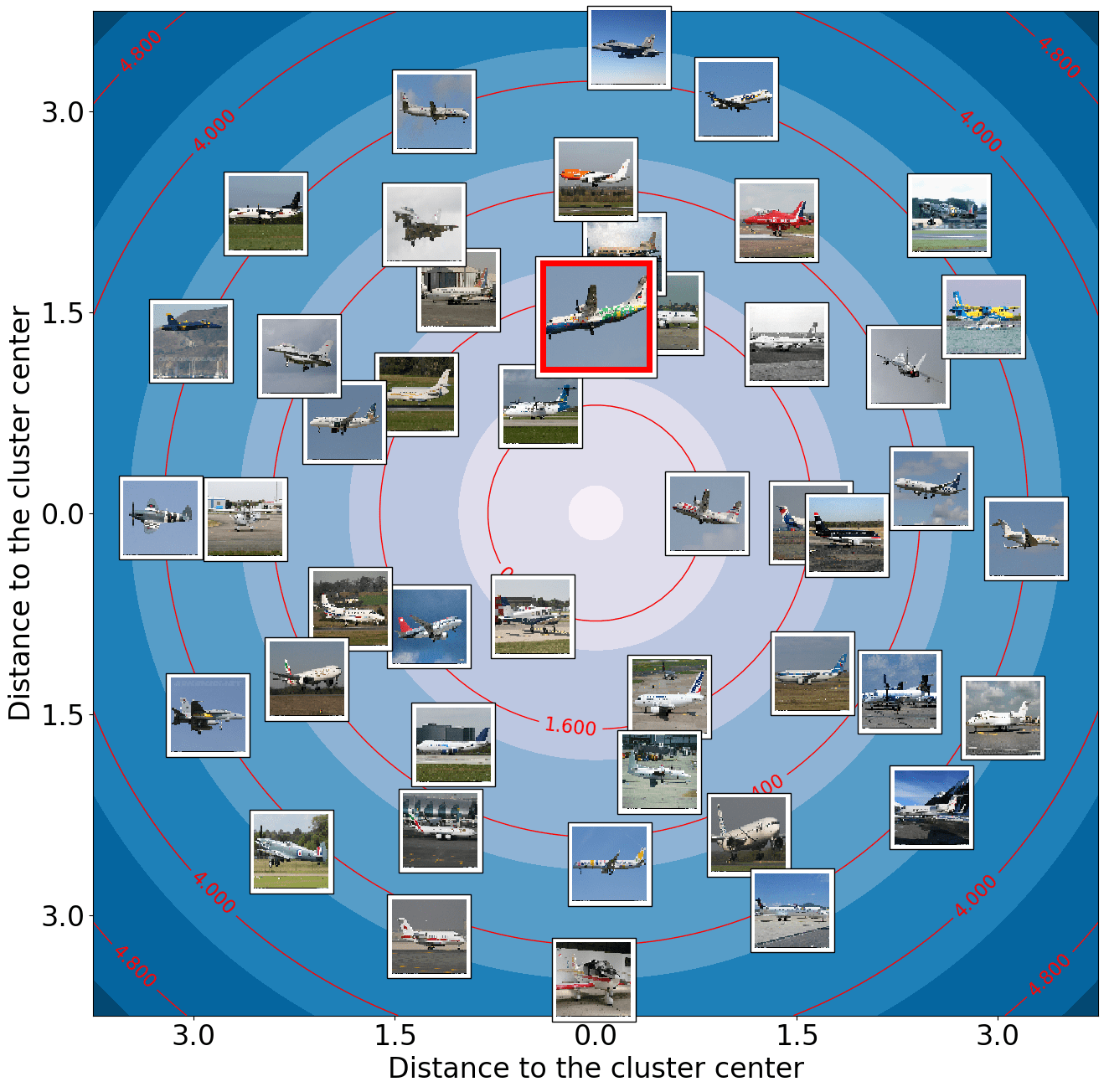} \\ \midrule
\includegraphics[width=0.175\linewidth]{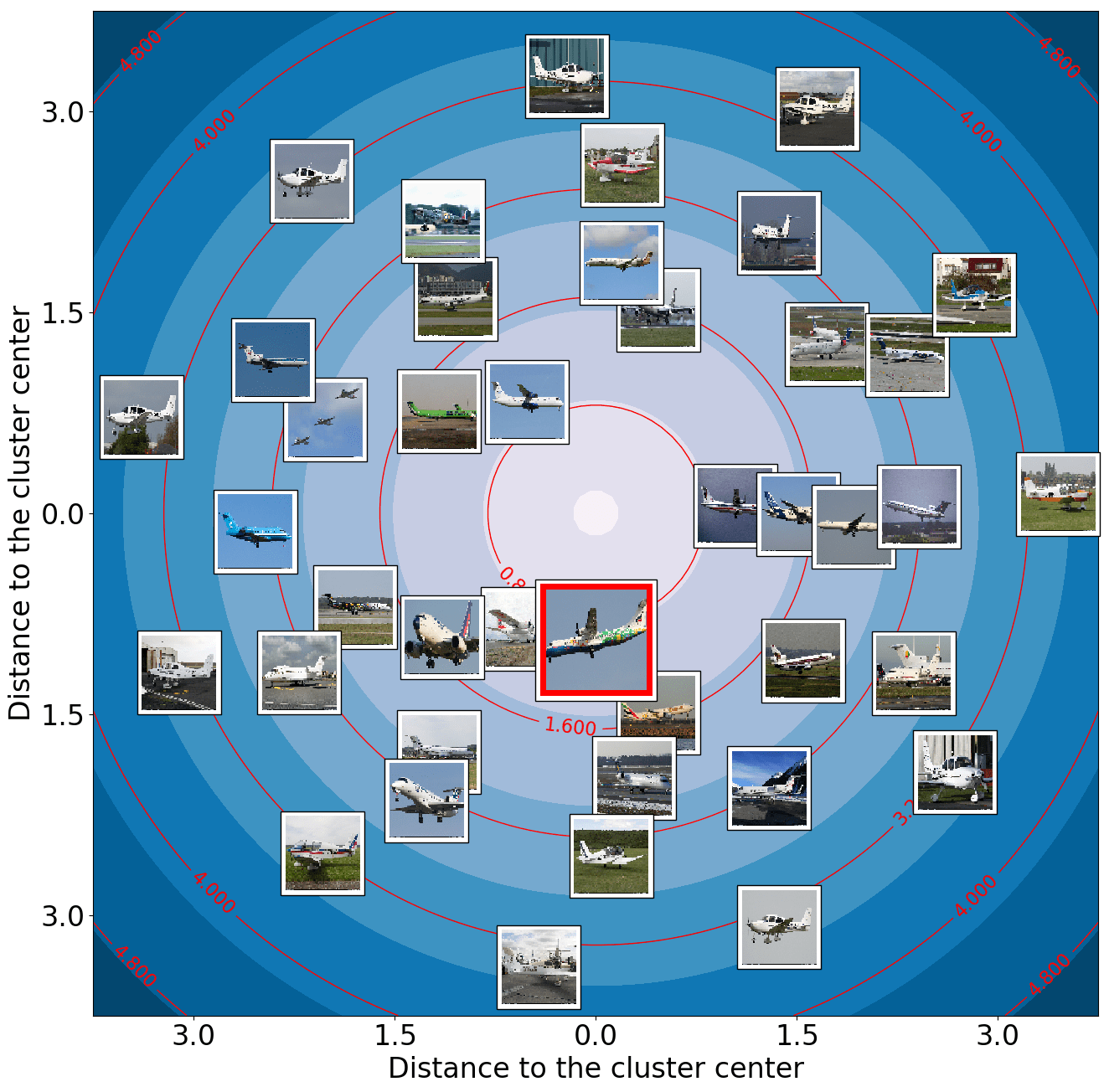} & \includegraphics[width=0.175\linewidth]{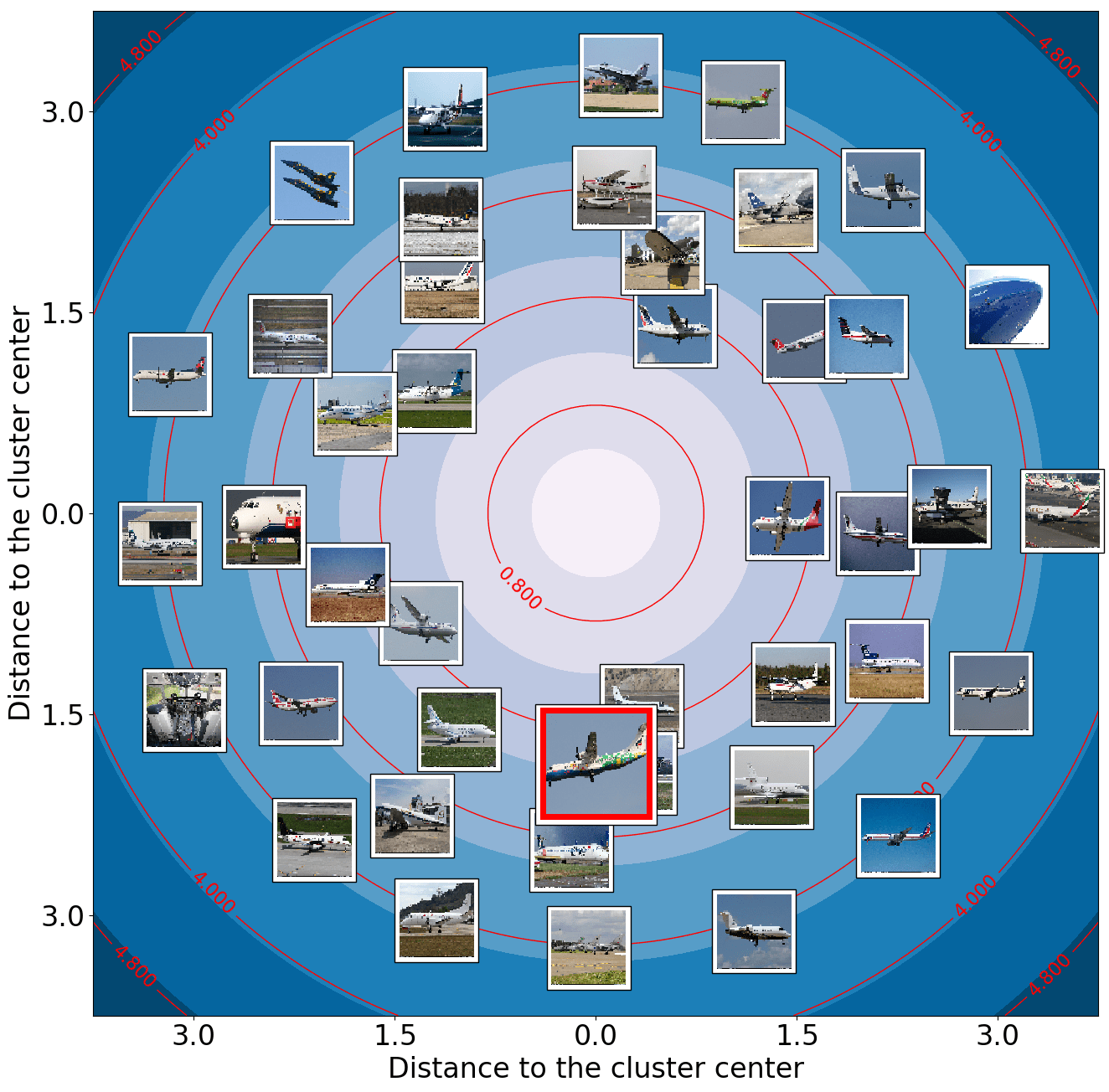} & \includegraphics[width=0.175\linewidth]{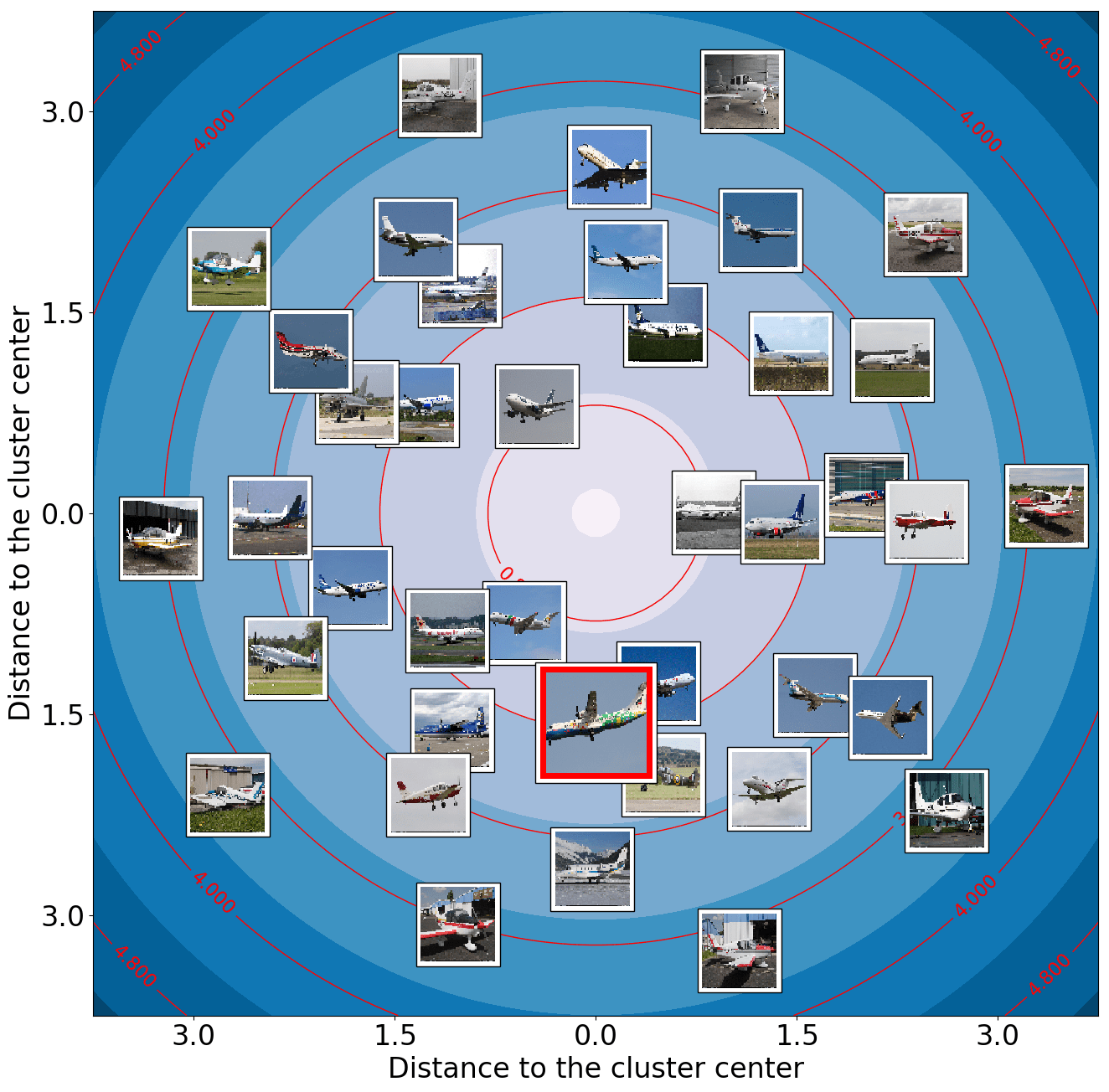} & \includegraphics[width=0.175\linewidth]{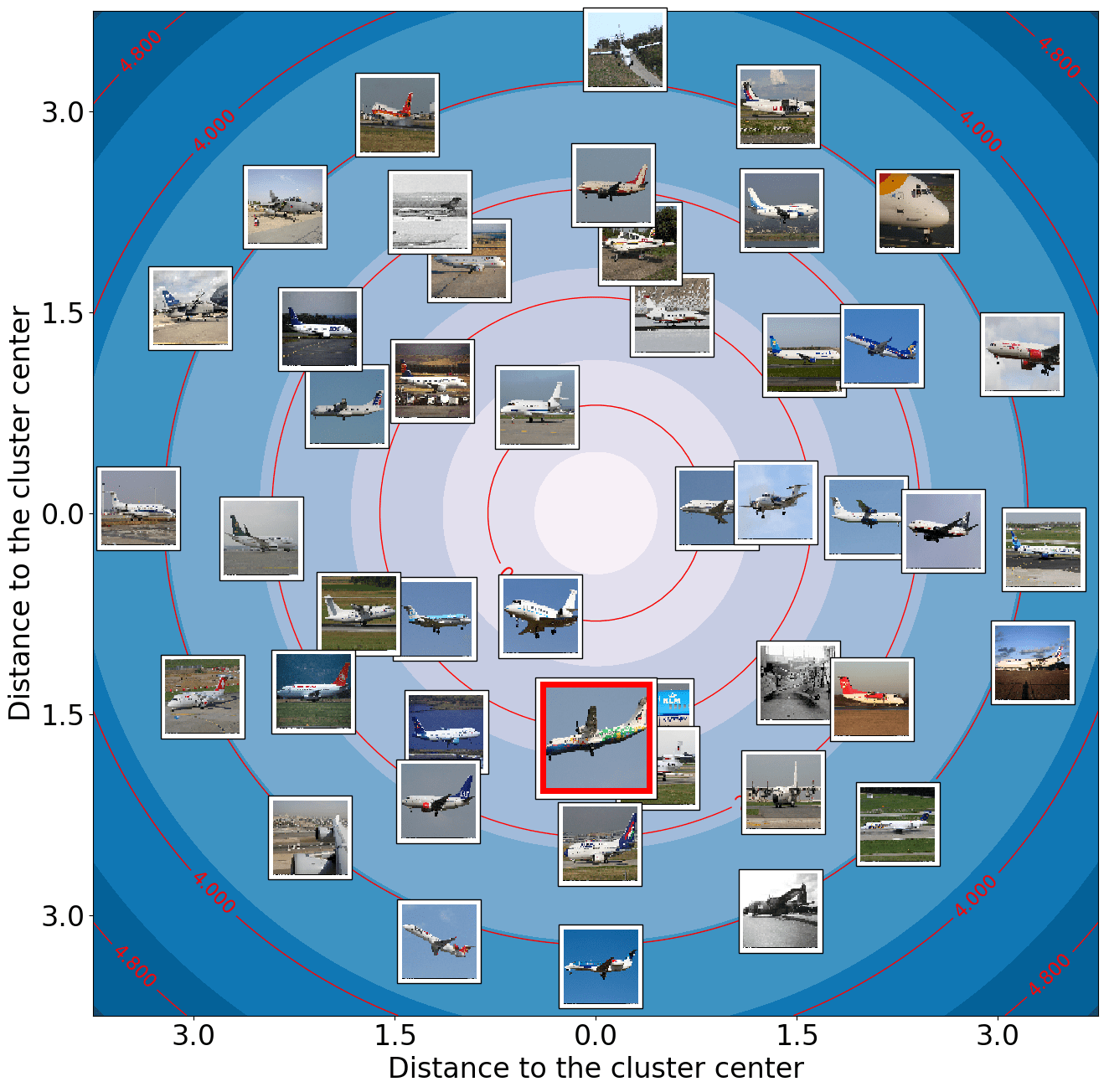} & \includegraphics[width=0.175\linewidth]{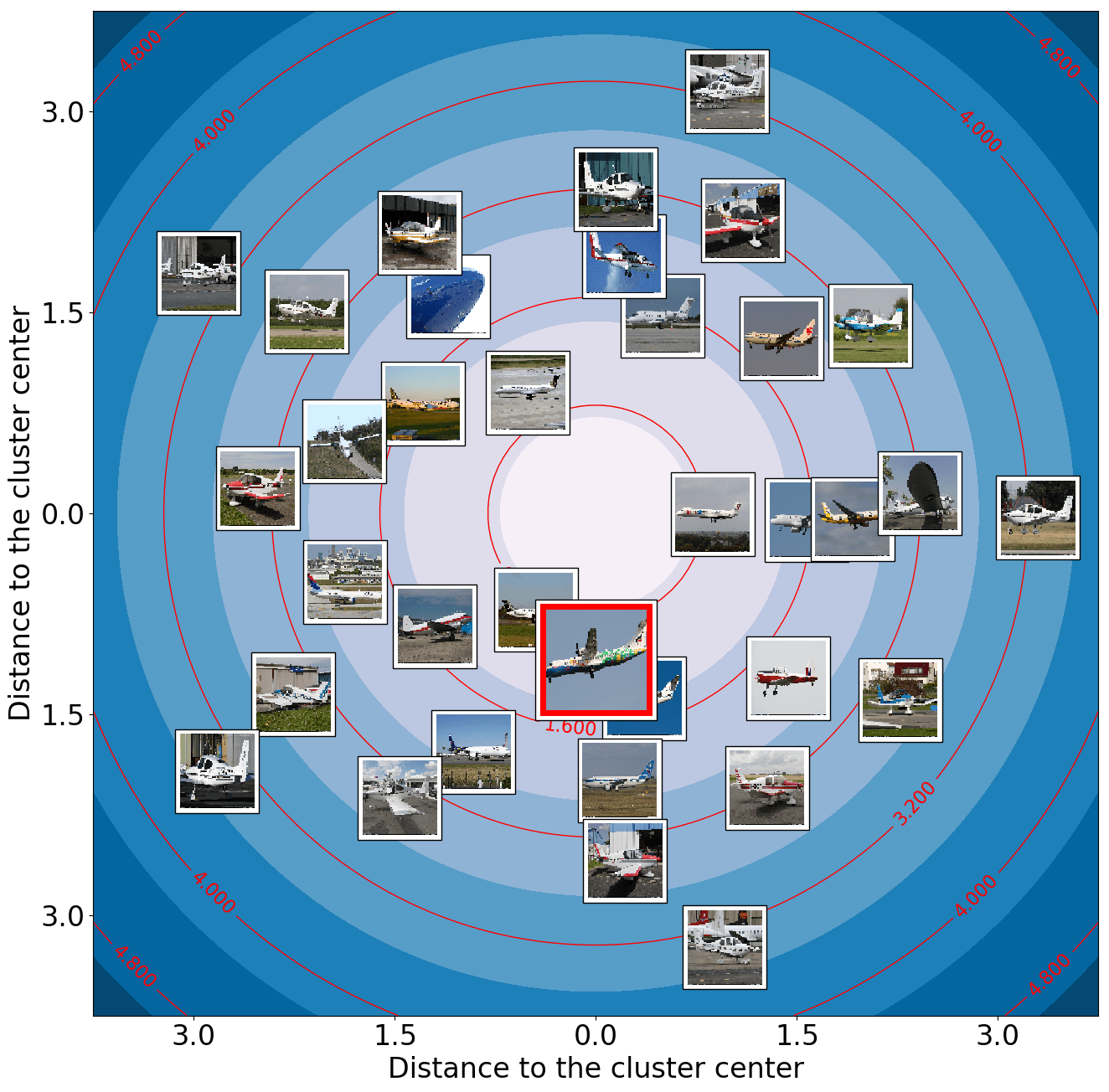} \\   
\toprule
\end{tabular}
     \end{center}
     \caption{The Figure illustrates the clusters contributing to a CNN-RBF network's correct class (top row) and the wrong class (bottom row). The larger image with red borders in each cluster representation is the test sample. Red circles show the distance of the samples to the cluster center, and the background is proportional to the activation values of the cluster. The brighter the activation value, the larger it is, and the maximum activation at the cluster center is equal to one (figure adopted from~\cite{amirian2020radial}).}
     \label{fig:corret_wrong_cluster}
\end{figure*}

The active clusters for every sample provide the reasoning behind the final decision of CNN-RBFs. The clusters can be depicted using the distance of images from their centers. Figure \ref{fig:corret_wrong_cluster} shows training samples and their distances from the cluster centers against a test sample. The product of activations and output weights determines the final decision of the RBFs. Thus, the importance of a cluster for a decision can be determined by sorting the product of activations and class weights. Figure~\ref{fig:corret_wrong_cluster} depicts the clusters with the highest contributions to the correct class (ground truth) and the wrong class based on this product. The wrong class here refers to the class with the second-highest confidence level.

\section{Discussions and Conclusions}
\label{chap:rbf_sec:conclusion}
This chapter presents fundamental architectural modifications to RBFs for integrating them into CNN architectures for computer vision. The experimental results indicate that integrating RBFs classifiers into CNN architectures achieves competitive performances in benchmark computer vision datasets by combining supervised and unsupervised learning. The proposed activation and training process is compatible with any arbitrary state-of-the-art CNN architecture, including inception blocks and residual connections. The small gap between the CNN-RBFs performance and the best CNN models is a subject for future research to find optimal regularization methods for RBF networks. Using RBF architectures with CNNs introduces two unique and network-specific opportunities for learning a similarity distance metric and interpreting the decision-making process in more detail. The classification of similar and dissimilar images found using a similarity distance metric trained by RBFs is interpretable by humans. The cluster representations are currently only used to trace the decision-making process. In the current research, the distribution of images around clusters is not visually conclusive because they are optimized in an unsupervised manner.

\chapter{Using Interpretability to Detect Adversarial Attacks for Robust CNNs}
\chaptermark{Interpretability to Detect Adversarial Attacks}
\label{chap:trace}

\label{chap:trace_sec:abstract}
The existence of adversarial attacks on convolutional neural networks (CNN) questions the fitness of such models for sensitive and serious practical applications. The adversarial attacks are minimal changes computed for a given input image, provoking a misclassification even though both images appear identical to a human observer; they are, therefore, difficult to detect. In a different context, backpropagated activations of CNNs' hidden layers for a given input image, so-called feature responses, have been helpful for humans to visualize and understand what the CNN looks at while computing its output class. This chapter presents a novel method to detect adversarial examples and identify manipulated images by tracking adversarial perturbations in feature responses. This method is fully human-explainable and allows the automatic detection of adversarial attacks using the average local spatial entropy of feature maps for a given input image without altering the original network architecture. Experiments confirm the validity and functionality of our approach for detecting state-of-the-art attacks on large-scale models trained on ImageNet. 

Alongside the contribution of this chapter to the robustness of computer vision models through the detection of adversarial attacks, this chapter presents one of the few applications of explainable artificial intelligence (XAI) for debugging models. Although feature response (maps) were initially intended to open the black box of vision models, they can also be used in practice for debugging, as presented in this chapter. This chapter demonstrates a successful application of vision model interpretability and XAI, which will hopefully inspire future work in the direction of debugging and designing models based on human-explainable methods. This chapter is adopted from the research published in~\cite{amirian2018trace}, licensed under CC BY 4.0~\footnote{\footnotesize{© 2018 Springer Nature Switzerland AG:} \url{https://creativecommons.org/licenses/by/4.0}}.


\section{Introduction}
\label{chap:trace_sec:introduction}
The success of deep neural nets for pattern recognition~\cite{schmidhuber2015deep} has been one of the primary drivers behind the recent surge of interest in AI. A substantial part of this success is due to the application of convolutional neural networks (CNNs)~\cite{lecun1998gradient,cirecsan2011committee} and their descendants on image recognition tasks. Moreover, the respective methods have reached a level of sophistication where they are now being used in business and industry~\cite{stadelmann2018beyondimagenet} and lead to a wide variety of deployed models for critical applications like automated driving~\cite{bojarski2016end} or biometrics~\cite{zhu2015multi}.

However, concerns regarding the reliability of deep neural networks have been raised after the discovery of so-called adversarial examples or attacks~\cite{szegedy2013intriguing}. These examples are specifically generated to fool a CNN into misclassifying visually very similar images or images that appear identical to the human eye with high confidence through the addition of barely visible perturbations~\cite{moosavi2016deepfool} (see Figure~\ref{chap:trace_fig:1}). The perturbations are computed using an optimization process on the input: the network weights are fixed, and the input pixels are optimized for the dual target of (a) classifying the input differently than the ground truth class and (b) minimizing the changes to the input. A growing body of literature confirms the importance of this discovery on practical applications of neural networks~\cite{akhtar2018threat}. The existence of adversarial attacks provokes questions on how CNNs achieve their performance and in what respect their decision-making differs from humans. In addition, adversarial attacks can threaten serious deployments of CNNs in the applications with the possibility of tailor-made attacks.

For instance, Su et al.~\cite{su2017one} report on successfully attacking neural networks by modifying a single pixel. This attack works without having access to the internal structure or the gradients in the network under attack. Moosavi-Dezfooli et al.~\cite{moosavi2017universal} further show the existence of universal adversarial perturbations that can be added to any image to fool a specific model.
Furthermore, the impact of similar attacks extends beyond classification~\cite{metzen2017universal}, attacks are transferable to other modalities~\cite{cisse2017houdini}, and also work on models distinct from neural networks~\cite{papernot2016transferability}. Finally, adversarial attacks have been shown to work reliably even after perturbed images have been printed and captured again via a mobile phone camera~\cite{kurakin2016adversarial}. Apparently, this research area touches on a weak spot concerning the robustness of CNNs in critical applications involving human privacy or security.

On the other hand, there is a recent interest in the explainability of AI agents, particularly using machine and deep learning models~\cite{vellido2012making,olah2018the}. It goes hand in hand with societal developments, like the new European legislation on data protection that affects any organization using algorithms on personal data~\cite{goodman2016eu}. While neural networks are publicly perceived as ``black boxes'' concerning how they arrive at their conclusions~\cite{DARPA2016explainable}, several methods have been developed recently to deliver insight into the representation and decision surface of trained models, improving interpretability. Prime candidates amongst these methods are feature response visualization approaches that provide information regarding operations in hidden layers of a CNN visible~\cite{zeiler2014visualizing,springenberg2014striving,olah2017feature}. They can thus serve a human engineer as a diagnostic tool in support of reasoning over the success and failure of a model on the task at hand.

This chapter presents a method for using a specific form of CNN feature visualization, namely feature response maps, using guided backpropagation technique~\cite{springenberg2014striving}, to not only \emph{trace} the effect of adversarial attacks but also to \emph{detect} them. This method traces the attacks on algorithmic decisions throughout CNNs. Moreover, it uses feature response maps as input to a novel automated detection approach based on a statistical analysis of feature maps' average local spatial entropy. The goal is to decide if a model is currently under attack by the given input. The proposed approach has the advantage over existing methods because it does not change the network architecture, i.e., it does not affect the classification accuracy but is explainable to humans. Experiments on the validation set of ImageNet~\cite{ILSVRC15} with VGG19 networks~\cite{simonyan2014very} show the validity of our approach for detecting various state-of-the-art attacks.

The remainder of this chapter is organized as follows: Section~\ref{chap:trace_sec:related_work} reviews related work in contrast to our approach. Section~\ref{chap:trace_sec:background} presents the background on adversarial attacks and feature response estimation before introducing the proposed approach in detail in Section~\ref{chap:trace_sec:detection}. Section~\ref{chap:trace_sec:experiments} reports on the experimental evaluations, and Section~\ref{chap:trace_sec:conclusions} concludes the chapter with an outlook on future work.

\section{Related Work}
\label{chap:trace_sec:related_work}

Work on adversarial examples for neural networks is a very active research field. Potential attacks and defenses are published at a high rate and have been surveyed by Akhtar and Mian~\cite{akhtar2018threat}. Amongst potential defenses directly comparable to our approach are those that focus solely on detecting a possible attack and not on additionally recovering correct classification. 

On the one hand, several detection approaches exist that exploit specific abnormal behavioral traces that adversarial examples leave while passing through a neural network. Liang et al.~\cite{liang2017detecting} consider the artificial perturbations as noise in the \emph{input} and attempt to detect it by quantizing and smoothing image filters. This method used a similar concept, which is the basis of SqueezeNet introduced by Xu et al.~\cite{xu2017feature}, which compares the network's \emph{output} on the raw and filtered input, and raises a flag if it detects a large difference between both. Feinman et al.~\cite{feinman2017detecting} observe the network's output confidence as estimated by dropout in the forward pass~\cite{gal2016dropout}, and Lu et al.'s SafetyNet~\cite{lu2017safetynet} looks for abnormal patterns in the ReLU activations of \emph{higher layers}. In contrast, the method presented in this chapter performs detection based on statistics of activation patterns in the complete \emph{representation learning} part of the network as observed in feature response maps, whereas Li~\cite{li2016adversarial} directly observes convolutional filter statistics there.

On the other hand, the second class of detection approaches trains sophisticated classifiers for directly sorting out adversarially optimized inputs: Meng and Chen's MagNet~\cite{meng2017magnet} learns the manifold of friendly images, rejects far away ones as hostile and modifies close outliers to be attracted to the friendly images' manifold before feeding them back to the network under attack. Grosse et al.~\cite{grosse2017statistical} enhance the output of an attacked classifier by an additional class and retrain the model to classify adversarial examples as such directly. Metzen et al.~\cite{metzen2017detecting} have a similar goal but target it via an additional subnetwork. In contrast, this chapter presents a method that uses a simple threshold-based detector and pushes all decision power to the human-explainable feature extraction via the feature response maps.

Finally, as shown in~\cite{akhtar2018threat}, different and mutually exclusive explanations for the existence of adversarial examples and the nature of neural network decision boundaries exist in the literature. Because our method enables a human investigator to trace attacks visually, it will be instrumental in taking this debate further. 

\section{Background}
\label{chap:trace_sec:background}

This section briefly presents adversarial attacks and the theory of feature response estimation before assembling both parts into the proposed detection approach in the next section.

\subsection{Adversarial Attacks}
\label{chap:trace_sec:adversarial_attacks}

\begin{figure}[htb!]
     \begin{center}
     \begin{tabular}{  c  c  c}
     \toprule

     Original  & Difference & Adversarial \\ \midrule

      \includegraphics[width=0.14\textwidth]{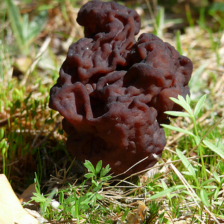} & \includegraphics[width=0.14\textwidth]{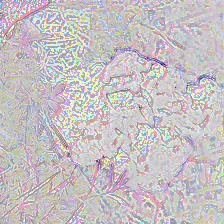} & \includegraphics[width=0.14\textwidth]{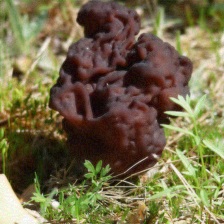}  \\ 
     Gyromitra  & Difference & Trafic light \\ \midrule
     
     \includegraphics[width=0.14\textwidth]{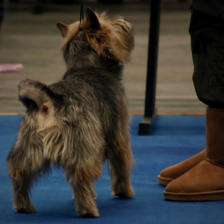} & \includegraphics[width=0.14\textwidth]{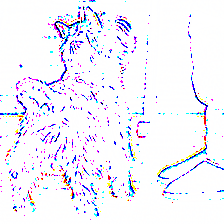} & \includegraphics[width=0.14\textwidth]{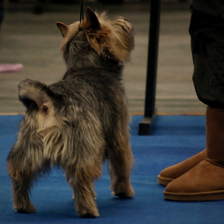}  \\ 
     Norwich terrier  & Difference & Lampshade \\ \midrule
     
      \includegraphics[width=0.14\textwidth]{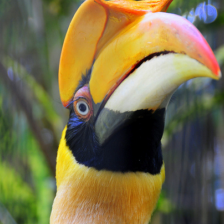} & \includegraphics[width=0.14\textwidth]{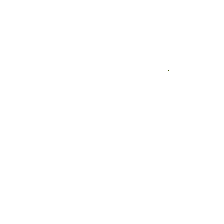} & \includegraphics[width=0.14\textwidth]{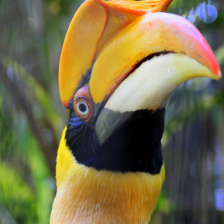}  \\ 
      Hornbill & Difference & Spotlight \\ \midrule
     
     \includegraphics[width=0.14\textwidth]{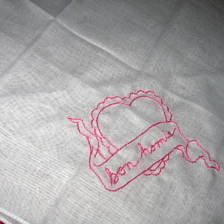} & \includegraphics[width=0.14\textwidth]{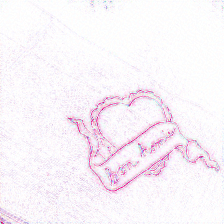} & \includegraphics[width=0.14\textwidth]{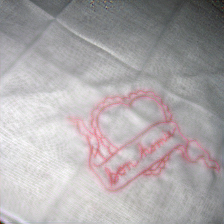} \\ 
     Handkerchief & Difference & Lampshade \\ \midrule

\includegraphics[width=0.14\textwidth]{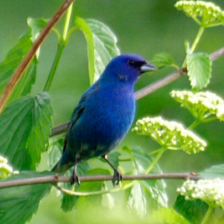} & \includegraphics[width=0.14\textwidth]{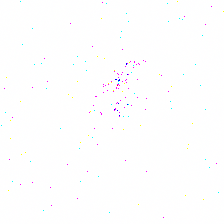} & \includegraphics[width=0.14\textwidth]{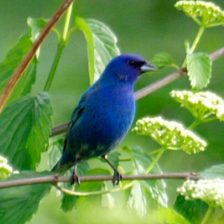}  \\ 
     Indigo bunting & Difference & Spotlight \\ \midrule     
      \end{tabular}
      \caption{Examples of different state-of-the-art adversarial attacks on a VGG19 model: original images and labels (left), perturbations (middle), and mislabeled adversarial examples (right). In the middle column, zero difference is encoded white, and the maximum difference is black because of visual enhancement (figure adopted from~\cite{amirian2018trace}).}
      \label{chap:trace_fig:1}
      \end{center}
\end{figure}

The principal idea behind adversarial attacks is to find a small perturbation for a given image that changes the prediction of a CNN. Pioneering work demonstrated that negligible and visually insignificant perturbations could lead to considerable deviations in the networks' output~\cite{szegedy2013intriguing}. The optimization problem of finding a perturbation $\boldsymbol{\eta}$ for a normalized clean image $\boldsymbol{I} \in \mathbb{R}^m$, where $m$ is the size (width$\times$height) of the image, is stated as follows \cite{szegedy2013intriguing}:
\begin{equation}
  \min_{\boldsymbol{\eta}} \parallel\boldsymbol{\eta}\parallel_2 \quad \text{s.t.} \quad \mathcal{C}(\boldsymbol{I}+\boldsymbol{\eta})\neq\ell\enspace;\quad \boldsymbol{I}+\boldsymbol{\eta} \in [0,1]^m
 \label{chap:trace_eq: optimization}
\end{equation}
where $\mathcal{C}(.)$ presents the classifier, and $\ell\enspace$ is the ground truth label. Szegedy et al.~\cite{szegedy2013intriguing} proposed a solution for the optimization problem of finding the perturbations in Equation~\ref{chap:trace_eq: optimization} for arbitrary labels $\ell^\prime$ that differ from the ground truth. However, they used box-constrained limited-memory Broyden–Fletcher–Goldfarb–Shanno (L-BFGS)~\cite{fletcher2013practical} to find perturbations satisfying Equation~\ref{chap:trace_eq: optimization}. Optimization based on the L-BFGS algorithm for finding adversarial attacks is computationally inefficient compared to gradient-based methods. Therefore, in this chapter, a few different gradient-based attacks, a one-pixel attack, and a boundary attack are used to compute adversarial examples, as explained in the following paragraphs (see Figure \ref{chap:trace_fig:1}).

\textbf{Fast gradient sign method (FGSM)}~\cite{goodfellow2014explaining} is a method suggested for computing adversarial perturbations based on the gradient $\nabla_{\boldsymbol{I}}J(\boldsymbol{\theta}, \boldsymbol{I}, \ell)$ of the cost function with respect to the original image pixel values:
\begin{equation}
\boldsymbol{\eta} = \epsilon \ \text{sign} (\nabla_{\boldsymbol{I}}J(\boldsymbol{\theta}, \boldsymbol{I}, \ell))
\label{chap:trace_eq: FGSM}
\end{equation}
where $\boldsymbol{\theta}$ represents the network parameters and $\epsilon$ is a constant factor that constrains the max-norm ($l_\infty$) of the additive perturbation ($\eta$). The ground truth label is presented by $\ell$ in Equation~\ref{chap:trace_eq: FGSM}. The $\text{sign}$ function is Equation~\ref{chap:trace_eq: FGSM} which computes the elementwise sign of the gradient of the loss function with respect to the input image. Optimizing the perturbation in Equation~\ref{chap:trace_eq: FGSM} in a single step is called the fast gradient sign method (FGSM) in the literature. This method is a white box attack, i.e. the algorithm for finding the adversarial example requires information on weights, gradients, and the network's architecture.

\textbf{Gradient attack} is a straightforward realization of finding adversarial perturbations in the FoolBox toolbox~\cite{rauber2017foolbox}. It optimizes pixel values of an original image to minimize the ground truth label confidence in a single step based on the gradient values instead of their sign proposed in the FGSM method.

\textbf{One pixel attack}~\cite{su2017one} is a semi-black box approach to compute adversarial examples using an evolutionary algorithm~\cite{storn1997differential}. The algorithm is not a white box since it does not need the gradient information of the classifier. However, it is not a full black box as it needs the class probabilities. The iterative algorithm starts with randomly initialized parent perturbations. The generated offspring compete with their parents at each iteration, and the winners advance to the next step. The algorithm stops when the ground truth label probability is lower than $5\%$.  

\textbf{DeepFool}~\cite{moosavi2016deepfool} is a white box iterative approach in which the closest direction to the decision boundary is computed in every step. It is equivalent to finding the corresponding path to the orthogonal projection of the data point onto the affine hyperplane, which separates the binary classes. The initial method for binary classifiers can be extended to a multi-class task by considering the task multiple one-versus-all binary classifications. After finding the optimal updates toward the decision boundary, the perturbation is added to the given image. The iterations continue estimating the optimal perturbation and apply it to the perturbed image from the last step until the network fails to predict the ground truth label.

\textbf{Boundary attack} is a black-box attack proposed by Brendel et al. in~\cite{brendel2017decision}. The algorithm starts with an adversarial image from another class compared with the target image and iteratively optimizes the distance between this image and the target image. It searches for an adversarial example with a minimum distance from the target image that keeps its original class throughoutd the optimization.

\subsection{Feature Response Estimation}
\label{chap:trace_sec:feature_response}
The technique of visualizing CNNs through feature responses is used to figure out which images' region leads to the final prediction of a network. Therefore, computing feature responses enhances the explainability of the classifiers. This chapter demonstrates how to use this visualization tool to trace the effect of adversarial attacks on CNNs' predictions and detect perturbed examples automatically.

\begin{table}[htb!]
     \begin{center}
     \resizebox{\textwidth}{!}{
     \begin{tabular}{  l | c  c | c c}
     \toprule
     One pixel attack \cite{su2017one}: & \includegraphics[width=0.14\textwidth]{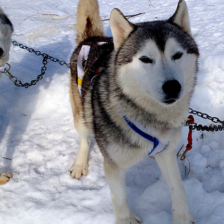} & \includegraphics[width=0.14\textwidth]{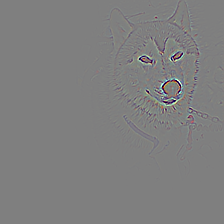} & \includegraphics[width=0.14\textwidth]{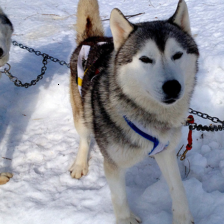} & \includegraphics[width=0.14\textwidth]{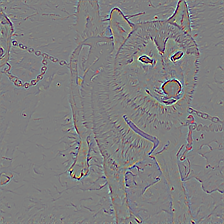} \\ 
     Predictions: & Eskimo dog & Feature response & Thimble & Feature response \\ \midrule
     FGSM \cite{goodfellow2014explaining}: & \includegraphics[width=0.14\textwidth]{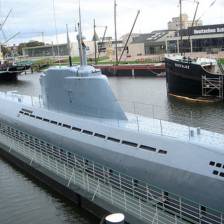} & \includegraphics[width=0.14\textwidth]{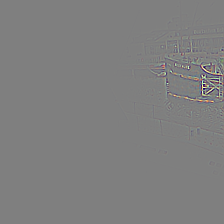} & \includegraphics[width=0.14\textwidth]{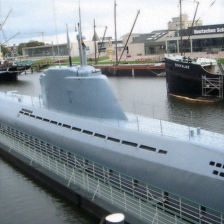} & \includegraphics[width=0.14\textwidth]{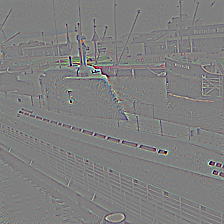}  \\ 
     Predictions: & Submarine & Feature response & Traffic light & Feature response \\ \midrule
     DeepFool \cite{moosavi2016deepfool}: & \includegraphics[width=0.14\textwidth]{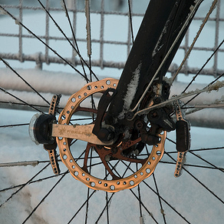} & \includegraphics[width=0.14\textwidth]{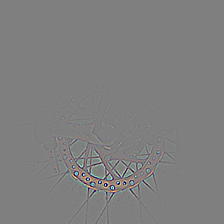} & \includegraphics[width=0.14\textwidth]{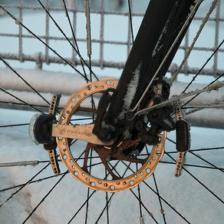} & \includegraphics[width=0.14\textwidth]{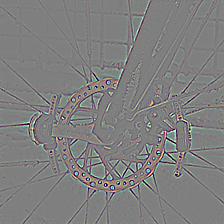} \\ 
     Predictions: & Disc brake & Feature response & Dome & Feature response \\ \toprule
      \end{tabular}}
      \caption{Effect of adversarial attacks on feature responses: (left) original images, and their feature responses, (right) perturbed versions, and their feature responses (figure adopted from~\cite{amirian2018trace}).}
      \label{chap:trace_fig:2}
      \end{center}
\end{table}

Erhan et al.~\cite{erhan2009visualizing} used backpropagation for visualizing feature responses of CNNs. They evaluate an arbitrary image in the forward pass and retain the activation values; then backpropagate from the last convolutional layer to the original image. As a result, the feature response maps have higher intensities in the regions that cause larger network activation values (see Figure~\ref{chap:trace_fig:2}). Moreover, the information on max-pooling layers in the forward pass can further improve the quality of these visualizations. Zeiler et al.~\cite{zeiler2014visualizing} proposed computing ``switches'', to consider the position of maximum values in all pooling layers, and then construct the feature response maps using transposed convolutional~\cite{dumoulin2016guide} layers.

Ultimately, Springenberg et al. \cite{springenberg2014striving} proposed a combination of both methods called ``guided backpropagation''. In this approach, the information of ``switches'' (max-pooling spatial information) is kept, and the activations are propagated backward with the guidance of the ``switch'' information. This method leads to the best performance in the visualization of the inner workings of the network. Therefore, guided backpropagation is used for computing feature response maps in this chapter.

\section{Explainable Adversarial Attacks Detection}
\label{chap:trace_sec:detection}

After reviewing the necessary background in the last section, this section presents this thesis's contribution to tracing adversarial examples in feature response maps, which inspired the novel approach to the automatic detection of adversarial perturbations in images. In this manner, visual representations of neural networks' inner workings also provide expert human guidance in developing CNNs that have increased reliability and explainability.    

\subsection{Tracing Adversarial Attacks}

The research question followed in this chapter is whether explainability methods can provide insight into the reasons behind the misclassification of adversarial examples. The effect of adversarial attacks in the feature response maps of CNNs is traced in Figure~\ref{chap:trace_fig:2}. The general phenomenon observed in all images is that the feature response maps' active region for adversarial examples is widely spread. In contrast, Figure~\ref{chap:trace_fig:2} demonstrates that the network looks at a smaller region of the image, i.e. is more focused, in the case of not manipulated samples.

The adversarial images are visually very similar to the original ones. However, they are not recognizable by deep CNNs. The original idea behind this study is that the focus of CNNs changes when facing an adversarial attack, leading to incorrect predictions. Conversely, the network makes the correct prediction once it focuses on the right region of the image. Visualizing the feature responses provides this and other exciting information regarding decision-making in computer vision models. For instance, the image of the submarine in Figure~\ref{chap:trace_fig:2} can be considered a good candidate for an adversarial attack since the CNN is making the decision based on an object in the background (see the feature response maps of the original submarine in Figure~\ref{chap:trace_fig:2}). 

\subsection{Detecting Adversarial Attacks}

\begin{table}[htb]
     \begin{center}
     \begin{tabular}{l | c c c c}
     \toprule
      & Original & Adversarial & Original & Adversarial \\ \midrule
      Image: & \includegraphics[width=0.14\textwidth]{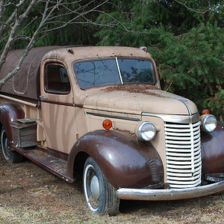} &                \includegraphics[width=0.14\textwidth]{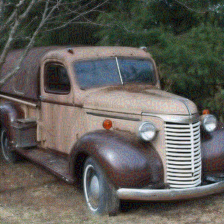} & \includegraphics[width=0.14\textwidth]       {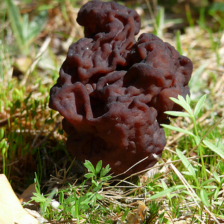} & \includegraphics[width=0.14\textwidth]{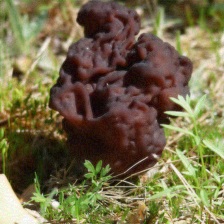} \\ \midrule
      Feature response: & \includegraphics[width=0.14\textwidth]{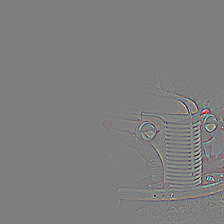} &                \includegraphics[width=0.14\textwidth]{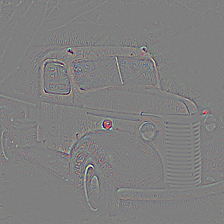} & \includegraphics[width=0.14\textwidth]       {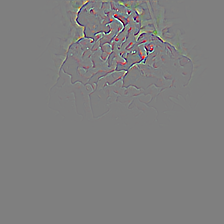} & \includegraphics[width=0.14\textwidth]{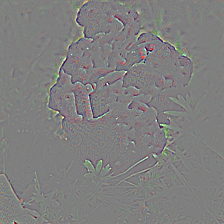} \\ \midrule
      Local spatial entropy: & \includegraphics[width=0.14\textwidth]{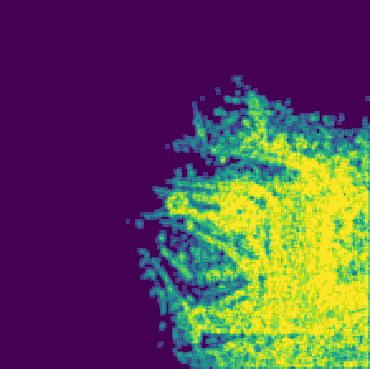} &                \includegraphics[width=0.14\textwidth]{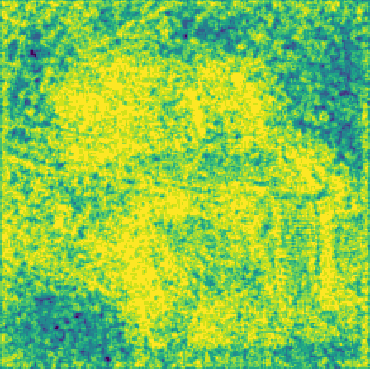} & \includegraphics[width=0.14\textwidth]       {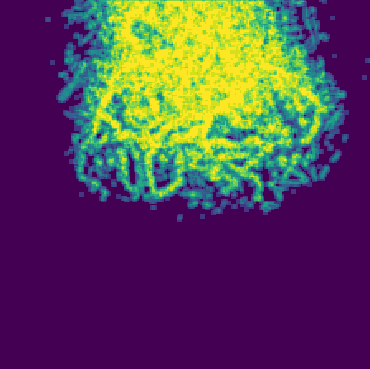} & \includegraphics[width=0.14\textwidth]{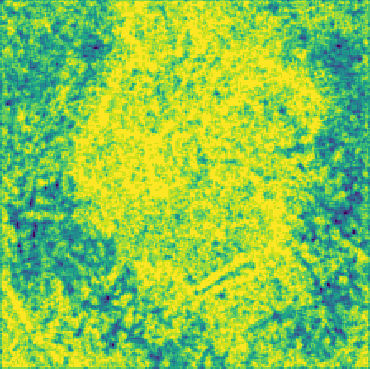} \\
      \toprule
      \end{tabular}
	  \caption{Input, feature response maps, and local spatial entropy for clean and perturbed images, respectively (table adopted from~\cite{amirian2018trace}).}
	  \label{chap:trace_fig:3}
      \end{center}
\end{table}

Experiments tracing the effect of adversarial attacks on feature response maps thus suggest that a CNN classifier focuses on a broader region of the input if it is deliberately perturbed. Figure~\ref{chap:trace_fig:2} demonstrates this connection for models' decision-making in the case of clean inputs compared with manipulated ones. The effect of adversarial manipulation is even more visible in the local spatial entropy of the grayscale feature response maps (see Figure~\ref{chap:trace_fig:3}). The feature response maps are initially converted to grayscale images, and local spatial entropies are computed based on the grayscale feature response maps as follows~\cite{chanwimaluang2003efficient}:
\begin{align}
S_k = - \sum_{i} \sum_{j} \boldsymbol{h}_k(i, j) \log_2 (\boldsymbol{h}_k(i, j))
\label{chap:trace_eq:en}
\end{align}
where $S_k$ is the local spatial entropy of a small part (patch) of the input image and $\boldsymbol{h}_k$ represents the normalized 2D histogram value of the $k^{th}$ patch. The indices $i$ and $j$ scan through the height and width of the image patches. The patch size is $3\times3$, the same as the filter size of the first layer of the VGG19 model used. The local spatial entropies of corresponding feature responses are presented in Figure~\ref{chap:trace_fig:3}, and their difference for clean and adversarial examples suggests a likely chance of detecting perturbed images based on these maps.

Accordingly, the proposed method in this chapter uses the average local spatial entropy of an image as the final single measure to decide whether an attack has occurred or not. The average local spatial entropy $\bar{S}$ is defined as:
\begin{align}
\bar{S} = \frac{1}{K} \sum_{k} S_k
\label{chap:trace_eq:en_ave}
\end{align}
where $K$ is the number of patches on the complete feature response maps and $S_k$ shows the local spatial entropy as defined in Equation~\ref{chap:trace_eq:en} and depicted in the last row of Figure~\ref{chap:trace_fig:3}. The proposed detector makes the final decision by comparing the average local spatial entropy from Equation \ref{chap:trace_eq:en_ave} with a selected threshold to measure the spatial complexity of the feature response maps.

\section{Experimental Results}
\label{chap:trace_sec:experiments}

\begin{figure}[htb!]
  \centering
  \subfloat[Histogram]{{\includegraphics[width=.45\textwidth]{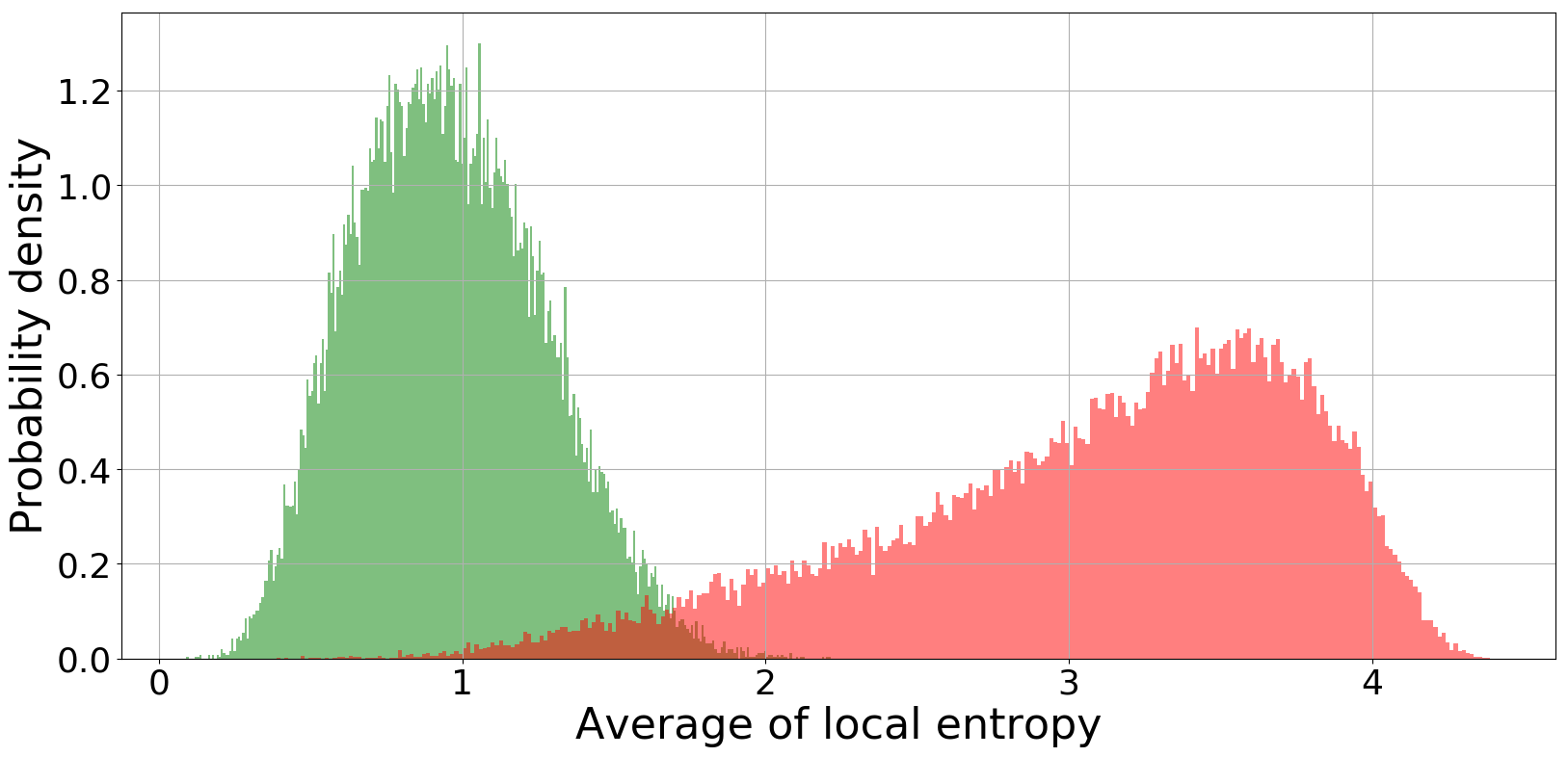} }}%
  \qquad
  \subfloat[ROC curves]{{\includegraphics[width=.45\textwidth]{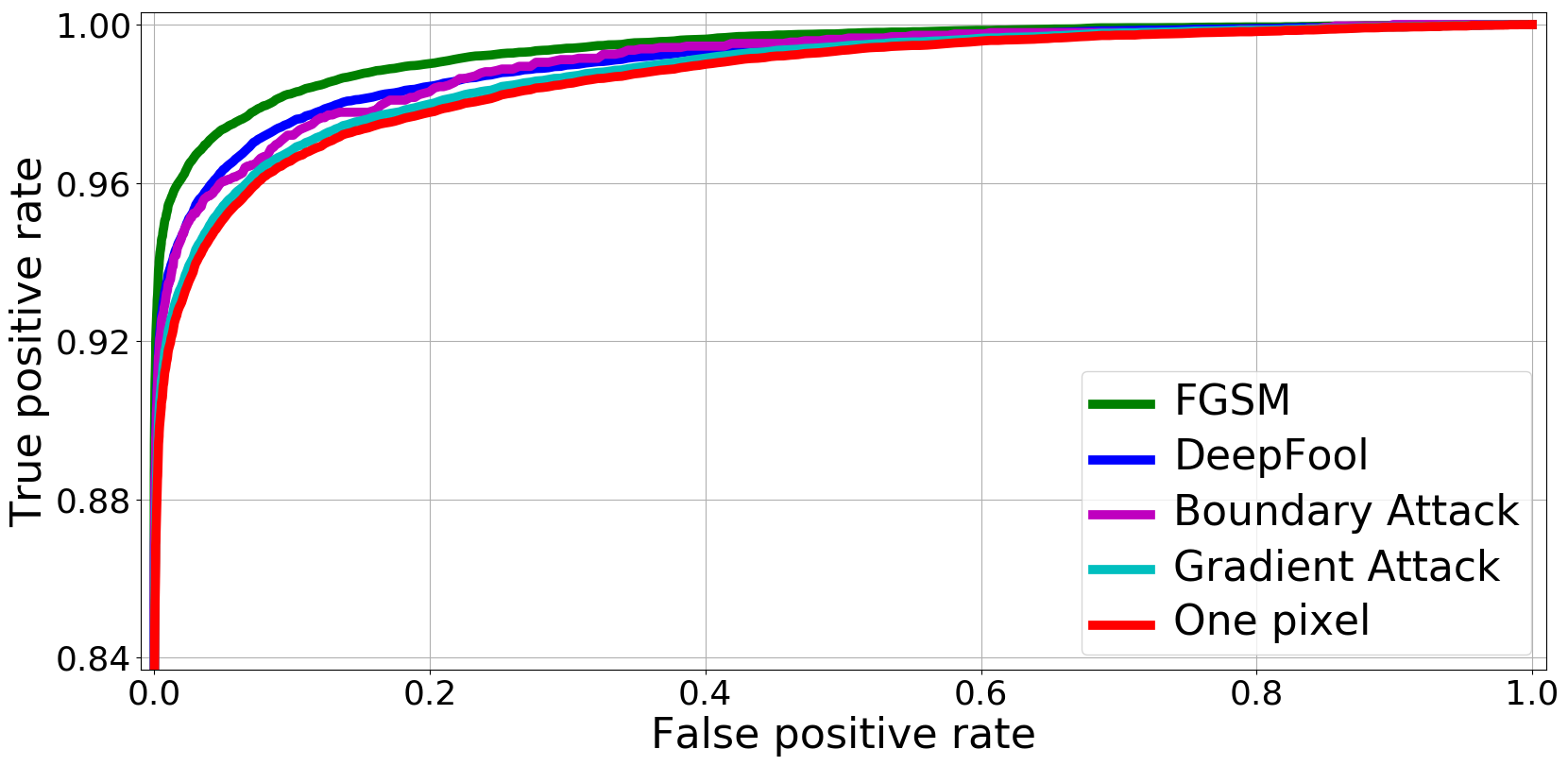} }}%
\caption{a) Distribution of average local spatial entropy in clean images (green) versus adversarial examples (red) as computed on the ImageNet validation set \cite{ILSVRC15}. b) Receiver operating characteristic (ROC) curve of the performance of the detection algorithm on different attacks (figure adopted from~\cite{amirian2018trace}).}
  \label{chap:trace_fig:hist_ROC}
\end{figure}

The first experiments visually compare the approximated distribution of the averaged local spatial entropy of feature response maps in clean and perturbed images to evaluate the value of the explained metric in Equation~\ref{chap:trace_eq:en_ave}. The validation set of ImageNet \cite{ILSVRC15} with more than $50,000$ images from $1,000$ classes is the subject of this study, and feature response maps are computed for the VGG19 model~\cite{simonyan2014very}. Perturbations for this experiment are computed using several different methods, and the distribution of average local spatial entropies is depicted for the fast gradient sign attack (FGSM). Figure~\ref{chap:trace_fig:hist_ROC} shows that the clean images are distinguishable from perturbed examples, although there is some overlap between the distributions.

Computing adversarial perturbations using evolutionary and iterative algorithms is demanding in terms of time and computational resources. To apply the proposed detector to a wide range of adversarial attacks, many images are randomly drawn from the validation set of ImageNet for each attack to evaluate the detection performance of the presented method. The selection of images for each attack is made sequentially by class and filename and comprises only the number of images per method that can be computed within a reasonable amount of time, using a reasonable number of resources (see Table~\ref{chap:trace_table:1}). The experiments are based on the FoolBox benchmarking implementation\footnote{\url{https://github.com/bethgelab/foolbox}}, and the attacks are computed using TitanX graphics processing unit (GPUs).

Figure~\ref{chap:trace_fig:hist_ROC}b presents the receiver operating characteristics (ROC) of the proposed detector; numerical evaluations are provided in Table~\ref{chap:trace_table:1}. Our detection method performs better for gradient-based perturbations compared to the single-pixel attack. Furthermore, Table~\ref{chap:trace_table:1} suggests that the best adversarial attack detection performance is achieved for FGSM and boundary attack perturbations, where the network confidences on the ground truth labels are changed the most after manipulating the images. This observation suggests that the proposed detector is more sensitive to stronger attacks in fooling the network with a more drastic effect on target class confidence. Using feature response maps, the proposed method detects more than $91\%$ of the perturbed images with a low false positive rate ($1\%$).

\begin{table}[htb]
\begin{center}
\resizebox{\textwidth}{!}{  
\begin{tabular}{l | c | c | c | c | c  c  c}
\toprule
\multirow{2}{*}{Adversarial attack} & \#Images & \multirow{2}{*}{Success rate} &  Ground truth   &  Target class & \multicolumn{3}{|c}{False positive rate} \\
 & (run time [days]) &   &  confidence  &  confidence & 1\% & 5\% & 10\% \\
\midrule
FGSM \cite{goodfellow2014explaining}  & $50,014$ (3) & $0.925$ & $\boldsymbol{0.022}$ & $\boldsymbol{0.588}$ & $\boldsymbol{0.954}$ & $\boldsymbol{0.974}$ & $\boldsymbol{0.983}$ \\
Gradient attack \cite{rauber2017foolbox} & $50,014$ (15) & $0.499$ & $0.052$ & $0.371$ & $0.922$ & $0.954$ & $0.969$ \\ 
One pixel attack \cite{su2017one}  & $50,014$ (32) & $0.620$ & $0.037$ & $0.463$ & $0.917$ & $0.951$ & $0.966$ \\  
DeepFool \cite{moosavi2016deepfool} & $47,858$ (42) & $0.606$ & $0.041$ & $0.446$ & $0.936$ & $0.963$ & $0.976$ \\
Boundary attack \cite{brendel2017decision} & $4,013$ (17) & $\boldsymbol{0.940}$ & $0.023$ & $0.583$ & $0.934$ & $0.960$ & $0.972$ \\
\toprule
\end{tabular}
}
\end{center}
\caption{The table describes the numerical evaluation of detection performance on different adversarial attacks. Column two gives the number of tested images and approximate elapsed run time. The success of an adversarial attack is defined if a perturbation changes the prediction. Columns four and five show average confidence values of the true (ground truth) and wrong (target) classes after a successful attack. Finally, the last columns show detection rates for different false positive rates (table adopted from~\cite{amirian2018trace}).}
\label{chap:trace_table:1}
\end{table}

\begin{table}[htb]
\begin{center}
\resizebox{\textwidth}{!}{  
\begin{tabular}{l | c | c | c | c  c  c}
\toprule
\multirow{2}{*}{Method} &  \multirow{2}{*}{Dataset} &  \multirow{2}{*}{Network}  &  \multirow{2}{*}{Attack} & \multicolumn{3}{|c}{Performance} \\
 &   &  &  & Recall & Precision & AUC \\\midrule
Uncertainty density estimation \cite{feinman2017detecting} & SVHN \cite{krizhevsky2009learning} & LeNet \cite{lecun1989backpropagation} & FGSM & - & - & $0.890$ \\
Adaptive noise reduction \cite{liang2017detecting} & ImageNet ($4$ classes) & CaffeNet & DeepFool & $0.956$ & $0.911$  & - \\
Feature squeezing \cite{xu2017feature} & ImageNet-1000 & VGG19 & Several attacks & $0.859$ & $0.917$ & $0.942$ \\ 
Statistical analysis \cite{grosse2017statistical} & MNIST & Self-designed & FGSM ($\epsilon=0.3$) & $\boldsymbol{0.999}$ & $\boldsymbol{0.940}$ & - \\ 
Feature response (our approach) & ImageNet validation & VGG19 & Several attacks & $0.979$ & $0.920$ & $\boldsymbol{0.990}$ \\
\toprule
\end{tabular}
}
\end{center}
\caption{This table describes the performance of similar state-of-the-art adversarial attack detection methods. The Area Under Curve (AUC) is the average value of all attacks in the third and last row (table adopted from~\cite{amirian2018trace}).}
\label{chap:trace_table:2}
\end{table}

In general, it is difficult to directly compare different studies on attack detection since they use a wide variety of neural network models, datasets, attacks, and experimental setups. Table~\ref{chap:trace_table:2} presents a short overview of the performances of current detection approaches. The approach presented in this chapter is most similar to the methods of Liang et al.~(\cite{liang2017detecting}) and Xu et al.~\cite{xu2017feature}. The detector proposed in this chapter outperforms both of the aforementioned methods based on the presented results in their work. However, it is not fully known if they used identical implementations and parameterizations of the attacks (e.g., a subset of images and learning rates for optimizing the perturbations). Similarly, adaptive noise reduction in the original publication~\cite{liang2017detecting} is only applied to four classes of the ImageNet dataset, and the method presented to detect adversarial attacks is implemented based on CaffeNet deep learning framework, which differs from our experimental setup.

\section{Discussion and Conclusion}
\label{chap:trace_sec:conclusions}
\begin{figure}[htb!]
     \begin{center}
     \begin{tabular}{ c  c  c  c}
     \toprule
     Original & Adversarial & Original & Adversarial \\ \midrule
     \includegraphics[width=0.14\textwidth]{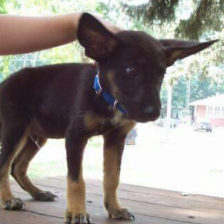} &             \includegraphics[width=0.14\textwidth]{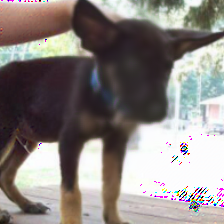} & \includegraphics[width=0.14\textwidth]{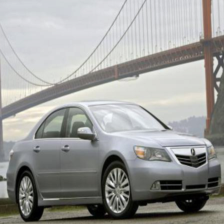} & \includegraphics[width=0.14\textwidth]{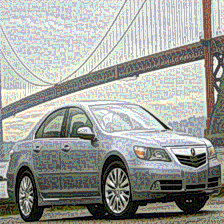}  \\ 
     \toprule
      \end{tabular}
      \caption{Successful adversarial examples created by DeepFool~\cite{moosavi2016deepfool} for binary and ternary classification tasks are only possible with noticeably visible perturbations (figure adopted from~\cite{amirian2018trace}).}
      \label{chap:trace_fig:6}
      \end{center}
\end{figure}

The results presented in this chapter demonstrate the relevance and importance of adversarial attacks and the necessity to improve the robustness of CNNs against such perturbations. However, preliminary experiments on binary (cat versus dog~\cite{parkhi2012cats}) and ternary (among three classes of cars~\cite{krause20133d}) classification tasks suggest that it is more challenging to find adversarial examples where marginal changes in the images are invisible to humans in such a setting. These tasks are proxies for the kind of few-class classification settings frequently arising in practice. Figure~\ref{chap:trace_fig:6} illustrates these experimental results.

This chapter offers an approach to detect adversarial attacks based on human-explainable feature response maps. The proposed method traces the effect of adversarial perturbations on the networks' focus region in original images, which inspired a simple yet robust approach for detecting adversarial attacks automatically. The proposed method is based on thresholding the averaged local spatial entropy of the feature response maps and detecting at least $91\%$ of state-of-the-art adversarial attacks with a low false positive rate on the validation set of ImageNet. However, the results are not directly comparable with state-of-the-art methods because of the diversity in the experimental setups and attack implementations.

Experimental results verify that feature response maps are informative in detecting specific failure cases in deep CNNs. Furthermore, the proposed detector uses the explainability of neural network decisions, an increasingly important topic for developing robust and reliable vision models. Future work, therefore, will concentrate on developing reliable and explainable image classification methods for practical use cases based on these preliminary results.


\chapter{Motion Compensation in Computed Tomography Using CNNs}
\chaptermark{Motion Compensation in CBCT Using CNNs}
\label{chap:motion}

This chapter presents this thesis's main applied contribution, motion artifact reduction, which enhances the quality of cone-beam computed tomography (CBCT) scans using 3D convolutional neural networks (3D-CNNs).
This application is relevant and exciting for two main reasons:
1) There is no analytical solution to the problem of motion compensation since reconstruction algorithms are developed based on the assumption of measurements from a constant volume.
2) Motion artifacts are relevant in CBCT scans because of their long acquisition time, and using CBCTs demonstrates improving cancer therapy via adaptive dose calculation and patient positing. 

This chapter offers a novel deep-learning (DL) based approach that significantly reduces motion artifacts and improves scan quality.

Because motion artifact reduction has no analytical solution, 3D deep convolutional neural networks (3D-CNNs) are employed as pre-and/or post-processing steps during CBCT reconstruction to target motion artifacts using a data-driven approach. The method described in the following paragraphs is performed either using the analytical Feldkamp-Davis-Kress (FDK) with filtered backprojection (FBP) reconstruction method or using the iterative algebraic reconstruction technique (iCBCT/ART). Based on refined UNet architectures, the neural networks are trained end-to-end via supervised learning. The dataset used in this chapter is generated from 4D computed tomography (CT) scans of lungs containing ten motion phases between inhalation and exhalation and patients' breathing curves with negligible motion artifacts. The training ground truth volumes are the averaged volume over all phases in 4D-CT or the volume at the average amplitude of the breathing curve. The trained networks are validated using real-world CBCT scans and quantitative image quality metrics. In addition, a qualitative evaluation from clinical experts is performed.

The novel approach in this chapter can generalize to unseen data and yield noticeable reductions in motion-induced artifacts and improvements in image quality compared with state-of-the-art CBCT reconstruction algorithms (up to 6.3dB and 0.19 improvements in PSNR and SSIM, respectively). The experimental findings from the simulation are confirmed by a clinical evaluation of real-world patients' scans (clinical experts reported at least a noticeable change in motion reduction over standard reconstruction in $74\%$ of cases). To the best of our knowledge, this is the first time that inserting deep neural networks as pre- and post-processing plugins in the existing CBCT reconstruction pipeline and end-to-end training demonstrates significant improvement in imaging quality and reducing motion artifacts in CBCT scans. This chapter is mainly adopted from the research published in~\cite{amirian2023mitigation}, licensed under CC BY-NC-ND 4.0~\footnote{© 2023 The Authors: \url{https://creativecommons.org/licenses/by-nc-nd/4.0}}. 


\section{Introduction}\label{chap:motion_sec:introduction}

Computed tomography (CT) has become a versatile radiology and radiation therapy imaging technique. It obtains detailed 3D scans of the human body for diagnostics and planning therapies. 
Cone-beam CT (CBCT), in particular, is used for reconstructing scans from measurements of radiation therapy treatment devices (linear accelerators). CBCT reconstruction techniques in image-guided radiation therapy (IGRT)~\cite{JAFFRAY20021337} and interventional radiology provide high spatial resolution in a cost-efficient manner~\cite{doi:10.3109/0284186X.2011.590525}.
IGRT treatment is performed in up to $40$ fractions. For each fraction, it is necessary to obtain the image of the day in order to position the patient accurately. Novel applications of CBCT imaging in IGRT, such as online adaptive replanning~\cite{yoon2020} or daily treatment planning and dose calculation~\cite{JAREMA2019E719}, are very well-known in the scientific community. 

There are two leading families of reconstruction algorithms used in modern CT scanners: ($i$) analytical techniques and ($ii$) algebraic reconstruction. The first group is based on filtered backprojection (FBP), most prominently represented by the Feldkamp-Davis-Kress (FDK) method~\cite{Feldkamp:84}. 
The second group consists of the algebraic reconstruction techniques (ART) family, which is based on reformulating and solving the reconstruction problem through an iterative optimization technique. Although the development of algebraic methods started in the late 1960s~\cite{osti_4102886}, they have only been deployed on CT scanners in the last 15 years\cite{safire} mainly because of their high computational cost. In recent years, this problem has disappeared due to the high availability and relatively low cost of GPUs. 

Iterative CBCT (iCBCT) reconstruction algorithms based on the ART family introduced in~\cite{paysan2018iterative} for Varian's Halcyon and TrueBeam addressed the need for superior image quality in terms of better noise suppression and improved contrast as compared with FDK and demonstrated in~\cite{GARDNER2019390,https://doi.org/10.1002/acm2.12747,doi:10.1177/1533033818823054,WASHIO2020109293}. However, there is still room for improvement in these methods, if they are to tackle real-world artifacts which are not a part of theoretical and analytical solutions. 

Imaging artifacts~\cite{Schulze2011ArtefactsIC} are still a prevalent feature in CBCT reconstruction. The main sources of artifacts are ($i$) electrical and photon count noise, ($ii$) photons from scattered X-rays, ($iii$) extinction and beam hardening effects (e.g., due to metal implants), ($iv$) approximations in the reconstruction (due to finite beam width and detector pixel size), ($v$) aliasing (due to finite pixel size and cone beam divergence), ($vi$) ring artifacts (due to defect or miscalibrated detector elements), and ($vii$) patient motion. 
 
Motion artifacts arise since the reconstruction methods assume that the scanned patient is stationary.
However, periodic respiratory and cardiac motions and non-periodic motions, such as gas bubbles in the abdomen caused by the digestive system, lead to acquiring projections from different states of motion in the body. Patients' motion leads to the appearance of evident and undesirable, typically streak-shaped artifacts in reconstructed scans.
 
The following motion compensation strategies are used so far in IGRT clinical routine: ($i$) 4D or gated CBCT based on an external breathing signal~\cite{Dillon_2020}, ($ii$) breath hold CBCT based on an external breathing signal and potential patient feedback, ($iii$) assisted breathing based on a ventilator system~\cite{doi:10.1080/0284186X.2020.1856408}, ($iv$) abdominal compression devices applied to the patient~\cite{doi:10.1080/0284186X.2022.2073186}, (v) internal breathing signal extraction~\cite{Mohd_Amin_Mokri_Ahmad_Ismail_Abd_Rahni_2021}.

This chapter presents a novel approach to mitigate motion artifacts in CBCT reconstruction using deep learning (DL). The CBCT reconstruction is integrated within a DL pipeline, where convolutional neural networks are employed as pre -and/or post-processing steps. These networks act on either the 2D X-ray projections (preprocessing), the reconstructed 3D volume (post-processing), or both. Next, the models are trained end-to-end in a supervised fashion using CBCT scans containing simulated motion artifacts and motion-free scans as ground truth. Finally, the trained models are validated quantitatively using various scan quality-related numerical metrics, and on an independent real-world patient CBCT dataset developed through qualitative clinical feedback.

\newpage
\section{Related Work}\label{chap:motion_sec:relatedwork}


Much work has been done~\cite{Schulze2011ArtefactsIC,boas2012,mar40ieee} regarding the characterization and mitigation of the various kinds of artifacts that negatively impact image quality in CT and CBCT scans.
In recent years, DL-based approaches have shown promising results, including applications for IGRT and adaptive radiation therapy~\cite{paysan_annpr2020}. 

W\"urfl et al.~\cite{wurfl2018} mapped the components of the FBP algorithm into a neural network by introducing a novel DL-enabled cone-beam back-projection layer. A forward projection operation efficiently computes the backward pass of the back-projection layer. This approach thus permits joint optimization and correction in the projection and volume domain. 
Moreover, Maier et al.~\cite{MaierLearningOperators2019} argued that implementing prior knowledge (such as the back-projection operation) in the form of (differentiable) known operators into DL algorithms reduces training error bounds while reducing the number of free parameters.

Limited-angle CT is employed to reduce acquisition time and decrease the radiation dose, which leads to a degradation of image quality and introduces sparseness artifacts. To overcome these issues, Wang et al. recently presented an encoder-decoder architecture based on the UNet model~\cite{unet_2015} to reconstruct high-quality scans with fewer projections. 
A UNet processes scans reconstructed by the simultaneous algebraic reconstruction technique (SART) to improve the imaging quality~\cite{sart_1984}. Experiments on chest and abdomen CT scans demonstrated the superiority of the proposed methods over existing approaches. Similarly, Schnurr et al. proposed UNet-based networks to correct limited-angle artifacts in circular tomosynthesis scans~\cite{Schnurr2019}.

DL-based approaches demonstrate success in metal artifact reduction (MAR)~\cite{Park2018,Zhang2018}. Lin et al. introduced a dual-domain architecture (DuDoNet) to jointly compensate for metal-induced artifacts in both projection and volume domains~\cite{lin2019dudonet}. Experimental results on the DeepLesion CT dataset~\cite{deep_lesion_db} showed that the proposed method outperformed both traditional and other DL-based approaches.
An improved model, DuDoNet++, was proposed to compensate for over-smoothed and distorted image reconstruction and led to improved artifact correction, especially for large metallic objects~\cite{dudonet_pp}. Furthermore, there have been recent efforts in MAR using unsupervised approaches, as explained in~\cite{Liao2020}.
The U-DuDoNet model~\cite{unpaired_dudonet} directly models the artifact generation and compensation process in both the projection and volume domains. More recently, the interactive~\cite{wang2021indudonet} and interpretable~\cite{wang2022idol} versions of DuDoNet have been introduced to improve the interpretability and enhance the interaction between projection and volume domains. 

DL-based approaches have been employed to improve sparseness artifacts generated by low-dose CT reconstruction~\cite{han2016deep,Jin2017,zhang2018sparse,Kofler2018}. Chen et al. present the AirNet model to fuse analytical and iterative CT reconstruction and integrate them into DL to improve sparse-data 3D and 4D CBCT reconstruction~\cite{Chen2020,chen20204d}.
In the projection domain, DL-based models can also correct signal degradation caused by X-ray photons scattering within the patients' body~\cite{Maier2018,Erath2019}.

Finally, motion artifact compensation using DL has received comparatively less interest. Paysan et al. present an initial study of a UNet-based artifact reduction method, but only in the volume domain~\cite{aapm2019conv}.
Su et al. used UNet architectures to reduce simulated motion artifacts in head CT scans based on simple simulated rigid (translations, rotations, oscillations) transformations. Finally, Lyu et al. used recurrent neural networks for cardiac motion artifact reduction in magnetic resonance imaging (MRI)~\cite{lyu2021cine}.


\section{Materials and Methods}
\label{chap:motion_sec:methods}
This section presents the preliminary knowledge and the related theory which lays the foundation of this chapter's main contributions and findings. First, this section starts with a more detailed explanation of the CBCT reconstruction techniques used in this chapter. Secondly, the motion simulation framework is explained, followed by a discussion on the simulated and real-world datasets and the clarification of their differences.

\subsection{CBCT Reconstruction}
\label{chap:motion_sec:reco}
Both analytical and iterative methods are considered for the reconstruction of 3D CBCT volumes from 2D cone-beam projections in this research work. \emph{Feldkamp-Davis-Kress}~\cite{Feldkamp:84} (FDK) is an analytical reconstruction method based on filtered back-projection (FBP). Although the \emph{Tuy} data-sufficiency conditions~\cite{tuy1983inversion} are not met for circular trajectories of a cone-beam source, FDK provides a fast and reliable analytical approximation of the inverse Radon transform, and it has become a golden standard for 3D CBCT reconstruction~\cite{BU08}.
\emph{Ram-Lak} filter compensates for the radial non-uniformity of the sampling density in FDK. Moreover, \emph{half-fan weighing} is necessary to avoid data duplication for the datasets acquired with half-fan geometry. The projections are acquired from the full 360$^\circ$ trajectory. However, the detector is shifted against the gantry to one direction to increase the field of view in half-fan geometry. Half-fan weighing is followed by \emph{cosine weighting} to decrease the longitudinal full-off effect due to the cone-beam geometry. Finally, the projections are down-sampled so that their resolution matches the cut-off frequency requirement given by the target resolution of the reconstructed volume.

Besides FDK, in this chapter, the iterative \emph{algebraic reconstruction technique} (ART) originally based on Kaczmarz algorithm~\cite{karczmarz1937angenaherte} is used for iterative CBCT (iCBCT) reconstruction. This method approximates the volume $\mathbf{f}$ by an iterative optimization of the data-fidelity cost function: $\parallel\mathbf{Af} - \mathbf{p}\parallel^2$, where $\mathbf{A}$ and $\mathbf{p}$ represent the forward-projection operator and projections in the attenuation space, respectively. In each iteration $k$, an update of the actual volume estimation is computed through back-projecting the gradient of the cost function, i.e.\,$\sum_\alpha \mathbf{A}^\top([\mathbf{Af}_k]_\alpha - \mathbf{p}_\alpha)$ where $\mathbf{p}_\alpha$ and $[\mathbf{Af}_k]_\alpha$ denote the projection under angle $\alpha$ and corresponding forward-projection of actual estimated volume $\mathbf{f}_k$, respectively, and $\mathbf{A}^\top$ represents the back-projection operator. One of the advantages of the iterative methods is allowing a straightforward integration of prior knowledge into the reconstruction process through a regularization term to augment the cost function during optimization.
The implementation used in this chapter employs the edge-preserving \emph{total variation} regularization, which helps to reduce the noise and cone-beam artifacts in the areas far from the iso-center.

In order to significantly reduce the computational cost, the GPU implementation of ART is further accelerated through the following approaches: First, the version of ART known as \emph{simultaneous ART} (SART) is used where the volume is updated in parallel for each input projection. Next, \emph{ordered subsets} (OS)~\cite{Kim:2015} and Nesterov \emph{momentum method}~\cite{nesterov2005smooth} are employed. 
Finally, a destination-driven approach~\cite{keck2009high} is employed in the forward projection of ART and backward projection of both ART and FDK. Further details about the TV-regularized OS-SART with momentum can be found in~\cite{peterlik2021reducing}, where the method is presented as a part of the iCBCT algorithm deployed clinically in Varian products.



\subsection{Motion Simulation}
\label{chap:motion_sec:motionsim}




It is necessary to simulate motion for volumes with available ground truth to train the models using supervised learning; hence, the motion simulation method aims to generate synthetic datasets of CBCT volumes with motion artifacts. The motion simulation method starts from the phase-gated 4D CT scans described in Section~\ref{chap:motion_subsec:datasets} and a set of recorded breathing curves. The method profits from the Deeds~\cite{HE13} algorithm to perform a deformable registration between CT volumes of the end-inhale and end-exhale phases and to create a patient-specific deformation vector field (DVF).

\begin{figure}[htb!]
    
    \begin{center}
        \includegraphics[width=0.49\linewidth]{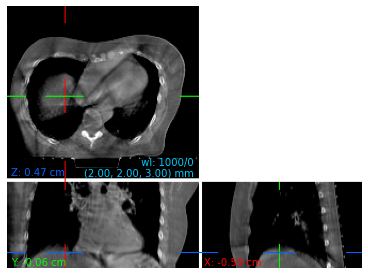}
        \includegraphics[width=0.49\linewidth]{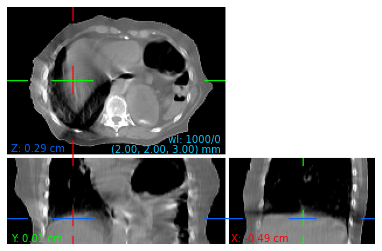}
        \caption{Motion Artifacts. Left: CBCT scans with motion artifacts from the test dataset. Right: Scan with artificially produced motion artifacts from the motion simulation. The scans are presented in HU with W/L=1000/0 (figure adopted from~\cite{amirian2023mitigation}).}
        \label{chap:motion_fig:compareMA}
    \end{center}
\end{figure}

A reconstructed CT scan, its DVF, and the patients' breathing curve are sufficient requirements to simulate the motion during the CBCT acquisition. The simulation method deforms the CT volumes by interpolating the DVFs according to the breathing amplitude. It creates a forward projection at each angular step by matching its timestamp with the relevant amplitude in the breathing curve. Each projection acquired through the described motion simulation method corresponds to a different respiratory state. Then, the CBCT volume is reconstructed using either the FDK or iCBCT reconstruction algorithms. Motion artifacts are evident in the volumes reconstructed from the explained motion-simulated CBCT projection acquisition technique. Figure~\ref{chap:motion_fig:compareMA} shows an example of typical motion artifacts created by patient motion in real-world (clinical) CBCT dataset (test dataset, see Section~\ref{chap:motion_subsec:datasets}) side-by-side with the emulated motion artifacts from our motion simulation.

The supervised learning approach presented in this chapter requires ground truth volumes without motion artifacts. The ground-truth volumes correspond either to a fixed motion state (average amplitude) or the average of all deformed volumes (average volume). Moreover, data augmentation is a crucial component of supervised learning pipelines to enhance optimization performance. Augmentation is realized by using different breathing curves to simulate motion with the same dataset of CT scan and DVF in this research work.

\vspace{-1cm}
\subsection{Datasets}\label{chap:motion_subsec:datasets}
For the training and validation of the different methods, a dataset of thoracic 4D CT scans from $80$ patients is split into fractions of $60\%$, $20\%$, and $20\%$ for training, validation, and testing, respectively. They were provided as input to the motion simulation described in Section~\ref{chap:motion_sec:motionsim}. The patient-specific anatomical correct deformations were extracted from the end inhale and exhale out of the ten breathing phases. To simulate plausible motion patterns during a virtual CBCT acquisition and to augment the training dataset, 400 recorded breathing curves were obtained via Varian's real-time position management (RPM) system. 

For testing the developed methods on real-world (clinical) patient CBCT scans, a dataset of 77 Halcyon scans was employed. All pre-processed projection data and reconstructed volumes were given at the same size, resolution, and geometry to ensure consistency: The projection size is 320x80 pixels (with a resolution of $1.344\times4.032$ mm), and the volume size is $256\times256\times48$ voxels ($2\times2\times3$ mm). The source-to-imager distance is 154 cm with a detector offset of 17.5 cm.

\section{Supervised Learning for Motion Artifact\protect\\Reduction}
This section presents the necessary underlying basics to start with supervised learning. First, the DL-enabled framework for reconstruction and refinement models, including UNets in the projection- and volume domains, is explained. Second, evaluation metrics for numerical analysis of simulated data are presented, and the section concludes by describing the experimental setup, including the hardware.    
\subsection{DL-Enabled CBCT Reconstruction}
\label{chap:motion_subsec:method}
Motion leads to inconsistencies in the acquired projections, which appear as artifacts in the volume domain after reconstruction. Therefore, motion corrections can be, in principle, applied before and/or after reconstruction. The models, estimating these correction steps, are implemented as trainable neural network architectures derived from 3D refined UNet architectures.

The reconstruction algorithm used is either FDK or iterative CBCT (iCBCT) reconstruction, as discussed in Section~\ref{chap:motion_sec:reco}. These algorithms are implemented based on forward- and back-projection layers implemented with custom compute unified device architecture (CUDA) code and interfaced as PyTorch modules. The analytical solution using filtered back-projection, inspired by FDK, is differentiable. Therefore, it is possible to back-propagate the gradient through this module and simultaneously optimize in both projection and volume domains, called dual-domain optimization. Dual-domain optimization requires a differentiable reconstruction method such as FDK and is not practical for iterative techniques.

The supervised learning approach uses the simulated motion dataset (Section~\ref{chap:motion_sec:motionsim}) for training the motion compensation networks, where the loss is calculated in the volume domain. The ground truth is either calculated as the motion-averaged volume (``average volume") or given as the volume corresponding to the fixed motion state matching the average breathing signal amplitude (``average amplitude").
The networks are validated on the validation and test portions of the simulated motion dataset and an independent real-world test dataset containing real-world clinical patient CBCT scans (see Section \ref{chap:motion_subsec:datasets}). In detail, the reconstruction pipeline consists of the following components:

\begin{figure}[htb!]
\centering
   \includegraphics[width=\linewidth,height=10.5cm]{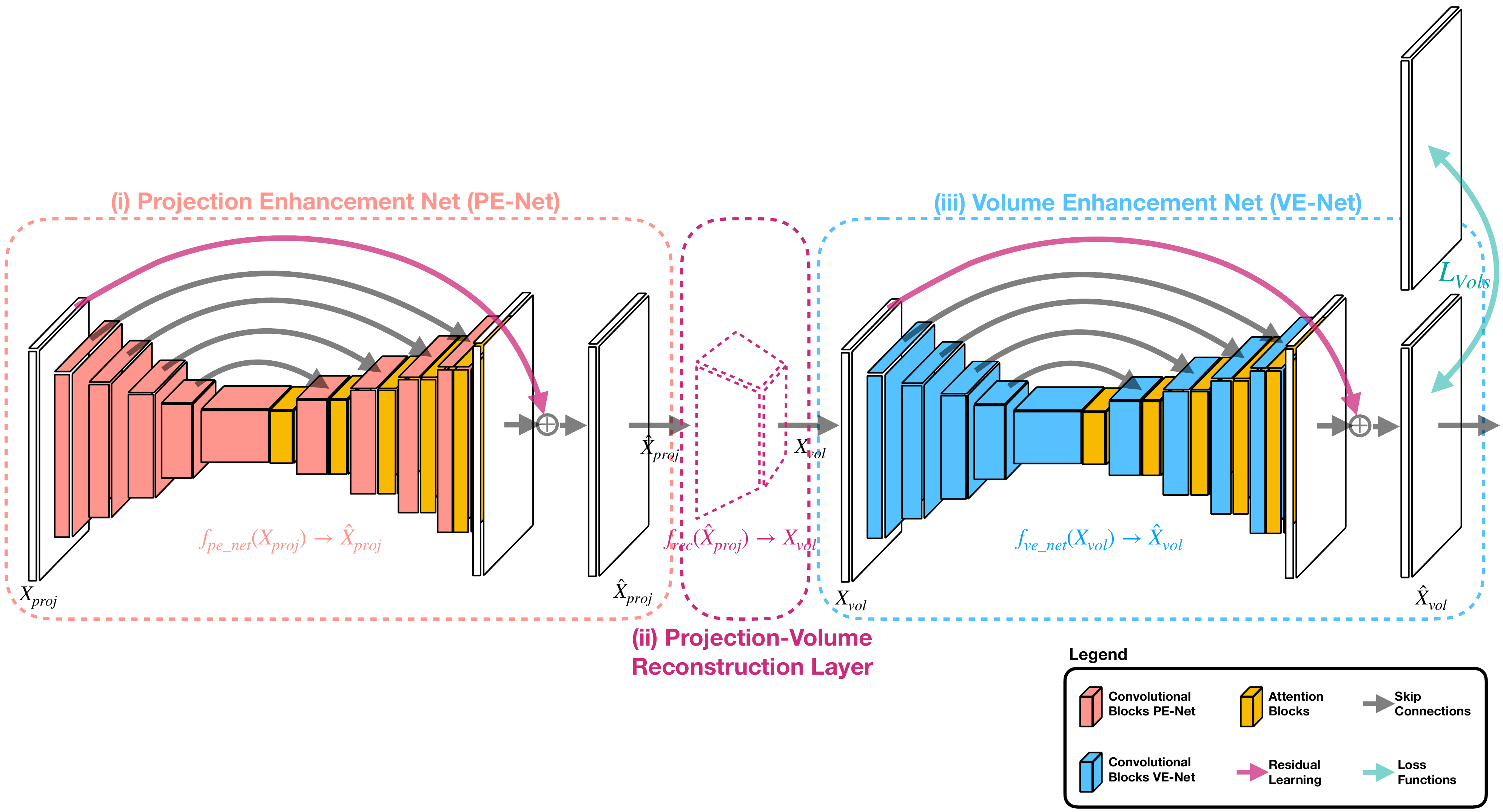}
    \caption{The architecture of the proposed dual-domain model for end-to-end optimization consists of the following components: ($i$) a projection enhancement network (PE-Net), ($ii$) a projection-to-volume reconstruction layer, and ($iii$) a volume enhancement network (VE-Net) (figure adopted from~\cite{amirian2023mitigation}).}
    \label{chap:motion_fig:architecture}
    \vspace{0.5cm}
\end{figure}

\textbf{Projection Enhancement Network (PE-Net)}: 
A convolutional neural network based on UNet architectures, explained in more detail in the next section, is deployed to mitigate motion-induced artifacts in the projection space. PE-Net receives as input the acquired projections $\{\mathcal{X}_{proj} \in \mathcal{R}^{H_{p}\times{W_{p}}\times{C_{p}}}\}$, and enhances these projections $\{\mathcal{\hat{X}}_{proj} \}$, 
i.e. ${\mathcal{\hat{X}}_{proj}} \approx f_{pe\_net}(\mathcal{X}_{proj})$ to remove motion artifacts in the projection space. Here, $H_{p}\times{W_{p}}\times{C_{p}}$ denotes the projection dimensions in terms of height, width, and number of projections.


\textbf{Projection-to-Volume Reconstruction Layer}: 
The projection-to-volume reconstruction layer $f_{rec}(\cdot)$ receives as input the (enhanced) projections $\{\mathcal{\hat{X}}_{proj}\}$ and outputs a reconstructed volume $\{\mathcal{X}_{vol} \in \mathcal{R}^{H_{v}\times{W_{v}}\times{C_{v}}}\}$, i.e. $f_{rec}(\mathcal{\hat{X}}_{proj})\rightarrow{\mathcal{X}_{vol}}: \mathcal{R}^{H_{p}\times{W_{p}}\times{C_{p}}} \rightarrow \mathcal{R}^{H_{v}\times{W_{v}}\times{C_{v}}}$, where $H_{v}\times{W_{v}}\times{C_{v}}$ represents the volume's height, width, and number of slices. 
This layer corresponds to the regular FDK or iCBCT reconstruction (Section \ref{chap:motion_sec:reco}).


\textbf{Volume Enhancement Network (VE-Net)}: 
The VE-Net $f_{ve\_net}(\cdot)$ is responsible for enhancing the reconstructed volume and compensating motion artifacts in the volume domain. As output, the VE-Net produces an enhanced volume $\{\mathcal{\hat{X}}_{vol} \in \mathcal{R}^{H_{v}\times{W_{v}}\times{C_{v}}}\}$, i.e. ${\mathcal{\hat{X}}_{vol}} \approx f_{ve\_net}(\mathcal{X}_{vol})$.

Our proposed dual-domain (end-to-end) model, shown in Figure~\ref{chap:motion_fig:architecture}, combines the above components for motion correction in both projection- and volume domains. It consists of three different modules: ($i$) a projection enhancement network (PE-Net), a ($ii$) projection-to-volume reconstruction layer, and a ($iii$) volume enhancement network (VE-Net).

The following paragraphs describe the different model blocks of the proposed architecture. Note that these blocks are used in both the projection enhancement (PE-Net) and volume enhancement (VE-Net) networks.

\textbf{Encoder Blocks}: The encoder blocks of the presented architecture in Figure~\ref{chap:motion_fig:architecture} consist of four similar submodules including 3D a convolutional layer with filters of size $3\times3\times3$, followed by an instance normalization~\cite{ulyanov2016instance}, the Swish activation function~\cite{ramachandran2017swish} and a 3D max-pooling layer of size $2\times2\times2$. The number of convolutional filters in the first block is doubled for every next layer. Hence, the latent representations of the input volume have a larger number of channels but a smaller spatial size with a higher receptive field after the first layer.      

\textbf{Decoder Blocks}: The decoder block aims at computing the motion corrections from latent representations and has four submodules starting with a trilinear upsampling followed by 3D convolutions of size $3\times3\times3$, instance normalization, and the Swish activation function. The number of convolutional filters is halved after each layer to make the entire model's architecture symmetric. 

\textbf{Attention mechanisms}: To further compensate for motion artifacts, the models rely optionally on attention mechanisms. More precisely, as part of the bottleneck and decoder blocks of both PE-Net VE-Net, there are channel-wise and spatial-wise attention layers~\cite{woo2018cbam} in 3D. The corresponding input feature maps are multiplied at each decoder layer with the generated attention maps to refine the original features. The model can focus on more relevant features using these attention layers. Models including attention layers, are denoted by ``Attn." in Table~\ref{chap:motion_tab:results}.

\textbf{Residual Learning and ResUNet}: Using residual learning is crucial to simplifying the learning task and improving the convergence speed. The architecture depicted in Figure~\ref{chap:motion_fig:architecture} uses two components to enhance the gradient flow and simplify the learning task. First, the proposed architecture profits from a direct residual connection from input to output (``residual learning") to optimize the required correction instead of reconstructing the ground truth.
The proposed architecture optionally includes internal residual connections between the input and output of the convolutional layers to improve the gradient flow as described in~\cite{res-unet}. Networks including ``ResUNet" layers are labeled as such in Table~\ref{chap:motion_tab:results}.





\subsection{Evaluation Metrics}\label{chap:motion_subsubsec:metrics}




The experimental results of motion compensation on the simulated dataset are reported based on numerical performances using several quantitative metrics~\cite{wang2004image} sensitive to the similarity of pairs of volumes $(x,x')$. The evaluation metrics are computed by summing up the differences over all components, voxels in volumes, as follows:

\begin{itemize}
\item root mean squared error: $\mathrm{RMSE}=\sqrt{\mathrm{MSE}}$\\
where $\mathrm{MSE}(x,x') =  \frac{1}{N} \sum_i \parallel x_i - x'_i \parallel^2 $
\item peak signal-to-noise ratio: $\mathrm{PSNR}=10\log_{10}\left(\frac{\mathrm{MAX}^2}{\mathrm{MSE}}\right)$
\item structural similarity index ($SSIM$)\cite{wang2004image}
\end{itemize}
%
%
%
%
%
%
In addition, Table~\ref{chap:motion_tab:results} reports the mean and standard deviation of the difference image $x-x'$ used for reducing (correcting) the motion artifacts.
All metrics are calculated in Hounsfield units (HU) from pairs of uncorrected or corrected body-masked volumes and their corresponding ground truth counterparts.

\subsection{Experimental Details}
This section describes the experimental setup, architectural variants, optimization settings, and implementation details used for training based on a motion-simulated dataset. 

\textbf{Experimental Setup}: The volume size is $256\times{256}\times{48}$ voxels based on the neural network architectures to optimize computational and memory allocation costs. Based on the training dataset discussed in Section~\ref{chap:motion_subsec:datasets}, $720$ projections are used per scan for training, and motion artifacts in CBCT scans are computed using motion simulation introduced in Section~\ref{chap:motion_sec:motionsim}. The reconstruction and forward-projection geometry are selected to match the real-world test dataset, which is used for clinical evaluation in Section~\ref{chap:motion_sec:results}.


\textbf{Data Augmentation}: 
Five different patient breathing curves per CT scan are added for motion simulation from each original CT scan in the training dataset.
Data augmentation through various breathing curves led to a considerable boost in the final performance of our motion correction models. 

\textbf{Model Architecture}: The baseline model, initially considered for motion correction, is a UNet with residual learning from input to output as depicted in Figure~\ref{chap:motion_fig:architecture}. A plain UNet~\cite{unet_2015} architecture without residual connections is already sufficient for correcting the artifacts in volume space; however, residual learning is necessary for the more complicated tasks, including correcting the projections and dual-domain optimization. Therefore, all of our models include residual learning. The baseline UNet model has a depth of $4$ and has $32$ filters in the first layer. The number of filters doubles after every layer up to the middle (model's bottleneck), and the architecture is reverted afterward. The same architecture is used for both PE-Net and VE-Net. For dual-domain optimization, two such models form the architecture together. For PE-Net, the models process the projections in chunks of $192$ due to memory limitations. PE-Net and VE-Net are extended with internal residual connections (``ResUnet") and/or channel-spatial attention (``Attn.") for different experiments presented in the next section.   

\textbf{Implementation and Optimization Settings}: The models used for motion compensation are implemented using the PyTorch~\cite{NEURIPS2019_9015} framework. The experiments were performed on NVIDIA V100 (A100) GPUs with $32 \ (40)$ GB of VRAM. Both projections and volumes are normalized using constant coefficients per dataset to the approximate range of zero and one before optimization. The loss function for optimizing the models is the difference between the predicted and reconstructed volume as computed by the $\ell_1-norm=\sum_i \parallel x_i - x'_i \parallel$. The AdamW~\cite{loshchilov2018decoupled} optimizer is used with a learning rate of $1.41 \cdot 10^{-6}$ and weight decay of $1.87 \cdot 10^{-8}$ in the projection domain, and a learning rate of $1.11 \cdot 10^{-4}$ and weight decay of $1.39 \cdot 10^{-8}$ in the volume domain. These parameters are the results of a joint hyperparameter sweep with other parameters, such as a number of convolutional filters, kernel size, and convolutional dilation. This experiment's batch size is $1$ (due to GPU memory limitations), and training continues for a total number of $300$ epochs. The model that reduces the validation loss the most during the training is selected for testing.

%



\section{Experimental Results}
\label{chap:motion_sec:results}

This section presents the experimental quantitative and qualitative results obtained by applying DL-based motion reduction techniques using 3D convolutional neural networks. First, the quantitative results obtained with the test portion of the simulated dataset are presented. Second, the qualitative results based on a clinical evaluation of the real-world test dataset are discussed. 


\subsection{Quantitative Results}

\begin{table}[p]
\centering
\resizebox{\textwidth}{!}{ 
\begin{tabular}{l|cccc}
\toprule
Model Architecture & RMSE $\downarrow$ & PSNR (dB) $\uparrow$ & SSIM $\uparrow$ & Mean±stdev \\
\hline

%
\multicolumn{5}{c}{\textbf{Baseline Performance of Average Volume}} \\
\hline
FDK                  & 77.8875 & 28.3802 & 0.8086 & - \\
iCBCT                & 76.2560 & 28.6741 & 0.8701 & - \\
\toprule
\multicolumn{5}{c}{\textbf{Baseline Performance of Average Amplitude}} \\
\hline
FDK                  & 86.9695 & 27.5059 & 0.7992 & -  \\
iCBCT                & 106.5914 & 25.6087 & 0.7304 & - \\
\hline


\hline
\multicolumn{5}{c}{\textbf{Volume Domain (Average Volume)}} \\
\hline

3D-UNet (FDK)           & \textbf{38.27(-39.62±9.06)} & \textbf{34.72(6.34±1.45)} &  \textbf{0.9585(0.1499±0.0412)} &  0.0154±38.2148   \\ 
3D-ResUNet (FDK)        & 39.86(-38.03±10.53) & 34.32(5.94±1.63) &  0.9495(0.1410±0.0457) &  -8.2486±38.8685 \\
3D-ResUNet+Attn.(FDK)   & 39.65(-38.24±8.58) & 34.35(5.97±1.17) &  0.9559(0.1473±0.0406) &  -1.9394±39.5164  \\
\hline
3D-UNet (iCBCT)\textsuperscript{\textdagger}         & \textbf{44.20(-32.05±14.65)} & \textbf{33.32(4.65±1.79)} &  \textbf{0.9481(0.0780±0.0400)} &  -3.7927±43.9936  \\
3D-ResUNet (iCBCT)      & 44.80(-31.46±14.67) & 33.22(4.54±1.80) &  0.9464(0.0763±0.0385) &  -1.9903±44.7111  \\
3D-ResUNet+Attn.(iCBCT) & 45.75(-30.50±15.01) & 33.05(4.37±1.89) &  0.9377(0.0676±0.0406) &  -6.0158±45.2901 \\
\hline

\hline
\multicolumn{5}{c}{\textbf{Volume Domain (Average Amplitude)}} \\
\hline

3D-UNet (FDK)           & 51.67(-35.30±11.08) & 32.10(4.59±1.10) &  0.9410(0.1418±0.0431) &  -3.5407±51.4552  \\ 
3D-ResUNet (FDK)        & \textbf{51.28(-35.69±11.87)} & \textbf{32.14(4.63±1.16)} &  \textbf{0.9417(0.1425±0.0432)} &  -2.9049±51.1370 \\
3D-ResUNet+Attn.(FDK)   & 51.87(-35.10±11.78) & 32.03(4.52±1.15) &  0.9326(0.1335±0.0456) &  -6.9976±51.2475  \\
\hline
3D-UNet (iCBCT)\textsuperscript{\textdagger}        & \textbf{55.42(-51.17±11.50)} & \textbf{31.42(5.81±1.33)} &  \textbf{0.9300(0.1996±0.0656)} &  0.7139±55.2177  \\
3D-ResUNet (iCBCT)      & 55.76(-50.83±12.06) & 31.35(5.75±1.39) &  0.9282(0.1979±0.0634) &  -4.0567±55.4900  \\
3D-ResUNet+Attn.(iCBCT) & 58.78(-47.81±11.28) & 30.88(5.27±1.28) &  0.9131(0.1828±0.0598) &  -11.9311±57.1327  \\
\hline



\multicolumn{5}{c}{\textbf{Projection Domain (Average Volume)}} \\
\hline
3D-UNet (FDK)          & 73.88(-4.01±1.88) & 28.89(0.51±0.33) &  0.8654(0.0569±0.0165) &  3.8085±73.5703 \\ 
3D-ResUNet (FDK)       & 67.91(-9.98±4.86) & 29.68(1.30±0.78) &  0.8931(0.0845±0.0224) &  -1.2820±67.7729 \\
3D-ResUNet+Attn.(FDK)  & \textbf{67.68(-10.21±7.28)} & \textbf{29.71(1.33±0.98)} &  \textbf{0.8940(0.0855±0.0232)} &  -1.5657±67.5189 \\
\hline

\multicolumn{5}{c}{\textbf{Dual-Domain (Average Volume)}} \\
\hline
3D-UNet (FDK)          & 49.19(-28.70±6.19) & 32.43(4.05±0.62) &  0.9377(0.1292±0.0349) &  -0.2131±48.9999 \\ 
3D-ResUNet (FDK)       & \textbf{45.51(-32.38±8.13)} & \textbf{33.07(4.69±0.73)} &  \textbf{0.9425(0.1339±0.0406)} & -8.9502±44.4396 \\
3D-ResUNet+Attn.(FDK)  & 45.65(-32.24±9.07) & 33.00(4.62±0.82) & 0.9396(0.1311±0.0425) &  -9.7962±44.3982 \\
\toprule
\end{tabular}
}
\caption{Presented are the quantitative results of DL-based motion correction for CBCT data with simulated motion. The table presents the performance of the proposed motion reduction framework based on the RMSE, PSNR, and SSIM metrics and reports the mean and standard deviation of the body-masked difference (correction) volumes. The metrics are calculated between the reconstructed and ground truth volumes, converted to HU with slope and intercept of $48200$ and $-1106$, respectively. All numerical values are averaged over the test set. The table shows the average metric together with the average gain (or loss) and the latter's standard deviation to clarify the contribution of the motion correction. For example, in the last row, the average PSNR is reported as $33.00$ dB, corresponding to an average improvement of $4.62$ dB, with a standard deviation of $0.82$ dB. The models noted by~\textdagger~are used for clinical evaluation (Section \ref{chap:motion_sec:clinical}) (figure adopted from~\cite{amirian2023mitigation}).}
\label{chap:motion_tab:results}
    \vspace{0.5cm}
\end{table}

In order to train the model architectures (see Figure \ref{chap:motion_fig:architecture}) in a supervised scenario, only the simulated motion dataset (Section~\ref{chap:motion_subsec:datasets}) is relevant. The training set is used for training the models, while the validation set results guide the optimization to select the best models and hyperparameters. Experimental results in this section consist of the final performance on the left-out test set during the training and parameter optimization.

Table~\ref{chap:motion_tab:results} presents the numerical performance of the various architectures discussed in Section~\ref{chap:motion_sec:methods} for two reconstruction methods FDK and iCBCT, with two different sets of ground truth volumes (``average volume" or ``average amplitude"). Three different neural network architectures are investigated: ``3D-UNet" (base architecture), ``3D-ResUNet" (UNet-based enhanced with ResUNet), and ``3D-ResUNet+Attn." (enhanced using both ResUNet and attention blocks). The ground truth volumes with average amplitude differ more from their corresponding uncorrected volume with motion artifacts than the ones with averaged volume. Therefore, the baseline RMSE is larger for average amplitude, and lower baseline performances in terms of PSNR and SSIM are reported in Table~\ref{chap:motion_tab:results}. Since computing the gradients in the backward pass of the reconstruction algorithm, which is required for training models in the projection domain, is only practical for the FDK reconstruction, Table~\ref{chap:motion_tab:results} does not present results based on iCBCT for optimizing in the projection domain and dual-domain. The numerical results are reported based on computing the metrics as introduced in Section~\ref{chap:motion_subsubsec:metrics} between the body-masked ground truth and reconstructed volumes, converted to HU. 

\begin{figure}[htb!]
    \centering
    \begin{tabular}{l|cc}
        \toprule
        Uncorrected Volume & \multicolumn{2}{c}{Ground Truth} \\
        \midrule
        \multirow{2}[2]{*}[12mm]{\includegraphics[width=0.30\linewidth]{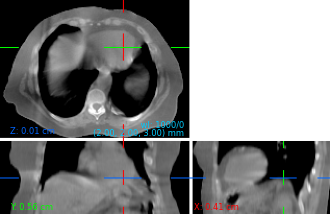}} &
        \includegraphics[width=0.30\linewidth]{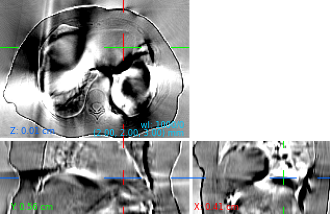} &
        \includegraphics[width=0.30\linewidth]{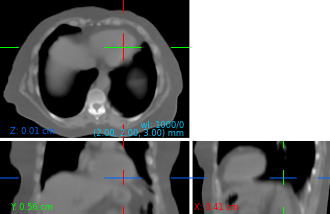} \\ \cmidrule{2-3}
        & \multicolumn{2}{c}{Corrected Volume} \\ \cmidrule{2-3}
        & \includegraphics[width=0.30\linewidth]{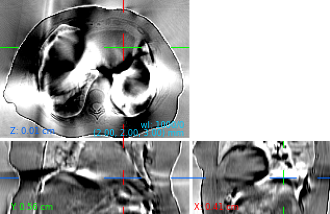} 
        & \includegraphics[width=0.30\linewidth]{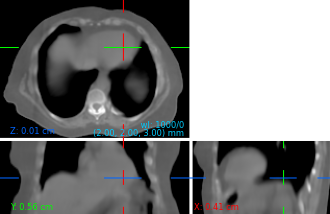}  \\
        \toprule
    \end{tabular}
    \caption{Example results for FDK reconstruction (volume domain optimization). Presented is the uncorrected volume using default reconstruction (left), the ground truth volume, both as difference and absolute image (``average volume", top right), as well as the corrected volume (bottom right). Images are presented in HU with W/L=1000/0 (figure adopted from~\cite{amirian2023mitigation}).}
    \label{chap:motion_fig:nice_fdk}   
    \vspace{0.5cm}
\end{figure}

The numerical evaluation demonstrates that training 3D-CNNs is consistently successful in compensating motion for both projection and volume domains, with the best performance being achieved in the volume domain. 
Numerically, it corresponds to a rise of $6.34$ dB in PSNR and $0.1499$ for SSIM for FDK with ``average volume" ground truth. The highest improvement reported for iCBCT is $5.81$ dB of PSNR and $0.1996$ in SSIM with ``average amplitude" ground truth. 
%
Table~\ref{chap:motion_tab:results} reports a very competitive performance in dual-domain optimization. However, most of the motion correction performance in the dual-domain setting is based on the volume domain corrections.
The maximum average gained PSNR in the case of pure projection domain optimization turned out to be $1.33$ dB. These results represent the first successful attempt at reducing motion artifacts in CBCT scans using deep neural networks.

The method proposed reduces motion artifacts for two reconstruction techniques (FDK and iCBCT) with several different architectures, including variants with added internal residual connections and/or channel-spatial attention. The motion compensation performance shows a small but consistent variance with the details of the neural network architecture. Reducing the motion artifacts in the projection domain is a subject for further research and optimization due to the more challenging optimization settings. Optimization in the projection domain relies on gradients propagated all the way through the CBCT reconstruction layer and suffers from the large volume of data in the projection space and current GPU memory limitations.

\begin{figure}[htb!]
    \centering
    \begin{tabular}{l|cc}
        \toprule
        Uncorrected Volume & \multicolumn{2}{c}{Ground Truth} \\
        \midrule
        \multirow{2}[2]{*}[12mm]{\includegraphics[width=0.30\linewidth]{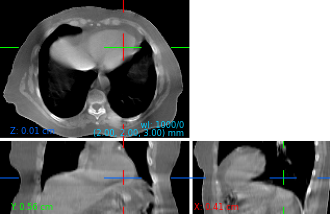}} &
        \includegraphics[width=0.30\linewidth]{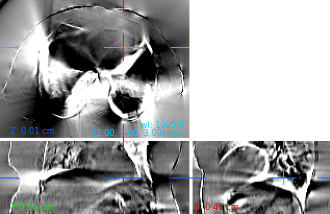} &
        \includegraphics[width=0.30\linewidth]{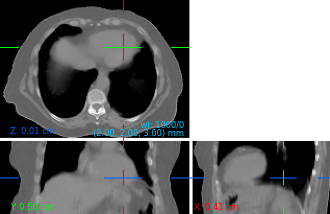} \\ \cmidrule{2-3}
         & \multicolumn{2}{c}{Corrected Volume} \\ \cmidrule{2-3}
         & \includegraphics[width=0.30\linewidth]{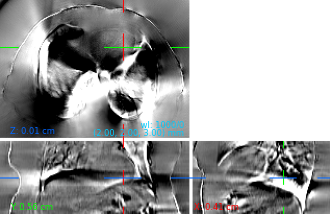} 
         & \includegraphics[width=0.30\linewidth]{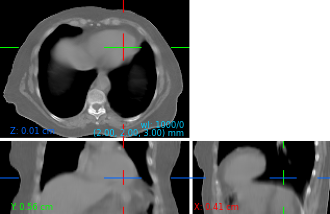}  \\
        \toprule
    \end{tabular}
    \caption{Example results for FDK reconstruction (volume domain optimization). The uncorrected volume using default reconstruction (left), the ground truth volume, both as difference and absolute image (``average volume", top right), as well as the corrected volume (bottom right) are depicted in the table. Images are presented in HU with W/L=1000/0 (figure adopted from~\cite{amirian2023mitigation}).}
    \label{chap:motion_fig:nice_icbct}   
   \vspace{0.5cm}
\end{figure}

Comparing the two CBCT reconstruction algorithms, iCBCT shows more robustness against motion during acquisition time, and a slightly lower drop in baseline performances is reported. In addition, artifact reduction using 3D-CNNs in the volume domain for iCBCT reconstruction is successful and shows the same results as FDK. 
Figures~\ref{chap:motion_fig:nice_fdk} and~\ref{chap:motion_fig:nice_icbct} present example visualizations of the observed motion artifact improvements seen in volume domain learning on top of the FDK and iCBCT reconstructions, respectively.

\subsection{Clinical Evaluation}
\label{chap:motion_sec:clinical}

\begin{figure}[htb!]
    \centering
    \vspace{0.5cm}
    \begin{tabular}{ccc}
        \toprule
        \multicolumn{3}{c}{Ground Truth: Average Volume} \\
        \midrule
        Uncorrected Volume & Residual Corrections & Corrected Volume \\ 
        \midrule
        \includegraphics[width=0.30\linewidth]{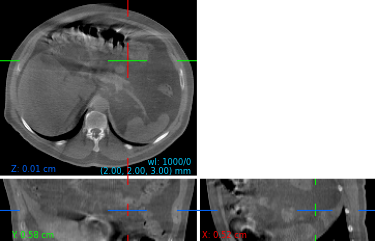} &
        \includegraphics[width=0.30\linewidth]{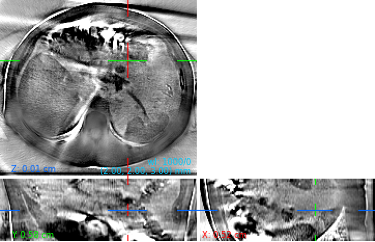} &
        \includegraphics[width=0.30\linewidth]{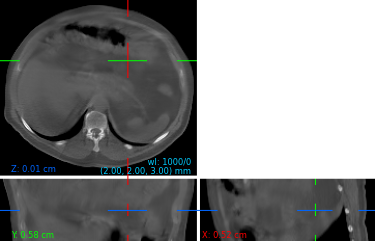} \\
        \toprule
        \multicolumn{3}{c}{Ground Truth: Average Amplitude} \\
        \midrule
        Uncorrected Volume & Residual Corrections & Corrected Volume \\ 
        \midrule
        \includegraphics[width=0.30\linewidth]{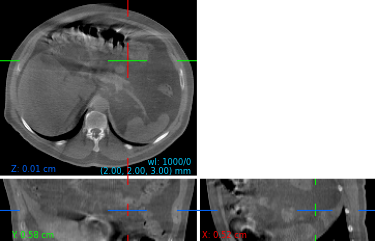} &
        \includegraphics[width=0.30\linewidth]{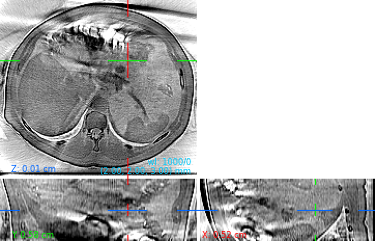} &
        \includegraphics[width=0.30\linewidth]{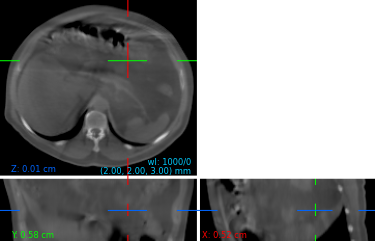} \\
        \toprule
    \end{tabular}
    \caption{The table shows example results for iCBCT reconstruction for real-world test dataset, using the two options for the choice of ground truth. The uncorrected volumes using default reconstruction (left), the residual corrections (middle), as well as the corrected volumes (right) are presented (figure adopted from~\cite{amirian2023mitigation}).}
    \label{chap:motion_fig:nice_washu}   
    \vspace{0.2cm}
\end{figure}

To validate the quantitative results of the previous section in a clinical setting, the trained motion compensation CNN models are applied to a real-world test dataset (see Section \ref{chap:motion_subsec:datasets} and Figure \ref{chap:motion_fig:nice_washu}). Finally, the performance of the motion correction models is evaluated based on the feedback obtained from clinicians.

The real-world CBCT scans used in this study are sufficiently different from the simulated training datasets, e.g., the projection count and HU calibration, to objectively judge the models' generalization capabilities.
The attenuation values of the real-world test dataset are rescaled to match the scale of the training dataset to compensate for the different calibrations.

The expert feedback was collected from a study where clinicians visually inspected $30$ pairs of iCBCT reconstructed and motion-corrected volumes, $15$ each for either a model trained using average amplitude or average volume ground truth. The best performing CNN architectures from Table~\ref{chap:motion_tab:results}, UNet in volume domain without residual connections or attention, were used for clinical validation. Subsequently, $20$ clinicians, including radiation oncologists, medical physicists, radiation technologists, and physicians, answered several questions about their preferences for using CNN models to reduce motion artifacts compared with the standard reconstruction. Each of the clinicians identified themselves as one of three general categories: physician ($26\%$), medical physicist ($37\%$), and dosimetrist/radiation technician ($37\%$).

Initial feedback on the iCBCT datasets indicated the presence of severe and mild unavoidable real-world artifacts besides motion in $34\%$ and $20\%$ of the scans, respectively. 
The study participants specified their level of agreement or preference concerning (a) a reduction of the observed motion artifacts and (b) the use for various applications, including dose calculation, patient positioning, and segmentation.

This clinical evaluation, the first of its kind to the best of our knowledge, faced the challenge of subjective assessments from experts with different clinical backgrounds. For example, physicians reported a noticeable or strong improvement in CNN-based motion artifact reduction using average volume ground truth in $80.00\%$ of scans, while medical physicists only reported this in $65.83\%$ of the scans. 
Nonetheless, medical physicists preferred CNN-corrected volumes in $63.33\%$ of the cases for dose calculation, while the physicians reported this in only $30.67\%$ of the cases.

\begin{table}[t]
\centering
\resizebox{\textwidth}{!}{ 
\begin{tabular}{l||ccc||ccc}
\hline
Ground Truth $\rightarrow$  & \multicolumn{3}{c||}{\textbf{Average Volume}} & \multicolumn{3}{c}{\textbf{Average Amplitude}} \\
$\downarrow$ Application / Preference $\rightarrow$ & CNN ($\%$) & Equal ($\%$) & Standard ($\%$) & CNN($\%$) & Equal($\%$) & Standard($\%$) \\
\hline \toprule
Motion artifact reduction                  & $\boldsymbol{74.00}$ & $26.00$ & - & $58.33$ & $ 41.67$ & -  \\
Plan adaptation and dose calculation       & $\boldsymbol{49.33}$ & $22.00$ & $28.67$ & $26.33$ & $17.33$ & $56.33$  \\
Soft-tissue-based patient positioning      & $23.00$ & $12.67$ & $64.33$ & $13.00$ & $7.00$ & $80.00$ \\
Manual and automatic tissue segmentation   & $24.33$ & $14.67$ & $61.00$ & $13.00$ & $10.33$ & $76.67$ \\
\hline
\end{tabular}
}
\caption{Results of the clinical evaluation. This table shows the preferences for CNN-based or default iCBCT reconstruction when using CNN models trained using either average volume or average amplitude ground truth concerning motion artifact reduction and potential applications such as plan adaptation and dose calculation, patient positioning and segmentation (table adopted from~\cite{amirian2023mitigation}).}
\label{chap:motion_tab:results_clinical}
    \vspace{0.5cm}
\end{table}

Table~\ref{chap:motion_tab:results_clinical} presents the average overall votes and the final clinical evaluation results. Despite the differences in the improvements reported by the different experts, there is a clear positive trend showing that the proposed CNN models are indeed able to reduce motion artifacts successfully. In addition, clinicians reported a tendency toward using CNN-corrected images (using average volumes ground truth) for plan adaptation and dose calculation. One area where clinical experts preferred to use images without CNN-based reconstruction is for soft-tissue-based patient positioning and manual or automatic tissue segmentation, as these images are typically sharper compared with the CNN-corrected ones. 

Nevertheless, quantitative evaluation to compute the level of agreement when applying an automatic segmentation algorithm using CBCT images with and without motion artifact correction leads to overwhelmingly positive results. The average dice score measures the automatic segmentation contours in original and motion-corrected volumes. This score is averaged over 18 organs or tissues, visible in most CBCT scans, including pulmonary arteries, breast, chest wall, lung, ribs, and spinal canal. The high dice score of $0.89$ ($0.88$) when using average volume (average amplitude) ground truth demonstrates a very high level of consistency between the obtained segmentation contours despite the low preference reported by clinical experts for using the motion-corrected images for segmentation. 

\section{Discussions and Conclusions}
\label{chap:motion_sec:discussion}

This chapter presents the first DL-based method for globally reducing motion artifacts in reconstructed 3D CBCT images, built on top of the two reconstruction algorithms FDK and iCBCT.
Neural network architectures which act either on the reconstructed CBCT volumes, the input X-ray projections or both were trained in a supervised way using a motion simulation framework to provide motion-free ground truth. The experimental results demonstrate that DL-based architectures can correct motion artifacts. So far, the best results have been obtained in the volume domain through the implementation of a refined U-net architecture.

The quantitative evaluations demonstrate that using DL through deep neural network architectures yields significant improvements in image quality and reduces motion-induced artifacts in CBCT scans. In addition, a clinical evaluation was performed, in which clinical experts confirmed the principal quantitative results for motion artifact reduction using a real-world test dataset. Clinicians confirmed that artifacts are reduced and expressed a preference for using CNN-corrected CBCT images for dose calculation. However, for patient positioning or segmentation, this could not yet be demonstrated.

There are several related avenues that could be explored in future research:

First, the presented results show promising improvements, mainly in the volume domain, independent of the acquisition parameters and reconstruction technique. However, there is room for improvement in the projection and dual-domain setting. One potential reason is the processing of the projections in batches due to GPU memory limitations, which leads to a loss of correlation between different projection batches separately processed by the neural network. In addition, great care is necessary to ensure the backpropagation of gradients through the CBCT reconstruction layer to provide a meaningful and noise-free learning signal in the projection domain.

Second, models trained using supervised learning typically suffer from generalization to data acquired in entirely different settings. Although the calibration technique used in this study successfully reduced the performance gap between the performance of the models on simulation and real-world datasets, generalization to highly different acquisition setups and other anatomies is not certain. This provides motivation for further investigation of unsupervised learning and/or domain adaptation techniques.

Third, the current motion simulation only simulates respiratory motion and does not include other effects, such as cardiac motion. Therefore, tackling cardiac motion in chest CBCT scans combined with respiratory motion remains an open problem. This method could also be extended to handle abdominal CBCT scans, including different motion effects.

In conclusion, while the initial results are very promising, future research will aim to improve adaptive treatment capabilities in IGRT, including patient positioning and tumor targeting, auto-segmentation, and dose calculation applications directly on the radiotherapy device.

\addtocontents{toc}{\protect\newpage}
\chapter{Applications in Affective Computing, Medical Imaging and Beyond}
\label{chap:applications}
\chaptermark{Applications of Machine and Deep Learning}
This thesis presents several contributions to diverse and interdisciplinary research niches among many applications of machine learning (ML) and deep learning (DL). Although the original papers in this section have much more extensive content, this chapter summarizes the most scientifically thrilling findings and lessons learned from applying ML and DL in the real world. This chapter is broader in terms of the various applications discussed, more diverse than previous chapters, more straightforward, and more interesting because of its diversity.

Despite the brief overviews, this chapter presents many interesting practical findings and insights that are beneficial in applied research and tuning ML and DL methods to their best performances. Furthermore, this chapter describes several practical problems in ML and DL, such as affective computing, pain estimation, and data homogenization, and identifies initial solutions to this particular research area. Finally, similar to the previous chapters, the remainder of this chapter discusses some exciting directions for future research.

The chapter is organized as follows: discussing solutions for facial expression estimation, emotion recognition, and findings on automated ML and DL focusing on bringing neural networks to their best performances. The chapter then presents two medical applications targeting signal processing for pain estimation and image processing for data homogenization for DL pipelines. The last two sections of this chapter introduce and elaborate on two well-known challenges of DL: fairness and robustness. 
\newpage
\section{Affective Computing}

This section presents two applications of machine learning in affective computing. It begins with explaining the application of support vector machines in facial expression estimation, followed by emotion recognition using audio-visual features.  

\subsection{Facial Expression Estimation}

\begin{figure*}[htb!]
\centering 
	\subfloat[AU1]{\includegraphics[width=0.14\linewidth]{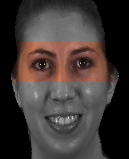}}
	\subfloat[AU4]{\includegraphics[width=0.14\linewidth]{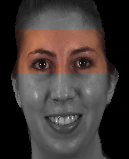}}
	\subfloat[AU6]{\includegraphics[width=0.14\linewidth]{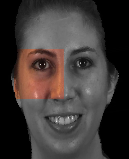}}
	\subfloat[AU10]{\includegraphics[width=0.14\linewidth]{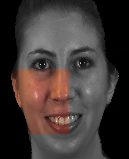}}
	\subfloat[AU12]{\includegraphics[width=0.14\linewidth]{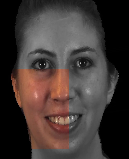}}
	\subfloat[AU14]{\includegraphics[width=0.14\linewidth]{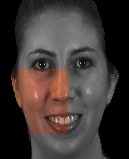}}
	\subfloat[AU17]{\includegraphics[width=0.14\linewidth]{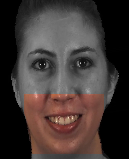}}
	\caption{Several action units (AUs) used for facial expression estimation (figure is adopted from~\cite{amirian2017support}).}
	\label{chap:applications_fig:AUs}
\end{figure*}

Human facial expressions can reveal information about their affective states~\cite{yang2019facial} or cognitive load\cite{kindsvater2016fusion}, which are crucial in human-computer interaction (HCI). There is extensive literature, and several surveys have been developed around facial expression estimation due to its importance. Researchers have divided the human face into several regions called action units (AUs) to quantify facial expressions. Figure~\ref{chap:applications_fig:AUs} shows a few such action units defined in the literature to measure facial expressions. The activity of AUs can be measured based on binary occurrence labels (active/deactive) or quantified in terms of intensity in discrete activation levels from zero to six. The predictions of AU intensities per frame can be considered a time series. One can compare measures such as root mean square error (RMSE) and Pearson correlation coefficient (PCC) between predictions and ground-truth labels. Since the labels are discrete, it is also possible to compute the mean intraclass correlation coefficient (ICC) as a performance measure~\cite{shrout1979intraclass}.

According to the usage of the ground-truth labels, AU activities can be predicted in binary occurrence or by estimating the intensity of a specific AU. Different types of neural networks can be used based on the goal of the facial expression task. For instance, a binary classifier can predict the occurrence of activation in an AU, or its level can be predicted using a multi-class classifier or a regression model.

\begin{figure*}[htb!]
\centering 
	\subfloat{\includegraphics[width=0.135\linewidth]{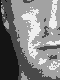}}\hspace{0.2em}%
	\subfloat{\includegraphics[width=0.135\linewidth]{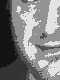}}\hspace{0.2em}%
	\subfloat{\includegraphics[width=0.135\linewidth]{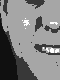}}\hspace{0.2em}%
	\subfloat{\includegraphics[width=0.135\linewidth]{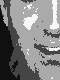}}\hspace{0.2em}%
	\subfloat{\includegraphics[width=0.135\linewidth]{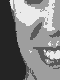}}\hspace{0.2em}%
	\subfloat{\includegraphics[width=0.135\linewidth]{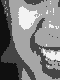}}\hspace{0.2em}%
	\subfloat{\includegraphics[width=0.135\linewidth]{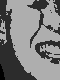}}\hspace{0.2em}%
	\caption{Several dictionary-learned facial templates used for facial feature extraction (figure is adopted from~\cite{amirian2017support}).}
	\label{chap:applications_fig:dict}
\end{figure*}

The presented method in this section for AU intensity estimation consists of multiple steps, starting with preprocessing for face alignment followed by training facial expression templates from data using K-SVD dictionary learning (see Figure~\ref{chap:applications_fig:dict}). Then, each input image's features are computed based on their projection onto dictionary elements to compute the facial features. Support vector machine (SVM) based classifiers and regression models are trained for AU occurrence and intensity estimation based on the computed features. Dictionary-learned features improved the baseline results using conditional random fields~\cite{valstar2017fera} by $35\%$. However, the DL-based method achieved superior performance on an unseen test dataset reported in the 3D facial expression recognition and analysis challenge (FERA 2017)~\cite{zhou2017pose}.   

The main practical insights of this section can be summarized as follows: 1) Preprocessing steps such as face alignment (especially when images are from different head poses) are vital for ML and DL-based approaches. 2) The performance of deep learning models has exceeded the ML-based methods for a long time; however, training regression models instead of classifiers imposes much more optimization effort and hyperparameter tuning overhead in DL than in ML. 3) Not all classes are located at an equal distance in the embedding space. Hence, using a hierarchical classifier to identify the occurrence and then predict the intensity of an AU improves the performance. 

More details on implementing dictionary-learned feature extraction techniques used with a support vector regression model for facial expression recognition are presented in~\cite{amirian2017support}.

\subsection{Emotion Recognition}
\label{chap:applications_sec:emotion}
Affective computing and, more specifically, human emotion recognition is an interdisciplinary field linking cognitive sciences, psychology, and computer science. It has recently attracted more attention in the context of human-computer interaction (HCI) to interpret and understand human behavior and emotions. The idea of automatic emotion recognition is useful for making various sensory measurements, including facial video, audio, and physiological signals to predict human affective status and emotions according to the changes in the sensory information.

Besides discrete emotional categories such as happiness, surprise, fear, anger, and disgust, there are two continuous dimensions of arousal and valance which can express human emotions. Arousal shows a human's level of activeness and engagement in a specific situation, while valance indicates the positiveness of a human's feelings. Arousal and valance levels can define different human feelings and affective statuses. For instance, happiness and excitement share a positive valance level with low and high arousal levels, respectively. Similarly, sadness and anger are positioned on the negative side of the valance scale with low and high arousal levels~\cite{russell1980circumplex}. Furthermore, researchers have also figured out that the affective status of humans in social interaction has another dimension described as dominance~\cite{kehrein2002prosody}.

Based on the briefly explained theory, Ringeval et al. designed an experiment and created the RECOLA multimodal corpus for emotion recognition~\cite{ringeval2013introducing}. The human subjects participating in this experiment were divided into two groups for interactive sessions. Then, they were presented with a task, such as surviving an air airplane crash in the middle of a jungle. The participants were given a list of tools and asked to rank the list of tools according to their preference independently. Then, the participants of each group were connected via video call to discuss their solutions and the organizers recorded the audio-visual information and physiological signals during this interactive session. According to the progress of the discussions, participants experienced different types of natural emotions with various intensities. After the interactive session, six psychologists rated the participants' emotions based on two dimensions: arousal and valance. Then, a gold standard label was computed based on the inter-rater agreements of the psychologist's ratings at every time step. Next, audio-visual and bio-physiological information processing occurred to quantify the participants' emotions automatically through the reproduction of the gold standard labels. 

Valstar et al. processed the raw audio-visual and bio-physiological data of the RECOLA dataset into features that are ready for ML-based pipelines for prediction~\cite{valstar2016avec}. They extracted local Gabor binary patterns from three orthogonal planes (LGBP-TOP)~\cite{almaev2013local} for appearance features and extracted facial landmarks to evaluate the face geometry~\cite{xiong2013supervised}. Moreover, they used the COVERED toolbox to extract voice quality and spectral features from audio~\cite{degottex2014covarep}. Bio-physiological recordings include electrocardiogram (ECG) and electrodermal activity (EDA) signals. All bio-physiological signals pass through band-pass filtering in the preprocessing step. The ECG signals are processed to extract heart rate (HR) and its measures of variability (HRV) as features for emotion recognition~\cite{ringeval2015prediction}. EDA signals are also decomposed to their rapid and transient component called skin conductance response (SCR) as well as a slower basal drift denoted as skin conductance level (SCL)~\cite{dawson2017electrodermal}. Valstar et al. computed four statistical features from both EDA components and their first derivative to use in the emotion recognition pipeline~\cite{valstar2016avec}.

\begin{figure*}[htb!]
    \centering
    \includegraphics[width=0.9\linewidth]{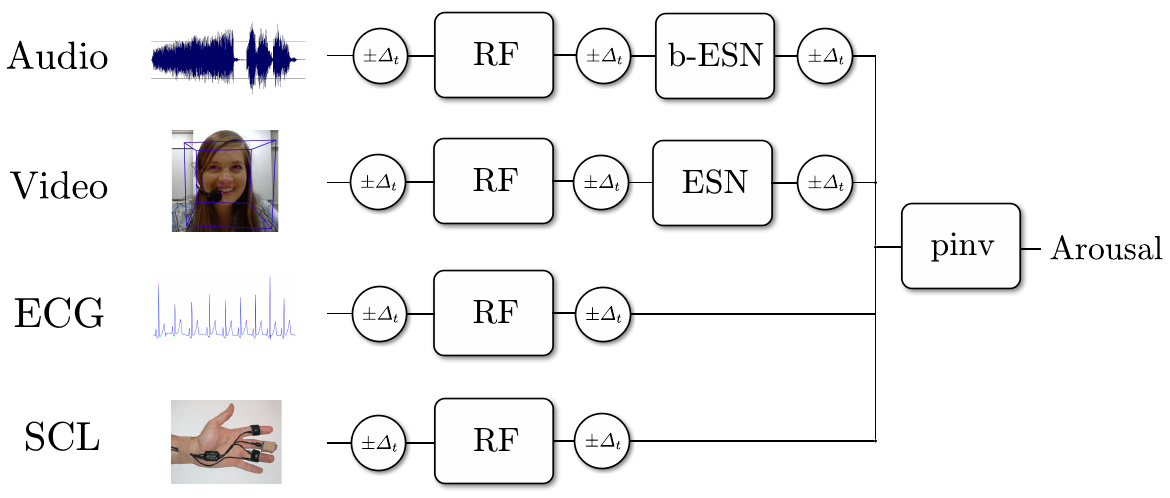}
    \caption{Presented is the proposed sequence of blocks for the automatic fusion of audiovisual and biophysiological information to predict arousal levels (figure is adopted from~\cite{amirian2016continuous}).}
    \label{chap:applications_fig:mutimodal}
\end{figure*}

This thesis offers an ML-based pipeline for processing the features' information and creating predictions from multiple data modalities. Figure~\ref{chap:applications_fig:mutimodal} demonstrates this pipeline, including random forests (RF~\cite{breiman2001random}) for training regression models, which predict the continuous labels in two dimensions of arousal and valance based on extracted features for the four data modalities of audio, video, ECG and EDA. Furthermore, since the psychologists evaluated the emotional status of the participants based on their audio-visual signals, these modalities contain more information, and RF models have created more precise and less noisy predictions using audio-visual features. 

\begin{figure*}[htb!]
    \centering
    \includegraphics[width=0.8\linewidth]{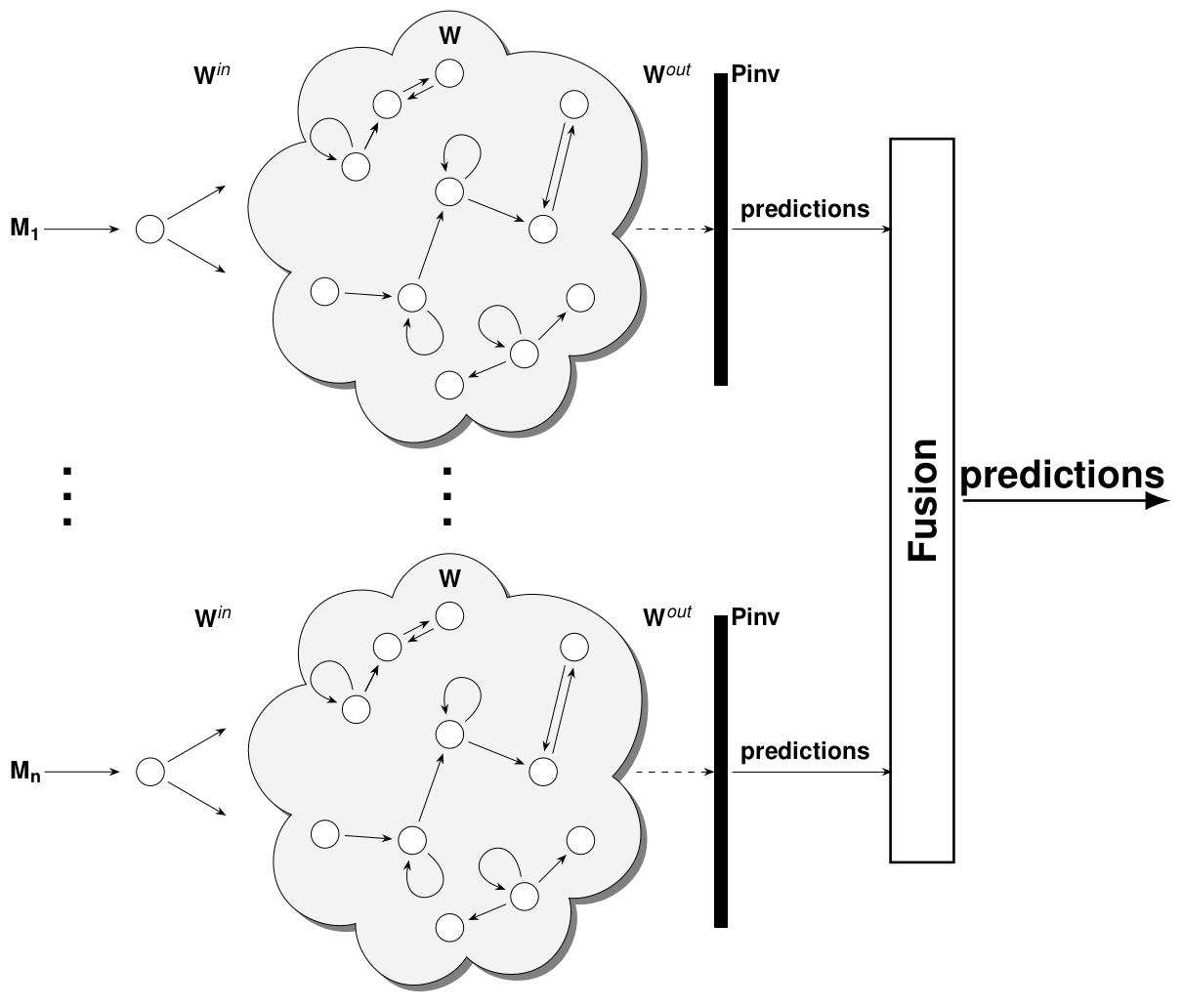}
    \caption{Presented are the echo state networks (ESNs) based architectures for modeling temporal information dependencies and fusing multimodal information. One ESN is trained for each modality, and the predictions of all modalities are combined using precomputed weights according to the importance of modalities per task.}
    \label{chap:applications_fig:esn}
\end{figure*}

Due to the precise prediction from audio-visual features, it was possible to train a recurrent model based on echo state networks (ESNs~\cite{jaeger2001echo}) to refine the prediction and consider temporal information. The idea of reservoir computing inspires ESNs. They are models with time series as input (here, the sequence of arousal or valance predictions) with internal states connected by random weights. The models' output weights are trained to minimize the loss function between the internal state of the reservoir and ground-truth labels (see Figure~\ref{chap:applications_fig:esn}). The emotion recognition pipeline described in this section uses a bi-directional ESN for audio and a standard ESN for video modalities which were selected based on their performance on each data modality. The internal state of the reservoir in bi-directional ESNs depends not only on previous samples in the time series but also on the samples that follow. An alternative to ESNs are long short-term memory (LSTM~\cite{hochreiter1997long}) models as well as their bi-directional version~\cite{schuster1997bidirectional}.

Multimodal information fusion is the final and essential component of the emotion recognition pipelines. Different modalities contain various levels of information for arousal and valance tasks. For instance, the arousal predictions from audio are more accurate, and video modalities include more information for the valance. Hence, averaging the predictions of all modalities does not lead to optimal predictions. The pipeline presented in Figure~\ref{chap:applications_fig:esn} uses Moore pseudo-inverse to minimize the mean square error of multimodal information fusion predictions and gold standard labels. Thus, the final information fusion block is a linear combination of the predictions from each modality. More details about numerical results and methodology are presented in a paper published in conjunction with the audio-visual emotion recognition challenge (AVEC)~\cite{valstar2016avec}.

\begin{figure*}[htb!]
    \centering
    \includegraphics[width=\linewidth]{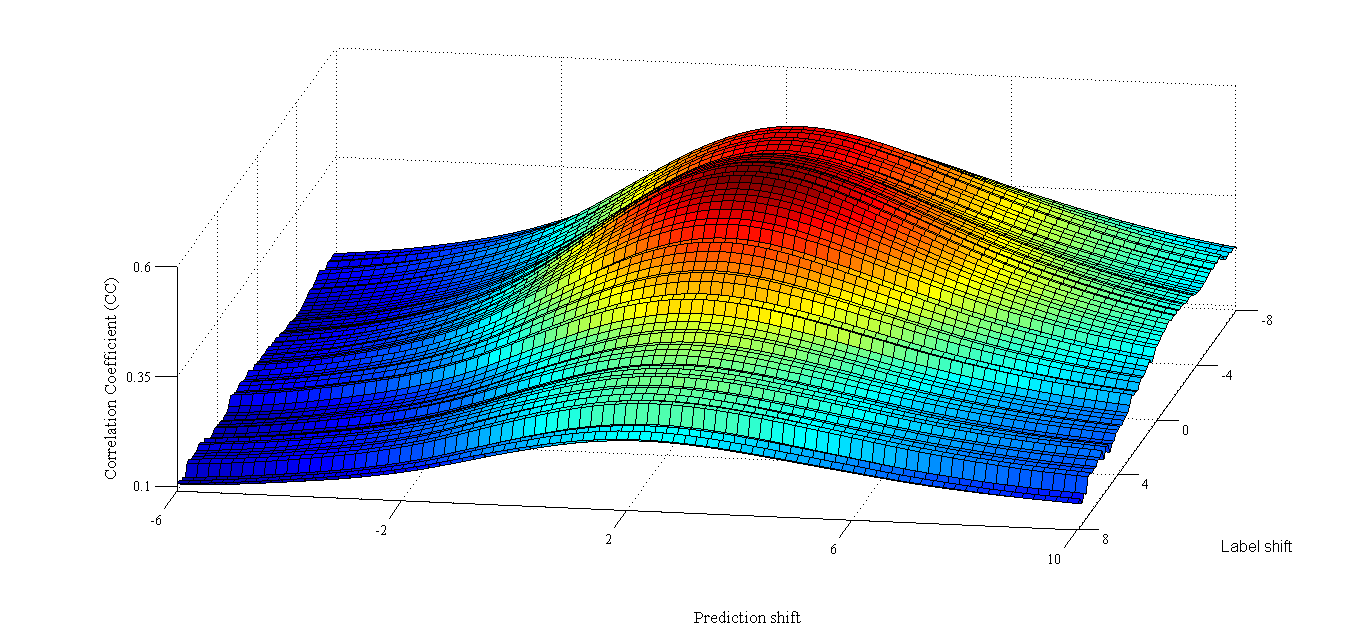}
    \caption{The figure depicts the temporal mismatch between audio features and gold standard labels. The average performance of predicting the arousal level of participants from audio increases considerably when features are aligned with gold-standard labels (figure is adopted from~\cite{amirian2016continuous}).}
    \label{chap:applications_fig:delay}
\end{figure*}

The most intriguing path for future research in emotion recognition is multimodal and temporal information fusion, in addition to multi-task learning. The information fusion technique presented in this section used universal weights to combine the modalities. However, the importance of different modalities can differ from time to time based on events in the audio-visual and bio-physiological signals. Thus, developing adaptive fusion techniques with attention mechanisms can significantly improve the fusion's performance. Furthermore, arousal and valance are predicted separately in this work. These tasks can be combined into a multi-task learning problem to train models that fine-tune the predictions of one emotion dimension by being aware of the other. The ratings from psychologists and gold standard labels are not necessarily temporally synchronized with physiological and audio-visual information. Tracing such temporal mismatches between features and labels is still cumbersome, which we tackle by finding the optimal shifts as a hyperparameter (see Figure~\ref{chap:applications_fig:delay}). Developing more sophisticated methods to model such temporal dependencies in high-dimensional features or video is another thrilling venue for research.

\newpage
\section{Automated Data Analysis}
The breakthroughs in ML and DL provide the opportunity to leverage an immense amount of data to approximate almost any function and draw arbitrary decision boundaries for classification and regression. However, the main consequence of such a large degree of freedom was the very challenging task of selecting suitable models with optimal generalization capabilities and a set of hyperparameters for a given dataset. Researchers used to tune the parameters empirically and put their inductive biases into models. With the rise of computing resources, searching architectures and hyperparameter spaces became feasible and popular. The body of literature focusing on automated ML (AutoML) and automated DL (AutoDL) is massive, and this section offers several insights found whilst researching these two subjects.      

\subsection{Automated Machine Learning}
The goal of automated machine learning is defined as solving the combined model section and hyperparameter (CASH) optimization problem. The main goal is to develop search algorithms that can adapt to the tasks based on new trials or even further leverage previous experience from other datasets (referred to as meta-learning in literature~\cite{vanschoren2019meta}). The AutoML challenge series provided a framework to compare methods targeting the CASH problem~\cite{guyon2017analysis}. The idea of this challenge is to provide two independent sets of benchmark datasets for training and testing with different tasks such as regression and various types of classification. The performance of methods developed for AutoML has been evaluated with strict limitations on time and resources.    
Several strategies have been adapted and used in the literature to solve the CASH problem automatically. The most straightforward strategy to target the CASH problem is the random search method. Although it is a naive search strategy, random search can achieve competitive results for a new task when no similar dataset or problem is available (see Table~\ref{chap:applications_table:automl}). Furthermore, it is possible to use evolutionary selection for tuning the choices of models and hyperparameters for the target datasets. The idea is to choose the subsequent models and hyperparameters based on the previous best-performing ones. Olson et al. proposed an evolutionary algorithm for the CASH problem called the tree-based pipeline optimization tool (TPOT~\cite{olson2016automating}). 

Several strategies have been adapted and used in the literature to solve the CASH problem automatically. The most straightforward strategy to target the CASH problem is the random search. Although it is a naive search strategy, the random search can achieve competitive results for the new task when no similar dataset or problem is available (see Table~\ref{chap:applications_table:automl}). Furthermore, it is possible to use evolutionary selection for tuning the choices of models and hyperparameters to the target datasets. The idea is to choose the subsequent models and hyperparameters based on the previous best-performing ones. Olson et al. proposed an evolutionary algorithm for the CASH problem called the tree-based pipeline optimization tool (TPOT~\cite{olson2016automating}). 

One method used to incorporate the previous experiences from other datasets is Bayesian optimization~\cite{feurer2014using}. The idea is simple, hence practical. The idea is to consider functions of choice (commonly Gaussian processes) with free parameters defining the space drawn by hyperparameters and objective functions. Free parameters of the Gaussian processes are updated after every trial of a set of HPs, and the performance on a given dataset is computed. The parameters of Gaussian processes, in this way, learn the connections between hyperparameters and performance on the task. Models trained based on meta-learning can transfer knowledge from training datasets to new target datasets. Auto-Sklearn is an example of such a method that not only selects the best model and hyperparameters based on the target dataset but also learns from previous runs on different datasets~\cite{feurer2015efficient}.

\begin{table*}[htb!]
	\centering
	\resizebox{\textwidth}{!}{\begin{tabular}{l l l|  c c | c c | c c}
        \toprule[1.5pt]
        & & &  \multicolumn{2}{c}{\textbf{Random-Search}} & \multicolumn{2}{c}{\textbf{Auto-Sklearn}} & \multicolumn{2}{c}{\textbf{TPOT}} \\
        \textbf{Dataset} & \textbf{Task} & \textbf{Metric}  & \textbf{Test} & \textbf{Time}  & \textbf{Test} & \textbf{Time} & \textbf{Test} & \textbf{Time}  \\
        \cmidrule(lr){1-9}

Cadata        & Regression                   & R2 (coefficient of determination)    &        0.7119        &      55.0    &       0.7327        &      54.9       &     \textbf{0.7989} &      54.6              \\
Christine     & Binary classification        & Balanced accuracy score              &        0.7146        &      99.4    &       0.7392        &      99.3       &       0.7442        &      105.1            \\
Digits        & Multiclass classification    & Balanced accuracy score              &        0.8751        &      201.2   &     \textbf{0.9542} &      201.2      &       0.9476        &      207.2         \\
Fabert        & Multiclass classification    & Accuracy score                       &        0.8665        &      77.5    &     \textbf{0.8908} &      77.4       &       0.8835        &      78.5         \\
Helena        & Multiclass classification    & Balanced accuracy score              &        0.2103        &      190.2   &       0.3235        &      216.4      &     \textbf{0.3470} &      197.5         \\
Jasmine       & Binary classification        & Balanced accuracy score              &     \textbf{0.8371}  &      24.1    &       0.8214        &      24.0       &       0.8326        &      25.9          \\
Madeline      & Binary classification        & Balanced accuracy score              &        0.7686        &      48.3    &     \textbf{0.8896} &      48.2       &       0.8684        &      53.0          \\
Philippine    & Binary classification        & Balanced accuracy score              &        0.7406        &      56.3    &       0.7634        &      56.2       &     \textbf{0.7703} &      56.4        \\
Sylvine       & Binary classification        & Balanced accuracy score              &        0.9233        &      28.9    &       0.9350        &      28.9       &     \textbf{0.9415} &      29.0          \\
Volkert       & Multiclass classification    & Accuracy score                       &        0.8154        &      122.3   &     \textbf{0.8880} &      122.2      &       0.8720        &      125.5          \\
\midrule
\multicolumn{3}{l|}{\textbf{Average Performance}}                                     &        0.7463        &\textbf{90.31} &      0.7938        &      92.85      &     \textbf{0.8006} &      93.26        \\
        \bottomrule[1.5pt]
    \end{tabular}}
    \caption{Performance of three automated machine learning algorithms with different paradigms on AutoML challenge datasets and their convergence time~\cite{guyon2017analysis} (table adopted from~\cite{tuggener2019automated}).}
	\label{chap:applications_table:automl}
\end{table*}

\subsection{Automated Deep Learning}
The mainstream research in AutoDL presented in Section~\ref{chap:theory_sec:ArchtictureSearch} focuses on developing novel vision architectures mainly based on the ImageNet dataset. However, the other open research question with more relevance to practical problems is finding the optimal architecture and set of hyperparameters for a given dataset that is not necessarily large in terms of the number of images and classes. Searching for solutions to the AutoDL problem inspired the series of AutoDL challenges to find lightweight models with hyperparameters that can quickly adapt to new but small datasets~\cite{liu2020towards}. The target of these challenges was the area under the learning curve (ALC) instead of the final or best performance. Hence, models converging faster outperform those with slow learning and better final performance based on the final evaluation metric. This evaluation metric highly favors lightweight models, which can be fine-tuned for new datasets very quickly.

\begin{figure*}[htb!]
     \centering
     \begin{tabular}{c c }
     \toprule
      \footnotesize{Chucky} & \footnotesize{Decal} \\ 
      \midrule
      \includegraphics[width=0.45\textwidth]{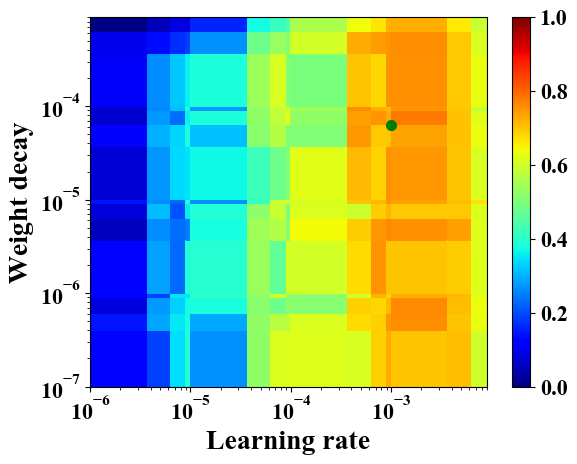} & 
      \includegraphics[width=0.45\textwidth]{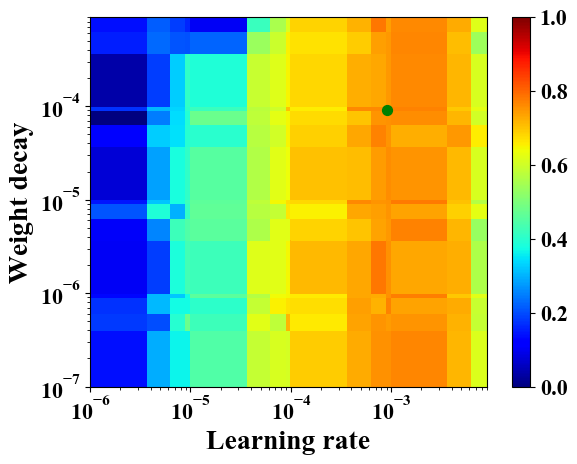}  \\
      \toprule
      \footnotesize{Hammer} & \footnotesize{Pedro} \\
      \midrule
      \includegraphics[width=0.45\textwidth]{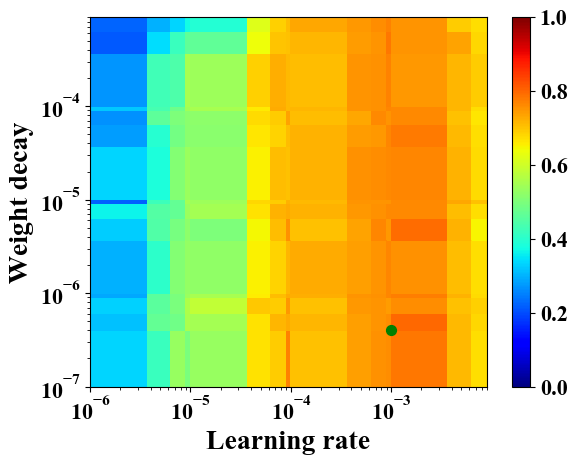} & 
      \includegraphics[width=0.45\textwidth]{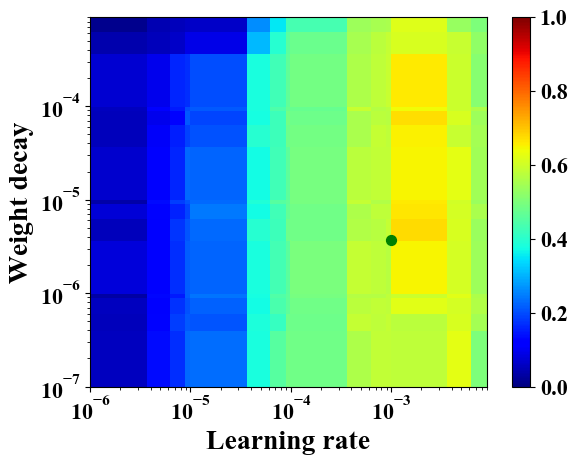}  \\
      \toprule
      \multicolumn{2}{c}{\footnotesize{Average}}\\
      \midrule
      \multicolumn{2}{c}{\includegraphics[width=0.45\textwidth]{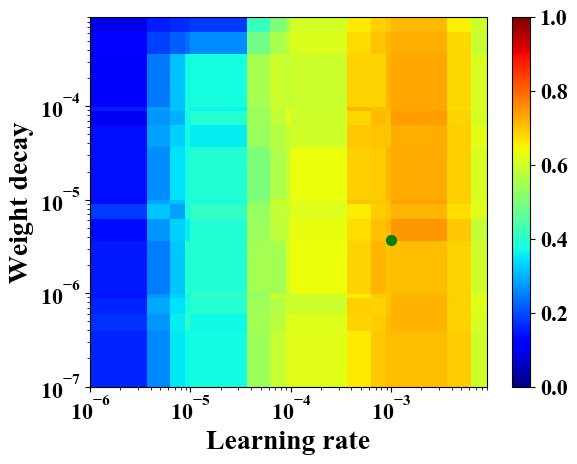}}\\
      \toprule
      \end{tabular}
      \caption{Performance of four different vision datasets in terms of ALC of MobileNetV2 as a function of weight decay and learning rate (top two rows) and averaged performance over all datasets (bottom row). The green dot indicates the best performance (figures are adopted from~\cite{tuggener2020design}).}          
      \label{chap:applications_fig:grid}
\end{figure*}

Deep convolutional neural networks (CNNs) outperformed the classical methods on AutoDL for vision. Due to their design, which is optimized to learn representations from a large dataset, pretraining on ImageNet is still an undeniable part of the model preparation. The performance of small models such as ResNet18~\cite{he2016deep} and MobileNet-V2~\cite{sandler2018mobilenetv2} which have been trained on a small dataset, show consistency by changing their learning rate for a fixed pipeline; hence, a fixed learning rate can be used for different datasets (see Figure~\ref{chap:applications_fig:grid}). However, regularization shows a more critical role in optimally fine-tuning the models to small new datasets. It is no wonder that the winning solutions of AutoDL contained the \textit{fast auto-augment} method to learn augmentation strategies tailored for a given dataset. 

Research in developing models for audio processing falls behind vision systems with respect to lightweight architecture searched models on large audio datasets. For example, a commonly used pretrained network for audio processing is VGGish~\cite{hershey2017cnn} trained on Youtube-8m dataset~\cite{Nagrani2017}, which is far from light-weight. Hence, searching for appropriate architectures is pivotal when searching for optimal models for small datasets. Similarly, augmentation strategies are not as well explored in audio processing, and a significant boost is expected with the development of more suitable or automated augmentation techniques.

Despite the differences between audio and video processing pipelines for research in AutoDL for audio-visual data, the block diagram used for pattern classification can be summarized with similar components as depicted in Figure~\ref{chap:applications_fig:grid}. Preprocessing the data is the first step which computes the spectrogram for audio data, augmentation for images, or selects key frames from videos. Then, the raw information can be processed into latent representations using convolutional backbones. Information fusion in the following steps combines the information along the axis of time via convolutions or spatially using global pooling. The last layer is a fully connected classifier to predict the patterns from the models' final embeddings. The main advantage of such a similar architecture is the possibility of mid-level information fusion between audio-visual modalities in applications such as emotion recognition, explained in Section~\ref{chap:applications_sec:emotion}.

\newpage
\section{Medical Diagnosis and Imaging}
This section presents two medical applications for machine and deep learning methods. First, the application of machine learning in pain detection in medicine through bio-physiological signal processing is explained. Second, data homogenization for medical images with future applications in merging datasets and image preprocessing in federated scenarios is discussed.

\subsection{Pain Estimation}
ML and DL methods have been widely used in medical applications such as pain estimation. Pain is an evolutionary mechanism developed in human bodies to stop and prevent external damaging stimuli or harmful behaviors. However, pain also appears as a consequence of operations in clinical settings. Not all patients, such as neonates, unconscious patients, or patients with cognitive or communicative impairments, are capable of communicating the location and level of pain when seeking treatment. Hence, automatic pain detection and intensity estimation have become more popular among researchers. 

Werner et al. introduced the Biovid heat pain database for automatic pain estimation from bio-physiological signals~\cite{walter2013biovid}. The idea of the experiment was to estimate stimulated pain using heat induced by a thermode. The experiment started with a calibration phase when the organizers measured the participants' pain perception and tolerance thresholds. Then, the experiment began with a cold thermode, which became increasingly hotter until the participant noticed the pain (perception threshold), and stopped when the heat became unbearable for a participant (tolerance threshold). Then, the temperature between these two thresholds was linearly divided into four levels, and the participants were stimulated with four pain levels during two parts of the experiment. Each part contains twenty episodes of pain stimulation with a duration of four seconds with a break of approximately eight seconds. The bio-physiological signals were recorded during the experiment for signal processing and automated pain estimation. The signals recorded in this experiment include several data modalities such as electromyography (EMG), electrocardiography (ECG), and electrodermal activities (EDA).     

After data collection, bio-physiological signals are preprocessed for feature extraction. Multiple time and frequency domain statistical features are available and computed for pain detection and pain level estimation~\cite{kachele2015multimodal}. The key component improving the pain estimation accuracy in this stage is the extraction of the modality-dependent features, especially in electrodermal activity (EDA) signals which contain the most relevant information for pain estimation~\cite{amirian2016using}. There needs for more research on bio-physiological pain estimation in order to be able to use supervised DL methods to optimize features (embeddings) automatically instead of computing hand-crafted features. However, this shortcoming is to some extent addressed using the \textit{unsupervised representation learning} for bio-physiological signals~\cite{thiam2020multimodal}.

Pain estimation based on the Biovid heat database can be considered a classification or regression task. Feature preprocessing considerably affects the accuracy of pain level quantification, and normalization of the features based on their mean and variance improves the performance of classifiers and regression models. Further improvements are achieved by normalizing the features per participant based on their baseline level of bio-physiological signals (feature personalization~\cite{kachele2015multimodal}). Moreover, personalization extends to another level by clustering people into several groups using Kullback-Leibler (KL) divergence and finding the closest subjects to train the model based on their data~\cite{kachele2017adaptive}. This improvement in estimating health-related measures using personalization hints at the fruitful direction of personalized information processing and treatment for research in health care. Different classifiers and regression models such as random forests (RF~\cite{breiman2001random}) and radial basis function networks (RBFs~\cite{broomhead1988multivariable}) showed a similar performance after tuning, and it is possible to predict the confidence of the estimated level of pain by combining the predictions from several individual models~\cite{kachele2017adaptive}.

\subsection{Data Homogenization}
Deep CNNs achieved great success in a wide range of computer vision tasks and improved state-of-the-art performances by a large margin; however, early on, they showed a weakness in generalization in the presence of a change in data distribution or concept drifts. This thesis offers an idea to deal with the changes in data distribution through data homogenization and merging multiple datasets. Deep learning and computer vision literature is full of attempts at domain adaptation~\cite{csurka2017comprehensive,wang2018deep,wilson2020survey} and style transfer research~\cite{gatys2016image,zhu2017unpaired}. However, merging a few datasets into a unified style using a preprocessing network is the novelty of the idea presented in this thesis.

\begin{figure*}[htb!]
    \centering
    \includegraphics[width=0.9\linewidth]{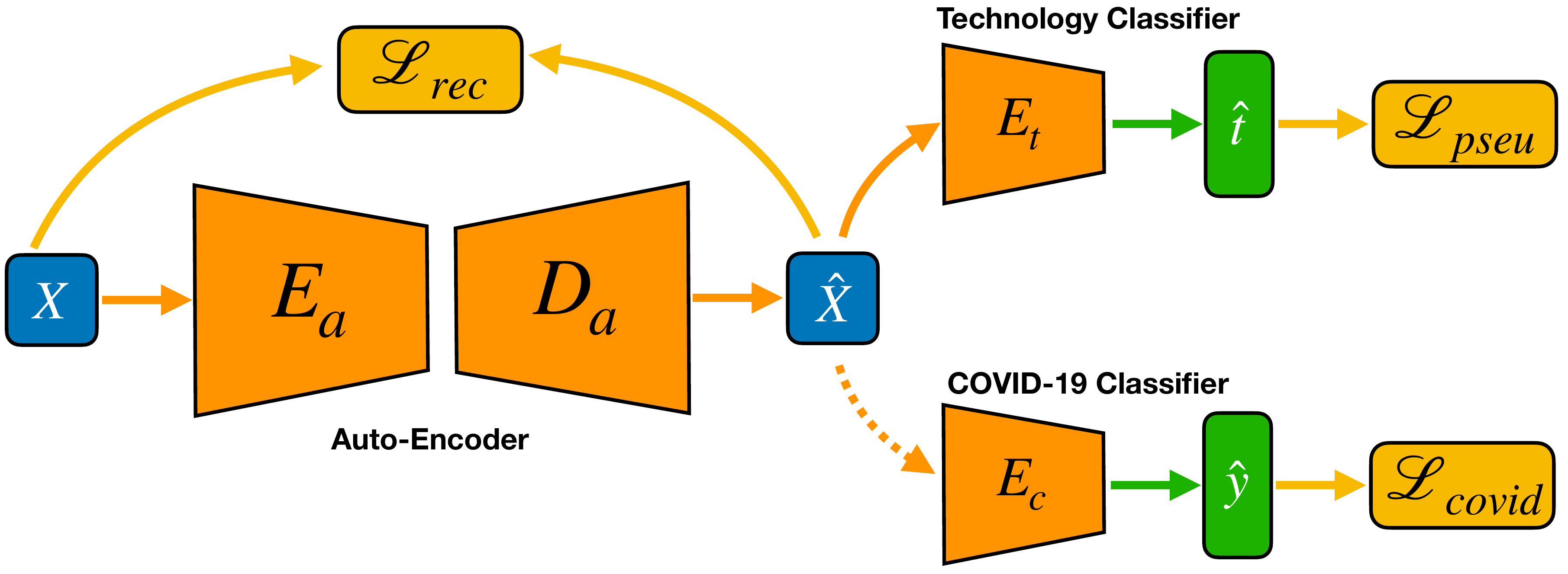}
    \caption{The proposed architecture for \textit{PrepNet} model with three modules: ($i$) an auto-encoder aims at CT dataset homogenizer; ($ii$) a multiclass classifier to recognized CT-datasets; and ($iii$) a binary classifier for diagnosis (COVID-19). The loss functions of the dataset classifier and auto-encoder were trained adversarially against each other. The binary classifier for diagnosis (COVID-19) was trained independently using the preprocessed scans by auto-encoder (figure adopted from~\cite{amirian2021prepnet}).}
    \label{chap:applications_fig:prepnet}
\end{figure*}

The research presented in this section is conducted in the context of COVID-19 detection from 2D chest computed tomography (CT) scans. This thesis presents a preprocessing network (PrepNet~\cite{amirian2021prepnet}) aiming at data homogenization with minimum changes in the original images. Accordingly, the proposed techniques have two main components: an autoencoder and a dataset/technology classifier. (see Figure~\ref{chap:applications_fig:prepnet}). The autoencoder-based CNN aims to find common ground for all the datasets and preprocesses them with minimal changes to fool the dataset classifier. The autoencoder and dataset classifier models are trained one after the other at each step of the training process. The preprocessing network aims at fooling the dataset classifier by erasing the differences between dataset samples, while the dataset classifier learns the differences between the new preprocessed scans. Two networks compete against each other to improve their performance in a similar optimization as generative adversarial networks (GANs). After sufficient training with the correct set of hyperparameters, the auto-encoder learns to bring the datasets into a joint distribution that looks similar to the human eye as well as CNNs. During the optimization, we minimize the reconstruction loss of the auto-encoder to keep the scan as unchanged as possible and only focus on erasing the dataset differences and reducing probable generative artifacts.     

\begin{table*}[htb!]
\centering
\resizebox{1.0\textwidth}{!}{
    \begin{tabular}{l|cccc|cccc|c|c|c}
        \toprule
        Test dataset $\rightarrow$ & \multicolumn{4}{c|}{SARS-COV-2} & \multicolumn{4}{c|}{UCSD COVID-CT} & Within Test & Cross-Dataset & Pre-trained  \\ 
        Dataset portion  & BA  & Sens     & Spec     & AUC      & Test  & Sens      & Spec     & AUC    & Average & Average & encoder\\
        \midrule
        \midrule
        & \multicolumn{11}{c}{\emph{COVID classifier}} \\
        \midrule
        SARS-COV-2      & $0.8924$  & $0.9292$ & $0.7876$ & $0.8584$ & $0.4433$ & $0.7835$  & $\boldsymbol{0.1262}$ & $0.4548$ & $\boldsymbol{0.8587}$ & $0.4159$ & \multirow{2}*{Yes}\\
        UCSD COVID-CT   & $0.3295$ & $0.3476$ & $0.2743$ & $0.3110$ & $\boldsymbol{0.8250}$  & $0.7113$  & $\boldsymbol{0.9320}$ & $\boldsymbol{0.8216}$ & (baseline) & (baseline) & \\ 
        \midrule
        \midrule
        & \multicolumn{11}{c}{\emph{AutoEncoder}} \\
        \midrule
        SARS-COV-2      & $0.8956$ & $\boldsymbol{0.9907}$ & $0.6460$ & $0.8183$ & $0.4983$ & $0.9175$   & $0.0970$ & $0.5073$   & $0.8555$ & $0.4836$ & \multirow{2}*{Yes}\\
        UCSD COVID-CT   & $0.49405$ & $0.6030$ & $\boldsymbol{0.3008}$ & $0.4519$ & $0.8154$  & $0.7216$ & $0.8846$ & $0.8031$ & ($-0.32\%$) & ($+6.77\%$) & \\
        \midrule
        \midrule
        & \multicolumn{11}{c}{\emph{PrepNet}} \\
        \midrule
         SARS-COV-2      & $\boldsymbol{0.9007}$ & $0.9353$ & $\boldsymbol{0.7982}$ & $\boldsymbol{0.8668}$ & $\boldsymbol{0.5157}$ & $\boldsymbol{0.9175}$  & $0.1067$ & $\boldsymbol{0.5121}$ & $0.8404$ & $\boldsymbol{0.5343}$ & \multirow{2}*{Yes}\\
        UCSD COVID-CT   & $\boldsymbol{0.5545}$ & $\boldsymbol{0.6446}$ & $0.1858$ & $\boldsymbol{0.4852}$ & $0.7800$ & $\boldsymbol{0.8556}$   & $0.7087$ & $0.7822$  & ($-1.83\%$) & ($+11.84\%$) & \\
        \toprule
    \end{tabular}}
    \caption{Test and cross-dataset performance of different methods. Using an adversarial loss to train a PrepNet improves the cross-dataset average performance (table adopted from~\cite{amirian2021prepnet}).}
    \label{chap:applications_tab:prepnet}
\end{table*}

\begin{figure*}[htb!]
    \centering
    \resizebox{\textwidth}{!}{\centering
\begin{tabular}{l | c | c c c c c}
\toprule
Dataset & COVID & Original & Histogram equalization & Normalization & Auto-encoder & \emph{PrepNet} \\
\toprule
SARS-COV-2 & Negative &
\includegraphics[width=3.5cm]{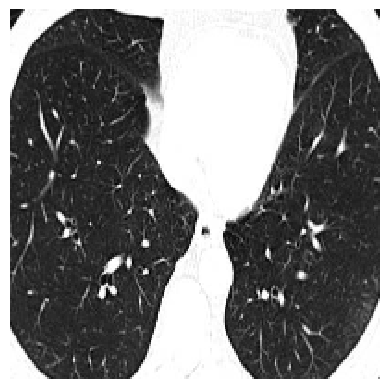} &
\includegraphics[width=3.5cm]{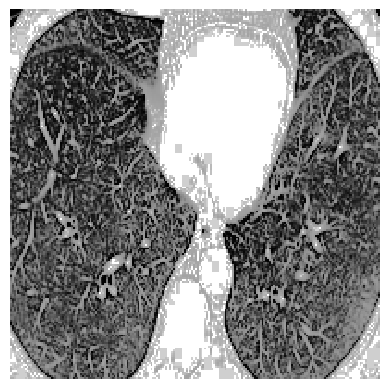} &
\includegraphics[width=3.5cm]{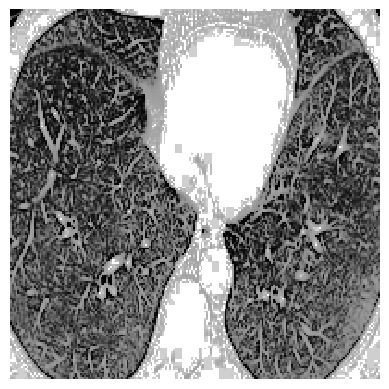} &
\includegraphics[width=3.5cm]{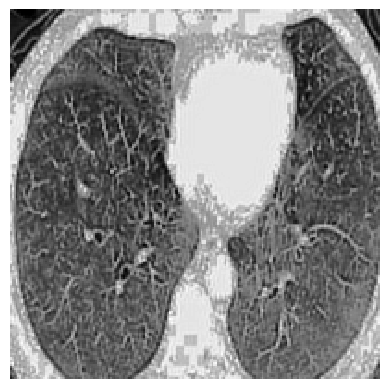} &
\includegraphics[width=3.5cm]{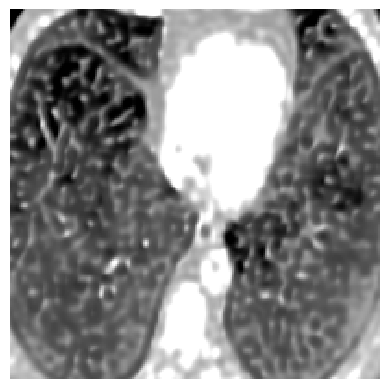}\\
\midrule
SARS-COV-2 & Positive &
\includegraphics[width=3.5cm]{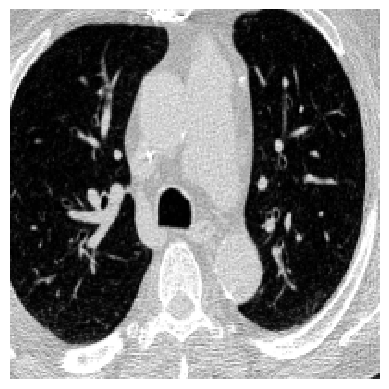} &
\includegraphics[width=3.5cm]{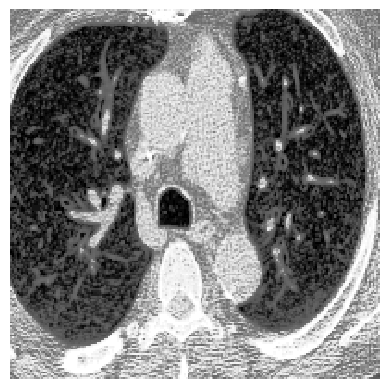} &
\includegraphics[width=3.5cm]{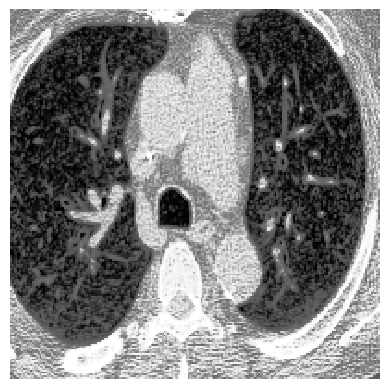} &
\includegraphics[width=3.5cm]{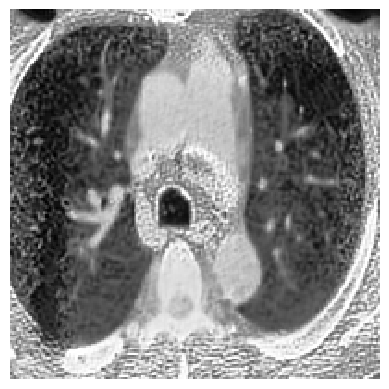} &
\includegraphics[width=3.5cm]{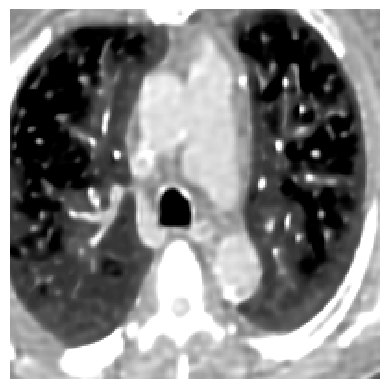}\\
\toprule
UCSD COVID-CT & Negative &
\includegraphics[width=3.5cm]{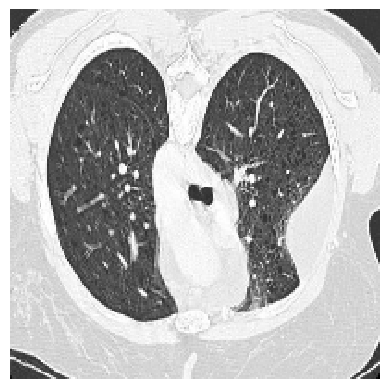} &
\includegraphics[width=3.5cm]{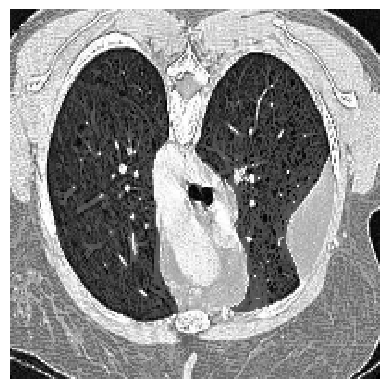} &
\includegraphics[width=3.5cm]{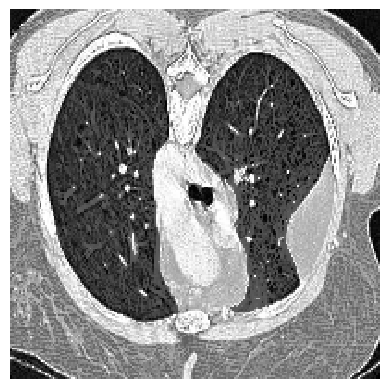} &
\includegraphics[width=3.5cm]{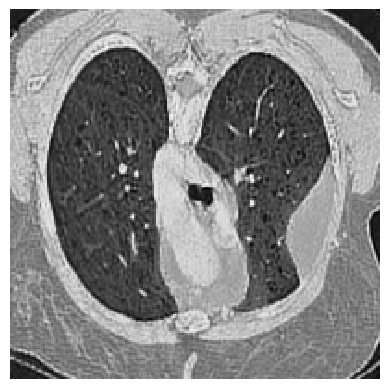} &
\includegraphics[width=3.5cm]{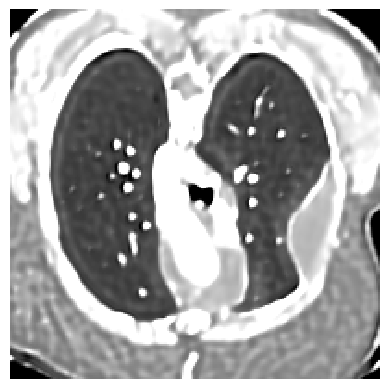}\\
\midrule
UCSD COVID-CT & Positive &
\includegraphics[width=3.5cm]{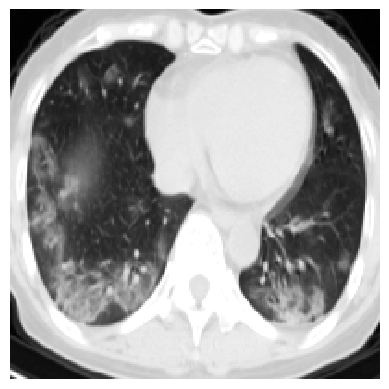} &
\includegraphics[width=3.5cm]{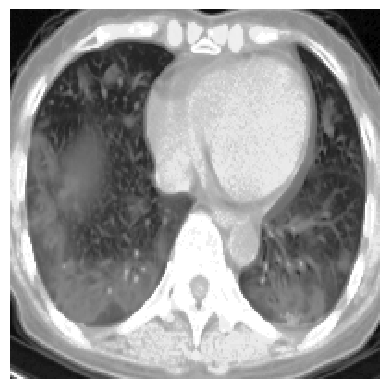} &
\includegraphics[width=3.5cm]{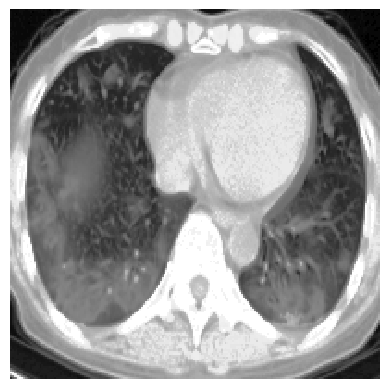} &
\includegraphics[width=3.5cm]{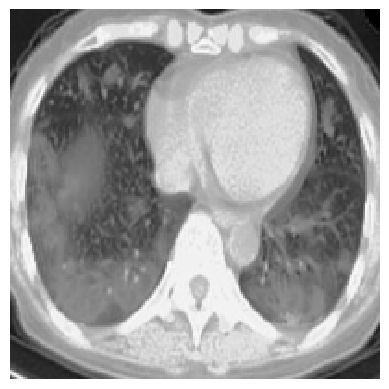} &
\includegraphics[width=3.5cm]{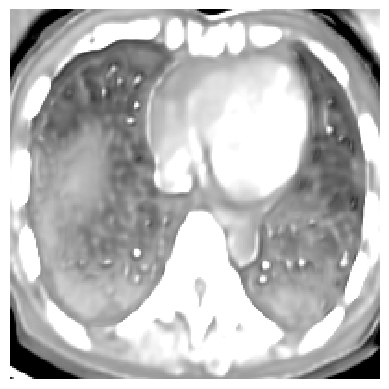}\\
\toprule     
\end{tabular}}
    \caption{Original images from the datasets with different prepossessing methods applied (figure adopted from~\cite{amirian2021prepnet}).}
    \label{chap:applications_fig:preprocessing}
\end{figure*}

The evaluation method proposed for PrepNet not only measures the intra-dataset test performance, but also focuses on cross-dataset performance. The ultimate goal of PrepNet is to homogenize datasets so that the model trained on one can be used for diagnosis on the other datasets. Two public datasets for COVID diagnosis from CT scans called SARS-COV-2\cite{DVN_SZDUQX_2020} and UCSD COVID-CT\cite{zhao2020covid} are the subjects of this study. Figure~\ref{chap:applications_fig:preprocessing} depicts the performance of our proposed PrepNet and visually compares its results with classical auto-encoders and other preprocessing techniques for chest CT scans. Table~\ref{chap:applications_tab:prepnet} shows the performance of the models trained on the original dataset, solely preprocessed using an auto-encoder learned in a self-supervised manner on reconstruction loss and preprocessed using PrepNet. PrepNet achieved the best cross-dataset generalization amongst all the presented methods with a minor drop in intra-dataset test performance. However, the gap between cross-dataset performance and intra-dataset performance is still significant, and there is considerable room for improvement in future research.

\newpage
\section{Face Recognition}
Face recognition (FR) and face matching technologies, especially in surveillance applications, were probably the most controversial models developed with (CNNs) for computer vision. The idea of identity matching and verification using images was so appealing for many applications such as online banking or intelligent surveillance that the research literature around developing models and collecting datasets for FR expanded rapidly~\cite{masi2018deep}. 

Researchers collected clean datasets for FR in academic research developments and datasets from real-world images~\cite{huang2008labeled}. Developing loss functions to compute generic embedding was a key component of extending FR to face matching on the face, which has not been seen in the training set. Triplet loss~\cite{sun2014deep,sun2015deeply,sun2016sparsifying} and more modern loss functions such as large margin cosine loss~\cite{wang2018cosface} and arccos loss with angular margin~\cite{deng2019arcface} are amongst such developments. Despite the scientific successes in this research area, social activists raised issues concerning fairness because of biases in FR systems. This section describes the issue of fairness and presents scientific findings in the context of FR systems. 

\subsection{Algorithmic Bias in FR Systems}
The research progress in FR technology was quick, and the models rapidly made their way into practical applications; however, multiple reports show some biases and inaccuracies against races that have fewer images in the training datasets\cite{hill2020wrongfully,mac2021facebook,lohr2018facial}. These incidences attracted much negative feedback from society, which was reflected in the news\footnote{\url{https://www.washingtonpost.com/technology/2021/02/17/facial-recognition-biden/}}\footnote{\url{https://www.bbc.com/news/technology-48276660}}. Another reaction followed this wave with companies starting to ban the FR technology\footnote{\url{https://www.banfacialrecognition.com/}}~\cite{krishna2020ibm}. As a result, using the FR technology started to be abandoned, and measuring algorithmic biases became more critical after these events~\cite {bellamy2019ai}.

\subsection{Measuring Bias and Awareness}
The main findings of our research are about methods of measuring and removing biases. After all the controversies regarding biases in FR technology, researchers quickly started to seek strategies for measuring the sources of such biases in FR. The pioneering research on collecting datasets with racial diversity rapidly exposed the gap in FR models' accuracy for different races, which causes limitations in service accessibility where FR technologies are involved and raises ethical issues regarding fairness. 

\begin{figure}[htb!]
    \vspace{-10pt}
    \centering
    \begin{tabular}{l|ccc}
        \toprule
        & VGGFace2 (128) & VGGFace2 (256) & VGGFace2 (2048)  \\
        \midrule
        \shortstack{Euclidean \\ distance}  & \includegraphics[width=3.45cm]{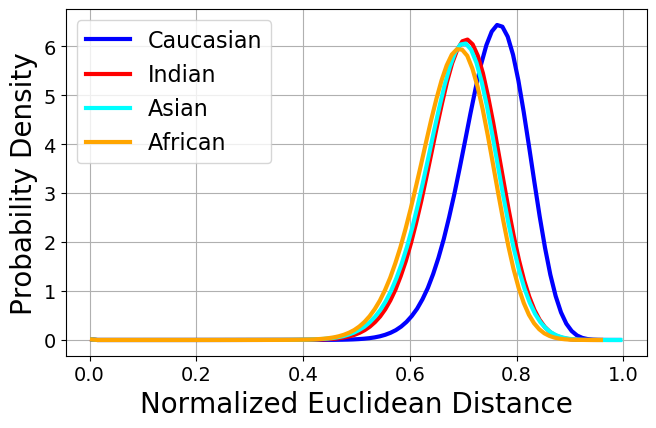} &
        \includegraphics[width=3.45cm]{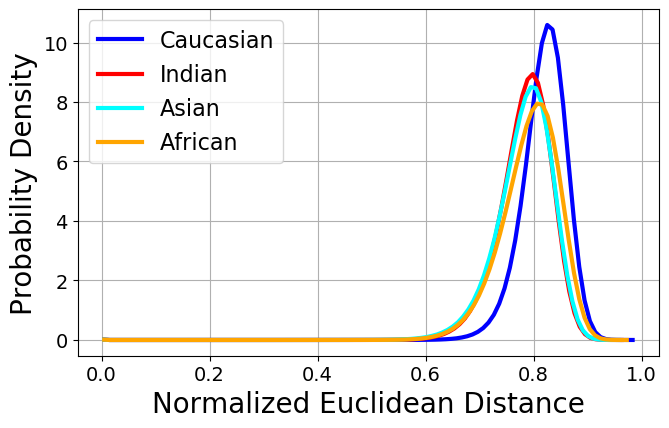} &
        \includegraphics[width=3.45cm]{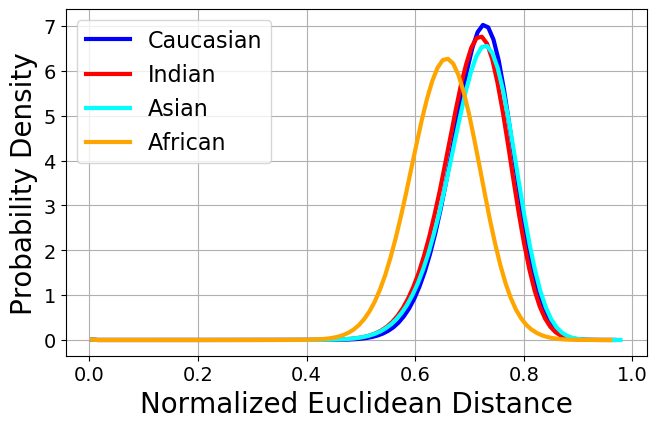} \\ \midrule
        \shortstack{Cosine \\ distance} &
        \includegraphics[width=3.45cm]{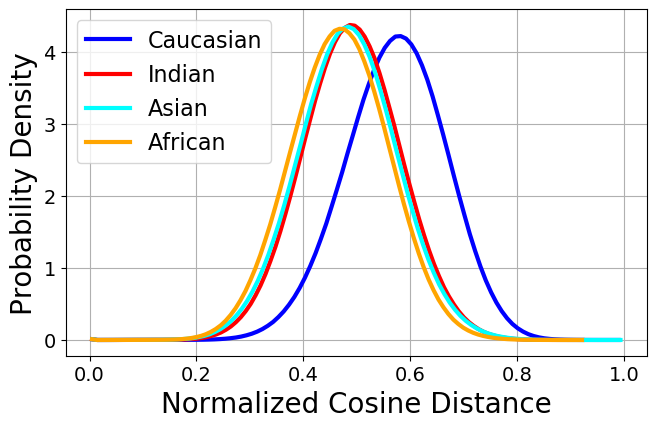} &
        \includegraphics[width=3.45cm]{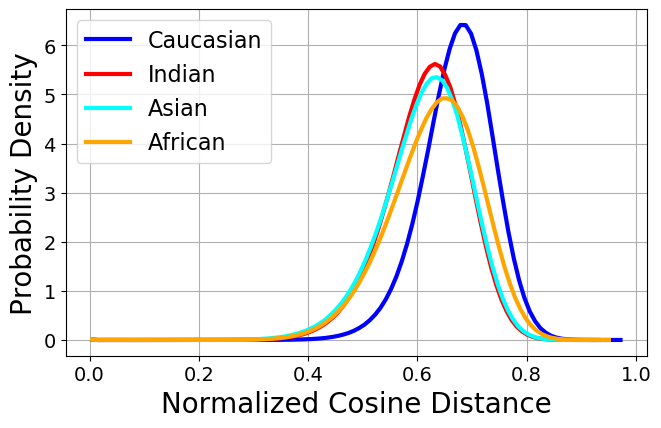} &
        \includegraphics[width=3.45cm]{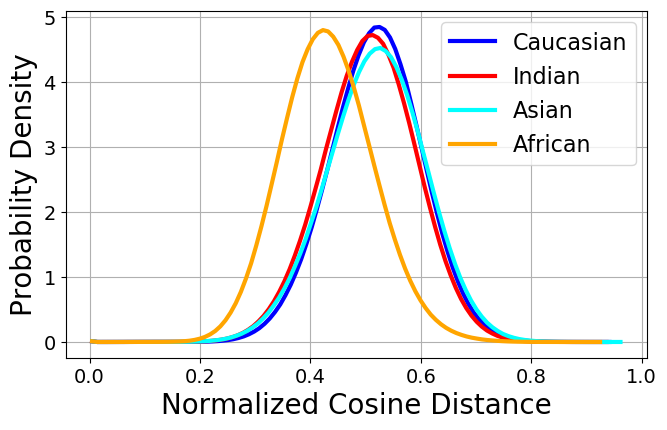} \\
        \toprule
    \end{tabular}
    \caption{Probability density distribution of pairwise (Euclidean and Cosine) distances between test images' embeddings of different races. The embeddings are computed using the VGG model fine-tuned for face recognition (VGGFace2~\cite{cao2018vggface2}) with different embedding dimensionalities ($128$, $256$ and $2048$). 
    The figure shows that the faces from the Caucasian race, which have the largest share of data samples, have a larger average distance than those of Africans, Asians, and Indians (figure adopted from~\cite{gluge2020not}).}
    \label{chap:applications_fig:FR}
    \vspace{-10pt}
\end{figure}

Researchers introduced racial awareness as a proxy for measuring biases in FR models and research showed that the FR models distribute the faces based on their ethnicity in the embedding space~\cite{wang2019racial}. Accordingly, several research works suggested adversarially removing the racial information as a solution to the problem of biases in FR systems~\cite{xu2018fairgan,kim2019learning,yucer2020exploring}. However, our research in measuring the biases demonstrated that the intuitive idea that racial clustering in embedding space is correlated with biases is not always true~\cite{gluge2020not}. Instead, the reason behind racial biases comes from how the faces of different races are distributed in the embedding space (see Figure~\ref{chap:applications_fig:FR}). Similarly, blinding the FR technologies from racial information in the embedding space does not necessarily lead to decreasing the racial bias~\cite{wehrli2021bias}. Hence, awareness and bias are two distinct though related issues in FR, and methods dealing with ethnicities individually, such as the research presented in~\cite{robinson2020face} does, are more appealing based on these findings.  

\newpage
\section{Rotation-Invariant Vision Transformers}
\label{chap:applications_sec:vits}
Inductive biases such as translation invariance undeniably accelerated the rapid advances of modern vision models based on convolutions through parameter sharing and improving sample efficiency. However, state-of-the-art models can only partially incorporate rotation invariance. Recent attempts to develop rotation-invariant techniques mainly face the challenge of high memory requirements or limiting the original model capacity. This section proposes an embedding layer method for vision transformers to leverage the invariance of self-attention layers to the order of tokens and train robust models against local and global rotation. The proposed image embedding technique requires negligible memory overhead to train rotation invariance models on large datasets such as ImageNet\cite{ILSVRC15}. Furthermore, the proposed method improves the robustness of vision transformers against rotation on the classification task. 

\subsection{Introduction and Problem Statement}

\begin{figure*}[htb]
    \centering
    \centering
    \resizebox{\textwidth}{!}{
    \subfloat[Classification]{\includegraphics[width=0.5\textwidth, valign=t]{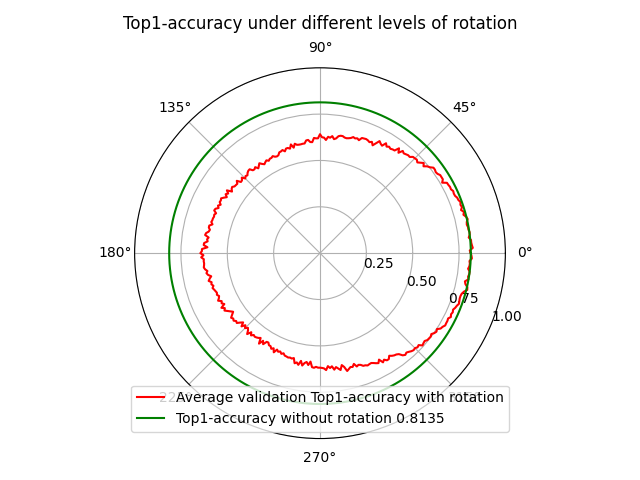}}
    \subfloat[Segmentation]{\includegraphics[width=0.5\textwidth, valign=t]{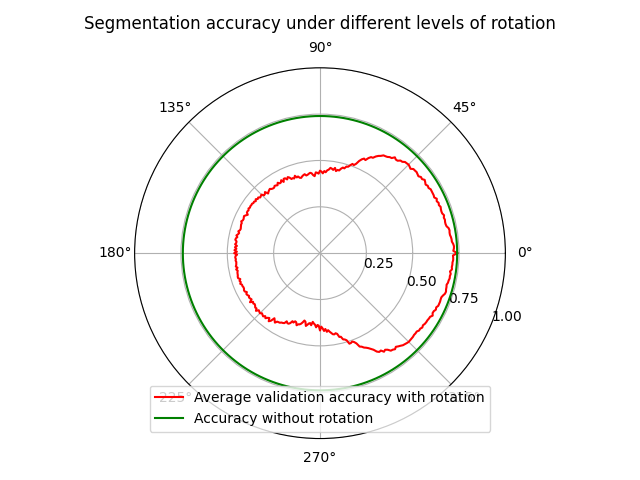}}
    \subfloat[Augmentation]{\includegraphics[width=0.5\textwidth, valign=t]{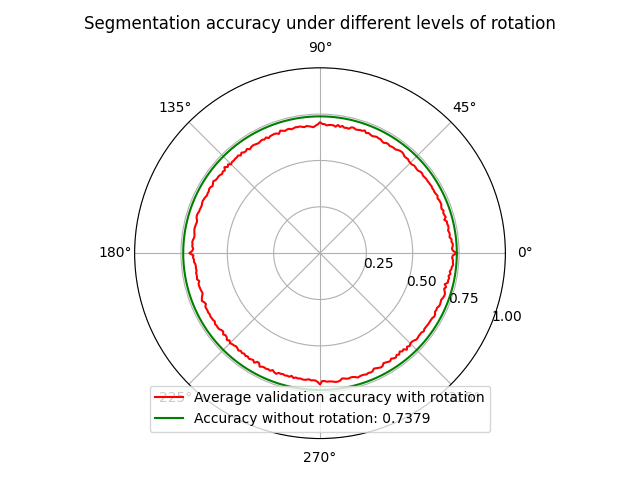}}}
    \caption{Classification and segmentation performance of vision transformers under different degrees of rotation. Augmentation improves the robustness of vision transformers against rotation; however, rotation invariance encoded in the method as inductive bias can improve the sample efficiency of the models.}
    \label{chap:applications_fig:vit_problem}
\end{figure*}

The performance of vision models, more specifically vision transforms (ViTs), drops when the input images are not presented to the models in the original pose. Rotation and scaling are two transformations that researchers found to be a reason for the decline in the performance of vision models from early works\footnote{\url{http://yann.lecun.com/exdb/lenet}}. This section presents a solution to rotation invariance in ViTs for object classification. Figure~\ref{chap:applications_fig:vit_problem} shows the decline in the performance of a ViT-based classifier and segmentation model after different degrees of rotation. Besides compensating for the drop in accuracy to improve the robustness of vision models, developing rotation equivariant methods was very appealing to add another inductive bias to enhance the vision models' sample efficiency and convergence speed.  

Several different techniques aim to improve the vision model's robustness against rotation. These methods can be divided into two categories: 1) Preprocessing data for training or evaluation. 2) Using equivariance or invariance inductive biases in vision models. The first group uses data augmentation, a conventional technique widely used in deep learning, to increase the size of the datasets artificially, improve the generalization, and train robust models \cite{simard2003best, sato2015apac}. The second group of the research can be summarized for CNNs and vision transforms (ViT) as follows:

\textbf{CNNs}: Cohen et al. introduced group equivariant convolutional neural networks (G-CNNs) to learn equivariant representation for discrete symmetry groups of rotations~\cite{cohen2016group}. Marcos et al. proposed the rotation of the convolutional filters instead of lifting the representation to the group and using the pooling and vector field representations of the input to achieve rotation invariant, covariant, and equivariant features~\cite{marcos2017rotation}. Lifting the representations or filters to the discrete Lie groups increases the memory consumption linearly with the group size. Alternatively, the convolutional filter can be designed to be equivariant to specific transformations. Esteves et al. train isotropic filters for rotation equivariant CNNs~\cite{esteves2018learning} and Weiler et al. proposed learning the models' weights which are expansion coefficients for the steerable function space~\cite{weiler2018learning}. Wiersma et al. presented a surface harmonic network with both invariant and equivariant features~\cite{wiersma2020cnns}. The main disadvantage of optimizing equivariant filters is limiting the capacity of models for learning the data.  

\textbf{ViTs}: Romero and Cordonnier used the group lifting concept to train equivariant vision transformers on a discrete group of image rotations~\cite{romero2020group}. Hutchinson et al. adapted a similar idea, generalized it to the continuous rotation and translation equivariant models, and applied their method to pattern recognition in point-could graphs, molecular property prediction, and chasing particle dynamics~\cite{hutchinson2021lietransformer}. Finally, Su et al. demonstrated that rotary positional encoding enhances the performance of natural language processing models~\cite{su2021roformer}. 

Next, this section reviews the main blocks and concepts used in ViTs. The explanation of rotation invariant and equivariant features is followed by self-attention and formulation under input rotation. 

\textbf{Rotation}:
Let $\boldsymbol{X}$ be a vectorized patch of an image with a given size, for example, $16\times16$. Then, we define a rotation matrix called $\boldsymbol{R}$ such that the transformed version of the original image $\boldsymbol{x}_\theta$ can be computed as follows:
\begin{equation}
    \boldsymbol{X}_\theta = \boldsymbol{R}\boldsymbol{X}
\end{equation}
where $\theta$ shows the angle of rotation. The rotation transformation is defined using a rotation matrix, and it can be computed as follows:
\begin{equation}
    \boldsymbol{R} = \boldsymbol{X}_\theta\boldsymbol{X}^\dagger
\end{equation}
The goal of the roto-translation equivariant models with a self-attention layer is computing a representation ($\mathcal{L}(\boldsymbol{x})$) in which the representations rotates with the same degree as the input image:
\begin{equation}
    \mathcal{L}(\boldsymbol{X}_\theta) \approx \mathcal{L}(\boldsymbol{X})
\end{equation}
Alternatively, we can consider the images as a spatial function in 2D space having three values (RGB vector) at every position and define the rotation on every pixel coordinate ($(x, y)$) as follows:
\begin{equation}
    r_\theta =
  \left[ {\begin{array}{cc}
    \cos{\theta} & -\sin{\theta} \\
    \sin{\theta} & \cos{\theta} \\
  \end{array} } \right] 
\end{equation}
Given the definition of the rotation matrix ($r_\theta $) based on pixel coordinates, the inverse of the rotation operator is equal to its transpose ($r_\theta r^T = \boldsymbol{I}$), and the following properties hold accordingly:
\begin{equation}
    \left[ {\begin{array}{c}
    x_\theta \\
    y_\theta \\
  \end{array} } \right] = r_\theta    
  \left[ {\begin{array}{c}
    x \\
    y \\
  \end{array} } \right] \\
\end{equation}

\begin{equation}
    \left[ {\begin{array}{c}
    x \\
    y \\
  \end{array} } \right] = r_\theta^T    
  \left[ {\begin{array}{c}
    x_\theta \\
    y_\theta \\
  \end{array} } \right]
\end{equation}

\textbf{Rotation invariance, covariance, and equivariance}: A representation ($\mathcal{L}(.)$) of an input pattern ($X$) is invariant to rotation ($R$) if it does not change with the rotation of the input. The equivariance representations rotate similarly with the input's rotation; However, covariant representations change according to the original representations based on a constant function ($f(.)$). These definitions can be shown in the following equations:

\begin{equation}
\begin{split}
    \textit{Invariant}:~~ &~~ \mathcal{L}(X_\theta) \approx \mathcal{L}(X) \\
    \textit{Equivariant}: &~~ \mathcal{L}(X_\theta) \approx \mathcal{L}_\theta(X) \\
    \textit{Covariant}:~~  &~~ \mathcal{L}(X_\theta) \approx f (\mathcal{L}(X))
\end{split}
\end{equation}

\textbf{Self-Attention}: The output of the self-attention layer for an image ($\boldsymbol(X) \in \rm I\!R ^{N\times T}$) converted to $N$ tokenized patches of length ($T$) can be written as follows:
\begin{equation}
\begin{split}
    \boldsymbol{Q} & := \boldsymbol{X}\boldsymbol{W}_q \\
    \boldsymbol{K} & := \boldsymbol{X}\boldsymbol{W}_k \\
    \boldsymbol{V} & := \boldsymbol{X}\boldsymbol{W}_v \\
    \boldsymbol{A} & := \boldsymbol{Q}\boldsymbol{K}^T \\
    \boldsymbol{Y} & := \textit{SA}(X) \\
                   & := softmax(\boldsymbol{A})\boldsymbol{V}\\
\end{split}
\end{equation}
where $\boldsymbol{K}$, $\boldsymbol{Q}$ and $\boldsymbol{V}$ shot the key, query and value. The linear weights used to compute the representations are denoted by $\boldsymbol{W}_k$, $\boldsymbol{W}_q$ and $\boldsymbol{W}_v$ for keys, queries and values, respectively. $\boldsymbol{A}$ shows the attention matrix and $\boldsymbol{Y}$ denotes the self-attention ($\textit{SA}$) layer. The softmax function, denoted by $softmax$, is defined as follows:
\begin{equation}
    softmax(\boldsymbol{x}_i) = \frac{exp(\boldsymbol{x}_i)}{\sum_{j}^{ }exp(\boldsymbol{x}_j)}
\end{equation}

\textbf{Self-Attention with Rotation}: 
Then, we can write the rotation invariant self-attention objective as follows:
\begin{equation}
\begin{split}
    \mathcal{L}(\boldsymbol{X}_\theta) &  = softmax(\boldsymbol{X}_\theta\boldsymbol{W}_q(\boldsymbol{X}_\theta\boldsymbol{W}_k)^T)\boldsymbol{X}_\theta\boldsymbol{W}_v \\
    & = softmax(\boldsymbol{X}_\theta\boldsymbol{W}_q\boldsymbol{W}_k^T\boldsymbol{X}_\theta^T)\boldsymbol{X}_\theta\boldsymbol{W}_v \\
    & = softmax((\boldsymbol{R}\boldsymbol{X})\boldsymbol{W}_q\boldsymbol{W}_k^T(\boldsymbol{R}\boldsymbol{X})^T)(\boldsymbol{R}\boldsymbol{X})\boldsymbol{W}_v \\
    & = softmax(\boldsymbol{R}\boldsymbol{X}\boldsymbol{W}_q\boldsymbol{W}_k^T\boldsymbol{X}^T\boldsymbol{R}^T)\boldsymbol{R}\boldsymbol{X}\boldsymbol{W}_v \\
    & \approx softmax(\boldsymbol{X}\boldsymbol{W}_q\boldsymbol{W}_k^T\boldsymbol{X}^T)\boldsymbol{X}\boldsymbol{W}_v \\
    & = softmax(\boldsymbol{X}\boldsymbol{W}_q(\boldsymbol{X}\boldsymbol{W}_k)^T)\boldsymbol{X}\boldsymbol{W}_v \\
    & = \mathcal{L}(\boldsymbol{X}) \\
\end{split}
\end{equation}

\subsection{Method and Experimental Results}
The mathematical formulation of self-attention with rotation suggests that it is possible to constrain the key, query and value matrices to make self-attention equivariant. The necessary condition is that both rotation matrices ($\boldsymbol{R}$ and $\boldsymbol{R}^T$) can commute\footnote{Two matrices $\boldsymbol{X}$ and $\boldsymbol{W}$ called to commute if $\boldsymbol{X}\boldsymbol{W}=\boldsymbol{W}\boldsymbol{X}$} through $\boldsymbol{X}\boldsymbol{W}_q\boldsymbol{W}_k^T\boldsymbol{X}^T$ and its softmax. This is the necessary condition to make self-attention equivariant ($\mathcal{L}(\boldsymbol{X}_\theta) = \boldsymbol{R}_\theta\mathcal{L}(\boldsymbol{X})$) which is more complicated than invariance, and it is also more appealing since equivariant features are useable in building invariant models. However, invariant models do not necessarily provide equivariant features.

 The problem of equivariant self-attention is an open problem for future research. However, this thesis offers a solution to rotation-invariant ViTs based on a fundamental property of self-attention. The self-attention mechanism is invariant to the order of the tokens, meaning the representations do not change when the order of the tokens is different. Therefore, if we transform the image so that rotation only changes the order of the tokens, then the ViT based on self-attention will be invariant and robust against rotation. 

\begin{figure*}[htb!]
    \centering
    \centering
    	\subfloat[Isotropical]{\includegraphics[width=0.4\textwidth, valign=t]{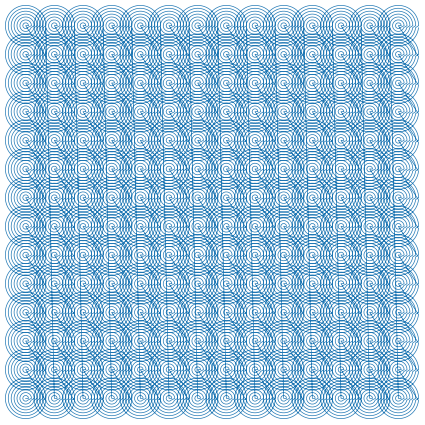}}
    \raisebox{-0.2cm}{\subfloat[Local radial]{\includegraphics[width=0.17\textwidth, valign=t]{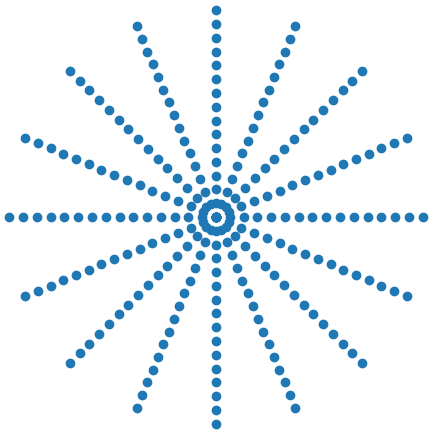}}}
    \subfloat[Radial]{\includegraphics[width=0.4\textwidth, valign=t]{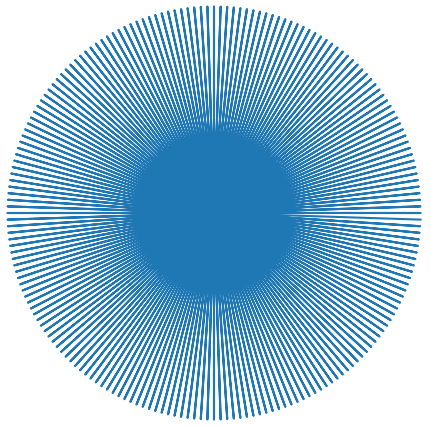}}
    \caption{The proposed patch embedding methods for vision transformers: a) Isotropic path embedding for the entire image. Every patch is sampled based on the circles around the center of the patch. b) Radial patch embedding in the patch level technique samples every patch based on the pixels on a circle radius positioned at the patch's center. c) Radial patch embedding for the entire image.}
    \label{chap:applications_fig:embedding}
\end{figure*}

The idea of rotation invariant ViTs can be realized using a radial tokenization technique presented in Figure~\ref{chap:applications_fig:embedding}. The idea is to take the tokens based on the polar coordinate and extract every token from the original image. The proposed method uses pixel values on a circle's radius placed at the center of the image instead of turning patches of size $16\times16$ into tokens. Using this embedding method, only the order of the tokens changes with the input rotations, and the whole ViT stays invariant to rotation. This idea works for global rotation; however, it can also be implemented at the patch level to tackle the local rotation of images' elements, which is more critical for medical applications~\cite{kumar2017dataset}.

Steerable Convolutions and isotropic filters inspire this section's other patch embedding techniques. Figure~\ref{chap:applications_fig:embedding} shows how isotropic patch embeddings turn the original image into patches. The idea here is to divide the original models into patches of size $16\times16$ and then sample them via circles around the center of the patch and project them into tokens afterward. Implementing the radial patch embedding technique at the patch level can train robust models against local rotations.  

\begin{figure}[htb!]
    \centering
    \centering
    \subfloat[Classification (top1)]{\includegraphics[width=0.33\textwidth, valign=t]{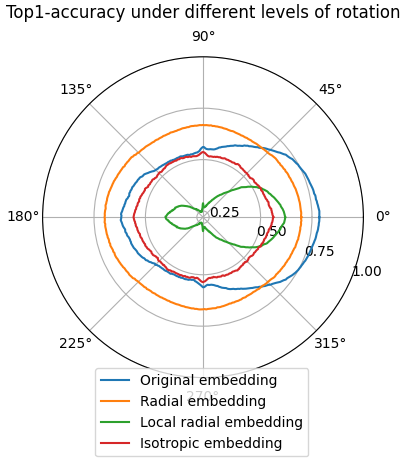}}~~~~~~~~~~
    \subfloat[Classification (top5)]{\includegraphics[width=0.33\textwidth, valign=t]{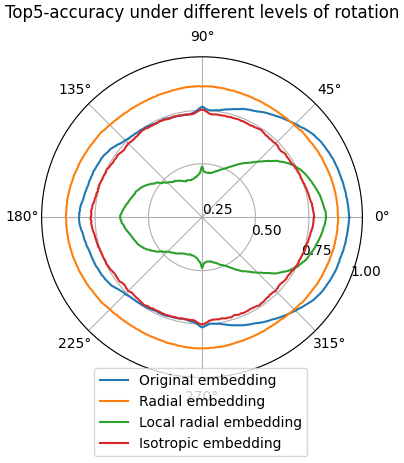}}
    \caption{Robust training against rotation using rotation invariant patch embedding techniques.}
    \label{chap:vit_fig:vit_robust}
\end{figure}

A ViT architecture based on Deit's baseline model\cite{touvron2021training} is optimized on the ImageNet dataset for pertaining, and initial evaluation shows the functionalities of the proposed methods. Figure~\ref{chap:vit_fig:vit_robust} depicts the performance of the different patch embedding methods used to improve the robustness of the ViTs against rotation. Radial and isotropic patch embeddings demonstrate considerably higher robustness against rotation compared with the original transformer. However, it is notable that training a transformer using data augmentation is a very competitive solution to the presented problem. Table~\ref{chap:applications_table:vit} shows that rotation invariant patch embedding models pretrained on ImageNet generalize to the other related object detection tasks with a drop in the performance compared to the original transformer method. 

This preliminary study shows that transformers can run in the rotation invariant mode without memory and compute overhead by adjusting the patch embedding techniques. Compared to similar methods such as group equivariant self-attention~\cite{romero2020group}, the proposed method profits from higher memory efficiency and angular resolution. The research questions regarding evaluating rotation invariance to ViT based on the initial motivations and goal, namely sample efficiency and training speed, are still open for further investigation.


\begin{table}[t!]
\begin{center}
\resizebox{0.8\textwidth}{!}{  
\begin{tabular}{ l | c | c c c c | c }
\toprule
\multirow{2}{*}{Dataset} &  & \multicolumn{4}{c|}{Base Vision Transormer} & Best \\ 
& Accuracy & Original & Isotropic & Radial & Local Radial & (SOTA) \\ 
\midrule
\multirow{2}{*}{Oxford-IIIT Pets~\cite{parkhi2012cats}}      
& top1 & $\boldsymbol{0.9305}$ & $0.6890$ & $0.8575$ & $0.8133$ & $0.9710$  \\ 
& top5 & $\boldsymbol{0.9926}$ & $0.9302$ & $0.9839$ & $0.9725$ & -         \\ \midrule
\multirow{2}{*}{Oxford Flowers~\cite{nilsback2008automated}} 
& top1 & $\boldsymbol{0.9167}$ & $0.8000$ & $0.8510$ & $0.8402$ & $0.9976$ \\ 
& top5 & $\boldsymbol{0.9696}$ & $0.9186$ & $0.9461$ & $0.9461$ & -        \\ \midrule
\multirow{2}{*}{FGVC Aircraft~\cite{maji2013fine}}           
& top1 & $\boldsymbol{0.7570}$ & $0.3378$ & $0.5794$ & $0.5425$ & $0.9490$ \\ 
& top5 & $\boldsymbol{0.9355}$ & $0.6439$ & $0.8599$ & $0.8428$ & -        \\ \midrule
\multirow{2}{*}{Caltech Birds~\cite{WelinderEtal2010}}       
& top1 & $\boldsymbol{0.7960}$ & $0.5221$ & $0.6933$ & $0.6262$ & $0.9548$ \\ 
& top5 & $\boldsymbol{0.9462}$ & $0.7886$ & $0.8952$ & $0.8714$ & -        \\ \midrule
\multirow{2}{*}{ImageNet~\cite{ILSVRC15}}                    
& top1 & $\boldsymbol{0.7884}$ & $0.5698$ & $0.7091$ & $0.6276$ & $0.9088$ \\
& top5 & $\boldsymbol{0.9370}$ & $0.7813$ & $0.8919$ & $0.8364$ & -        \\ \toprule

\end{tabular}}
\end{center}
\caption{The performance of rotation invariant vision transformers on several vision benchmark vision datasets. Rotation invariant patch embedding increases the robustness of ViTs at the expense of a decrease in performance.}
\label{chap:applications_table:vit}
\end{table}


\chapter{Conclusions}
\label{chap:conclusion}

DL breakthroughs and computer vision models developed based on DL revolutionized the research areas of image, video, and information processing in the last decade. Deep CNNs have become so popular that it is incredibly cumbersome and rare to find cases in which classical approaches can still outperform CNNs on academic datasets. Despite the undeniable breakthroughs, DL methods have faced arduous challenges for deployment in practical applications. This thesis discussed many such challenges and presented scientific developments to tackle these challenges. Nonetheless, there is still considerable room for further research to accelerate the entrance of DL-based techniques into practical applications. This chapter briefly summarizes the thesis, revisits the research challenges such as trustworthiness, explainability, robustness, optimization, and fairness, points out this thesis's contribution, and draws an outline for future work.

\section{Summary of Thesis}
This thesis has been motivated by the hindrances of using DL, specifically vision models, in practical applications. After describing the challenges and laying the theoretical foundation in the first two chapters, the thesis presented an alternative to classical multilayer perceptrons (MLPs) and instead uses radial basis function networks (RBFs) as classifiers for convolutional neural networks (CNNs). RBFs have been in the scientific literature for a long time. However, they have not been optimized for CNNs before because of the complications in the optimization. This thesis offered theoretical breakthroughs to adapt RBFs for CNNs to improve the robustness and interpretability of the classification. 

The interpretability of CNNs has been at the center of attention in many research works recently. However, methods such as guided-backpropagation~\cite{springenberg2014striving} developed in this context have mostly been used to monitor the models' behavior~\cite{reyes2020interpretability}. The fourth chapter of this thesis extended the idea of understanding vision models and putting them into action for debugging CNNs and detecting adversarial attacks with the hope of inspiring more such research in the future. 
This thesis's fifth and sixth chapters focused on ML and DL applications. Chapter~\ref{chap:motion} described how a problem without analytical solutions can be addressed using data-driven methods and simulation. It presented motion compensation in cone-beam computed tomography (CBCT) scans using 3D-CNNs. Chapter~\ref{chap:applications} reviewed several different applications of ML and DL in affective computing and health care, and pointed at the findings of this thesis targeting fairness in facial recognition systems and robustness of vision transformers (ViTs). The optimization process is crucial in bringing vision models to performance and affects their behavior in terms of robustness, generalization, and data requirements. Chapter~\ref{chap:applications} also offered findings in hyperparameter and model optimization gained by employing ML and DL in several applications and formulating the best practices and patterns in the automated search for best ML and DL models.
 
\section{Future Research Work}

Researchers' long-term vision of applying ML and DL in medical applications and autonomous driving systems is only feasible by establishing human trust in reliable and robust artificial intelligence (AI). Thus, \textit{trustworthiness} and \textit{reliability} are the overarching themes in the research community for practical AI applications with maximum performance and minimum negative impact~\cite{kaur2022trustworthy}. It is intuitively clear that a model or algorithm used in applications involving human privacy or service access has to be reliable and trustworthy. Furthermore, trustworthiness is in demand in medical applications and autonomous driving systems involving human life and security. 

Despite the demand for trustworthiness being intellectually evident, best practices for engineering trustworthy models for a specific application is an open problem and requires further investigation~\cite{serban2021practices}. The importance of trustworthiness is also highly dependent on the application. For instance, robustness against spoofing or adversarial attacks is more relevant to person identification problems, while adaptation to the new vendors and image acquisition parameters emerge in medical image processing. Since the term trustworthiness is generic and includes many aspects, researchers break it down into several categories with more specific definitions where it is also possible to evaluate the performance based on acceptable common-sense explanations or mathematical metrics. 

The long-term vision of AI research (reliability and trustworthiness) can be divided into smaller actionable blocks that current research addresses. Explainability, robustness, and fairness are the requirements of the trustworthy AI concepts investigated in this thesis. The remainder of this section explains this thesis's contributions to the components of trustworthy AI and opportunities for mid-term research in these areas.

\textit{Explanability}: Answering the following three questions is the target of the research around explainable AI in computer vision: 1) How do models learn? 2) What do models learn? 3) How do models predict? The first question is the most complicated to answer. The research literature addressing this area is meager, but includes studies that use information theory to explain the behavior of models during optimization~\cite{saxe2019information, shwartz2017opening}. The second and third questions are more pragmatic, relevant for practical applications, well-studied, and more connected to each other~\cite{springenberg2014striving, gilpin2018explaining, reyes2020interpretability}. The features learned in vision models for decision-making are mainly evaluated using feature visualization techniques. These techniques compute the region of input images that the models look at to make a decision based on reverting the forward path or treating the models as a black box using iterative optimization. Moreover, researchers investigated the behavior of models as black boxes via post-hoc analysis to identify why the models predict a specific class. An alternative to black-box analysis is using methods such as Bayesian inference, which are more transparent by design. The contribution of this thesis to explainable AI research is revisiting radial basis function networks (RBFs) and adapting them as classifiers for CNNs by solving a few architectural hindrances. The proposed models compute a similarity metric between test and training images and derive visual clues about the decision-making process of the vision models. This research is the first to use RBFs on top of the traditional computer vision backbones. Evaluating the robustness of models using RBF classifiers against anomalies and adversarial attacks is an open question for future research.

\textit{Robustness}: Researchers very quickly discovered robustness issues in computer vision models. CNN performance shows a decline in the presence of different lighting conditions and variability in the pose of the input images. The robustness problem had even more impact in the medical domain because of manual changes in image acquisition parameters, different image acquisition vendors, and frequent imaging software and hardware updates. Domain adaptation and life-long learning in the presence of concept drift are the offsprings of the robustness and generalization issues and have tremendous exciting research potential. This thesis presented a method for data homogenization that enhances merging data from different datasets and it is practical for domain adaptation. One of the hot topics threatening the validity of CNN's for vision problems is adversarial attacks. Researchers have found that images which appear identical to the human eye can be optimized to fool vision models into making an incorrect decision. This thesis offered a method based on reverting the CNNs to visualize the models' feature response and detect adversarial attacks with very high accuracy. This research can be extended to use black-box feature visualization to detect attacks on any model and optimize the input to reduce the adversarial effects in future work. Moreover, this thesis presented a novel embedding technique for rotation invariant vision transformers to improve model robustness against input rotation. Applying rotation invariant transformers to small datasets, especially aerial images and histology datasets, to leverage the rotation invariance as inductive bias is another promising research offspring of this thesis.

\textit{Fairness}: Neural networks became very popular because of their strength in approximating arbitrary functions for classification or regression solely from data without any knowledge of the task. Although neural networks provide the opportunity to learn with minimal inductive biases, the optimization process instead follows the most efficient direction in parameters space to minimize the optimization objective (loss function) based on the existing biases in the datasets. Using these biases helped to solve the problems that researchers had not found any analytical solution to before, such as motion artifact reduction presented in Chapter~\ref{chap:motion}. However, social activists rapidly discovered the drawbacks of learning from data in the social fairness aspect of face recognition (FR) systems for surveillance. The collected datasets were biased, in that the majority of the images were of white male celebrities, which was reflected in the trained models when they returned a higher accuracy for the majority race in the datasets. Studies showed that the models produce a lower accuracy for racial minorities, and that this inequality was even visible when comparing the models' accuracy for recognizing females and children with males. This thesis offered relevant research and findings about a standard method of measuring biases in FR systems and showed that racial awareness and bias are not necessarily correlated. The research concerning fairness is also quite an exciting and simultaneously challenging area. Data-driven techniques are an option for reducing biases by collecting datasets with equal populations from all sensitive features, such as race and gender. A balanced dataset is a solution to the problem faced by FR systems. However, problems such as recruitment and job application processing confront more challenges due to biases in ground truth labels based on previous hiring decisions, which opens a lot of fascinating topics for future research.   


\section{Practical Discussions}

Alongside all the debates about the trustworthiness of AI models for applications where human safety and privacy are involved~\cite{kaur2022trustworthy, ahmed2021intelligent, huang2019lightweight}, AI-based models have also found their way into less critical applications~\cite{stadelmann2018deep}. However, AI projects still suffer from a very high failure rate in development and post-deployment due to problems such as concept drift. This section briefly discusses the content of this thesis related to applications and optimization.        

\textit{Applications}: Despite the early challenges in deploying ML and DL techniques, this thesis has shown several successful examples of ML and DL in real-world applications. Data preprocessing and cleaning before training a model is one of the most critical components of any ML pipeline. Face alignment for FR systems or facial expression estimation is an example of data preparation before training. Although DL-based models are unrivaled for vision problems, their performance is highly dependent on the quality of the data. Determining the mutual information of the data samples and target pattern requires further research; a visual review of the datasets before model development is the key to success in applied projects~\cite{Karpathy2019Recipe}. Neural networks are applicable and highly recommended to approximate classical methods that are computationally expensive (such as the iterative reconstruction of computed tomography scans) or enhance their performance where analytical solutions do not exist (for example, in motion artifact reduction). This thesis offered an application of three-dimensional CNNs in reducing motion artifacts in volumetric cone-beam computed tomography (CBCT) scans with great success. This research path was extraordinarily successful and gained positive feedback and attention from clinical experts. The particular area of research is novel and ripe for further research in similar applications, such as sparseness artifact reduction, auto-segmentation, and dose calculation from CBCT scans for cancer therapy.    

\textit{Optimization}: ML and DL present the opportunity to explore and search among a family of neural networks to model all possible problems in computer vision. However, this large degree of freedom appears at the expense of the vast search space of parameters and potential models. Optimization concentrates on techniques that are key to neural architecture search, hyper-parameter (HP) tuning, and finding the shortest path to a stable minimum for a given dataset and model. This thesis presented the observed patterns for model and HP optimization based on ML and DL algorithms for small datasets and proposed combining supervised and unsupervised learning to enable the optimization of RBFs as classifiers for conventional CNN architectures. So far, neural architecture search has been aimed at minimizing the number of flops and latency in inference regardless of sample efficiency. Sample efficiency is another challenge in practical applications where data or labels are scarce. Hence, architectures searched for the highest sample efficiency are critical for practical applications. 
Other directions for future research include limiting search space and constraining optimization techniques to more explainable and robust methods that serve the purposes of trustworthy AI. The current top-performing computer vision models are derived from automatically searched architectures that target latency optimization and disregard models' explainability. There is a belief in a trade-off between accuracy and explainability in the scientific community~\cite{yang2022unbox}. The drop in accuracy occurs when predictive complexities are removed to make the models more explainable. However, another critical research piece refers to this trade-off as a myth~\cite {rudin2019stop} and encourages researchers to optimize intrinsically interpretable models to the same level of performance as black box models. Neural architecture search in the space of intrinsically interpretable and explainable models clarifies this controversial research area to show the correctness of these contradicting opinions.



\bibliographystyle{plain}
\bibliography{references.bib}

%




\end{document}